\documentclass{article}

\usepackage{palatino}
\usepackage[letterpaper,margin=1in]{geometry}

\usepackage{natbib}

\usepackage{graphicx}
\usepackage{subfigure}
\usepackage{booktabs}
\usepackage{hyperref}

\usepackage{amsmath}
\usepackage{amsthm}

\usepackage{amssymb,amsopn,algorithm,algorithmic,float,bbm,bm,enumerate,color,multirow,gensymb}

\usepackage{epsfig,subfigure,graphicx}
\usepackage{comment}
\usepackage{authblk}
\usepackage{afterpage}
\usepackage{thmtools, thm-restate}

\usepackage[dvipsnames]{xcolor}
\usepackage[]{color-edits}
\allowdisplaybreaks[4]

\addauthor{ab}{magenta}
\addauthor{yz}{blue}
\addauthor{tc}{purple}
\newcommand{\beq}{\begin{equation}}
\newcommand{\eeq}{\end{equation}}

\newtheorem{theo}{Theorem}

\newtheorem{lemm}{Lemma}

\newtheorem{corr}{Corollary}

\theoremstyle{definition}

\newtheorem{example}{Example}

\newcommand\I{\mathbb{I}}

\newcommand\R{\mathbb{R}}

\newcommand\E{\mathbb{E}}

\renewcommand\P{\mathbb{P}}

\newcommand\1{\mathbbm{1}}

\newcommand{\x}{\mathbf{x}}
\newcommand{\y}{\mathbf{y}}

\newcommand{\cD}{{\cal D}}

\newcommand{\cH}{{\cal H}}

\newcommand{\cN}{{\cal N}}
\newcommand{\cP}{{\cal P}}
\newcommand{\cQ}{{\cal Q}}

\newcommand{\cX}{{\cal X}}
\newcommand{\cY}{{\cal Y}}

\newcommand{\cZ}{{\cal Z}}

\newcommand{\vertiii}[1]{{\left\vert\kern-0.25ex\left\vert\kern-0.25ex\left\vert #1
    \right\vert\kern-0.25ex\right\vert\kern-0.25ex\right\vert}}

\newcommand{\myref}[1]{(\ref{#1})}

\DeclareMathOperator{\tr}{Tr}

\DeclareMathOperator{\diag}{diag}

\newcommand{\eqed}{\nolinebreak[1]~~~\hspace*{\fill} \rule{5pt}{5pt}\vspace*{\parskip}\vspace*{1ex}}

\newcommand {\commentout}[1] {}

\def\ints{{{\rm Z} \kern -.35em {\rm Z} }}  
\def\smallints{{{\rm Z} \kern -.3em {\rm Z} }}  
\def\pints{{{\rm I} \kern -.15em {\rm N} }}      
\newcommand{\reals}{\mathbb R}

\def\cplx{{{\rm I} \kern -.45em {\rm C} }}       
\def\l2{\rm {\mathcal L}^{2}(\reals)}            

\newcommand{\nr}{\nonumber}
\newcommand{\be}{\begin{eqnarray}}
\newcommand{\ee}{\end{eqnarray}}
\newcommand{\bea}{\begin{eqnarray}}
\newcommand{\eea}{\end{eqnarray}}
\newcommand{\beaa}{\begin{eqnarray*}}
\newcommand{\eeaa}{\end{eqnarray*}}
\newcommand{\bnad}{\begin{nad}}
\newcommand{\enad}{\end{nad}}

\newcommand{\sign}{{\mbox{\rm sign}}}

\title{De-randomized PAC-Bayes Margin Bounds: \\
Applications to Non-convex and Non-smooth Predictors}
           
\author{Arindam Banerjee}
\author{Tiancong Chen}
\author{Yingxue Zhou}

\affil{Department of Computer Science \& Engineering\\ 
University of Minnesota, Twin Cities\\ 
\texttt{ \footnotesize{banerjee@cs.umn.edu,\{chen6271,zhou0877\}@umn.edu}}}

\date{}
\begin{document}

\maketitle

\begin{abstract}
In spite of several notable efforts, explaining the generalization of deterministic non-smooth deep nets, e.g., ReLU-nets, has remained challenging. Existing approaches for deterministic non-smooth deep nets typically need to bound the Lipschitz constant of such deep nets but such bounds are quite large, may even increase with the training set size yielding vacuous generalization bounds. In this paper, we present a new family of de-randomized PAC-Bayes margin bounds for deterministic non-convex and non-smooth predictors, e.g., ReLU-nets. Unlike PAC-Bayes, which applies to Bayesian predictors, the de-randomized bounds apply to deterministic predictors like ReLU-nets. A specific instantiation of the bound depends on a trade-off between the (weighted) distance of the trained weights from the initialization and the effective curvature (`flatness') of the trained predictor. 

To get to these bounds, we first develop a de-randomization argument for non-convex but smooth predictors, e.g., linear deep networks (LDNs), which connects the performance of the deterministic predictor with a Bayesian predictor. We then consider non-smooth predictors which for any given input realized as a smooth predictor, e.g., ReLU-nets become some LDNs for any given input, but the realized smooth predictors can be different for different inputs. For such non-smooth predictors, we introduce a new PAC-Bayes analysis which takes advantage of the smoothness of the realized predictors, e.g., LDN, for a given input, and avoids dependency on the Lipschitz constant of the non-smooth predictor. After careful de-randomization, we get a bound for the deterministic non-smooth predictor. We also establish non-uniform sample complexity results based on such bounds. Finally, we present extensive empirical results of our bounds over changing training set size and randomness in labels.
\end{abstract}

\tableofcontents

\newpage

\section{Introduction}
\label{sec:intro}


Recent years have seen several notable efforts to explain generalization of deterministic deep networks, e.g., ReLU-nets, Res-nets, etc. \citep{bafo17, gora18, nebh18,lose19, frei2019algorithm}. The classical approach to generalization bounds typically considers two terms \citep{bame02,bama99,ssbd14,mohri_foundations_2018}: a first term characterizing the empirical performance often at a certain margin and a second term characterizing the capacity/complexity of the class of predictors under consideration. The classical approach has so far struggled to explain the empirical performance of deep nets which perform surprisingly well on the training set even with random labels but is capable of generalizing well on real problems \citep{zhbh17}. Such struggles have led to calls for rethinking the classical approach to generalization \citep{zhbh17}, the need to consider implicit bias~\citep{nets15,soho18}, and concerns regarding the effectiveness of using uniform convergence for such analysis \citep{nako19b}.

The literature has broadly two types of generalization bounds for deep nets: results which apply to the original non-smooth deterministic network \citep{nako18,nebh18,bafo17,lilu18,gora18} and results which apply to a modified and/or restricted network possibly with suitable restrictions on the learning algorithm \citep{cagu19a, arora_stronger_2018, ardu19, duzh2019,soho18,guls2018implicit}. 
The focus of the current work is on the first type of bounds. Notable advances have been made for such bounds in recent years including approaches based on bounding the Rademacher complexity \citep{bafo17,gora18,lilu18} or de-randomized PAC-Bayes bounds \citep{nako18,nebh18}, among others \citep{lose19}. Getting suitable margin bounds for non-smooth deep nets from such approaches typically require a characterization of the Lipschitz constant of the deep net~\citep{nebh18,bafo17}. Existing bounds on the Lipschitz constant are based on product of layer-wise spectral norms which can be quite large, and can yield vacuous bounds~\citep{nako19b}.


In this paper, we present margin bounds on the generalization error of deterministic non-convex and non-smooth deep nets 
based on a new de-randomization argument on PAC-Bayes bounds. At a high level, there are three key aspects to our analysis. First, we show that for {\em detereministic smooth predictors}, one can de-randomize PAC-Bayes bounds to get suitable margin bounds. Second, we show that for {\em deterministic non-smooth predictors} such as ReLU-nets, one can carefully extend the de-randomization strategy for smooth predictors to get deterministic margin bounds for such non-smooth predictors. The bounds we present are {\em non-uniform bounds}, which holds with high probability for all predictors, but the exact bound is different for each predictor. Third, the PAC-Bayesian bounds we consider are based on Gaussian posterior and prior distributions, and the generalization error bound depends on the KL-divergence between these distribution. We give examples of different choices of the prior distributions and posterior distributions and the corresponding deterministic generalization bound. A key example considers an anisotropic Gaussian posterior distribution, where the anisotropy depends inversely on the Hessian of the loss~\citep{denker1991transforming,mackay1992practical}, suitably defined for non-smooth predictors. 
We highlight key facets of each of these aspects below.


\subsection{De-Randomization for Smooth Predictors}
First, we establish a de-randomized PAC-Bayes margin bound for non-convex but smooth predictors, e.g., linear deep networks (LDNs) with a fixed structure. The analysis is inspired by a classical de-randomization argument for linear predictors \citep{langford2003pac,mcda03} suitably generalized to smooth non-convex predictors. 
For a training set $S=\{(x_i,y_i), i=1,\ldots,n\}$, let $\phi^{\theta^\dagger}$ be a smooth predictor learned from $S$ where $\theta^\dagger \in \R^p$ denotes the learned parameters of the predictor. Consider any prior Gaussian distribution $\mathcal{P}$ over the parameters chosen before training, and let posterior $\mathcal{Q}$ be a multivariate Gaussian distribution with mean $\theta^{\dagger}$. Then, ignoring constants and certain other details, an informal version of the generalization bound for smooth predictors is as follows: with high probability, for any $\gamma > \gamma_0$ and all $\theta^\dagger$, we have 
$$
\ell_{0}\left(\phi^{\theta^{\dagger}}, D\right) \leq  \ell_{\gamma}\left(\phi^{\theta^{\dagger}}, S\right)+\frac{1}{2 n} K L(\mathcal{Q} \| \mathcal{P})+ c_0 \exp \left(-\min \left(c_{2} \gamma^{2}, c_{1} \gamma\right)\right)
$$
where $D$ denotes the true underlying distribution for $(x,y)$, $\ell_{\gamma}(\phi^{\theta^\dagger},W)$ denotes the margin loss~\footnote{Margin loss is defined at the end of Section~\ref{sec:intro} under notation.} at $\gamma$ with samples drawn following distribution $W$, and $\gamma_0,c_0,c_1,c_2$ are constants. The de-randomization argument yields:
\begin{itemize}
\item a bound for the deterministic predictors $\phi^{\theta^\dagger}$, which makes it different from standard PAC-Bayes bounds which only apply to Bayesian predictors, 
\item the bound holds with high probability for all such $\phi^{\theta^\dagger}$, but 
\item the bound is potentially different for each predictor because $\cQ$ depends on $\theta^\dagger$, which makes it different from standard uniform bounds, e.g., based on Rademacher complexity.
\end{itemize}

As a special example of the posterior, we consider an anisotropic Gaussian with diagonal covarince depending inversely on the curvature of the loss at $\theta^\dagger$, i.e., diagonal elements of the Hessian $\cH_{l,\phi}^{\theta^{\dagger}} \triangleq \frac{1}{n} \sum_{i=1}^n \nabla^2 l(y_i,\phi^{\theta^{\dagger}}(x_i))$, where $l$ denotes the cross-entropy loss for classification. The technicalities behind modeling the posterior covariance with the inverse Hessian has been extensively studied in the literature \citep{denker1991transforming,mackay1992practical}.
%
%
Let $\sigma^2 \in \R_{++}, \theta_0 \in \R^p$ be chosen before seeing the training set. Then, 
ignoring constants and certain other details, an informal version of the generalization bound for smooth predictors is as follows: with high probability, for any $\gamma > \gamma_0$ and all $\theta^\dagger$, we have 
\begin{equation*}
\ell_{0}(\phi^{\theta^{\dagger}},D)  \leq \ell_{\gamma}(\phi^{\theta^{\dagger}},S) + 
\frac{1}{2n} \bigg(  \underbrace{\sum_{j =1}^{p} \ln \frac{\max\{ \cH_{l,\phi}^{\theta^{\dagger}}[j,j], 1/\sigma^2\}}{1/\sigma^2}}_{\textup{effective curvature}}  +  \underbrace{\frac{\|\theta^{\dagger}-\theta_0\|_2^2}{\sigma^2}}_{\textup{$L_2$ norm}} \bigg) + c_0 \exp\left( -\min(c_2 \gamma^2, c_1 \gamma) \right) ~,
\end{equation*}
Note that the first term considers the empirical margin loss at a certain $\gamma$ and the last term has a mixed tail decay in terms of $\gamma$. The (weighted) $L_2$ norm term is related to the distance from the initialization  which typically shows up in certain existing bounds, especially ones using Gaussian distributions in PAC-Bayes~\citep{nebh17,nebh18,dzro18}. As we discuss in the sequel, the bound straightforwardly extends to the more flexible setting of anisotropic prior. The effective curvature term plays an important role in the bound. Empirically, only few of the Hessian diagonal elements are large, i.e., cross the threshold  $1/\sigma^2$ (see Section~\ref{sec:expt}), and the contribution from all the other terms is 0. Thus, although the term has a summation over all $p$ dimensions, only the parameters with sharp curvature contribute to the effective curvature term. 
The effective curvature term and the $L_2$ norm terms also illustrate a trade-off: a small value of $\sigma$ sets the threshold $1/\sigma^2$ to be high, possibly making the effective curvature term small or even completely wiping out the term, but the $L_2$ norm term would be larger due to small $\sigma$; a larger $\sigma$ would have the opposite impact. The bounds can be made more flexible by using anisotropic priors, and we discuss this aspect in Section~\ref{ssec:sec1ns}.


\subsection{De-randomization for Non-smooth Predictors}
\label{ssec:sec1ns}
Second, we establish a bound for non-convex and non-smooth predictors, e.g., ReLU-nets, Res-nets, etc., by utilizing the bound for smooth predictors. The crux of the argument utilizes a self-evident but tricky-to-use fact: for any specific input, a deterministic ReLU-net (and many other deep nets) effectively becomes a linear deep net (LDN) which only includes the active edges from the ReLU-net. For a deep net $\psi^\theta$ with parameters $\theta \in \R^p$, the structure of the realized LDN can be represented as a binary vector $\xi_x^\theta \in \{0,1\}^p$ for input $x$, where 1 denotes an active edge, and the realized LDN will have parameters $\theta \odot \xi_x^\theta$, where $\odot$ denotes the Hadamard project.
Utilizing such realized LDNs for analysis seems doomed from the start because the realized LDN depends on the input $x$ and the structure $\xi_x^\theta$ of such LDNs, being discrete objects, are not even a continuous functions of the input. 
To make progress, for any fixed input $x$, we first develop a de-randomization argument which connects
\begin{itemize}
\item the performance of a Bayesian predictor over LDNs with parameters $\theta \odot \xi_x^\theta$, where the posterior over $\theta$ is a (potentially anisotropic) Gaussian distribution with mean $\theta^\dagger$ to 
\item the performance of a deterministic predictor whose parameter is $\theta^\dagger \odot \xi_x^{\theta^\dagger}$,
\end{itemize}
where $\theta^\dagger$ is the learned parameter of the network and the mean of the Gaussian posterior for the Bayesian predictor. 
The de-randomization argument establishes and utilizes a novel martingale difference sequence (MDS) associated with $\xi_x^\theta$ where the layers of the deep net serve as steps of the MDS. The analysis for the single input $x$ is subsequently extended to all inputs by taking suitable expectations, yielding a bound for the deterministic non-smooth predictor. The analysis avoids having to explicitly bound the Lipschitz constant of non-smooth ReLU-nets \citep{nako18,nebh18,bafo17,gora18} and establishes an interesting connection with LDNs~\citep{arco18,labr18}. 

Based on the above perspective, for any input $x$, the non-smooth predictor 
\begin{equation*}
    \psi^{\theta^\dagger}(x) = \phi^{\theta^\dagger \odot \xi_x^{\theta^\dagger}}(x)~, 
\end{equation*}
where $\phi$ denotes a LDN with parameters $\theta^\dagger \odot \xi_x^{\theta^\dagger}$. 
Then, ignoring constants and certain other details, an informal version of the generalization bound for non-smooth predictors is as follows: with high probability for any $\gamma > \gamma_0$ and all $\theta^\dagger$, we have 
$$
\ell_{0}\left(\psi^{\theta^{\dagger}}, D\right) \leq  \ell_{\gamma+2 \varrho_{k}}\left(\psi^{\theta^{\dagger}}, S\right)+\frac{1}{2 n} K L(\mathcal{Q} \| \mathcal{P})+c_0 \exp \left(-\min \left(c_{2} \gamma^{2}, c_{1} \gamma\right)\right)
$$
where the subscript $k$ is the depth of ReLU-nets and $\varrho_k$ is an additional margin due to the non-smoothness which depends on $\| \theta^\dagger\|_2$. Such a dependence does not become an issue because the margin is unnormalized and depth $k$ ReLU-nets are positively homogeneous of degree $k$, so the extra margin can be handled by suitable scaling the parameters, as we show in the sequel.

As before, as a special example of the posterior, we consider an anisotropic Gaussian with diagonal covarince depending inversely on the curvature of the loss at $\theta^\dagger$, i.e., diagonal elements of the Hessian $\cH_{l,\psi}^{\theta^{\dagger}} \triangleq \frac{1}{n} \sum_{i=1}^n \nabla^2 l(y_i,\phi^{\theta^{\dagger} \odot \xi_{x_i}^{\theta^\dagger}}(x_i))$, where $l$ denotes the cross-entropy loss for classification \citep{denker1991transforming,mackay1992practical}. 
%
%
As before, let $\sigma^2 \in \R_{++}, \theta_0 \in \R^p$ be chosen before seeing the training set. Then, ignoring constants and certain other details, an informal version of the generalization bound for non-smooth predictors is as follows: with high probability 
\begin{equation*}
\ell_{0}(\psi^{\theta^{\dagger}},D)  \leq \ell_{\gamma+2\varrho_k}(\psi^{\theta^{\dagger}},S) + 
\frac{1}{2n} \bigg(  \underbrace{\sum_{j =1}^{p} \ln \frac{\max\{ \tilde \cH_{l,\psi}^{\theta^{\dagger}}[j,j], 1/\sigma^2\}}{1/\sigma^2}}_{\textup{effective curvature}}  +  \underbrace{\frac{\|\theta^{\dagger}-\theta_0\|_2^2}{\sigma^2}}_{\textup{$L_2$ norm}} \bigg) + c_0 \exp\left( -\min(c_2 \gamma^2, c_1 \gamma) \right) ~,
\end{equation*}
Further, one can choose different prior $\omega_j \leq \sigma, j=1,\ldots,p$ corresponding to each parameter and the bound straightforwardly extends  to such anisotropic prior based on such $\omega_j$. Note that for such anisotropic prior, we get the following form for the middle terms:
\begin{equation*}
 \underbrace{\sum_{j =1}^{p} \ln \frac{\max\{ \cH_{l,\psi}^{\theta^{\dagger}}[j,j], 1/\omega_j^2\}}{1/\omega_j^2}}_{\textup{effective curvature}}  +  \underbrace{\sum_{j=1}^p \frac{(\theta_j^{\dagger}-\theta_{0,j})^2}{\omega_j^2}}_{\textup{$L_2$ norm}} ~.
\end{equation*}

While we do not consider quantitatively tightening the bounds in the current work, one can possibly do that by suitable choices of $\omega_j,\theta_0$, e.g., based on differential privacy \citep{dzda17,dziugaite2018entropy,dzro18}. With such choices, sharper bounds would have the following qualitative behavior:
\begin{itemize}
    \item $\theta_j^\dagger \approx \theta_{0,j}$: For parameters which have not moved much during training, i.e., lazy parameters, the bound can be made less dependent on the curvature $\cH_{l,\psi}^{\theta^{\dagger}}[j,j]$, e.g., one can suitably choose a small $\omega_j$ making the threshold $1/\omega_j^2$ high which reduces the effective curvature term. In other words, even for sharp bounds, it is ok for lazy parameters to have some amount of curvature after training, i.e., they need not be along `flat' directions.
    \item $\theta_j^\dagger \not \approx \theta_{0,j}$: For parameters which have moved a lot during training, i.e., active parameters, there is more dependence on the curvature $\cH_{l,\psi}^{\theta^{\dagger}}[j,j]$, e.g., one can suitably choose a large $\omega_j$ to reduce the (weighted) Euclidean distance term, thereby making the threshold $1/\omega_j^2$ small which increases the effective curvature term. In other words, for sharp bounds, the curvature for active parameters need to be small after training, i.e., they need to be along 'flat' directions. 
\end{itemize}
We extensively study such qualitative insights empirically in Section~\ref{sec:expt}. In particular, with increase in the fraction of random labels, we observe that both the (weighted) $L_2$ norm term and the effective curvature term increases, yielding larger bounds whereas the training set error stays at zero.

We extend the above analysis and establish sample complexity bounds corresponding to the above bound, i.e., given any $\epsilon, \delta$, how many samples $n_0(\epsilon,\delta)$ do we need so that with probability at least $(1-\delta)$, the true error rate of any ReLU-net on the underlying distribution $D$ is at most $\epsilon$ more than the empirical error rate of the ReLU-net? A unique aspect of the bound as outlined above is that, unlike uniform bounds say based on Rademacher complexity, the bound is specific to each $\phi^{\theta^\dagger}$, and different for different predictors, relying on the corresponding (weighted) $L_2$ norm and effective curvature. As a result, the associated sample complexity result is {\em non-uniform}, i.e., the sample complexity $n_0(\epsilon,\delta,\psi^{\theta^\dagger})$ depends on the predictor $\psi^{\theta^\dagger}$. Such non-uniform bounds~\citep{benedek_nonuniform_1988,benedek_nonuniform_1994} are in sharp contrast to the more widely used uniform bounds~\citep{kopa00,bame02}, where the bound and the resulting sample complexity is the same for all predictors in a hypothesis class and depends only on properties such as VC dimensions and Rademacher complexities of the hypothesis class. 



\subsection{Posterior for PAC-Bayes: Anisotropy using the Hessian}
Third, for both the smooth and non-smooth setting, we establish the de-randomized PAC-Bayes margin bounds by considering suitable anisotropic posteriors based on the diagonal elements of the Hessian of the loss corresponding to the learned ReLU-net. For smooth predictors, we directly use the average Hessian of the loss with the smooth predictor. For non-smooth predictors, as outlined above, we consider the average Hessian of the loss with the realized LDN for each input. Such a Hessian  is well defined and the diagonal elements are computable from the training set. 
In spite of the dependency on the Hessian, which can be changed based on re-parameterization without changing the function~\citep{smle18,dipb17}, the bound itself is scale-invariant since KL-divergence is invariant to such re-parameterizations~\citep{kl2011,ligu19}. As discussed above, the resulting bounds depend on a trade-off between the effective curvature and the (weighted) $L_2$ norm, i.e., (weighted) Euclidean distance from the initialization of the learned parameters. The trade-off gets determined by the marginal prior variance $\sigma$ (or $\sigma_j$). Qualitatively, the bound will be small if parameters which have moved away from the initialization have small curvature whereas parameters which have stayed close to the initialization can have large curvature. The bound will be really small, implying good generalization, if after training the parameters stay close to the initialization and have small curvature. The bound provides a concrete realization of the notion of `flatness' in deep nets \citep{smle18,hosc97b,kemn17} and illustrates a trade-off between curvature and distance from initialization. The bounds leaves the room open for further quantitative sharpening using ideas which are getting explored in the recent literature~\citep{dzda17,dzro18}. 



The bounds we propose passes several sanity checks both in theory and through experiments. First, the bounds apply to the original non-smooth deterministic deeps net~\citep{gora18,nebh18}, e.g., ReLU-net, Res-net, CNNs, etc.
Second, empirically, the bound decreases with an increase in the number of training samples and the behavior holds up across changes in depth, width, and mini-batch size. This is in contrast with certain existing bounds which may even increase with an increase in the number of training samples~\citep{nako19b,bafo17,gora18,nebh18}. Third, the bound increases with increase of random labels although the training set error goes to zero \citep{zhbh17,nebh17}. Empirically, both the effective curvature and the distance from the initialization increases with increase in number of random labels.  
Finally, without any optimization, the bounds are meaningful and non-vacuous, and can be quantitatively sharpened based on recent advances in PAC-Bayes bounds \citep{yasu19,dzro18}. 

The rest of the paper is organized as follows. In Section~\ref{sec:related}, we review existing theory for the generalization bound of deep neural networks and the study of the geometry of the Hessian.
In Section~\ref{sec:margin2nc_updated}, we present bounds for deterministic non-convex but smooth predictors (proofs in Appendices~\ref{app:margin2nc_updated} for 2-class, multi-class in Appendix~\ref{app:margink}). 
In Section~\ref{sec:margin2ns}, we present bounds for deterministic non-convex and non-smooth predictors (proofs in Appendix~\ref{app:margin2ns}). 
We present experimental results in Section~\ref{sec:expt} and conclude in Section~\ref{sec:conc}.

\noindent {\bf Notation.} For ease of exposition, we present results for the 2-class case, and relegate the $k$-class case to Appendix~~\ref{app:margink}. 
For 2-class, we denote smooth predictors $\phi: \R^p \times \R^d \mapsto \R$ as $\phi^{\theta}(x), \theta \in \R^p, x \in \R^d$. The true labels $y \in \{-1,+1\}$ and predicted labels $\hat{y}=\sign(\phi^{\theta}(x))$. We denote the training set as $S$ and true data distribution as $D$. For any distribution $W$ on $\cX \times \cY$ and any $\beta \in \R$, we define the margin loss as
\begin{align*}
\ell_{\beta}(\phi^{\theta},W) \triangleq \P_{(x,y) \sim W}\left[  y\phi^{\theta}(x) \leq \beta\right]~.
\end{align*}
For a Bayesian predictor, we maintain a distribution $\cQ$ over the parameters $\theta$, and the corresponding margin loss is
$\ell_{\beta}(\cQ,W)  \triangleq \E_{\theta \sim \cQ}[\ell_{\beta}(\phi^{\theta},W)]$.
For $k$-class, with $\phi: \R^p \times \R^d \mapsto \R^k$, the predictions $\phi^{\theta}(x) \in \R^k$. Further, the margin loss 
\begin{align*}
\ell_{\beta}(\phi^{\theta},W) \triangleq \P_{(x,y) \sim W}\left[  \phi^{\theta}(x)[y] \leq \max_{\tilde y \neq y} \phi^{\theta}(x)[\tilde y] + \beta\right]~,
\end{align*}
and, as before, $\ell_{\beta}(\cQ,W)  \triangleq \E_{\theta \sim \cQ}[\ell_{\beta}(\phi^{\theta},W)]$. 
We denote non-smooth predictors as $\psi^{\theta}(x)$ with the rest of the notation inherited from the smooth case. $c_0$ denotes an absolute constant noting that $c_0$ can change across equations.


\section{Related Work}
\label{sec:related}
Since traditional approaches that attribute small generalization error either to properties of the model family or to the regularization techniques fail to explain why deep neural networks generalize well in practice \citep{zhbh17, nebh17}, 
several different theories have been suggested to characterize the generalization error of deep nets. These rely on measures such as the PAC-Bayes theory \citep{mcda99}, Rademacher complexity \citep{bame02}, `flat minima' \citep{hosc97b}, algorithmic stability \citep{hare16}, and more. In this section, we review existing theory and bounds for characterizing the generalization error of deep neural networks and the study of the geometry of the Hessian of the loss function. 

\noindent\textbf{PAC-Bayesian Bound.} The PAC-Bayesian theory has been proven useful in various areas, including classification \citep{langford2003pac,parrado2012pac,lacasse2007pac,germain2009pac},
high-dimensional sparse regression \citep{alquier2013sparse,guedj2013pac},
algorithmic stability \citep{london2014pac,london2017pac},
and many others. The first PAC-Bayesian inequality was introduced by \cite{mcda99,mcallester1999some}, based on the earlier work by \cite{shawe1997pac} which introduced the first PAC style analysis of a Bayesian style classification estimator. This inequality has been further extended to the KL-divergence between the in-sample and out-sample risk by \citep{langford2001bounds,seeger2002pac,langford2005tutorial}. Later on, the general framework of PAC-Bayesian theorem, which unifies the distance between in-sample and out-sample risk by a convex function, was introduced by \cite{begin2014pac,begin2016pac}. Many useful results are under the PAC-Bayesian framework including the classical theorem by \cite{mcallester1999some}, \cite{langford2001bounds}, the `fast-rate' form  \citep{cato07} adopted in this work, and others \citep{alquier2016properties}. 
Most recent works on PAC-Bayesian theory has seen a growing interest in data-dependent priors \citep{dziugaite2018entropy,dzro18}, which are also connected to stability \citep{boel02}. To our specific goal of interests in explaining neural networks, \cite{langford2002not} started the thread by applying PAC-Bayesian bounds to two-layer stochastic neural networks. Recently, PAC-Bayesian theory has been widely explored in explaining generalization for deep nets \citep{ nebh18,nako18}. The bottleneck of these work is the natural property of PAC-Bayesian frameworks which only works for stochastic predictors. Most recent works \citep{nebh17,nebh18,nako18,dzro18} provided generalization guarantees for the deterministic networks by different de-randomization methods, which are further shown vacuous on real world datasets \citep{nako19b}.

\noindent\textbf{Rademacher Complexity.} The measurement of Rademacher complexity \citep{bame02} has been explored to derive the generalization bound of deep nueral networks which depends on the size of the neural network, i.e., depth, width and norm of weights in each layer (usually Frobenius norm and spectral norm). 
\cite{nets15} established results that characterize the generalization bounds 
in terms of the depth and Frobenius norms of 
weights in each layer which scales exponential with the depth even assuming the Frobenius norms of weights is bounded. \cite{bafo17}
used the tool of covering numbers to directly upper bound the Rademacher complexity. Although this bound has no explicit exponential dependence on the depth, there is a unavoidable polynomial dependence on the depth.
Recently, \cite{gora18} showed that exponential depth dependence in Rademacher complexity-based analysis \citep{nets15} can be avoided by applying contraction to a slightly different object. 
Thus, one can improve the results in \cite{nets15} from exponential dependence on depth to polynomial dependence. They also provided nearly size-independent bounds, assuming some control over the norm of the parameter matrices (which includes the Frobenius norm and the trace norm as special cases). 
This size-independent bounds are developed by
the technique showing that the neural network can be approximated by the composition of a shallow network and univariate Lipschitz functions. 
Recently, \cite{lilu18} established a generalization error bound for a general family of
deep neural networks including CNNs, ResNets by bounding the empirical Rademacher complexity through introducing a new Lipschitz analysis for deep neural networks. Their bound also scales with the product of the spectral norm of the weights in each layer. The lower bound in \cite{gora18} showed that such dependence on the product of norms across layers is generally inevitable for the Rademacher complexity analysis. 

Note that other than \cite{lilu18}, there are other studies on generalization bound for CNNs. \cite{du2018many} proved bounds for CNNs in terms of the number of parameters, for two-layer networks. \cite{arco18} analyzed the generalization of networks
output by a compression scheme applied to CNNs. \cite{zhou2018understanding} provided a generalization
guarantee for CNNs satisfying a constraint on the rank of matrices formed from their kernels.  \cite{lee2019wide} provided a size-free bound for CNNs in a general unsupervised learning framework that includes PCA learning. \cite{lose19} proved bounds on the generalization error of CNNs in terms of the training loss, the number of parameters, the Lipschitz
constant of the loss and the distance from the weights to the initial weights.

\noindent\textbf{`Flat Minima'.} The concept of generalization via achieving flat minima was first proposed in \cite{hosc97b}. Based on minimum description length (MDL) principle, they suggested that the `flat' minima of the objective function generalizes well, because the flat minimum corresponds to `simple' networks and low expected overfitting. Motivated by such an idea, \cite{chaudhari2019entropy} proposed the Entropy-SGD algorithm which biases the parameters to wide valleys to guarantee generalization. \cite{kemn17} showed that small batch size can help SGD converge to flat minima, which validates their observations that neural networks trained with small batch generalize better than  those trained with large batch. However, for deep nets with positively homogeneous activation functions, most measures of sharpness/flatness and norm are not invariant to re-scaling of the network parameters (`$\alpha$-scale transformation' \citep{dipb17}). This means that the measure of flatness/sharpness can be arbitrarily changed through re-scaling without changing the generalization performance, rendering the notion of `flatness' meaningless.  
To handle the sensitivity to reparameterization, 
\cite{smle18} explained the generalization behavior through `Bayesian evidence', which penalizes sharp minima but is invariant to model reparameterization. 

\noindent\textbf{Algorithmic Bound.} Stochastic gradient descent (SGD) method and its variants are algorithms of choice for many Deep Learning tasks. Different algorithmic choices for optimization such as the initialization, update
rules, learning rate, and stopping condition, will lead to different minima with different generalization behavior \citep{nebh17}. Generalization behavior depends implicitly on the algorithm used to minimize the training error. Thus, training algorithms has been studied to explain the generalization ability of neural networks. \cite{nets15,guls2018implicit,soho18} considered implicit bias
of gradient descent as the cause of good generalization of neural network. \cite{hare16} discussed how stochastic
gradient descent ensures uniform stability, thereby helping generalization for convex objectives. Recently,  
\cite{ardu19} proposed a generalization bound of SGD for training neural networks independent of network size, using a data-dependent complexity measure, namely ` Gram matrix', to explain why true labels give faster convergence rate and better generalization behavior than random labels. Specifically, they gave new analysis for overparameterized two-layer neural networks with ReLU activation trained by gradient descent, when the number of neurons in the hidden layer is sufficiently large. Later, \cite{cagu19a,frei2019algorithm} presented generalization error bound of stochastic gradient descent for learning over-parameterized deep nets.

\noindent\textbf{PAC Learning Model and Uniform/Non-uniform Convergence.} PAC learning
was introduced by \cite{vali84}, which gives sample complexity for a certain hypotheses class.
The history of uniform convergence can date back to 1930s when \cite{glivenko1933sulla,cantelli1933sulla} proved the first uniform convergence result, and provided the classes of functions for which uniform convergence holds, which are also called Glivenko-Cantelli classes. The relation of PAC learnability and uniform convergence are thoroughly studied in \cite{vapnik1992principles,vapnik1999overview,vapnik2013nature}, where they characterized PAC learnability of classes of binary classifiers using VC-dimension introduced by \cite{vapnik1968uniform}. Except for binary classification problems, there is no equivalence between learnability and uniform convergence in general \citep{shalev-shwartz_learnability_2010}. The use of Rademacher complexity for bounding uniform convergence is due to \cite{kopa00,bame02}, which has become the primary approach to provide generalization guarantees \citep{bousquet2002concentration,boucheron2005theory,bartlett2005local}. \cite{nako19b} provided both theoretical and empirical evidence that existing uniform bounds toolbox are vacuous on both real world and artificial datasets, which posted questions on the power of uniform bounds. As a response, \cite{negrea_defense_2019} extended the classical Glivenko-Cantelli classes to structural Glivenko-Cantelli classes to formalize the specific failure of uniform convergence. Different from the mainstream usage of uniform convergence, non-uniform convergence did not get too much attention. The concept of non-uniform bounds was introduced by \cite{benedek_nonuniform_1988}. Some additional details can be found in the extended version by \cite{benedek_nonuniform_1994}. Chapter 7 in \cite{ssbd14} discussed the non-uniform learnability and the computational aspects for countable hypothesis classes.

\noindent\textbf{Geometry of Hessian.} The empirical analysis of the Hessian of the Neural Networks has drawn attention in the deep learning community. \cite{sabo16,sagun_empirical_2017} studied the spectrum of the Hessian for two layer feed forward network. They showed that the eigenvalues are composed of a `bulk' concentrated around zero which includes most of the eigenvalues and a few outliers emerging from the bulk. Later on, \cite{papyan_full_2018,papyan2019measurements} observed a similar structure of the Hessian when training larger neural networks such as VGG, Res-nets on MNIST and CIFAR-10 datasets.
They analyzed such a structure by decomposing the Hessian with the covariance matrix of the stochastic gradients and the averaged Hessian of predictions. \cite{ghorbani_investigation_2019} introduced a spectrum
estimation methodology and captured the same Hessian behavior on ImageNet dataset. Inspired by the Hessian structure, \cite{ligu19} studied the connection between the generalization of neural networks and Hessian structures.

\section{Bounds for Smooth Predictors}
\label{sec:margin2nc_updated}

We consider smooth predictors $\phi^{\theta}(x)$ and focus on the 2-class case, and delegate similar analysis for the $k$-class case to the supplementary. 
The smoothness of interest in the context of our analysis is that w.r.t.~$\theta$ rather than $x$, i.e.,
for any fixed $x \in \cX$, for any $\theta_1, \theta_2$, we assume 
\begin{equation}
    \begin{split}
    \phi^{\theta_1}(x) =  ~\phi^{\theta_2}(x) + \langle \theta_1 - \theta_2,\nabla_{\theta_2} \phi^{\theta_2}(x) \rangle  
    + \frac{1}{2} (\theta_1-\theta_2)^T H_{\phi}^{\tilde{\theta}}(x) (\theta_1 - \theta_2),
\end{split}
\label{eq:smooth}
\end{equation}
where $\tilde{\theta} = \tau \theta_1 + (1-\tau) \theta_2$ for some $\tau \in [0,1]$ and $H_{\phi}^{\theta}(x) = \nabla^2_{\theta} \phi^{\theta}(x)$ denotes the Hessian of the predictor. 
We make the following assumption for our analysis:


\begin{restatable}{asmp}{asmpgen}
\label{asmp:gen}
	$\phi^{\theta}(x)$ is smooth as in \eqref{eq:smooth} such that
	\begin{enumerate}
		\item  the gradients have bounded $L_2$-norm, i.e., $\| \nabla \phi^{\theta}(x) \|_2^2 \leq G^2$ for all $\theta, x$; and
		\item the Hessian  $H_\phi^{\theta}(x) = \nabla_{\theta}^2 \phi^{\theta}(x)$ is bounded, i.e.,
       $-H \preceq H_\phi^{\theta}(x) \preceq H$, where $H$ is positive semi-definite with spectral norm $\|H\|_2\leq \zeta$. 
	\end{enumerate} 
We denote the stable rank  with $\kappa:=\frac{\|H\|_F^2}{\|H\|_2^2}$ and the intrinsic dimension with $\alpha :=\frac{\tr(H)}{\|H\|_2}$. 
\end{restatable}
Definitions of stable rank and intrinsic dimension mildly differ in the literature. We follow the definitions in \cite{vers18}.

\subsection{Bounds for Stochastic vs.~Deterministic Smooth Predictors}
\label{ssec:dws}

The PAC-Bayes analysis needs suitable choices for prior $\cP$ and posterior $\cQ$. 
We consider Gaussian prior $\cP$
chosen before training. 
With $\theta^{\dagger}$ denoting the learned parameters after training on $S$, we choose $\cQ = \cN(\theta^{\dagger},\Sigma_{\theta^{\dagger}})$, 
an anisotropic Guassian with marginal variances bounded by some $\sigma >0$, i.e.,
$\Sigma_{\theta^\dagger} = \diag(\nu^2_j)$ with $\nu^2_j = \min\{\sigma^2, \sigma^2_j\}, \forall j \in [p]$, for some
suitable choices for the marginal variances $\sigma_j$. We will discuss the choices of $\sigma_j$ and the prior distribution $Q$ in subsequent parts.

The crux of the de-randomization argument is to relate margin bounds corresponding to the stochastic predictor $\theta \sim \cQ$ and the deterministic predictor with parameter $\theta^{\dagger}$:

\begin{restatable}{theo}{theosmarg}
\label{theo:s_marg}
	Let $W$ be any distribution on pairs $(x,y)$ with $x \in \R^d$ and $y \in \{-1,+1\}$. For any $\theta^{\dagger} \in \R^d$, let $\cQ$ be a multivariate Gaussian distribution with mean $\theta^{\dagger}$ and covariance $\Sigma_{\theta^\dagger} = \diag(\nu^2_j)$ with $\nu^2_j = \min\{\sigma^2, \sigma^2_j\}, \forall j \in [p]$,  for some $\sigma^2 > 0$.
	Under Assumption~\ref{asmp:gen}, for $\tilde \gamma > 2$ and any $\beta \in \R$, 
	we have 
\begin{align} 
\ell_{\beta}(\cQ, W) &\leq \ell_{\beta + \varrho}(\phi^{\theta^{\dagger}},W)  + 4\exp(- \min(c_2\tilde \gamma^2,c_1\tilde \gamma))
\label{eq:s_aniso_swd}\\
\ell_{\beta}(\phi^{\theta^{\dagger}},W)  &\leq  \ell_{\beta +\varrho }(\cQ,W) + 4\exp(- \min(c_2\tilde \gamma^2,c_1\tilde \gamma))
\label{eq:s_aniso_dws}
\end{align}
where $\varrho=\sigma^2\zeta \alpha\tilde \gamma$, constant $c_2 =  \min\left[\frac{\sigma^2\zeta^2 \alpha^2}{2G^2},\frac{ \alpha^2}{8\kappa}\right]$, $c_1= \frac{ \alpha}{4}$ and $G$, $\kappa$, $\alpha$, $\zeta$ are as in Assumption \ref{asmp:gen}.
\end{restatable} 


We highlight key aspects of the proof, especially the dependence on the smoothness of $\phi^{\theta}(x)$ w.r.t.~$\theta$ but not the smoothness w.r.t.~$x$. While this aspect is not critical for smooth predictors, it will be key when analyzing non-smooth predictors in Section~\ref{sec:margin2ns}. 
For establishing \eqref{eq:s_aniso_swd}, we focus on the set
\begin{equation*}
\cZ_{\beta + \frac{1}{2}\sigma^2 \alpha \tilde \gamma}^{(>)}(\theta^\dagger) \triangleq \left\{ (x,y) \in \cX \times \cY | y \phi^{\theta^{\dagger}}(x) > \beta +\frac{1}{2} \sigma^2 \alpha \tilde \gamma \right\}~,
\end{equation*}
the set of points where the deterministic predictor $\phi^{\theta^\dagger}$ achieves a margin more than $\beta +\frac{1}{2} \sigma^2 \alpha \tilde \gamma$.
For any $z = (x,y) \in  \cZ_{\beta +\frac{1}{2} \sigma^2 \alpha \tilde \gamma}^{(>)}(\theta^\dagger)$, we show that
\begin{equation}
\P_{\theta \sim \cQ} \left[ y \phi^{\theta}(x) \leq \beta | z \in \cZ^{(>)}_{\beta + \frac{1}{2}\sigma^2\alpha \tilde \gamma}(\theta^{\dagger} )\right] \leq 4 \exp(- \min(c_2 \tilde \gamma^2,c_1\tilde\gamma))~.
\label{eq:smbnd1}
\end{equation}
In other words, if the deterministic predictor $\phi^{\theta^{\dagger}}(\cdot)$ has a large margin of at least $(\beta + \frac{1}{2}\sigma^2\alpha \tilde \gamma)$, then the probability that the stochastic predictor $\phi^{\theta}(x), \theta \sim \cQ$ will have a small margin of at most $\beta$ is exponentially small, i.e., $4  \exp(- \min(c_2 \tilde \gamma^2,c_1\tilde\gamma))$. The analysis utilizes the smoothness of $\phi^{\theta}(x)$ w.r.t.~$\theta$ as in \eqref{eq:smooth}, and is for a specific $z = (x,y)$.  The random linear and quadratic terms resulting from the Taylor expansion in \eqref{eq:smooth} are respectively bounded with suitable applications of the Hoeffding and Hanson-Wright inequalities~\citep{boucheron_concentration_2013,vers18}.

Further, for $z \not \in \cZ^{(>)}_{\beta + \frac{1}{2}\sigma^2\alpha \tilde \gamma}(\theta^{\dagger} )$, we simply have 
\begin{equation}
\P_{\theta \sim \cQ}\left[ y \phi^{\theta}(x) \leq \beta | z \not \in \cZ^{(>)}_{\beta + \frac{1}{2}\sigma^2\alpha \tilde \gamma}(\theta^{\dagger} )\right] \leq 1~,
\label{eq:smbnd2}
\end{equation}
where the result is still for a specific $z =(x,y)$. 
Based on the law of total probability, taking expectations w.r.t.~$z \sim W$ and utilizing \eqref{eq:smbnd1} and \eqref{eq:smbnd2}  above yields~\eqref{eq:s_aniso_swd}. The analysis for establishing \eqref{eq:s_aniso_dws} is similar.

Finally, note that the condition $\tilde \gamma > 2$ in Theorem~\ref{theo:s_marg} is not restrictive since the result is in terms of the unnormalized margin. Predictors such as LDNs are positively homogeneous of degree $k$, where $k$ is the depth of the network, so that for any $\lambda > 0$, $\phi^{\lambda \theta}(x) = \lambda^k \phi^{\theta}(x)$, i.e., the unnormalized margin can be suitably scaled by scaling the parameters. We get into the details of this aspect in Section~\ref{sec:margin2ns} (Theorem~\ref{theo:sampcomp}) when we establish sample complexity results.

\subsection{Main Result: Deterministic Smooth Predictors}
The two-sided relationships between the stochastic and deterministic predictors in Theorem~\ref{theo:s_marg} can now be used to get bounds on the deterministic predictor $\phi^{\theta^{\dagger}}(x)$. With $\gamma = \sigma^2 \zeta\alpha \tilde \gamma$, respectively choosing $\beta=0, W=D$ for \eqref{eq:s_aniso_dws} and $\beta = \gamma/2, W=S$ for \eqref{eq:s_aniso_swd}, we have
\begin{align*}
\ell_{0}(\phi^{\theta^{\dagger}},D) & \leq  \ell_{\gamma/2}(\cQ,D) +  4 \exp \left( -\min(c_2 \gamma^2,c_1\gamma) \right), \\ 
\ell_{\gamma/2} (\cQ, S)  & \leq \ell_{\gamma}(\phi^{\theta^{\dagger}},S) + 4 \exp \left( -\min(c_2 \gamma^2,c_1\gamma) \right).
\end{align*} 

With probability at least $(1-\delta)$, PAC-Bayes gives 
\begin{equation*} 
KL_B(\ell_{\gamma/2} (\cQ, S) \| \ell_{\gamma/2}(\cQ,D)) \leq \frac{KL(\cQ \| \cP) + \log \frac{1}{\delta}}{n}~, 
\end{equation*}
where $KL_B$ denotes the Bernoulli KL-divergence. For any $\eta \in (0,1)$, we unpack $KL_B$ using the `fast rate' form \citep{cato07}[Theorem 1.2.6], \citep{yasu19} to get
\begin{equation*} 
~\ell_{\gamma/2}(\cQ,D)) \leq a_{\eta} \ell_{\gamma/2} (\cQ, S) + b_{\eta} \frac{KL(\cQ \| \cP) + \log \frac{1}{\delta}}{n}~, 
\end{equation*}
where $a_{\eta} = \frac{\log (1/\eta)}{1-\eta}, b_{\eta} = \frac{1}{1-\eta}$ are the same constants in classical regret bounds for online learning \citep{bane06}. While $a_{\eta} > 1$, the above form usually yields quantitatively tighter bounds for predictors which have low margin loss $\ell_{\gamma/2} (\cQ, S)$ because of the dependence on $\frac{1}{n}$. Our bounds can also be done with the `slow rate' $\frac{1}{\sqrt{n}}$ dependence \citep{mcda03}. Lining up these bounds yields the following result:

\begin{restatable}{theo}{theosmooth}
\label{theo:smooth}
Consider any prior Gaussian distribution $\cP$ over the parameters chosen before training, and let $\theta^{\dagger} \in \mathbb{R}^p$ be the parameters of the model after training. Let $\cQ$ be a multivariate Gaussian distribution with mean $\theta^{\dagger}$ and covariance $\Sigma_{\theta^\dagger} = \diag(\nu^2_j)$ with $\nu^2_j = \min\{\sigma^2, \sigma^2_j\}, \forall j \in [p]$ for some $\sigma^2 > 0$.
Under Assumption~\ref{asmp:gen},
we have
with probability at least $1-\delta$, for any $\theta^{\dagger}$, $\eta \in (0,1), \gamma > 2\sigma^2 \zeta\alpha$, we have the following scale-invariant bound for the deterministic smooth predictor $\phi^{\theta^\dagger}$:
\begin{align*}
\ell_{0}(\phi^{\theta^{\dagger}},D)  \leq a_{\eta} \ell_{\gamma}(\phi^{\theta^{\dagger}},S) + 
\frac{b_{\eta}}{2n} KL(\cQ\|\cP)
 +d_{\eta}\exp\left( -\min(c_2 \gamma^2, c_1 \gamma) \right) +  b_{\eta} \frac{\log (\frac{1}{\delta})}{n},
\end{align*}
where $a_{\eta} = \frac{\log (1/\eta)}{1-\eta}, b_{\eta} = \frac{1}{1-\eta}$, $d_{\eta} = 4(a_{\eta}+1)$, $c_2 =  \min\left[\frac{1}{2\sigma^2 G^2},\frac{1}{8\sigma^4\kappa\zeta^2}\right],~c_1=\frac{1}{4\sigma^2\zeta}$ and $G$, $\zeta$, $\alpha$, $\kappa$ are as in Assumption \ref{asmp:gen}.
\end{restatable}


Theorem \ref{theo:smooth} shows that the generalization error of a trained deterministic model can be bounded by the empirical margin loss and the KL-divergence between prior $\cP$ and posterior $\cQ$, with additional terms.  Since empirical margin loss can be small via training, the generalization error boils down to the KL-divergence $KL(\cQ\|\cP)$. We provides the following examples of the choice of $\cP$ and covariance of $\cQ$ and give detail bounds on the KL-divergence $KL(\cQ\|\cP)$.

\begin{example} \label{exa:general_sm}
If we choose $\cP$ be an anisotropic Gaussian prior $\cN(\theta_0, \Sigma_0)$ with $\Sigma_0 = \diag(\omega_j^2)$, where $\omega_j >0,~\forall j \in [p]$ chosen before training. Note that we have defined 
$\cQ = \mathcal{N}\left(\theta^{\dagger}, \Sigma_{\theta^{\dagger}}\right),$  with  $\Sigma_{\theta^{\dagger}}=\mathrm{diag}(\nu_{i}^{2})$ where $\nu_{i}^{2}=\min \left\{\sigma^{2}, \sigma_{i}^{2}\right\}, \forall j \in[p],$ for some suitable choices for the marginal variances $\sigma_{j}$.  
Then the $KL$-divergence term $KL(\cQ\|\cP)$ becomes
\begin{align*}
KL(\cQ\|\cP) = \sum_{j=1}^p \left(\frac{\nu_j^2}{\omega_j^2} + \ln \frac{\omega_j^2}{\nu_j^2} - 1\right) +\sum_{j=1}^p\frac{(\theta_j^\dagger -\theta_{0,j})^2}{\omega_j^2}. \vspace{-3mm}
\end{align*}
Note that the first term is the Itakura-Saito distance between the posterior and prior variances. \eqed
\end{example}

\begin{example}\label{exa:hess_sm}
For the posterior 
$\cQ$, recall that we consider  $\mathcal{Q}=\mathcal{N}\left(\theta^{\dagger}, \Sigma_{\theta^{\dagger}}\right)$ where $\Sigma_{\theta \dagger}=\operatorname{diag}\left(\nu_{j}^{2}\right)$ with $\nu_{j}^{2}=\min \left\{\sigma^{2}, \sigma_{j}^{2}\right\}, \forall j \in[p]$. 
Thus we consider the covaraince  that acknowledges the curvature at $\theta^\dagger$, i.e., with cross-entropy loss for  $\phi^{\theta^{\dagger}}(x)$ at $\left(x_{i}, y_{i}\right)$ denoted by $l(y_{i}, \phi^{\theta^{\dagger}}\left(x_{i}\right))$, we consider  $\sigma_j^2 = 1/\cH_{l,\phi}^{\theta^{\dagger}}[j,j]$, where $\cH_{l,\phi}^{\theta^{\dagger}} \triangleq \frac{1}{n} \sum_{i=1}^n \nabla^2 l(y_i,\phi^{\theta^{\dagger}}(x_i))$ is the Hessian of the loss function.  
The anisotropy in the posterior can be understood as follows: for parameters $\theta^{\dagger}_j$ having high curvature $\cH_{l,\phi}^{\theta^{\dagger}}[j,j]$, the posterior variance $\nu_j^2$ is small 
so that we will not deviate too far in the $j$-th component while sampling from the posterior; on the other hand, for parameters $\theta^{\dagger}_j$ with small curvature, i.e., `flat' directions, the posterior variance is $\nu_j^2 = \sigma^2$. 
We also consider the isotropic Gaussian prior $\cN(\theta_0, \sigma^2 \mathbb{I})$  where $\sigma >0$ chosen before training. 
Let $\nu_j^2 = \min\{\sigma^2,\sigma_j^2\}$, where $\sigma_j^2 =  \frac{1}{\cH_{l,\phi}^{\theta^{\dagger}}[j,j]}$  for all $j \in [p]$, the $KL$-divergence term $KL(\cQ\|\cP)$ becomes 
\begin{equation}
KL(\cQ\|\cP) \leq 
\underbrace{\sum_{j=1}^{p} \ln \frac{\max \left\{\mathcal{H}_{l, \phi}^{ \dagger}[j, j], 1 / \sigma^{2}\right\}}{1 / \sigma^{2}}}_{\text {effective curvature }}+\underbrace{\frac{\left\|\theta^{\dagger}-\theta_{0}\right\|_{2}^{2}}{\sigma^{2}}}_{L_{2} \text { norm }}.
\end{equation}

The `effective curvature' term depends on the diagonal elements of the Hessian $\cH_{l,\phi}^{\theta^{\dagger}}$ of the loss, see also \citep{denker1991transforming,mackay1992practical}. One concern in using the Hessian $\cH_{l,\phi}^{\theta^{\dagger}}$ is its scale-dependence \citep{dipb17}, but we prove that our bound is scale-invariant. The reason is that the prior and posterior use the same basis, i.e., each dimension corresponds to a parameter, so that scaling based reparameterizations affects both the prior and posterior the same way, and does not change the KL-divergence~\citep{kl2011,ligu19}. While the anisotropic posterior  could have been constructed from the eigen-values~ rather than the diagonal elements of the Hessian $\cH_{l,\phi}^{\theta^{\dagger}}$, the resulting bound would have been dependent on the scaling of parameters~\citep{dipb17} and hence undesirable. Further, the diagonal elements of the Hessian are much easier to numerically compute compared to the eigen-values.

The diagonal elements have an interesting empirical behavior (Section~\ref{sec:expt}): a small number of diagonal elements have relatively high values and most have quite small values. Such behavior aligns well with recent results on the eigen-spectrum of the Hessian \citep{ligu19,sabo16,papyan_full_2018,papyan2019measurements,ghorbani_investigation_2019}. Further, all the diagonal elements decrease as more samples are used for training. 

The trade-off between the `effective curvature' term and the `$L_2$ norm' term in the bound comes because of our use of anisotropic posterior $\cQ$. Choosing a higher value for $\sigma^2$ diminishes the dependency on the `$L_2$ norm' term and increases the dependency on the `effective curvature' term; and vice versa. There has been recent advances in suitably choosing the prior for PAC-Bayes analysis~\citep{dzda17,dzro18}, and such advances can be applied here to get quantitatively tighter bounds. \eqed
\end{example} 
The result can be straightforwardly extended to consider an anisotropic prior for the PAC-Bayes analysis as in the following example.

\begin{example}\label{exa:hess_sm}
One can consider an anisotropic Gaussian prior with mean $\theta_0$ and covaraince $\Sigma_0 = \diag(\omega_j^2)$, where $ 0< \omega_j \leq \sigma^2,~\forall j \in [p]$ chosen before training. Let $\nu_j^2 = \min\{\omega_j^2,\sigma_j^2\}$, where $\sigma_j^2 =  \frac{1}{\cH_{l,\phi}^{\theta^{\dagger}}[j,j]}$  for all $j \in [p]$, the $KL$-divergence term $KL(\cQ\|\cP)$ becomes 
\begin{equation}
KL(\cQ\|\cP) \leq 
\underbrace{\sum_{j=1}^{p} \ln \frac{\max \left\{\mathcal{H}_{l, \phi}^{\theta^{\dagger}}[j, j], 1 / \omega_{j}^{2}\right\}}{1 / \omega_{j}^{2}}}_{\text {effective curvature }}+\underbrace{\sum_{j=1}^{p} \frac{\left(\theta_{j}^{\dagger}-\theta_{0, j}\right)^{2}}{\omega_{j}^{2}}}_{L_{2} \text { norm }} =\underbrace{\sum_{\ell =1}^{\tilde p} \ln \frac{\omega^2_{(\ell)}}{\tilde{\nu}^2_{(\ell)}}}_{\textup{effective curvature}}  +  \underbrace{\sum_{j=1}^{p}\frac{(\theta^{\dagger}_j -\theta_{0,j})^2}{\omega_j^2}}_{\textup{$L_2$ norm}},
\end{equation}
where $\tilde p = | \{ j : \cH_{l,\phi}^{\theta^{\dagger}}[j,j] > 1/ \omega_j^2 \}|$, and  $\{ \tilde \nu_{(1)}^2, ..., \tilde \nu_{(\tilde p)}^2\}$ be the subset of $\cH_{l,\phi}^{\theta^{\dagger}}[j,j]$.
\end{example}


The `effective curvature' term depends on the diagonal elements of the Hessian $\cH_{l,\phi}^{\theta^{\dagger}}$ of the loss. In essence, the effective curvature only considers components which have high curvature, i.e.,  for each parameter $\theta_j$, if the curvature $\cH_{l,\phi}^{\theta^{\dagger}}[j,j] > \frac{1}{\omega^2_j}$, then we get a non-zero contribution from that term.
For the $L_2$-norm term, the distance from the initialization is scaled by the marginal variance $\omega_j^2$. Thus, we essentially get the  trade-off between the effective curvature and the $L_2$-norm with a fine grained control based on $\omega_j^2$ specific to each term. \eqed

\section{Bounds for Non-Smooth Predictors}
\label{sec:margin2ns}
The challenge in developing bounds for deterministic deep nets has primarily been for the non-smooth predictors.
For concreteness, we focus on ReLU-nets, denoted as $\psi^{\theta}(x)$, noting that argument extends seamlessly to other deep nets such as CNNs and Res-nets. An interesting property of such $\psi^{\theta}(x)$ is that for a given $x$, there is a linear deep net (LDN) $\phi(x)$ with structure $\xi$, i.e., the set of edges that are active given the input, such that $\psi^{\theta}(x) = \phi(x)$. More precisely, let $\xi_x^\theta \in \{0,1\}^p$ denotes a bit vector where a 0 indicates that edge is inactive for input $x$ for a ReLU-net with parameter $\theta$. Then, the LDN has parameters $\theta \odot \xi_x^\theta$, and we have:
$\psi^{\theta}(x) = \phi^{\theta \odot \xi_x^\theta}(x)$. 
The challenge in using such a property is that the realized structure $\xi_x^\theta$ depends on $x$. We develop a PAC-Bayes analysis which maintains distributions over $\theta$, do the analysis in terms of the LDNs $\phi^{\theta \odot \xi_x^\theta}(x)$, and subsequently get margin bounds for deterministic ReLU-nets by de-randomization.

\subsection{Bounds for Stochastic vs.~Deterministic Non-Smooth Predictors}
Our strategy for getting a bound on the deterministic non-smooth predictor is as follows:
we de-randomize the stochastic LDNs $\phi^{\theta \odot \xi_x^\theta}(x), \theta \sim \cN(\theta^{\dagger},\Sigma_{\theta^{\dagger}})$ to get a margin bound on 
$\phi^{\theta^\dagger \odot \xi_x^{\theta^\dagger}}(x)$ which is exactly the deterministic non-smooth predictor $\psi^{\theta^{\dagger}}(x)$, i.e., 
\begin{equation}
\psi^{\theta^{\dagger}}(x) = \phi^{\theta^\dagger \odot \xi_x^{\theta^\dagger}}(x).
\label{eq:ns_key}
\end{equation}
Our analysis will de-randomize $\theta \sim \cQ$ for any fixed $z = (x,y)$ by carefully handling the binary random vector $\xi_x^\theta$ for $\theta \sim \cN(\theta^{\dagger},\Sigma_{\theta^{\dagger}})$, and then extend the analysis to any $z \sim \cD$. With our choice of $\cQ$, we have the following result:
\begin{restatable}{theo}{theonsmarg}
\label{theo:ns_marg}
Let $W$ be any distribution on pairs $(x,y)$ with $x \in \R^d$ and $y \in \{-1,+1\}$. Let $\cQ$ be a multivariate Gaussian distribution with mean $\theta^{\dagger}$ and covariance $\Sigma_{\theta^\dagger} = \diag(\nu^2_j)$ with $\nu^2_j = \min\{\sigma^2, \sigma^2_j\}, \forall j \in [p]$, which is absolutely continuous w.r.t.~$\cP$.
Under Assumption~\ref{asmp:gen}, for any 
$\tilde \gamma > 2$ 
and any $\beta \in \R$, for a depth $k > 1$ ReLU-net, we have
\vspace{-2mm}
\begin{align}
\ell_{\beta}\left(\cQ, W\right) & \leq \ell_{\beta+ \frac{\gamma}{2}
+\varrho_k}(\psi^{\theta^{\dagger}},W) + 6\exp ({-\min(c_2\tilde\gamma^2, c_1\tilde \gamma)}),
\label{eq:ns_aniso_swd} \\
\ell_{\beta}(\psi^{\theta^{\dagger}},W)  &\leq  \ell_{\beta + \frac{\gamma}{2}
+\varrho_k}(\cQ,W)+ 6\exp({-\min(c_2\tilde\gamma^2, c_1\tilde \gamma)}),
\label{eq:ns_aniso_dws}
\end{align}
where $\gamma = 3\sigma^2\zeta\alpha\tilde \gamma$, for $k>2$, $\varrho_k = G\|\theta^\dagger\|_2+\frac{1}{2}\zeta\|\theta^\dagger\|_2^2$, for $k=2$, $\varrho_k = \frac{3}{2}G\|\theta^\dagger\|_2$, $c_2 =  \min\left[\frac{\sigma^2\zeta^2\alpha^2}{2G^2},\frac{\sigma^2\alpha^2}{ 2\|\theta^\dagger\|_2^2}, \frac{\alpha^2}{8\kappa}\right]$, $c_1= 
\frac{\alpha}{4}$, and
$G$, $\alpha$, $\zeta$, $\kappa$ are as in Assumption \ref{asmp:gen}.
\end{restatable} 
\vspace*{-3mm}

It is instructive to compare Theorem~\ref{theo:ns_marg} for non-smooth deep nets with the corresponding result, Theorem~\ref{theo:s_marg}, for smooth predictors. The key difference is the margins: for depth $k > 2$, the margin is
\begin{equation*}
\beta + \sigma^2\alpha\tilde \gamma ~~~\text{(smooth)} \quad \qquad \text{vs.} \quad \qquad    \beta + \frac{3}{2}\sigma^2\alpha\tilde \gamma+G\|\theta^\dagger\|_2+\frac{1}{2}\zeta \|\theta^\dagger\|_2^2 ~~~\text{(non-smooth)}~,
\end{equation*}
so that the price of non-smoothness is the additional term $G\|\theta^\dagger\|_2+\frac{1}{2}\zeta  \|\theta^\dagger\|_2^2$, which only depend on $\|\theta^\dagger\|_2$. A similar comparison can be done for the depth $k=2$ case. In Theorem~\ref{theo:sampcomp}, we discuss sample complexity for getting an error rate $\epsilon$ with probability at least $(1-\delta)$,  we show that price of the extra margin can be handled by utilizing the fact that a depth $k$ ReLU-net $\psi^\dagger$ is positively homogeneous of degree $k$, so that this additional margin does not impede generalization.

We highlight key aspects of the proof (see Appendix~\ref{app:margin2ns} for details). For establishing \eqref{eq:ns_aniso_swd}, for depth $k > 2$, we focus on the set
\begin{equation*}
 \tilde \cZ_{\beta +\frac{3}{2} \sigma^2\zeta\alpha\tilde \gamma+ \varrho_k}^{(>)}(\theta^\dagger) \triangleq \{ (x,y) \in \cX \times \cY | y \phi^{\theta^{\dagger} \odot \xi_x^{\theta^\dagger}}(x) > \beta + \frac{3}{2} \sigma^2\zeta\alpha\tilde \gamma+\varrho_k \}
 \end{equation*}
 where $\varrho_k =  G \| \theta^\dagger\|_2 + \frac{1}{2}\zeta \| \theta^\dagger \|_2^2$.
For any $z = (x,y) \in  \tilde \cZ_{\beta +\frac{3}{2} \sigma^2\zeta\alpha\tilde \gamma+ \varrho_k}^{(>)}(\theta^\dagger)$, we show
\begin{equation}
\P_{\theta \sim \cQ}\left[ y \phi^{\theta \odot \xi_x^\theta}(x) \leq \beta | z \in \cZ^{(>)}_{\beta + \frac{3}{2} \sigma^2\zeta\alpha\tilde \gamma+\varrho_k}(\theta^{\dagger} )\right] \leq 6 \exp(- \min(c_2 \tilde \gamma^2,c_1\tilde\gamma))~.
\label{eq:nsbnd1}
\end{equation}
In other words, if the deterministic predictor $\psi^{\theta^{\dagger}}(x)=\phi^{\theta^{\dagger} \odot \xi_x^{\theta^{\dagger}}}(x)$ has a large margin of at least $(\beta + \varrho_k)$, then the probability that the stochastic predictor $\phi^{\theta \odot \xi_x^\theta}(x), \theta \sim \cN(\theta^{\dagger},\Sigma_{\theta^{\dagger}})$ will have a small margin of at most $\beta$ is exponentially small, i.e., $6 \exp(- \min(c_2 \tilde \gamma^2,c_1\tilde\gamma))$. 
Note that for a given $z$, the comparison here is in between a deterministic LDN with structure $\xi_x^{\theta^\dagger}$ and stochastic LDNs with structure $\xi_x^{\theta}$ where $\theta$ is drawn from the posterior $\cQ$. For a given $z$, since these are all LDNs with different parameters (with some components being zero), the smoothness of LDNs w.r.t.~the parameters can be utilized for the analysis. 

There are technical intricacies in the comparison analysis stemming from the fact that the random structures $\xi_x^{\theta} \in \{0,1\}^p$ have dependencies across components. For any given $z = (x,y) \in  \tilde \cZ_{\beta + \frac{3}{2} \sigma^2\zeta\alpha\tilde \gamma+\varrho_k}^{(>)}(\theta^\dagger)$, the analysis compares the margin of the detereministic LDN with parameter $\theta^\dagger \odot \xi_x^{\theta^\dagger}$ and random LDNs with parameters $\theta \odot \xi_x^\theta$. Noting that $\theta = \theta^\dagger + \delta$, where $\delta \sim \cN(0,\Sigma_{\theta^\dagger})$ by construction, at a high level, the analysis can be viewed as a large deviation bound of the margin of a deterministic LDN with parameter $\theta^\dagger \odot \xi_x^{\theta^\dagger}$ and random LDNs with parameters 
\begin{align}
(\theta^\dagger + \delta) \odot \xi_x^{\theta^\dagger + \delta}
= \theta^\dagger \odot \xi_x^{\theta^\dagger + \delta} +  \delta \odot \xi_x^{\theta^\dagger + \delta}~.
\label{eq:decomp1}
\end{align}
Recall that in \eqref{eq:nsbnd1}, the additional margin in the stochastic LDNs compared to the deterministic LDN is $\varrho_k = G \| \theta^\dagger\|_2 + \frac{1}{2}\zeta \| \theta^\dagger \|_2^2$ for $k > 2$, with a similar but simpler term for $k=2$. Following \eqref{eq:decomp1}, a comparison between 
\begin{align}
    \theta^\dagger \odot \xi_x^{\theta^\dagger}~~~\text{(deterministic)} \qquad \text{and} \qquad \theta^\dagger \odot \xi_x^{\theta^\dagger + \delta}~~~\text{(stochastic, first term in \eqref{eq:decomp1})}
\label{eq:decomp2}
\end{align}
yields the additional margin terms $G \| \theta^\dagger\|_2 + \frac{1}{2}\zeta \| \theta^\dagger \|_2^2$ in $\varrho_k$. In essence, the comparison here effectively gets an exact upper bound of the deviation since the randomness is only in the structure, not the parameters. The $G \| \theta^\dagger\|_2$ and $\frac{1}{2}\zeta \| \theta^\dagger \|_2^2$ terms in the additional margin correspond respectively to the first order and second order terms in the Taylor expansion for the smooth LDN predictors.

The more challenging aspect of the analysis stems from the second term $\delta \odot \xi_x^{\theta^\dagger + \delta}$ in \eqref{eq:decomp1}, which has randomness both in the parameters and structure, and the components of $\xi_x^{\theta^\dagger + \delta}$ are not independent. While the term also occurs in \eqref{eq:decomp2}, that analysis can be simplified by utilizing the fact that the components of $\xi_x^{\theta^\dagger + \delta}$ are in $\{0,1\}$. For doing the margin analysis for the random vector $ \delta \odot \xi_x^{\theta^\dagger + \delta}$, we need to establish large deviation bounds for (a) linear forms of the random vector corresponding to the first order term in the Taylor expansion, and (b) quadratic forms of the random vector corresponding to the second order term in the Taylor expansion. For random vectors of the form $ \delta \odot \xi_x^{\theta^\dagger + \delta}$, if the binary random vector $\xi_x^{\theta^\dagger + \delta} \in \{0,1\}^p$ has arbitrary dependencies, a Hoeffding-type inequality~\citep{boucheron_concentration_2013} on linear forms of $ \delta \odot \xi_x^{\theta^\dagger + \delta}$ need not hold. Further, if the binary random vector $\xi_x^{\theta^\dagger + \delta} \in \{0,1\}^p$ has arbitrary dependencies, a Hanson-Wright type inequality~\citep{hska12,ruve13} on quadratic forms of $ \delta \odot \xi_x^{\theta^\dagger + \delta}$ need not hold. In fact, the Hanson-Wright inequality~\citep{hska12,ruve13} is only known to hold for random vectors with independent components~~\citep{hska12,ruve13}. 

The challenges outlined above get resolved by paying close attention to the nature of dependency among the components of $\xi_x^{\theta^\dagger+\delta}$. 
We order the components of $\xi = \xi_x^{\theta^\dagger+\delta}$ layerwise, so that if there are $k$ layers and $p_h, h=1,\ldots,k$ parameters in each of the layers,
\begin{itemize}
    \item $\xi_{1:p_1}$ correspond to the structure of edges in the first layer, 
    \item $\xi_{(p_1+1):(p_1+p_2)}$ correspond to the structure of edges in the second layer, and so on till 
    \item $\xi_{\left(\sum_{h=1}^{k-1} p_h+1\right):p}$ correspond to the structure of edges in the last layer.
\end{itemize}
The exact ordering of indices for edges in a given layer is unimportant. 
The key observation is that with such an ordering of indices, the status of a specific edge 
$\xi_x^{\theta^\dagger+\delta}[i]$ only depends on parameters and status of edges preceding the specific edge; in fact, the dependency is only on parameters and status of edges till the previous layer. In particular, for $i=1,\ldots,p$, we have\footnote{the statement can be refined by noting that dependency is only on parameters and status of edges till the previous layer, and all edges in the first layer are typically present (i.e., linear model, with no ReLU), but such refinements are not needed for our analysis.}
\begin{equation}
    \xi_x^{\theta^\dagger + \delta}[i] = f_i((\theta^\dagger + \delta)_{1:(i-1)}, \xi_{1:(i-1)},x)~,
    \label{eq:recdn}
\end{equation}
for some suitable function $f_i$. In other words, whether an edge will be active or inactive for a given input $x$ depends on the earlier parameters 
$(\theta^\dagger + \delta)_{1:(i-1)}$ and their active/inactive status $\xi_{1:(i-1)}$. In fact, for a ReLU-net, if $\xi_i$ is in layer $h, h=1,\ldots,k$, then $\xi_i$ only depends on parameters $(\theta^\dagger+\delta)_{i'}$ and status $\xi_{i'}$ for edges (connections) in the earlier layers of the ReLU-net, i.e., layers $h'=1,\ldots,(h-1)$. In particular, such $\xi_i$ do not depend on parameters $\delta_{i'}$ and status $\xi_{i'}$ for edges in the same layer or subsequent layers. 

The above seemingly simple observation is a direct consequence of the structure of ReLU-nets (and also CNNs, ResNets, etc.), and gives enough structure to establish Hoeffding-type and Hanson-Wright-type inequalities. In particular, while the components of $\delta \odot \xi_x^{\theta^\dagger + \delta}[i] = \delta_i \xi_i$ is a product of two random variables $\delta_i$ and $\xi_i$, conditioned on the history $1:(i-1)$, $\xi_i$ is in fact deterministic and $\delta_i$ is zero mean and independent of $\xi_i$. As a result, we can establish a Hoeffding-type inequality for linear forms of $\delta \odot \xi_x^{\theta^\dagger + \delta}$ using a Azuma-Hoeffding type analysis, by viewing the linear form as a Martingale Difference Sequence (MDS). Further, a Hanson-Wright-type inequality for bounding quadratic forms of $\delta \odot \xi_x^{\theta^\dagger + \delta}$ is also established using the specific dependency structure in the components of $\delta \odot \xi_x^{\theta^\dagger + \delta}$. Putting all of these together completes the analysis for the case $z \in \cZ^{(>)}_{\beta + \frac{3}{2}\sigma^2\zeta\alpha \gamma+\varrho_k}(\theta^\dagger)$.

Finally, as in the smooth case, for $z \not \in \cZ^{(>)}_{\beta +\frac{3}{2} \sigma^2\zeta\alpha\tilde \gamma+ \varrho_k}(\theta^\dagger)$, we simply have 
\begin{equation}
\P_{\theta \sim \cQ}\left[ y \phi^{\theta \odot \xi_x^\theta}(x) \leq \beta | z \not \in \tilde \cZ^{(>)}_{\beta + \frac{3}{2} \sigma^2\zeta\alpha\tilde \gamma+\varrho_k}(\theta^{\dagger} )\right] \leq 1~.
\label{eq:nsbnd2}
\end{equation}
While no fancy analysis is needed here, the result is still for a specific $z =(x,y)$. 
Based on the law of total probability, taking expectations w.r.t.~$z \sim W$ and utilizing the two results \eqref{eq:nsbnd1} and \eqref{eq:nsbnd2} above yields~\eqref{eq:ns_aniso_swd}. The analysis for establishing \eqref{eq:ns_aniso_dws} is similar. The analysis for depth $k=2$ is the same with the $\zeta \| \theta^\dagger\|_2^2$ term dropping out. 



\subsection{Main Result: Deterministic Non-Smooth Predictors}

Theorem~\ref{theo:ns_marg} can now be used to get bounds on the deterministic predictor $\phi^{\theta^{\dagger} \odot \xi_x^{\theta^\dagger}}(x) = \psi^{\theta^{\dagger}}(x)$. With 
respectively choosing $\beta=0, W=D$ for \eqref{eq:ns_aniso_dws} and $\beta = \gamma/2 + \varrho_k, W=S$ for \eqref{eq:ns_aniso_swd}, we have
\begin{align*}
\ell_{0}(\psi^{\theta^{\dagger}},D)  & \leq  \ell_{\gamma/2+\varrho_k}(\cQ,D) +  6 \exp(- \min(c_2  \gamma^2,c_1\gamma)), \\
\ell_{\gamma/2+\varrho_k} (\cQ, S) & \leq \ell_{\gamma+2\varrho_k}(\psi^{\theta^{\dagger}},S) + 6 \exp(- \min(c_2 \gamma^2,c_1\gamma)). 
\end{align*} 
Recall that with probability at least $(1-\delta)$, PAC-Bayes gives
\begin{equation*}
KL_B(\ell_{\gamma/2+\varrho_k} (\cQ, S) \| \ell_{\gamma/2+\varrho_k}(\cQ,D)) \leq \frac{KL(\cQ \| \cP) + \log \frac{1}{\delta}}{n},
\end{equation*}
where $KL_B$ denotes the Bernoulli KL-divergence. For any $\eta \in (0,1)$, we again unpack $KL_B$ using the `fast rate' form \citep{cato07}[Theorem 1.2.6],\citep{yasu19} to get
\begin{equation*}
\ell_{\gamma/2+\varrho_k}(\cQ,D)) \leq a_{\eta} \ell_{\gamma/2+\varrho_k} (\cQ, S) + b_{\eta} \frac{KL(\cQ \| \cP) + \log \frac{1}{\delta}}{n},
\end{equation*}
where $a_{\eta} = \frac{\log (1/\eta)}{1-\eta}, b_{\eta} = \frac{1}{1-\eta}$ are constants. The bounds can also be done with the `slow rate' $\frac{1}{\sqrt{n}}$ dependence \citep{mcda03}. Lining up these bounds yields the following result:
\begin{restatable}{theo}{nonsmooth}
\label{theo:non_smooth}
Consider any Gaussian prior distribution $\cP$ 
chosen before training, and let $\theta^{\dagger}$ be the parameters of the model after training. 
Let $\cQ$ be a multivariate Gaussian distribution with mean $\theta^{\dagger}$ and covariance $\Sigma_{\theta^\dagger} = \diag(\nu^2_j)$ with $\nu^2_j = \min\{\sigma^2, \sigma^2_j\}, \forall j \in [p]$, which is absolutely continuous w.r.t.~$\cP$.
Under Assumption \ref{asmp:gen}, with probability at least $1-\delta$, for any $\psi^{\theta^{\dagger}}$, $\eta \in (0,1), \gamma > 6\sigma^2 \zeta\alpha$, we have the following scale-invariant bound:
\begin{align*}
\hspace*{-5mm} 
\ell_{0}(\psi^{\theta^{\dagger}},D) \leq  ~a_{\eta} \ell_{\gamma + 2\varrho_k}(\psi^{\theta^{\dagger}},S)  + \frac{b_{\eta}}{2 n} KL(\cQ \| \cP)+ d_{\eta}\exp(-\min(c_2\gamma^2,c_1\gamma)) +  b_{\eta}\frac{\log (\frac{1}{\delta})}{n},
\end{align*}
where for $k>2$, $\varrho_k = G\|\theta^\dagger\|_2+\frac{1}{2}\zeta\|\theta^\dagger\|_2^2$, for $k=2$, $\varrho_k=\frac{3}{2}G\|\theta^\dagger\|_2$, $a_{\eta} = \frac{\log (1/\eta)}{1-\eta}, b_{\eta} = \frac{1}{1-\eta}$, $d_{\eta} = 6(a_{\eta}+1)$, $c_2 =  \min\left[\frac{1}{18\sigma^2G^2},\frac{1}{18\sigma^2\zeta^2 \|\theta^\dagger\|_2^2},\frac{1}{72\sigma^4\kappa\zeta^2}\right]$, $c_1= 
\frac{1}{12\sigma^2\zeta}$,  $G$, $\alpha$, $\zeta$, $\kappa$ are as in Assumption \ref{asmp:gen}.
\end{restatable}
The result above is essentially the same as Theorem~\ref{theo:smooth} for smooth predictors with an additional margin of $2\varrho_k$. Since empirical margin loss can be small via training, the generalization error boils down to the KL-divergence $KL(\cQ\|\cP)$. We provides the following examples of the choice of $\cP$ and covariance of $\cQ$ and give detail bounds on the KL-divergence $KL(\cQ\|\cP)$.


\begin{example}\label{exa:general_ns}
As the Example \ref{exa:general_sm}, we first consider the most general case, i.e., $\cP$ is an anisotropic Gaussian prior $\cN(\theta_0, \Sigma_0)$ with $\Sigma_0 = \diag(\omega_j^2)$, where $\omega_j >0,~\forall j \in [p]$ chosen before training. 
Note that we have defined 
$\cQ = \mathcal{N}\left(\theta^{\dagger}, \Sigma_{\theta^{\dagger}}\right),$  with  $\Sigma_{\theta^{\dagger}}=\mathrm{diag}(\nu_{i}^{2})$ where $\nu_{i}^{2}=\min \left\{\sigma^{2}, \sigma_{i}^{2}\right\}, \forall j \in[p],$ for some suitable choices for the marginal variances $\sigma_{j}$.  
Then we have the following bound on the KL-divergence. 
\begin{align*}
KL(\cQ\|\cP) = \sum_{j=1}^p \left(\frac{\nu_j^2}{\omega_j^2} + \ln \frac{\omega_j^2}{\nu_j^2} - 1\right) +\sum_{j=1}^p\frac{(\theta_j^\dagger -\theta_{0,j})^2}{\omega_j^2}.
\end{align*}
\end{example}
Note that the first term is the Itakura-Saito distance between the posterior and prior variances. \eqed

\begin{example}\label{exa:ios_ns}
One can also consider a special case of the posterior that utilize the curvature of the parameter. We denote 
$\tilde{\mathcal{H}}_{l, \phi}^{\theta^{\dagger}} \triangleq \frac{1}{n} \sum_{i=1}^{n} \nabla^{2} l\left(y_{i}, \phi^{\theta^{\dagger} \odot \xi_{x_{i}}^{\theta \dagger}}\left(x_{i}\right)\right)$ as the Hessian of the loss.
Note that $\tilde{\cH}_{l,\phi}^{\theta^{\dagger}}$ is well defined and commutable based on the training set. Then we have the following example of the posterior, i.e., $\nu_j^2 = \min\{\omega_j^2,\sigma_j^2\}$, where
$\sigma_j^2 =  \frac{1}{\tilde \cH_{l,\phi}^{\theta^{\dagger}}[j,j]}$ for all $j \in [p]$ and $\omega_j^2 \leq \sigma^2$.
One can consider an isotropic Gaussian prior $\cN(\theta_0, \sigma^2 \mathbb{I})$  where $\sigma >0$ chosen before training. 
Then the $KL$-divergence term $KL(\cQ\|\cP)$ becomes
\begin{align*}
KL(\cQ\|\cP) = 
\underbrace{\sum_{j=1}^{p} \ln \frac{\max \left\{\tilde{\mathcal{H}}_{l, \psi}^{\theta \dagger}[j, j], 1 / \sigma^{2}\right\}}{1 / \sigma^{2}}}_{\text {effective curvature }}+\underbrace{\frac{\left\|\theta^{\dagger}-\theta_{0}\right\|_{2}^{2}}{\sigma^{2}}}_{L_{2} \text { norm }}.
\vspace{-3mm}
\end{align*}
\end{example}

The trade-off between the `effective curvature' term and the `$L_2$ norm' term in the bound comes from our use of anisotropic posterior $\cQ$. Choosing a higher value for $\sigma^2$ diminishes the dependency on the `$L_2$ norm' term and increases the dependency on the `effective curvature' term; and vice versa. \eqed


\begin{example}
One can also consider the special cases of the posterior covariance that considers the curvature of the loss function at $\theta^\dagger$, i.e., $\nu_j^2 = \min\{\omega_j^2,\sigma_j^2\}$, where
$\sigma_j^2 =  \frac{1}{\tilde \cH_{l,\phi}^{\theta^{\dagger}}[j,j]}$ for all $j \in [p]$ and $\omega_j^2 \leq \sigma^2$. Then the $KL$-divergence term becomes 
\vspace{-2mm}
\begin{align}
KL(\cQ\|\cP) &\leq 
\underbrace{\sum_{j=1}^{p} \ln \frac{\max \left\{\mathcal{H}_{l, \psi}^{\theta^{\dagger}}[j, j], 1 / \omega_{j}^{2}\right\}}{1 / \omega_{j}^{2}}}_{\text {effective curvature }}+\underbrace{\sum_{j=1}^{p} \frac{\left(\theta_{j}^{\dagger}-\theta_{0, j}\right)^{2}}{\omega_{j}^{2}}}_{L_{2} \text { norm }} 
= \underbrace{\sum_{\ell =1}^{\tilde p} \ln \frac{\omega^2_{(\ell)}}{\tilde{\nu}^2_{(\ell)}}}_{\textup{effective curvature}}  +  \underbrace{\sum_{j=1}^{p}\frac{(\theta^{\dagger}_j -\theta_{0,j})^2}{\omega_j^2}}_{\textup{$L_2$ norm}},
\end{align}
where $\tilde p = | \{ j : \tilde \cH_{l,\phi}^{\theta^{\dagger}}[j,j] > 1/ \omega_j^2 \}|$, and  $\{ \tilde \nu_{(1)}^2, ..., \tilde \nu_{(\tilde p)}^2\}$ be the subset of $\cH_{l,\phi}^{\theta^{\dagger}}[j,j]$ larger than $1/\omega_j^2$.
\end{example}
While we do not consider quantitatively tightening the bounds in the current work, one can possibly tighten the bound by suitable choices of $\omega_j,\theta_0$, e.g., based on differential privacy \citep{dzda17,dziugaite2018entropy,dzro18}.
\eqed


\subsection{Non-uniform Bounds}
 
A unique aspect of the bound in Theorem~\ref{theo:non_smooth} is that the result holds with probability $(1-\delta)$ for any $\psi^{\theta^\dagger}$, but the actual bound is different for different $\psi^{\theta^\dagger}$, i.e., Theorem~\ref{theo:non_smooth} is a non-uniform bound~\citep{benedek_nonuniform_1994,blumer_learnability_1989,ssbd14}. 
At a high level, recall that uniform bounds take the following form: with probability at least $(1-\delta)$, for all predictors $\psi^{\theta^\dagger}$ in a hypothesis class $\cH$, i.e., $\forall \psi^{\theta^\dagger} \in \cH$, we have
\begin{align}
    \ell_0(\psi^{\theta^\dagger}, D) \leq  \ell_0(\psi^{\theta^\dagger}, S) + \frac{C(\cH)}{\sqrt{n}} + c \sqrt{\frac{\log 1/\delta}{n}}~\qquad \text{(Uniform bound)}~,
    \label{eq:uniform}
\end{align}
where $C(\cH)$ is a suitable measure of the complexity of the hypothesis class~\citep{bame02,ssbd14,mohri_foundations_2018}, e.g., VC dimension, Rademacher complexity, etc. Uniform bounds became the primary approach towards generalization bounds following an influential set of papers around two decades back~\citep{kopa00,bame02}. In contrast, non-uniform bounds take the following form: with probability at least $(1-\delta)$, for any predictor $\psi^{\theta^\dagger}$ in a hypothesis class $\cH$, we have
\begin{align}
    \ell_0(\psi^{\theta^\dagger}, D) \leq  \ell_0(\psi^{\theta^\dagger}, S) + \frac{f(\psi^{\theta^\dagger})}{\sqrt{n}} + c \sqrt{\frac{\log 1/\delta}{n}}~\qquad \text{(Non-uniform bound)}~,
    \label{eq:nonuniform}
\end{align}
where $f(\psi^{\theta^\dagger})$ depends only on that specific predictor $\psi^{\theta^\dagger}$ and not the entire hypothesis class $\cH$. Note that while the exact bound on the right hand side is different from each predictor $\psi^{\theta^\dagger}$, these bounds hold simultaneously for all predictors $\psi^{\theta^\dagger}$ with probability at least $(1-\delta)$. In terms of sample complexity, while uniform bounds lead to the same sample complexity for all predictors in the hypothesis class $\cH$, non-uniform bounds understandably lead to different sample complexity of each predictor depending on $c(\psi^{\theta^\dagger})$. The concept of  non-uniform bounds was introduced by~\cite{benedek_nonuniform_1988} as an extension to Valiant's PAC learning framework~\citep{vali84}; the abstract of their 1988 paper starts off as:
\begin{quote}
    The learning model of Valiant is extended to allow the number of examples to depend on the particular concept to be learned, instead of requiring a uniform bound for all concepts of a concept class.

   This extension, called {\em nonuniform learning}, enables learning many concept classes not learnable by the previous definitions. $\cdots$
\end{quote}
Additional details on non-uniform learnability can be found in~\cite{benedek_nonuniform_1994}; also see \cite{blumer_learnability_1989} and Chapter 7 in \cite{ssbd14}. 

The result in Theorem~\ref{theo:non_smooth} is a non-uniform margin bound, and the margin has a dependency on $\| \theta^\dagger\|_2$, which is a property of the predictor $\psi^{\theta^\dagger}$. More generally, the bound is not quite in the form~\eqref{eq:nonuniform}. Rather than getting a bound in the form~\eqref{eq:nonuniform}, our next result directly characterizes the non-uniform sample complexity, i.e., for a given $\delta,\epsilon$, how many samples $n_0(\delta,\epsilon,\psi^{\theta^\dagger})$ do we need such that with probability at least $(1-\delta)$, for any $\psi^{\theta^\dagger}$,
$\ell_{0}(\psi^{\theta^{\dagger}},D) \leq  \ell_0(\psi^{\theta^\dagger},S) + \epsilon$.
Unlike the case on uniform bounds, note that the non-uniform sample complexity $n_0(\delta,\epsilon,\psi^{\theta^\dagger})$ is specific to each predictor $\psi^{\theta^\dagger}$~\citep{ssbd14}. Further, the proof illustrates that the dependence on $\psi^{\theta^\dagger}$ includes aspects such as effective curvature and not just $\|\theta^\dagger\|_2$. Finally, the dependence of the margin on $\|\theta^{\dagger}\|_2$ does not become an issue since that margin is unnormalized and can be controlled by scaling the predictor and utilizing the fact that a depth $k$ ReLU-net is $k$-homogeneous.  
\begin{restatable}{theo}{sampcomp}
Under Assumption \ref{asmp:gen}, in the setting of Theorem~\ref{theo:non_smooth}, with probability at least $(1-\delta)$, for any $\psi^{\theta^{\dagger}}$, there exists $n_0 = n_0(\psi^{\theta^\dagger},\sigma,\epsilon,\delta)$ such that for any $n \geq n_0$, we have 
\begin{equation}
\ell_{0}(\psi^{\theta^{\dagger}},D) \leq  \ell_0(\psi^{\theta^\dagger},S) + \epsilon~. 
\label{eq:sc1}
\end{equation}
$n_0$ has a polynomial dependency on $\sigma$ and $\log(1/\delta)$. Further, with margin function $g(\gamma) \triangleq  \ell_{\gamma}(\psi^{\theta^\dagger},S) - \ell_0(\theta^\dagger,S)$, if $g(\gamma) \geq c_1 \gamma^{c_2/q}$ for some finite integer $q$, then $n_0$ has a polynomial dependency on $1/\epsilon$.
\label{theo:sampcomp}
\end{restatable}
In essence, the result says that if the margin function $g(\gamma)$ does not increase too slowly (e.g., logarithmic), the predictor dependent sample complexity $n_0(\psi^{\theta^\dagger},\sigma,\epsilon,\delta)$ has polynomial dependency on $\sigma, \log (1/\delta), 1/\epsilon$. In traditional uniform convergence theory, a deep neural network with ReLU activation has been proven to be within VC class \citep{goldberg1995bounding,bartlett_nearly-tight_2019}, which is in turn proven to have polynomial sample complexity (or so-called PAC-learnable) with ERM algorithm from Sauer's lemma \citep{sauer1972density}. In essence, the relations of traditional PAC learnability and nonuniform learnability can be characterized by: a hypothesis class of binary classifiers is nonuniformly learnable if and only if it is a countable union of PAC learnable hypothesis classes, from \cite{ssbd14} Theorem 7.2. On the aspect of effective bound, our sample complexity can be easily extended to multi-class case and, in the modern high-dimensional settings, the sample complexity bounds based on VC dimension have an undesired dependency of square root of dimension of parameters.
More interestingly, unlike uniform sample complexities, the non-uniform sample complexity in Theorem~\ref{theo:sampcomp} depends on the predictor $\psi^{\theta^\dagger}$ and the nature of dependency is not just on $\| \theta^\dagger\|_2$, but also on more subtle aspects such as effective curvature. 

For technical reasons, the proof of the sample complexity result in Theorem~\ref{theo:sampcomp} utilizes Assumption~\ref{asmp:gen} under parameter scaling, i.e., $\theta^\dagger$ replaced by $\lambda \theta^\dagger$ for $\lambda \geq 1$. We review Assumption~\ref{asmp:gen} in its original form and how it is used under such parameter scaling. In its original form, Assumption~\ref{asmp:gen} can be ensured algorithmically, i.e., for a suitable choice of $G$, the gradients of the (realized) smooth predictors can be truncated to ensure $\| \phi^\theta(x)\|_2^2 \leq G^2$; and for suitable choices of $H$, the Hessian of the (realized) smooth predictors can be truncated to ensure $-H \preceq \nabla_\theta^2 \phi^\theta \preceq H$. Note that the Hessian is not considered in learning deep nets based on SGD, but is considered in stochastic quasi-Newton or other second order algorithms.
Under parameter scaling, i.e., $\theta^\dagger$ replaced by $\lambda \theta^\dagger$ for $\lambda \geq 1$, we can assume that $G, H$ stay the same, and the resulting proof of Theorem~\ref{theo:sampcomp} will be relatively simple.  However, for $\lambda \geq 1$, many more $\phi^\theta$ will have truncated gradients and Hessians compared to the $\lambda=1$ case. Qualitatively, such additional truncation is undesirable. The proof of Theorem~\ref{theo:sampcomp} proceeds by not needing such additional truncations, but allowing the constants $G, H$ as well as derived constants to grow with $\lambda$.
Note that 
\begin{align}
    \nabla \phi^{\lambda \theta} = \lambda^{k-1} \nabla \phi^{\theta} \qquad \text{and} \qquad  \nabla^2 \phi^{\lambda \theta} = \lambda^{k-2} \nabla \phi^{\theta}~,
\end{align}
where the gradients are w.r.t.~the scaled parameters $\lambda \theta$. As a result, in Assumption~\ref{asmp:gen}, it suffices to have the constant related to the first order gradient to be $G_{\lambda} = \lambda^{k-1} G$, and the constant related to the Hessian, viz.~$\zeta$, to be scaled by $\lambda^{k-2}$. The proof of Theorem~\ref{theo:sampcomp} works with such scaled constants as needed so as to avoid additional truncations due to parameter scaling.

\section{Experimental Results}
\label{sec:expt}
We discuss a variety of experiments based on training ReLU-nets on MNIST and CIFAR-10. For the experiment, we consider the isotropic Gaussian prior with $\theta_0 = 0$ and $\omega_j = \sigma$ for all $j \in [p]$. Thus the KL-divergence term becomes $\frac{b_{\eta}}{2 n}\left(\sum_{\ell=1}^{\tilde{p}} \ln \frac{1/\tilde{\nu}_{\ell}^{2}}{1 / \sigma^{2}}+\frac{\left\|\theta^{\dagger}-\theta_{0}\right\|^{2}}{\sigma^{2}}\right)$. In practice, it has been empirically observed that 1) the generalization error (test error rate) decreases as the training sample size increases \citep{nako19b}, and 2) the generalization error increases when the randomness in the label increases \citep{zhbh17}. To examine whether our bound efficiently capture the above observations, we divide our
experiments into two sets to address questions: (i) How does our bound behave as we increase the number of random labels? (ii) How does our bound behave with an increase in the number of training samples? 
We evaluate
these questions empirically in Sections \ref{exp:main_rand} and \ref{exp:main_sample}, respectively. We also evaluate the effect of parameter $\sigma$ and $\gamma$ in our bound and compare the spectral norm with $L_2$ norm in Section \ref{exp:add_result}.
To thoroughly evaluate our generalization bound, we consider  variants of setting such as depth $\{2, 4, 6, 8\}$, width $\{128, 256, 384, 512\}$, micro-batch (size 16) \citep{nako19b} and mini-batch (size 128) training. We present representative results here, with details of the setup and additional results are in Appendix \ref{app:expt}. 

\begin{figure*}[t] 
\centering
 \subfigure[Test Error Rate.]{
 \includegraphics[width =  0.28\textwidth]{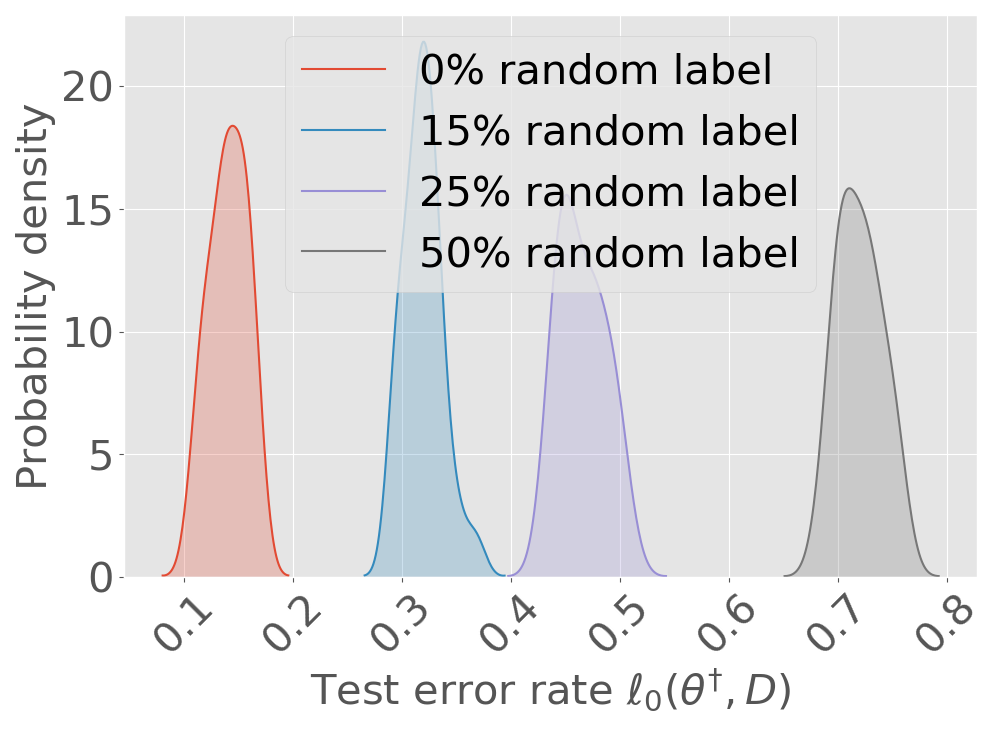}
 } 
 \subfigure[Diagonal Elements of Hessian.]{
 \includegraphics[width = 0.28 \textwidth]{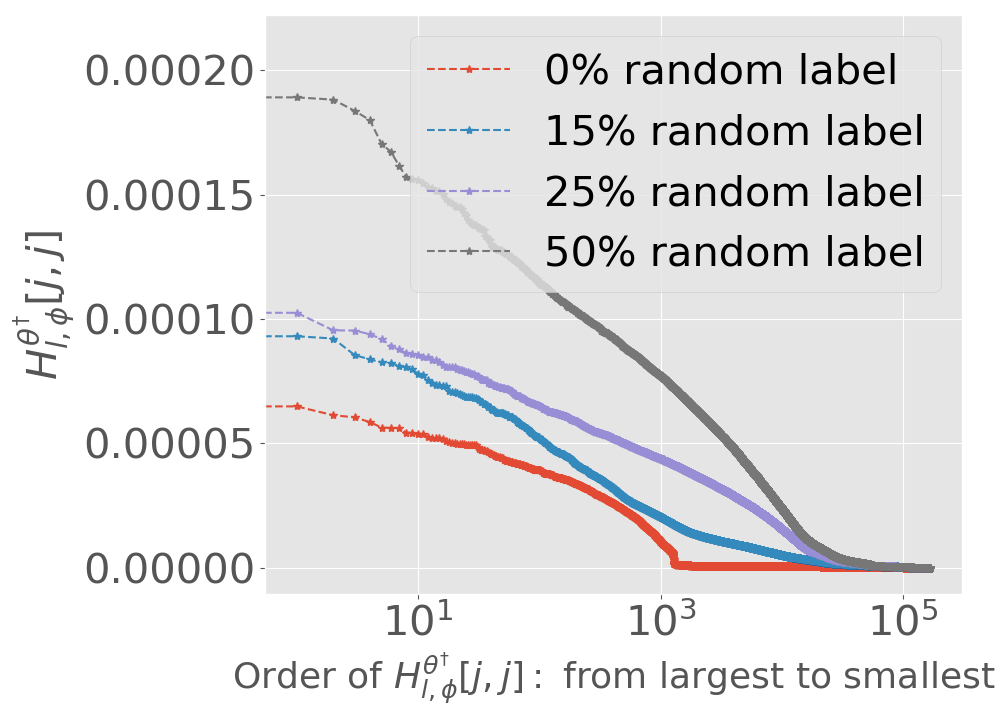}
 } 
 \subfigure[Effective Curvature.]{
 \includegraphics[width = 0.28 \textwidth]{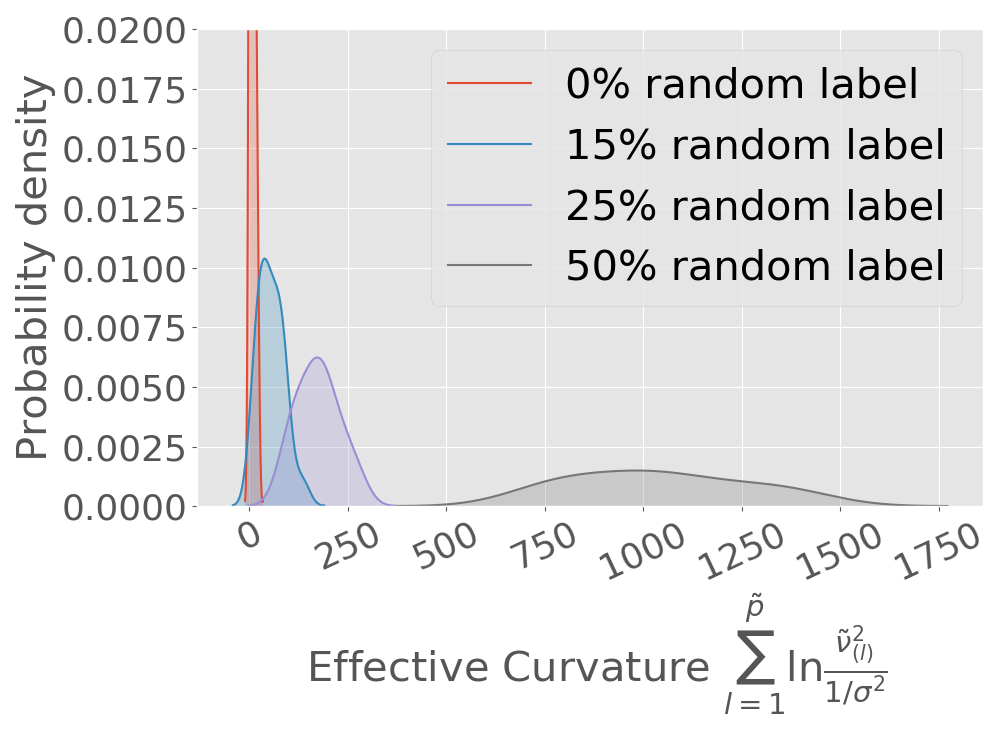}
 } 
 \subfigure[$L_2$ norm.]{
 \includegraphics[width = 0.28 \textwidth]{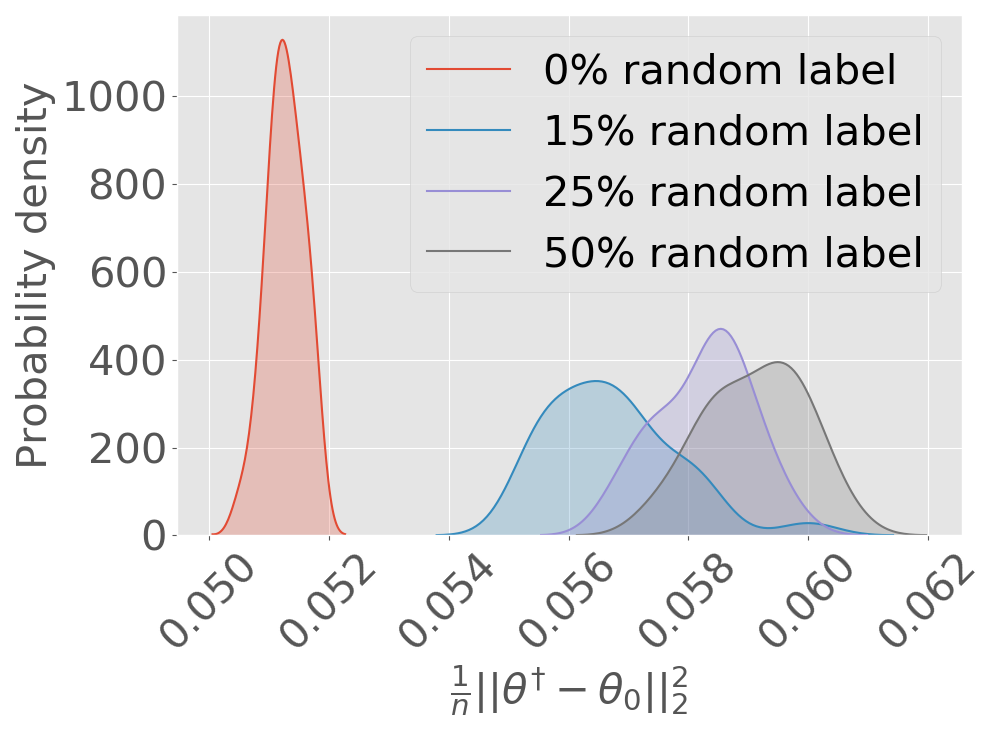}
 } 
  \subfigure[Margin Loss.]{
 \includegraphics[width = 0.28 \textwidth]{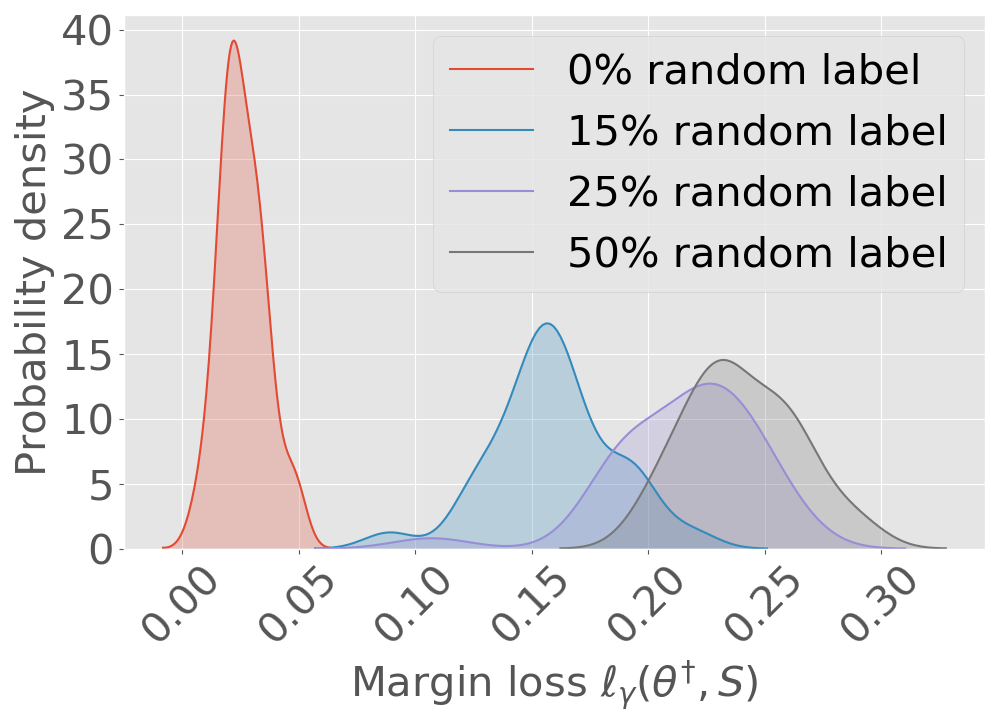}
 } 
   \subfigure[Generalization Bound.]{
  \includegraphics[width = 0.28 \textwidth]{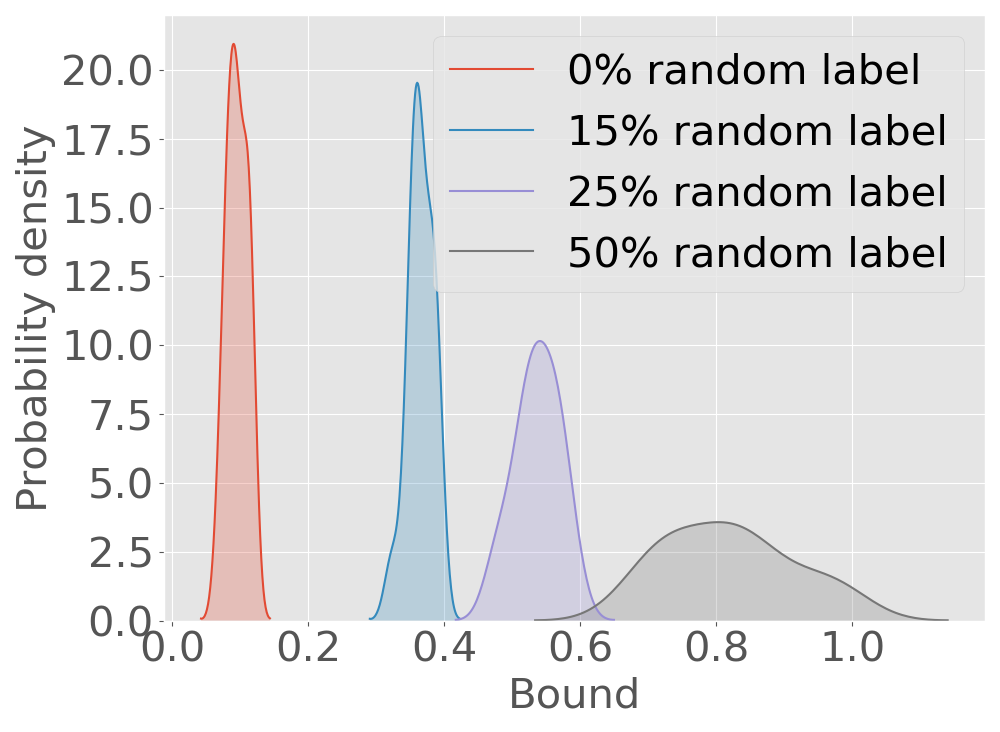}
  } 
\vspace{-2mm}
\caption[]{Results for ReLU-nets with depth = 4, width =128, total 167,818 parameters, trained on 1000 samples from MNIST with batch size = 128 and a increase in number of random labels (30 runs each) from 0\% to 50\%. (a) test set error rate; (b) diagonal elements (mean) of $\tilde H_{l,\phi}^{\theta^\dagger}$; (c) effective curvature with $\sigma^2 = 20000$; d) $L_2$ norm of $\theta^{\dagger}$; (e) margin loss with margin $\gamma = 9$; (f) generalization bound with $\eta = 0.1$.
Increasing percentage of random labels, the generalization bound as well as the components (effective curvature, $L_2$ norm, margin loss) increase, and the bound in (f) stays valid for the test error rate in (a).
}
\label{fig:main_mnist_rand}
\vspace*{-4mm}
\end{figure*}

\begin{figure*}[!t] 
\centering
 \subfigure[Test Error Rate]{
 \includegraphics[  width =  0.28\textwidth, height = 0.22\textwidth]{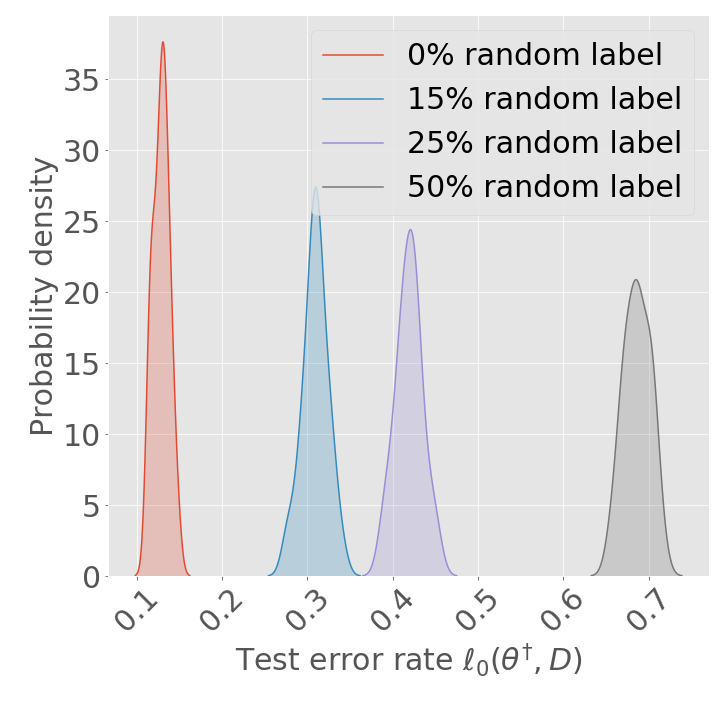}
 }  \vspace{-2mm}
 \subfigure[Diagonal Elements of  Hessian.]{
 \includegraphics[  width =  0.28\textwidth, height = 0.22\textwidth]{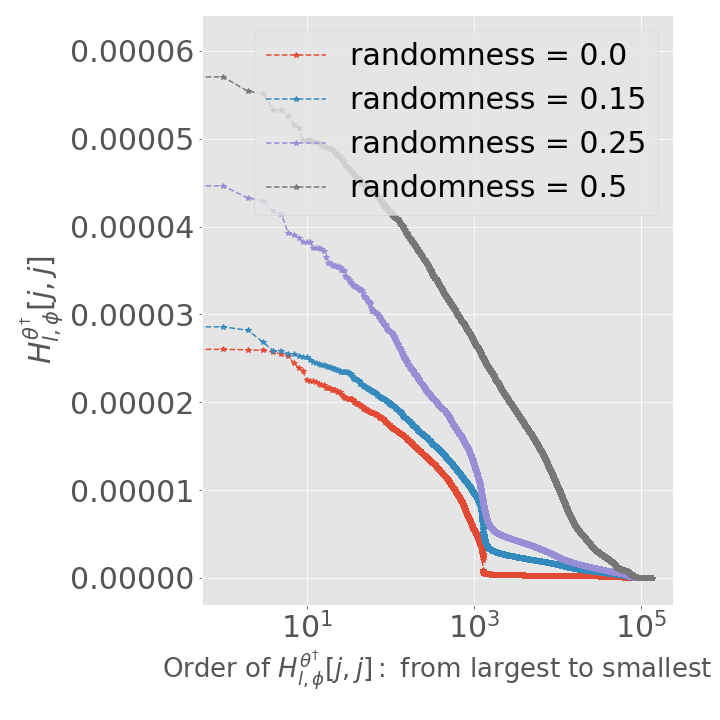}
 }  \vspace{-2mm}
 \subfigure[Effective Curvature.]{
 \includegraphics[  width =  0.28\textwidth, height = 0.22\textwidth]{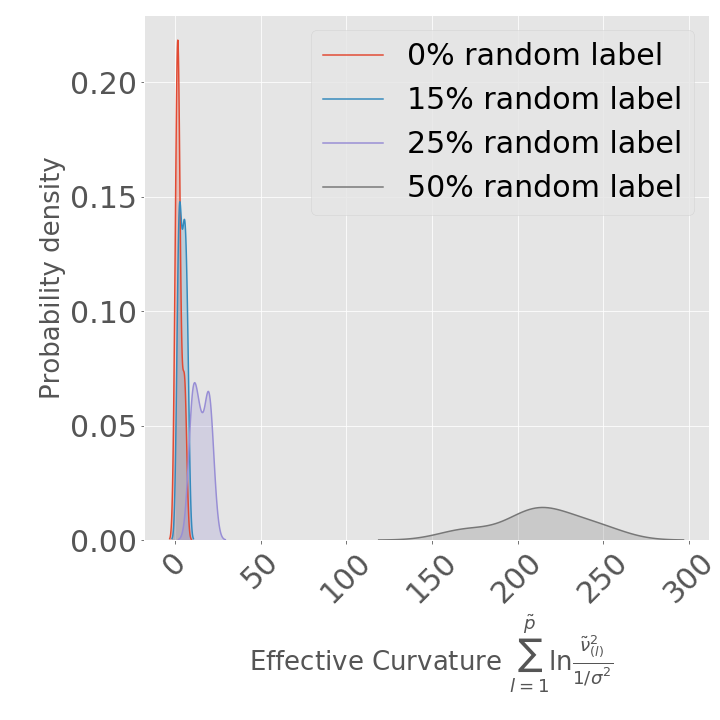}
 }  \vspace{-2mm}
 \subfigure[$L_2$ norm / no. sample.]{
 \includegraphics[  width =  0.28\textwidth, height = 0.22\textwidth]{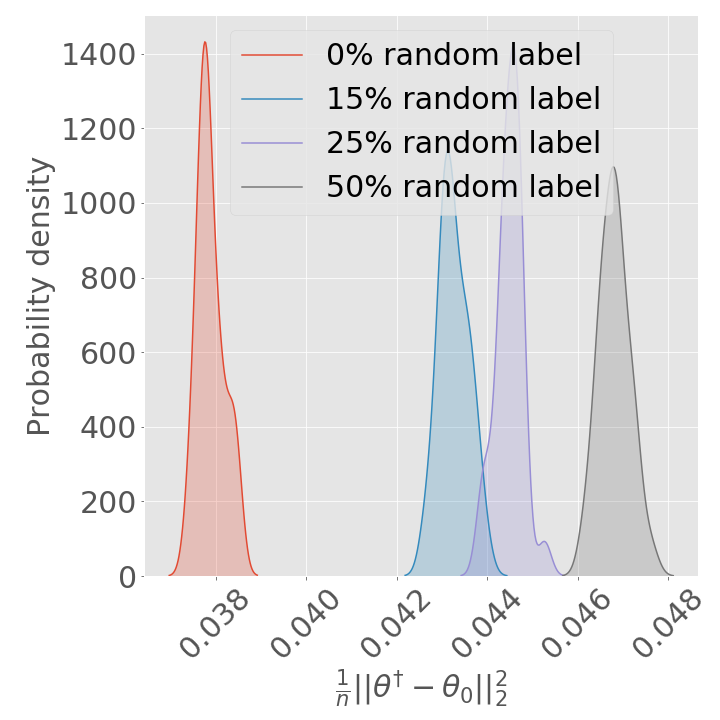}
 } 
  \subfigure[Margin Loss.]{
 \includegraphics[  width =  0.28\textwidth, height = 0.22\textwidth]{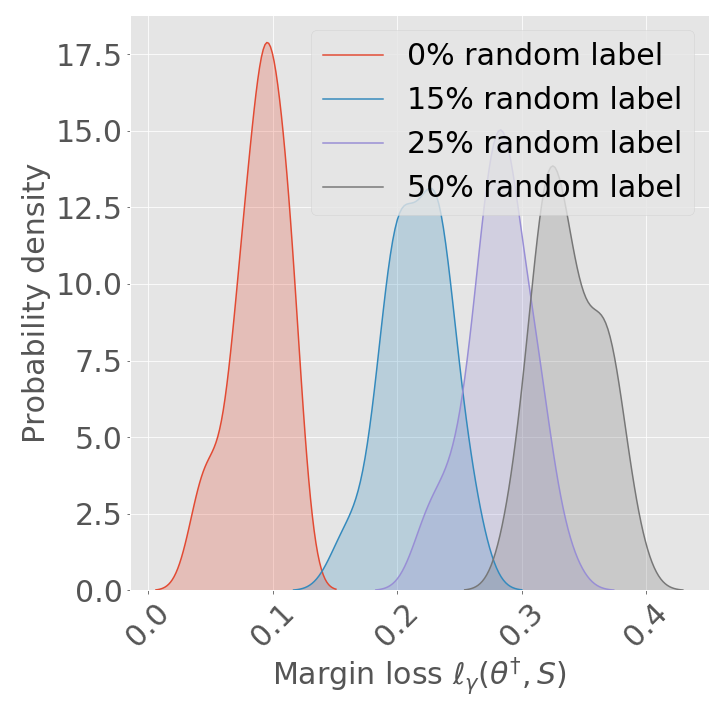}
 } 
   \subfigure[Generalization Bound.]{
  \includegraphics[  width =  0.28\textwidth, height = 0.22\textwidth]{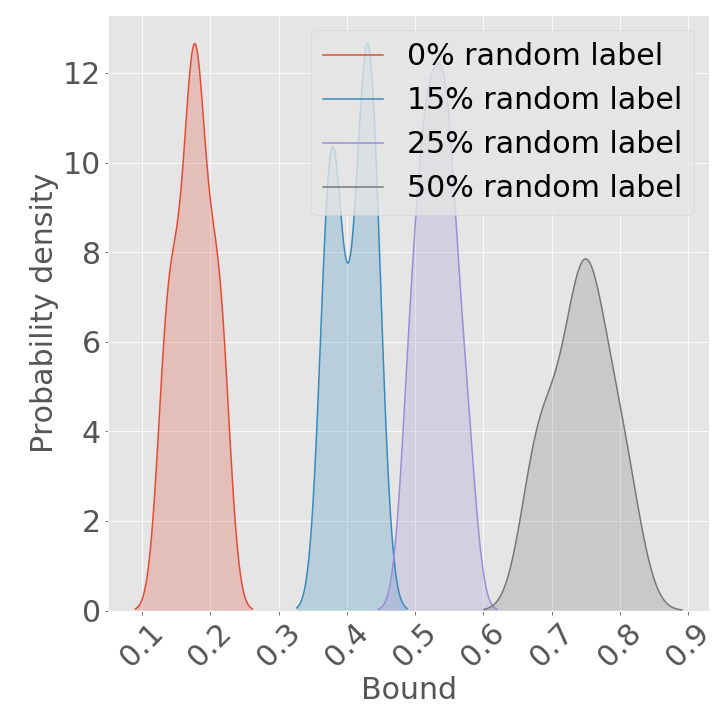}
  } 
\vspace{-2mm}
\caption[]{Results for ReLU-nets with depth = 2, width =128, total 134,794 parameters, trained on 1000 samples from MNIST with batch size = 128 and a increase in number of random labels (30 runs each) from 0\% to 50\%. (a-f) refer to Figure \ref{fig:main_mnist_rand}.
Increasing percentage of random labels, the generalization bound as well as the components (effective curvature, $L_2$ norm, margin loss) increase, and the bound in (f) stays valid for the test error rate in (a).
}
\vspace*{-4mm}
\label{fig:main_mnist_d2_l128}
\end{figure*}

\begin{figure*}[t] 
\centering
 \subfigure[Test Error Rate]{
 \includegraphics[  width =  0.28\textwidth, height = 0.22\textwidth]{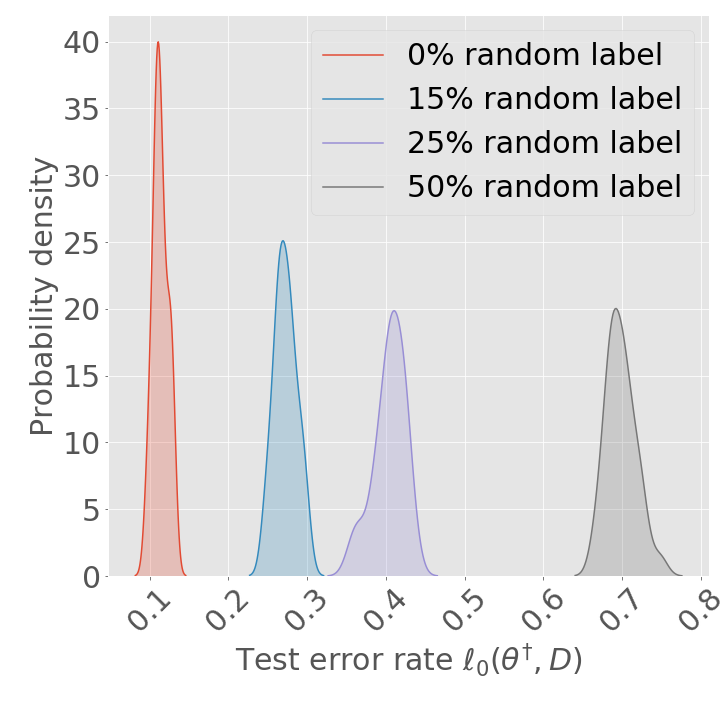}
 } \vspace{-2mm}
 \subfigure[Diagonal Elements of  Hessian.]{
 \includegraphics[  width =  0.28\textwidth, height = 0.22\textwidth]{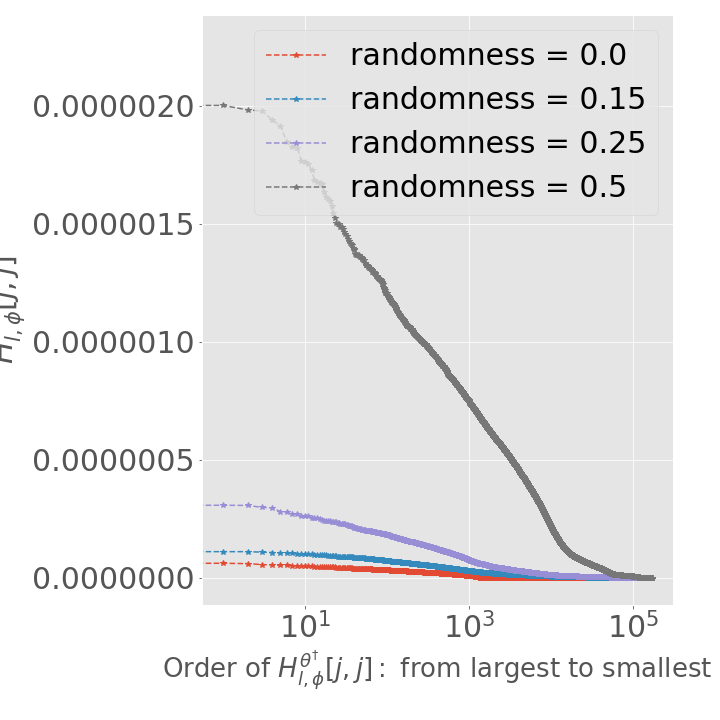}
 }  \vspace{-2mm}
 \subfigure[Effective Curvature.]{
 \includegraphics[  width =  0.28\textwidth, height = 0.22\textwidth]{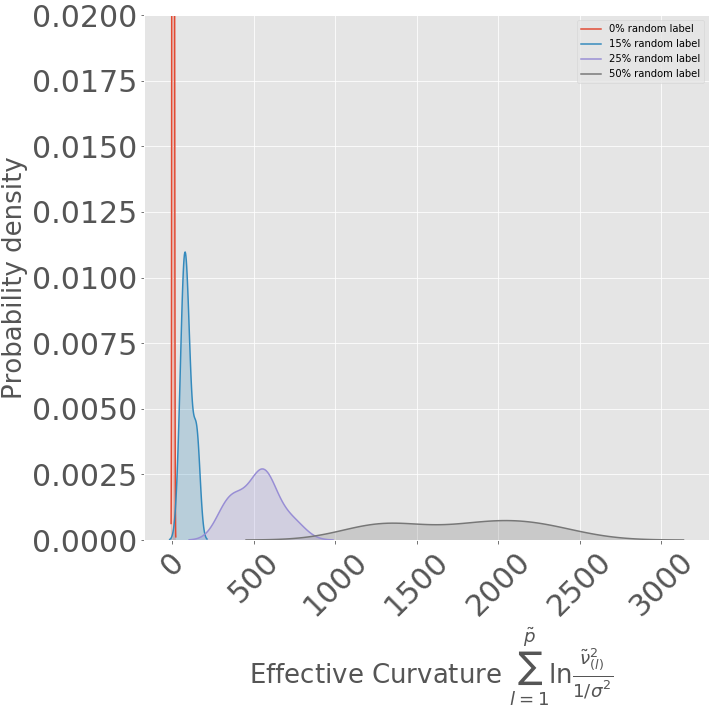}
 }  \vspace{-2mm}
 \subfigure[$L_2$ norm / no. sample.]{
 \includegraphics[  width =  0.28\textwidth, height = 0.22\textwidth]{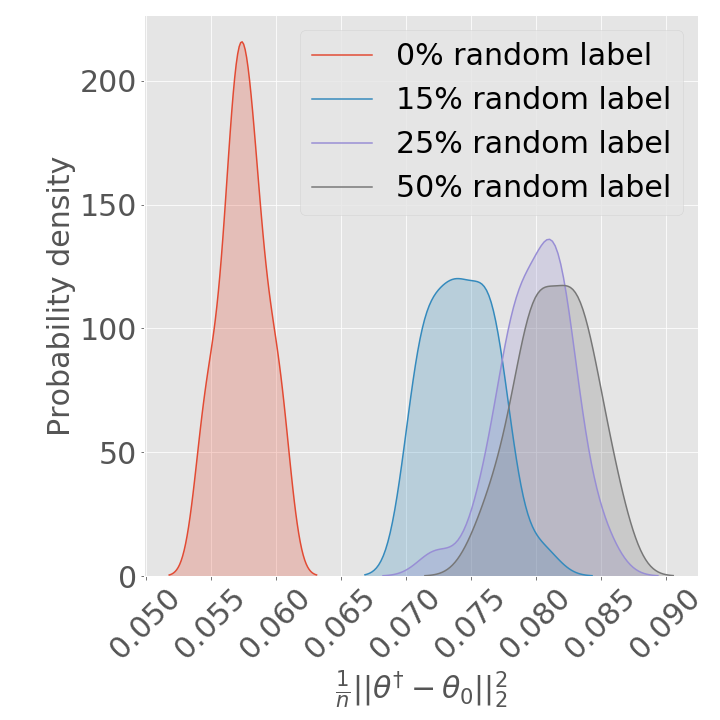}
 } 
  \subfigure[Margin Loss.]{
 \includegraphics[  width =  0.28\textwidth, height = 0.22\textwidth]{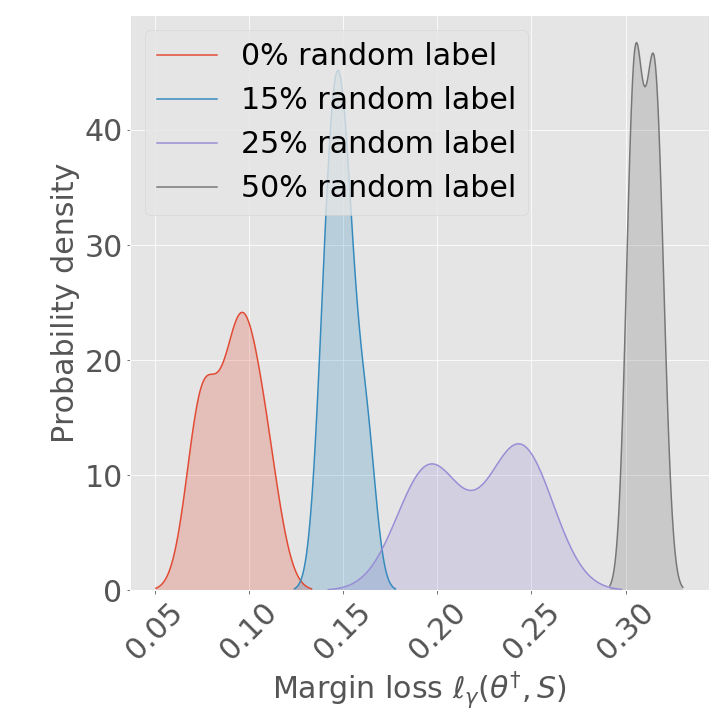}
 } 
   \subfigure[Generalization Bound.]{
  \includegraphics[  width =  0.28\textwidth, height = 0.22\textwidth]{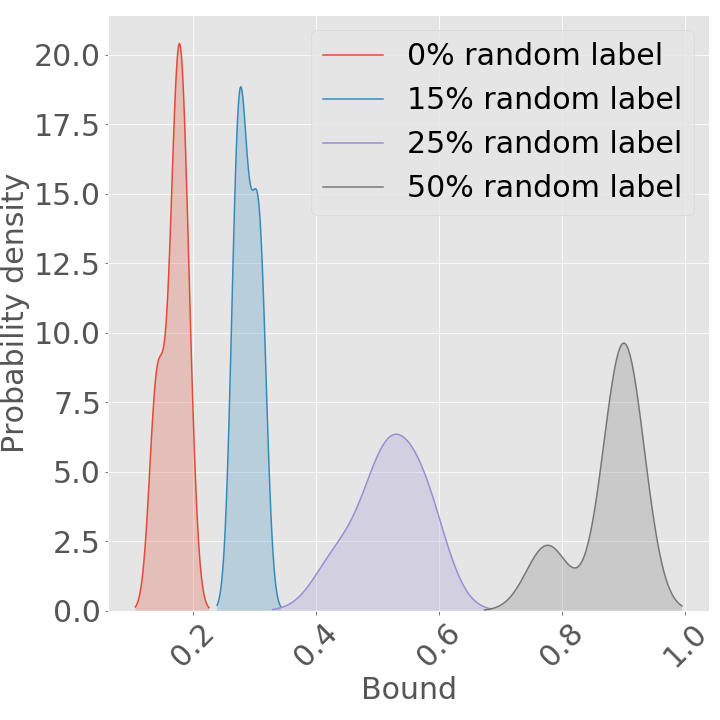}
  } 
\vspace{-2mm}
\caption[]{Results for ReLU-nets with depth = 4, width =128, total 167,818 parameters, trained on 1000 samples from MNIST with batch size = 16 and a increase in number of random labels (30 runs each) from 0\% to 50\%. (a-f) refer to Figure \ref{fig:main_mnist_rand}.
Increasing percentage of random labels, the generalization bound as well as the components 
stay valid for the test error rate in (a).
}
\label{fig:main_mnist_d4_l128_bs16}
\end{figure*}

\begin{figure*}[!t] 
\vspace*{-2mm}
\centering
 \subfigure[Test Error Rate]{
 \includegraphics[width =  0.28\textwidth, height = 0.22\textwidth]{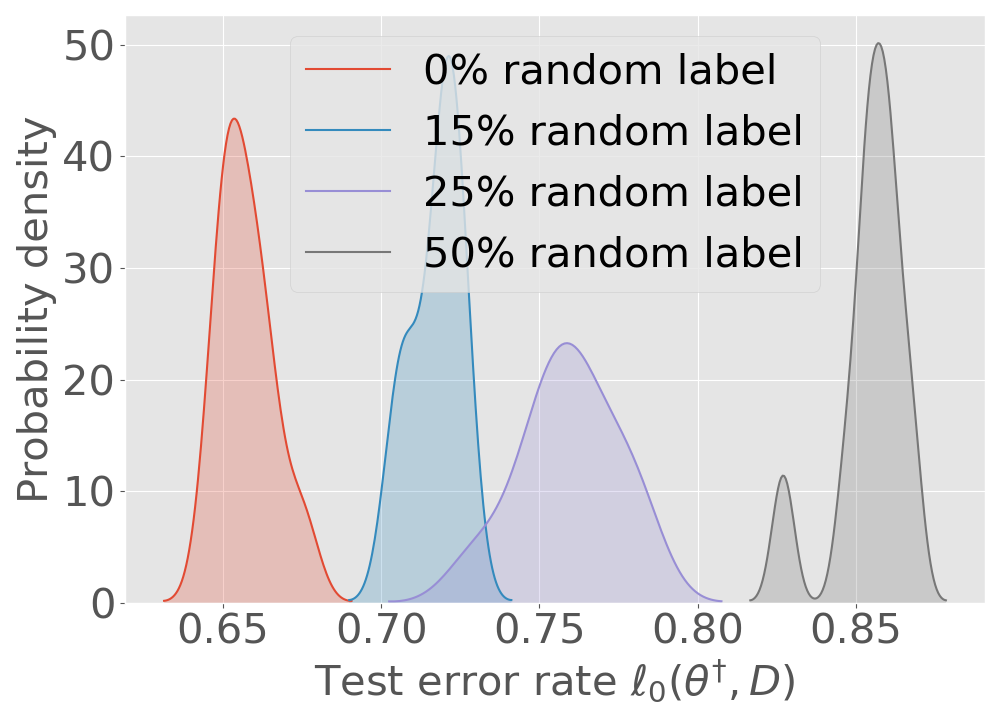}
 } 
 \subfigure[Diagonal Elements of  Hessian.]{
 \includegraphics[  width =  0.28\textwidth, height = 0.22\textwidth]{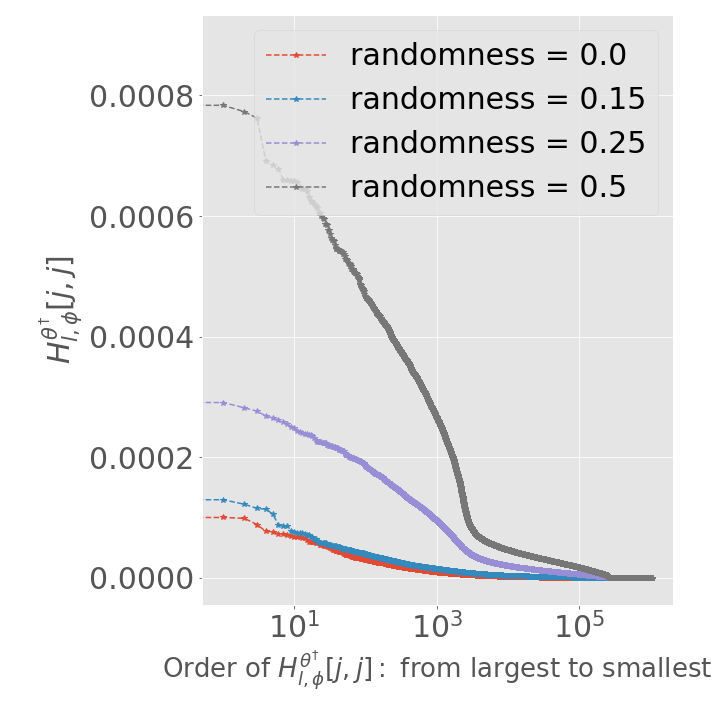}
 } 
 \subfigure[Effective Curvature.]{
 \includegraphics[  width =  0.28\textwidth, height = 0.22\textwidth]{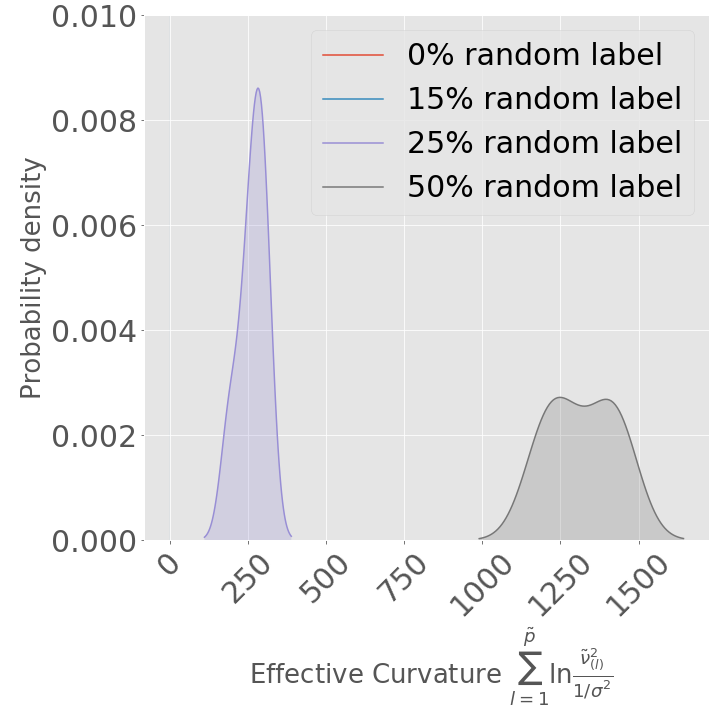}
 } 
 \subfigure[$L_2$ norm / no. sample.]{
 \includegraphics[  width =  0.28\textwidth, height = 0.22\textwidth]{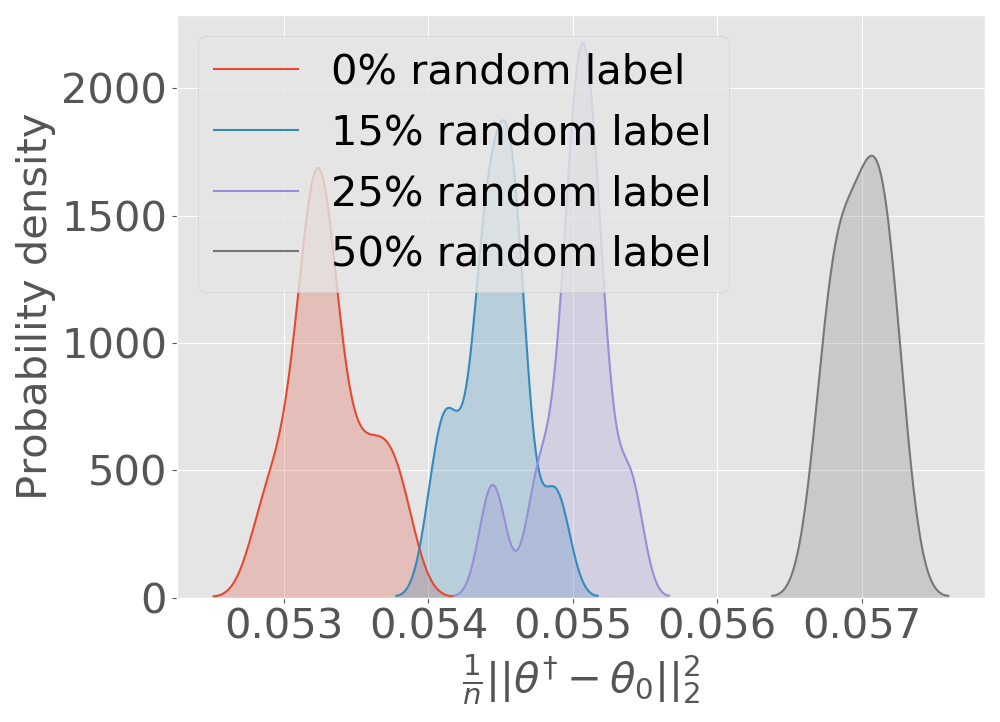}
 } 
  \subfigure[Margin Loss.]{
 \includegraphics[  width =  0.28\textwidth, height = 0.22\textwidth]{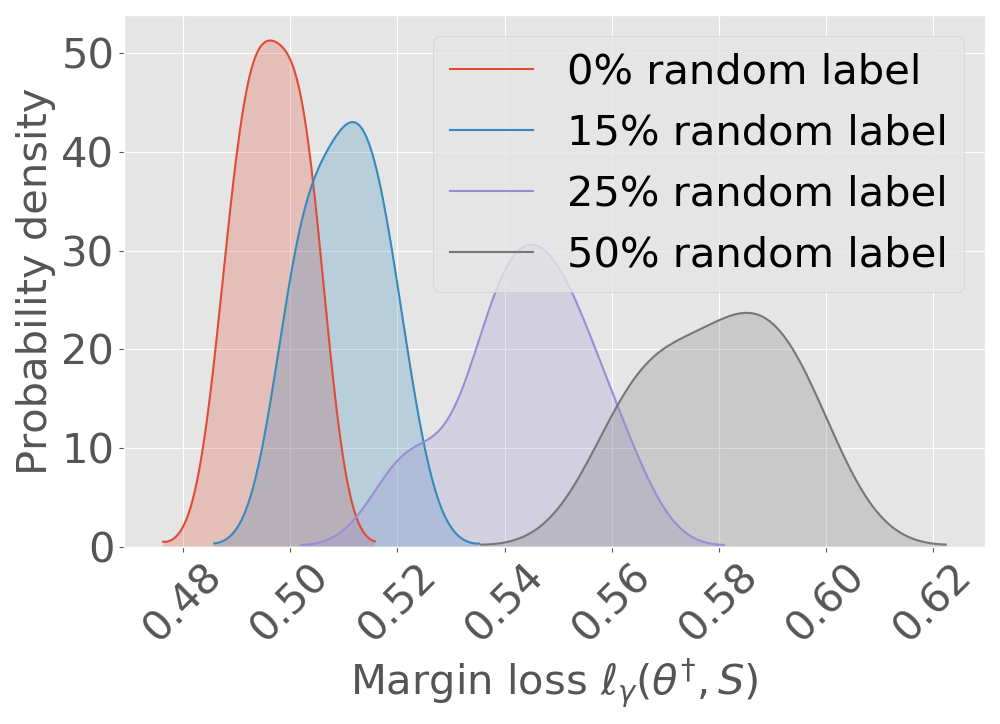}
 } 
   \subfigure[Generalization Bound.]{
  \includegraphics[  width =  0.28\textwidth, height = 0.22\textwidth]{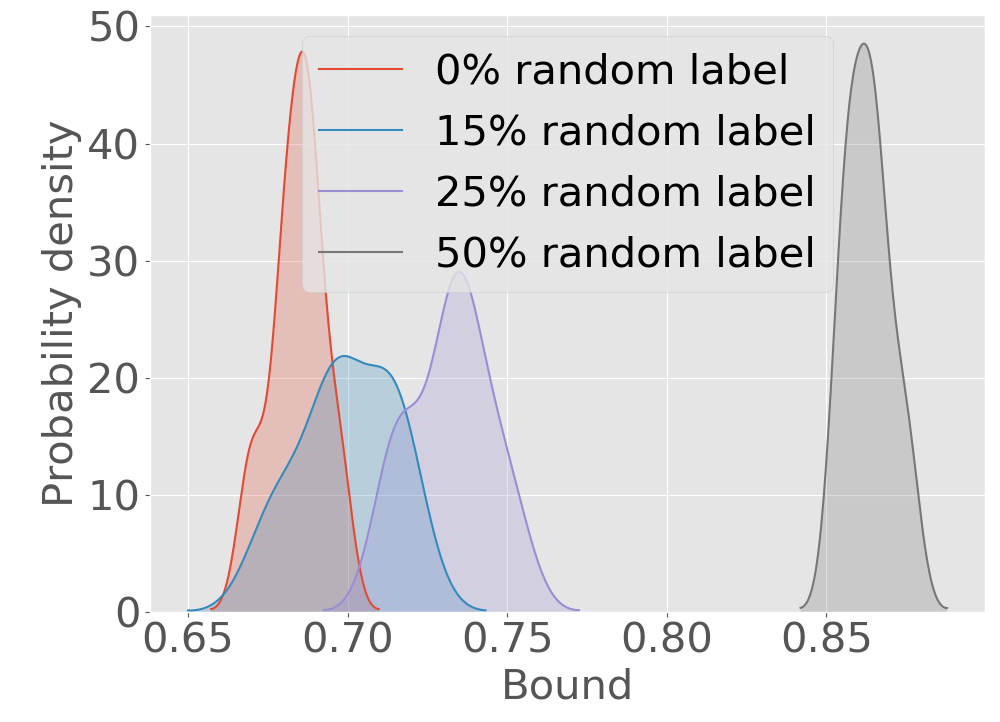}
  } 
\vspace{-4mm}
\caption[]{Results for ReLU-nets with depth = 4, width =256, total 1,052,426 parameters, trained on 1000 samples from CIFAR-10 with batch size = 128 and a increase in number of random labels (20 runs each) from 0\% to 50\%. (a-f) refer to Figure \ref{fig:main_mnist_rand}. In (c), the effective curvature for 0 \% and 15\% random label is zero.
Increasing percentage of random labels, the generalization bound as well as the components 
stay valid for the test error rate in (a).
}
\label{fig:main_cifar_d4_l256}
\end{figure*}

\begin{figure*}[t] 
\vspace{-3mm}
\centering
 \subfigure[Test Error Rate]{
 \includegraphics[  width =  0.28\textwidth, height = 0.22\textwidth]{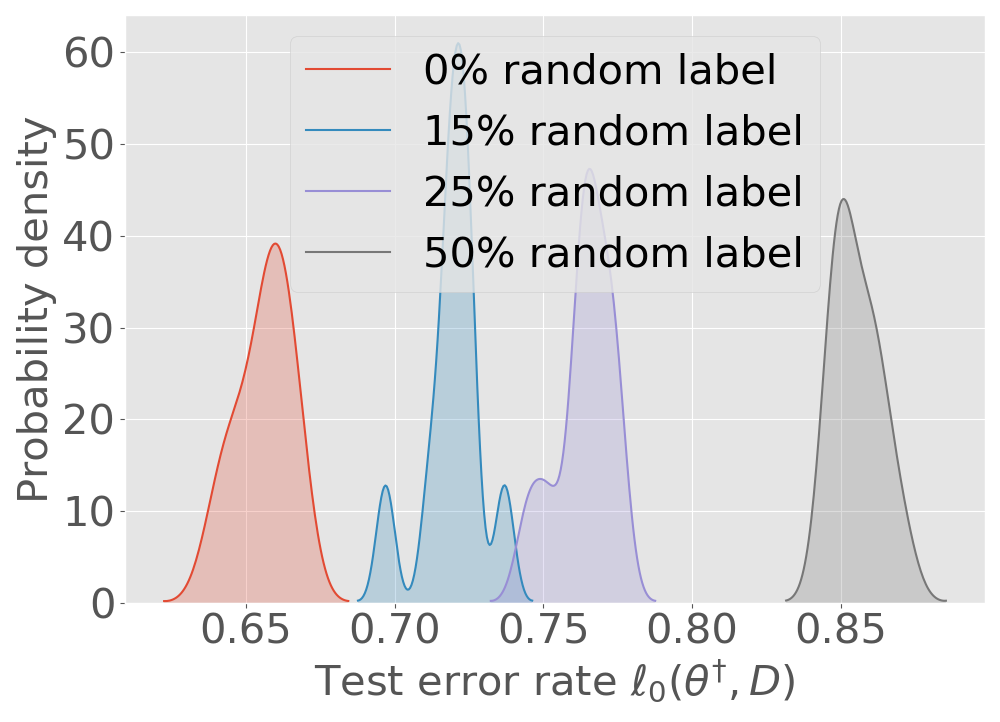}
 }  \vspace{-2mm}
 \subfigure[Diagonal Elements of  Hessian.]{
 \includegraphics[  width =  0.28\textwidth, height = 0.22\textwidth]{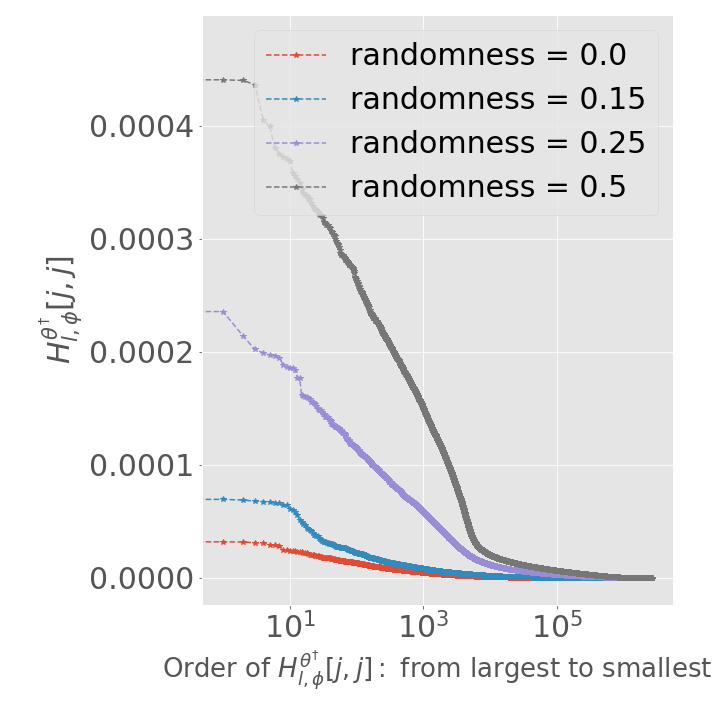}
 } 
 \subfigure[Effective Curvature.]{
 \includegraphics[  width =  0.28\textwidth, height = 0.22\textwidth]{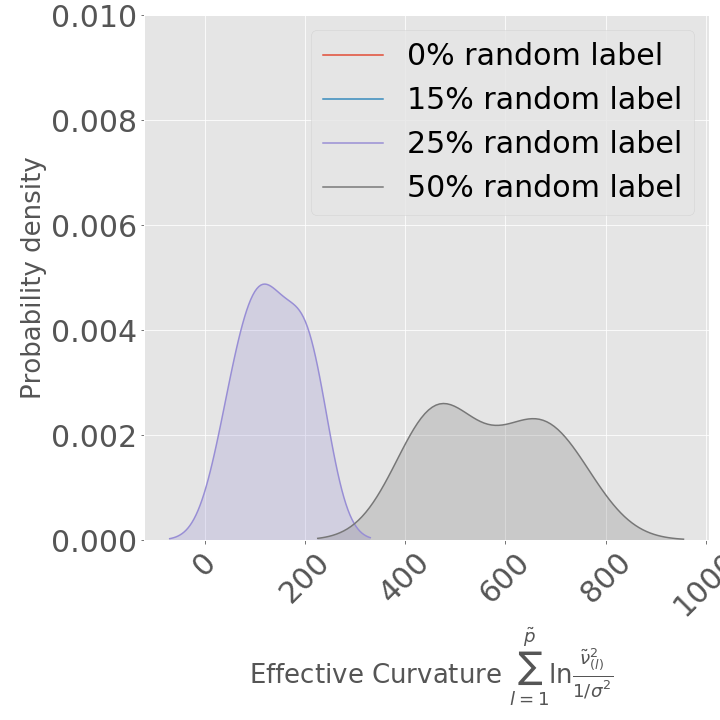}
 } 
 \subfigure[$L_2$ norm / no. sample.]{
 \includegraphics[  width =  0.28\textwidth, height = 0.22\textwidth]{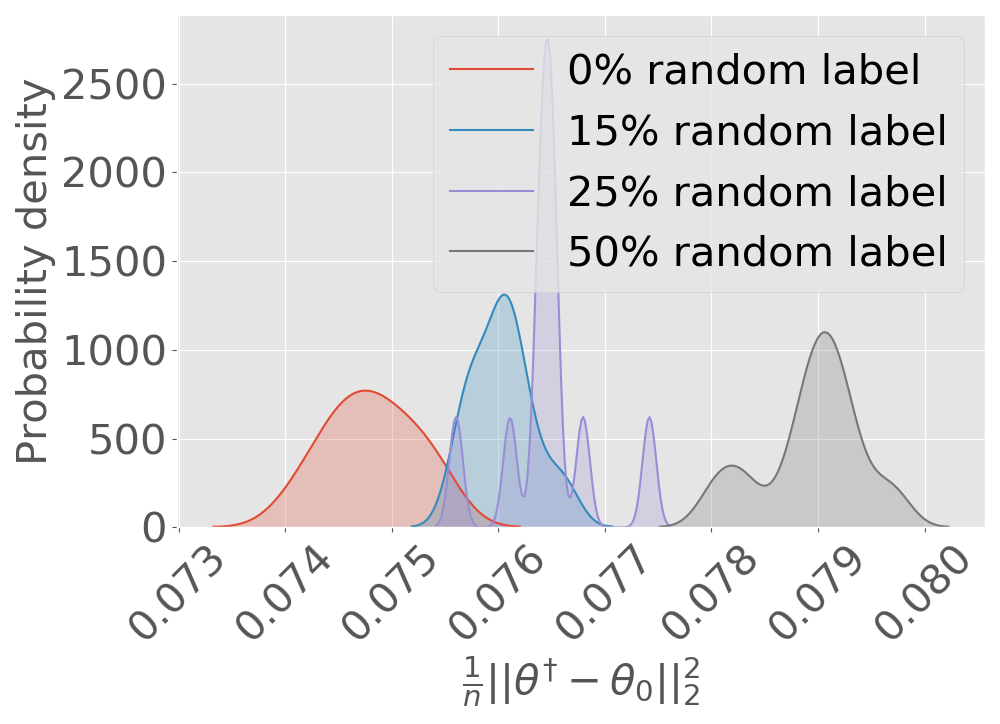}
 } 
  \subfigure[Margin Loss.]{
 \includegraphics[  width =  0.28\textwidth, height = 0.22\textwidth]{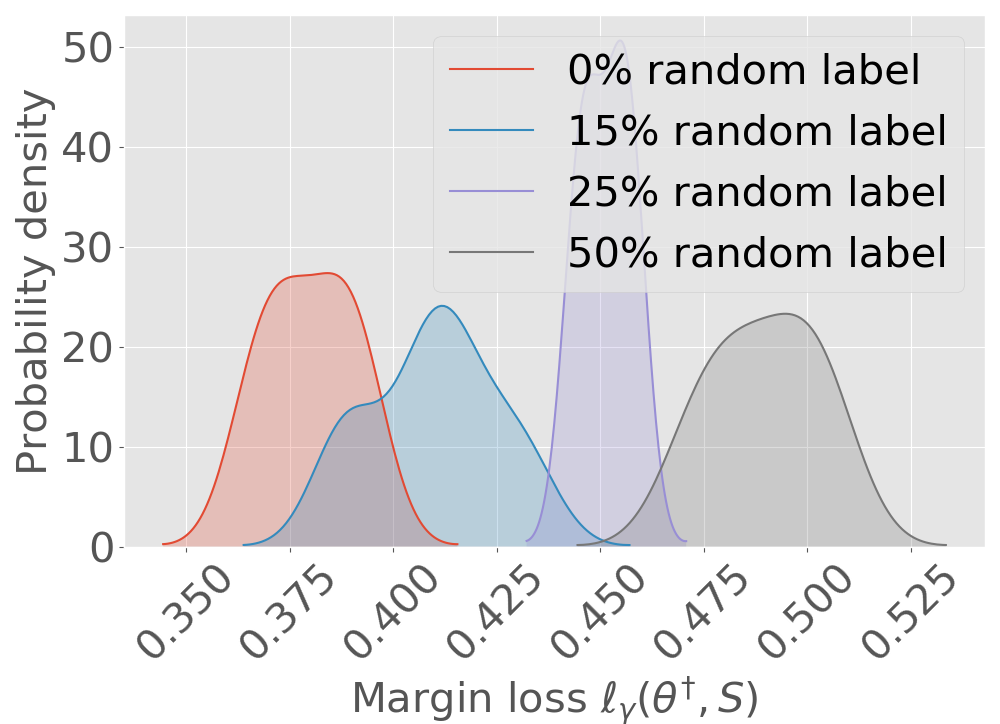}
 } 
   \subfigure[Generalization Bound.]{
  \includegraphics[  width =  0.28\textwidth, height = 0.22\textwidth]{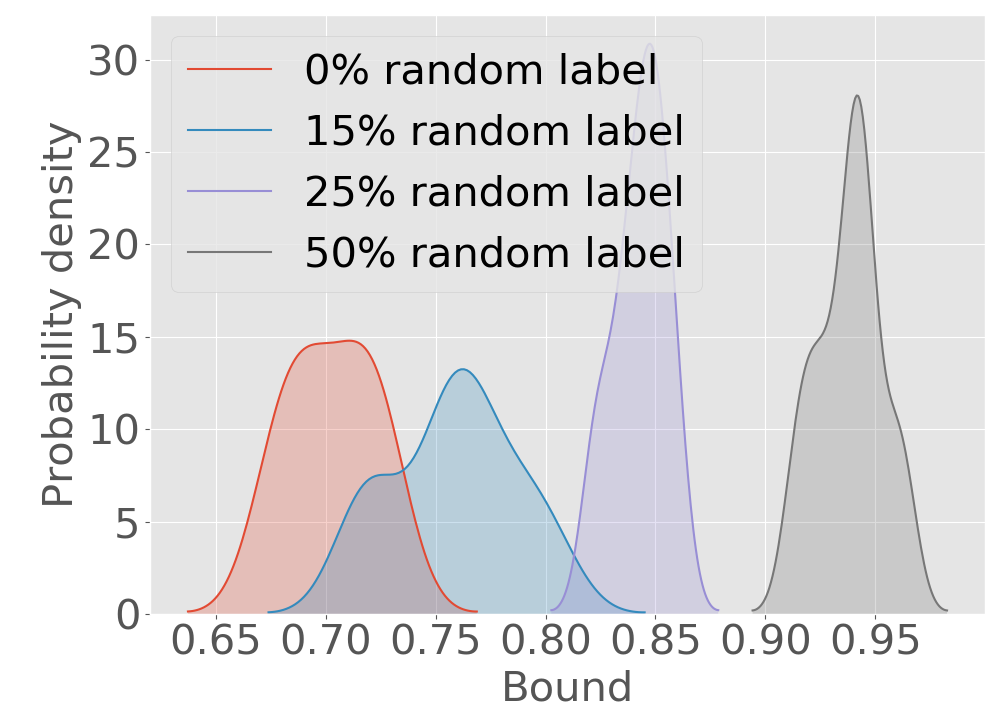}
  } 
\vspace{-4mm}
\caption[]{Results for ReLU-nets with depth = 4, width =512, total 2,629,130 parameters, trained on 1000 samples from CIFAR-10 with batch size = 128 and a increase in number of random labels (20 runs each) from 0\% to 50\%. (a-f) refer to Figure \ref{fig:main_mnist_rand}. In (c), the effective curvature for 0 \% and 15\% random label is zero.
Increasing percentage of random labels, the generalization bound as well as the components (effective curvature, $L_2$ norm, margin loss) increase, and the bound in (f) stays valid for the test error rate in (a).
}
\label{fig:main_cifar_d4_l512}
\end{figure*}

\begin{figure*}[!t] 
\centering
 \subfigure[Test Error Rate]{
 \includegraphics[  width =  0.28\textwidth, height = 0.22\textwidth]{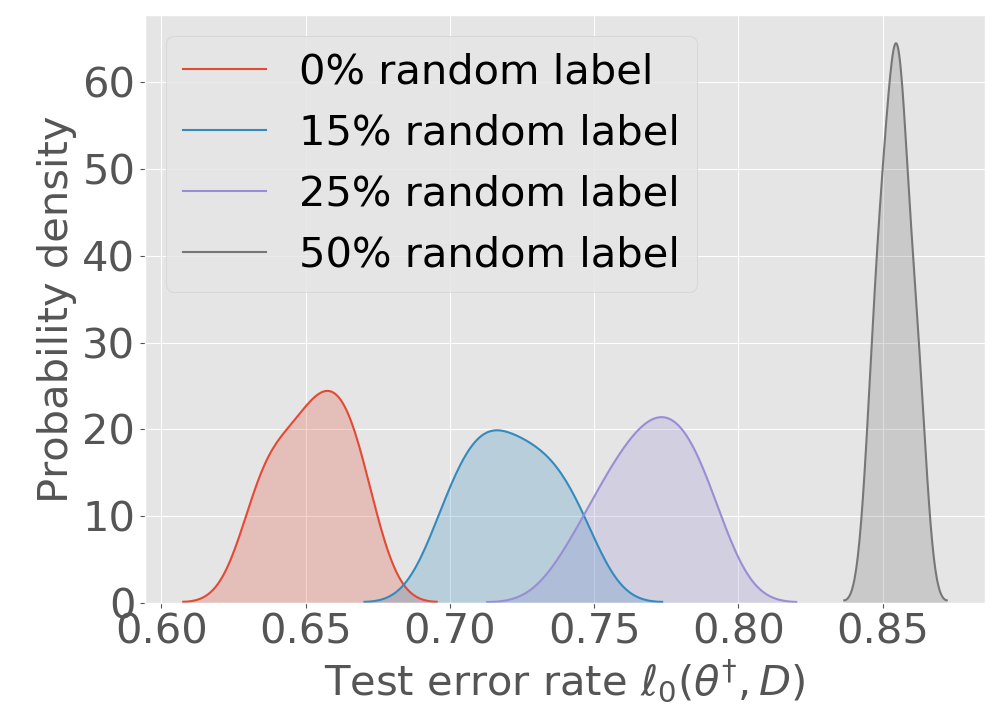}
 } 
 \subfigure[Diagonal Elements of  Hessian.]{
 \includegraphics[  width =  0.28\textwidth, height = 0.22\textwidth]{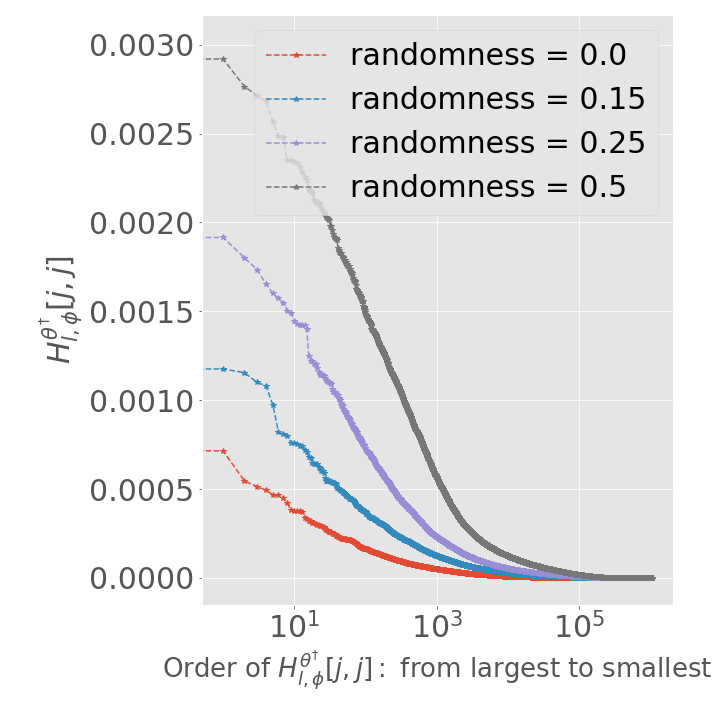}
 } 
 \subfigure[Effective Curvature.]{
 \includegraphics[  width =  0.28\textwidth, height = 0.22\textwidth]{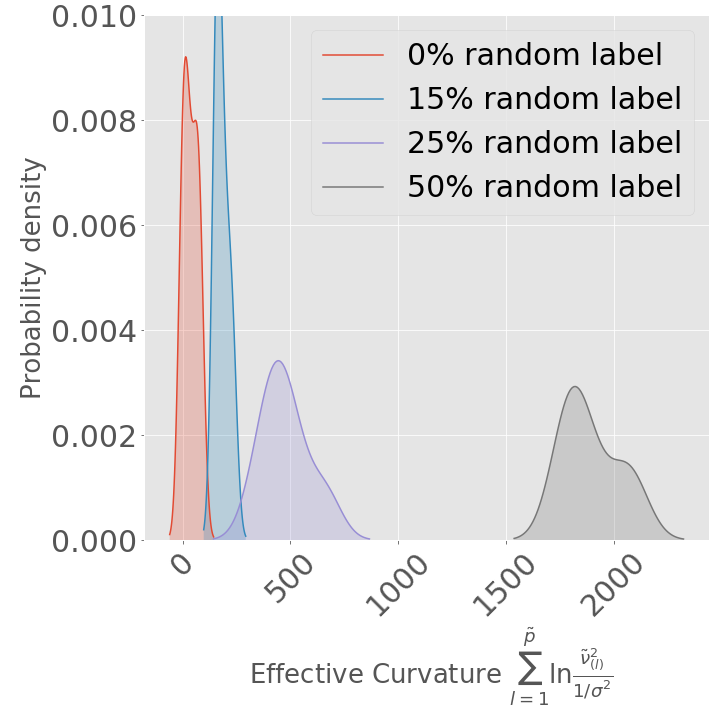}
 } 
 \subfigure[$L_2$ norm / no. sample.]{
 \includegraphics[  width =  0.28\textwidth, height = 0.22\textwidth]{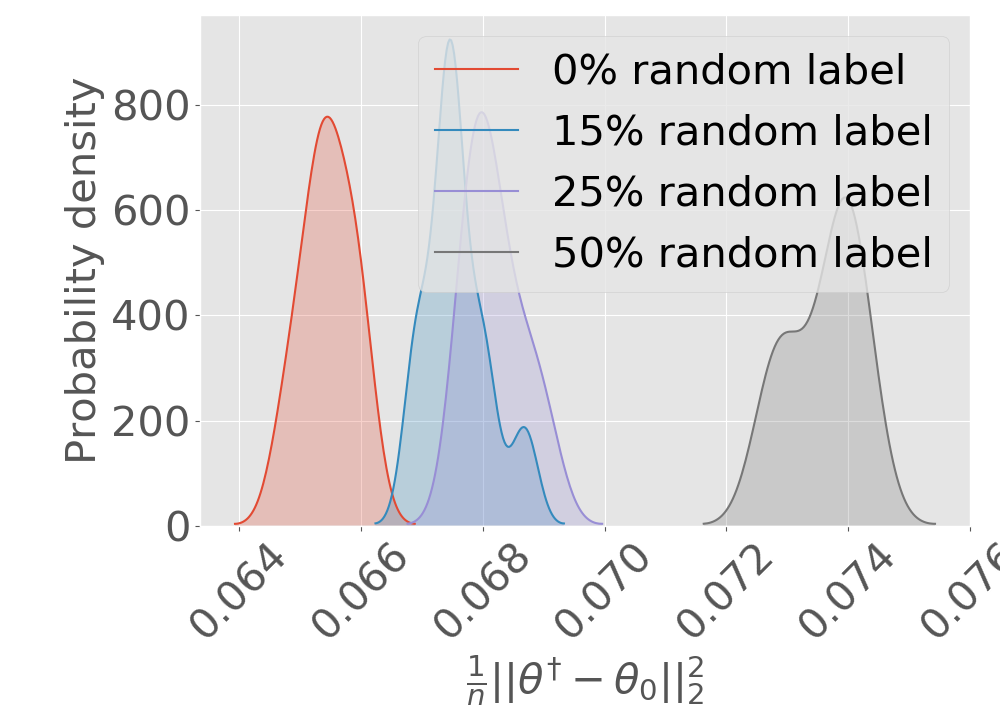}
 } 
  \subfigure[Margin Loss.]{
 \includegraphics[  width =  0.28\textwidth, height = 0.22\textwidth]{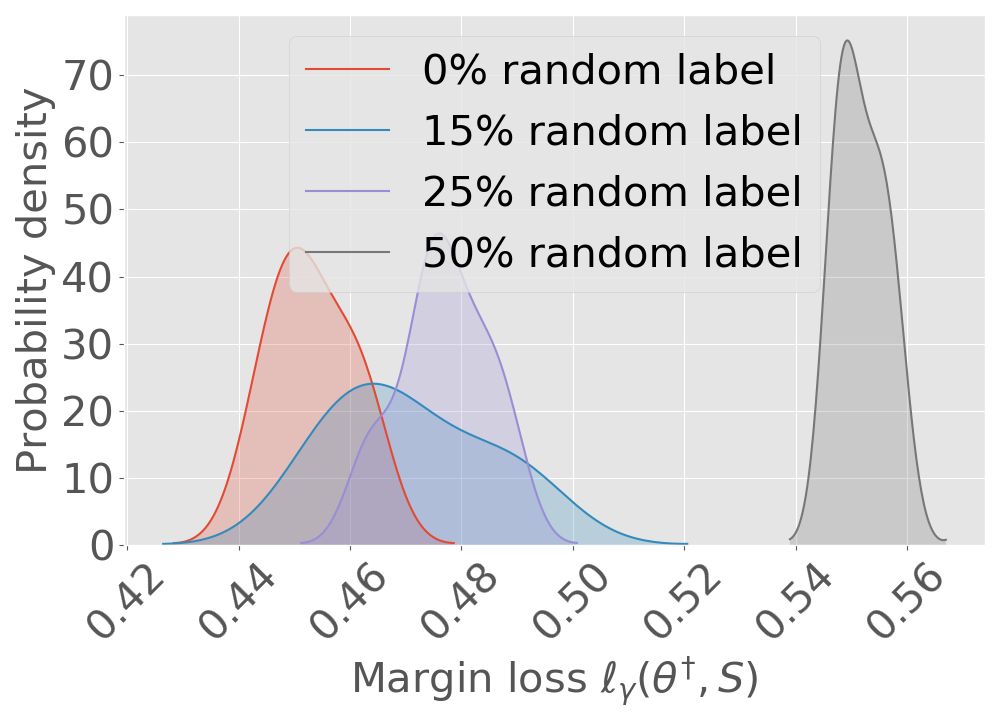}
 } 
   \subfigure[Generalization Bound.]{
  \includegraphics[  width =  0.28\textwidth, height = 0.22\textwidth]{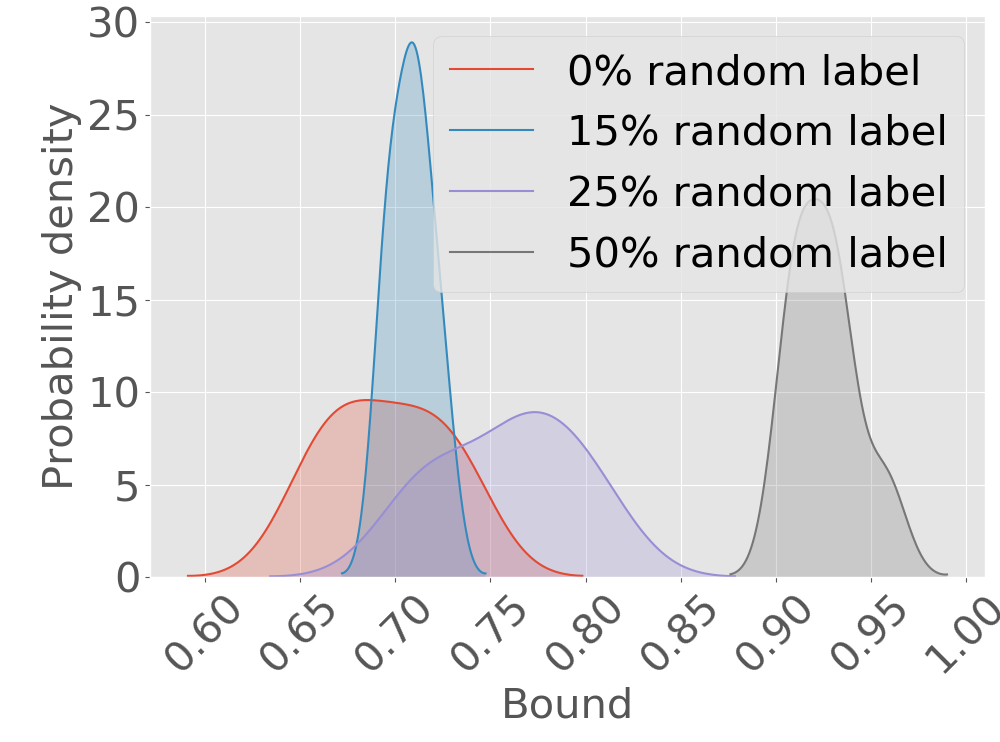}
  } 
\vspace{-4mm}
\caption[]{Results for ReLU-nets with depth = 4, width =256, total 1,052,426 parameters, trained on 1000 samples from CIFAR-10 with batch size = 16 and a increase in number of random labels (20 runs each) from 0\% to 50\%. (a-f) refer to Figure \ref{fig:main_mnist_rand}. 
Increasing percentage of random labels, the generalization bound as well as the components (effective curvature, $L_2$ norm, margin loss) increase, and the bound in (f) stays valid for the test error rate in (a).
}
\label{fig:main_cifar_d4_l256_bs16}
\vspace*{-4mm}
\end{figure*}



\begin{figure*}[t] 
\centering
 \subfigure[Test Error Rate]{
 \includegraphics[width = 0.28 \textwidth]{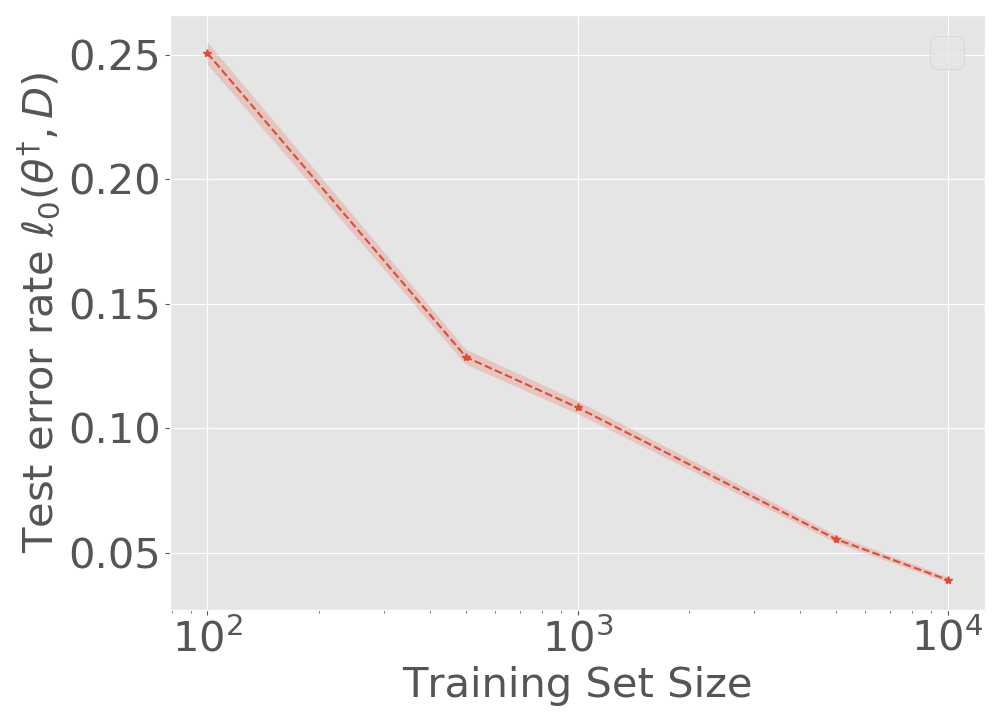}
 }  
 \subfigure[Diagonal Elements of Hessian.]{
 \includegraphics[width = 0.28 \textwidth]{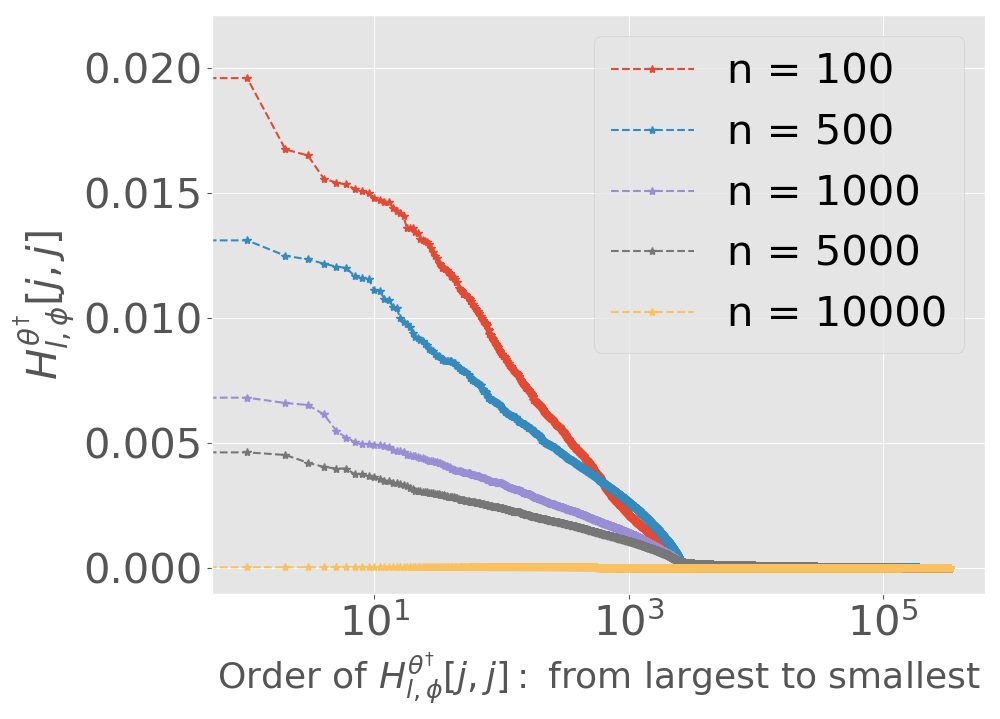}
 }  
 \subfigure[Effective Curvature.]{
 \includegraphics[width = 0.28 \textwidth]{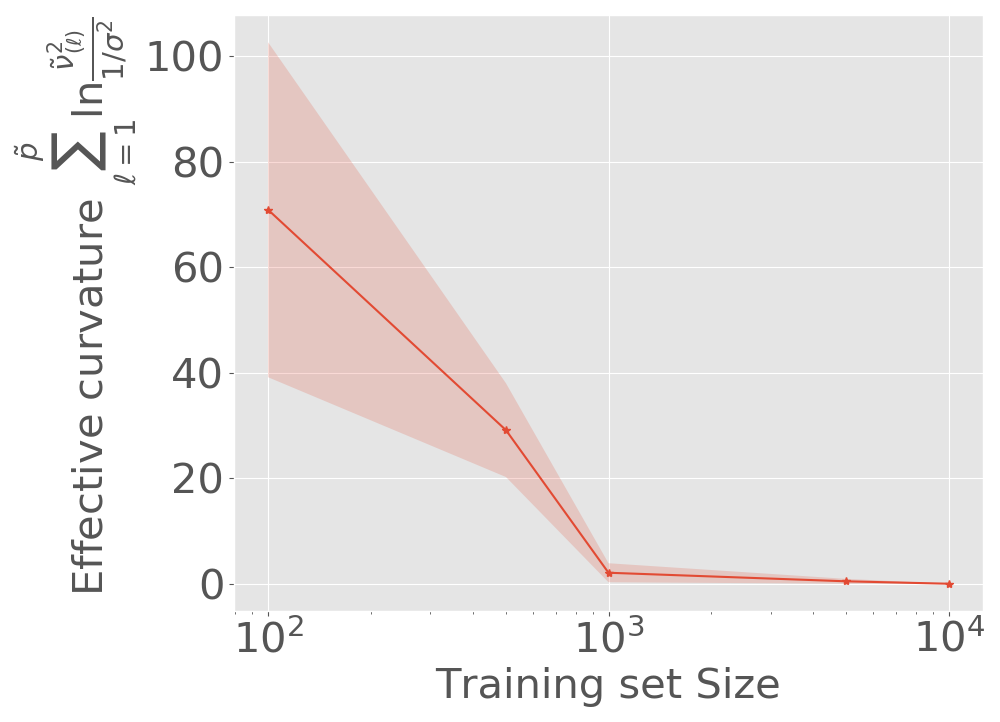}
 }  
 \subfigure[$L_2$ norm.]{
 \includegraphics[width = 0.28 \textwidth]{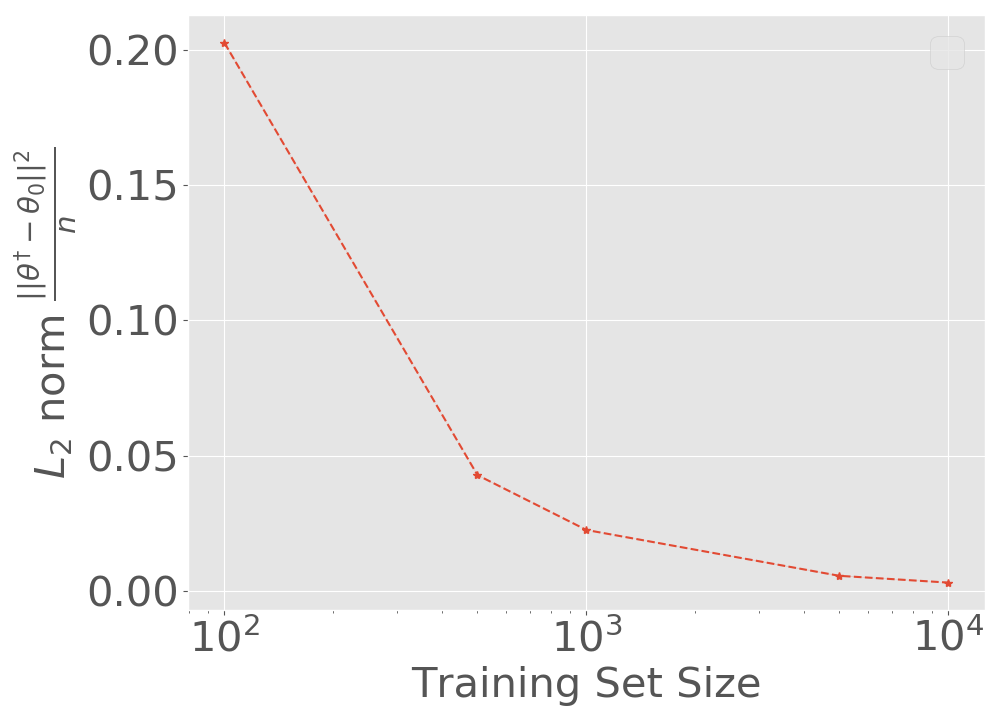}
 } 
  \subfigure[Margin Loss.]{
 \includegraphics[width = 0.28 \textwidth]{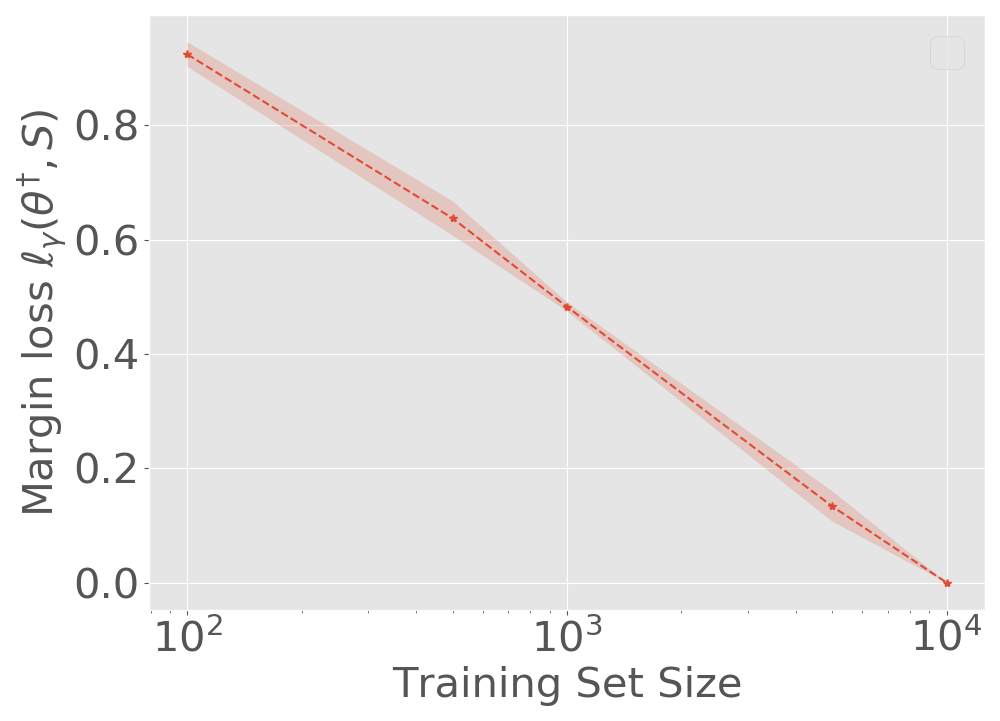}
 } 
  \subfigure[Generalization Bound.]{
 \includegraphics[width = 0.28 \textwidth]{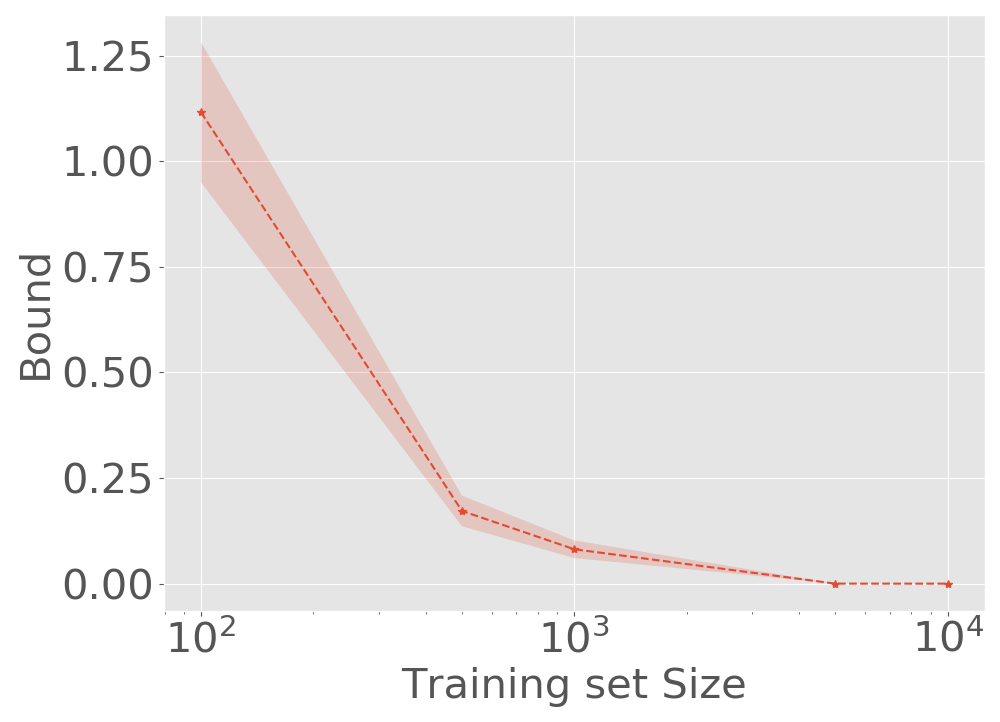}
 } 
\caption[]{Results for ReLU-nets with depth = 4, width =256, total 399,872 parameters, trained on MNIST with batch size = 128 and with increasing training set size $n$ (5 runs for each) from 100 to 10,000. (a) test set error rate; (b) diagonal elements (mean) of $\tilde \cH_{l,\phi}^{\theta^{\dagger}}$; (c) effective curvature; (d) $L_2$ norm of $\theta^{\dagger}$; (e) margin loss; (f)  generalization bound. The bound and all its components decrease with increase in $n$ from 100 to 10,000.
}
\label{fig:main_mnist_sample}
\end{figure*}

\begin{figure*}[!t] 
\centering
 \subfigure[Test Error Rate]{
 \includegraphics[width = 0.28 \textwidth]{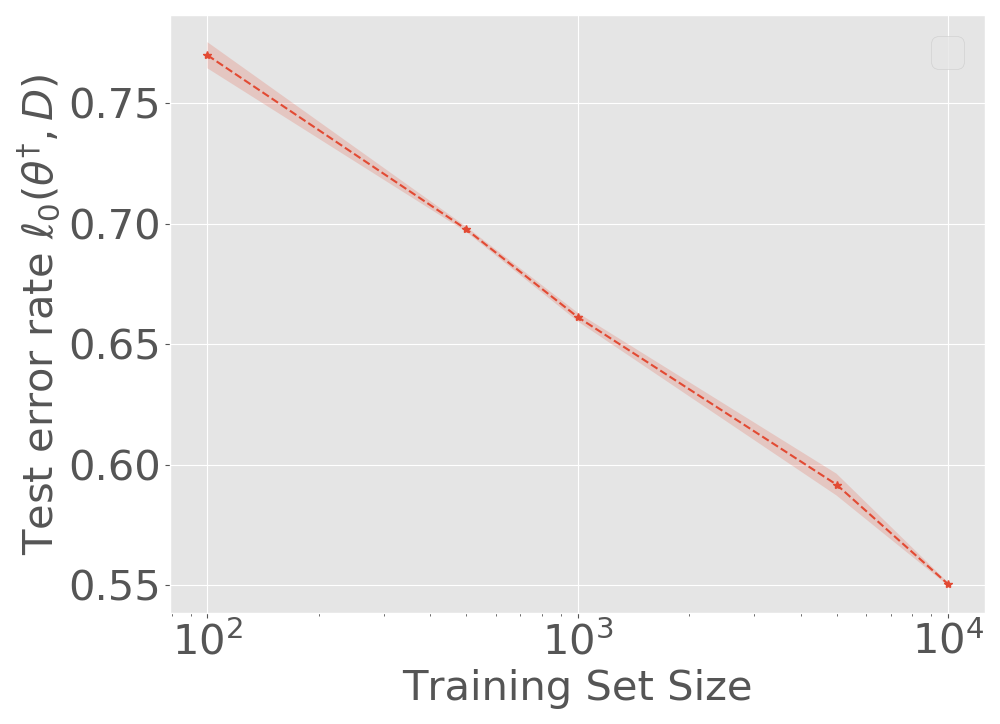}
 }  
 \subfigure[Diagonal Elements of Hessian.]{
 \includegraphics[width = 0.28 \textwidth]{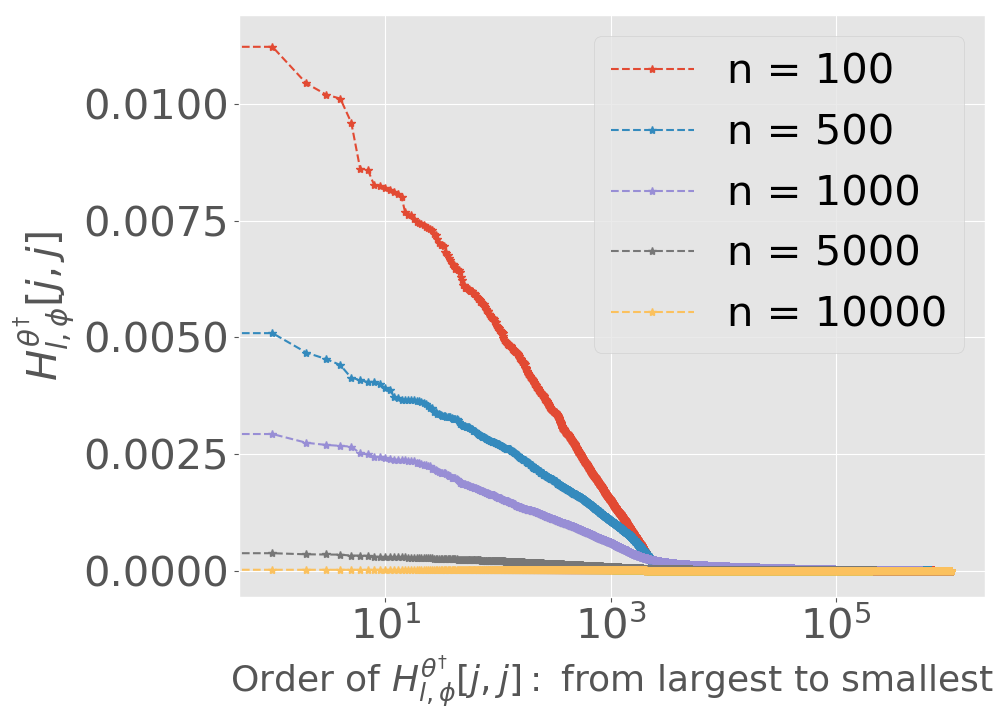}
 }  
 \subfigure[Effective Curvature.]{
 \includegraphics[width = 0.28 \textwidth]{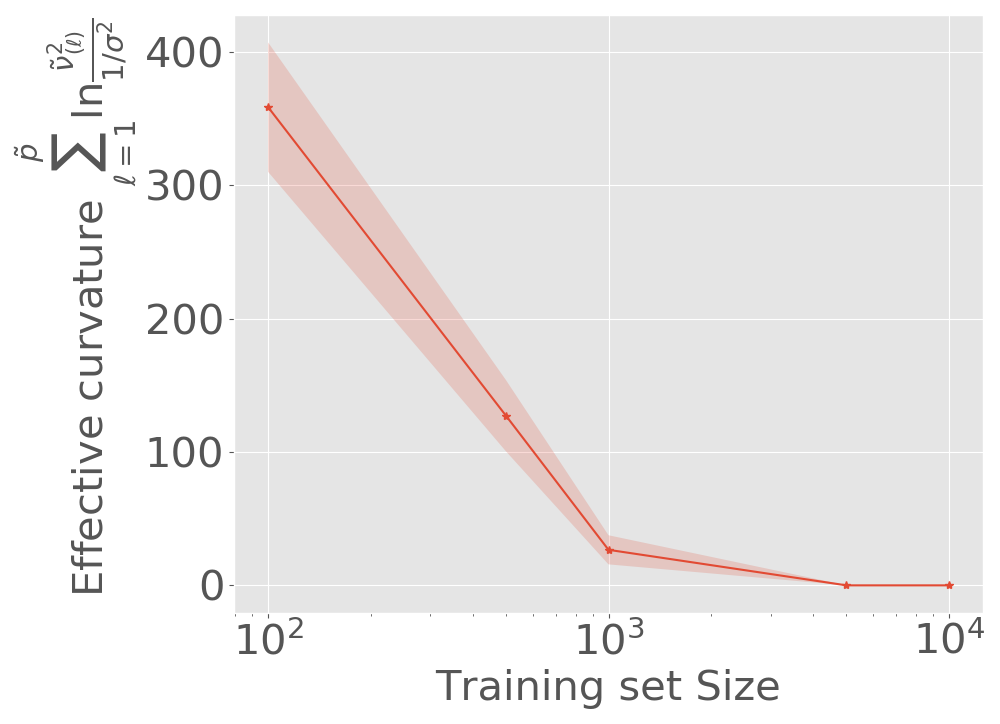}
 }  
 \subfigure[$L_2$ norm.]{
 \includegraphics[width = 0.28 \textwidth]{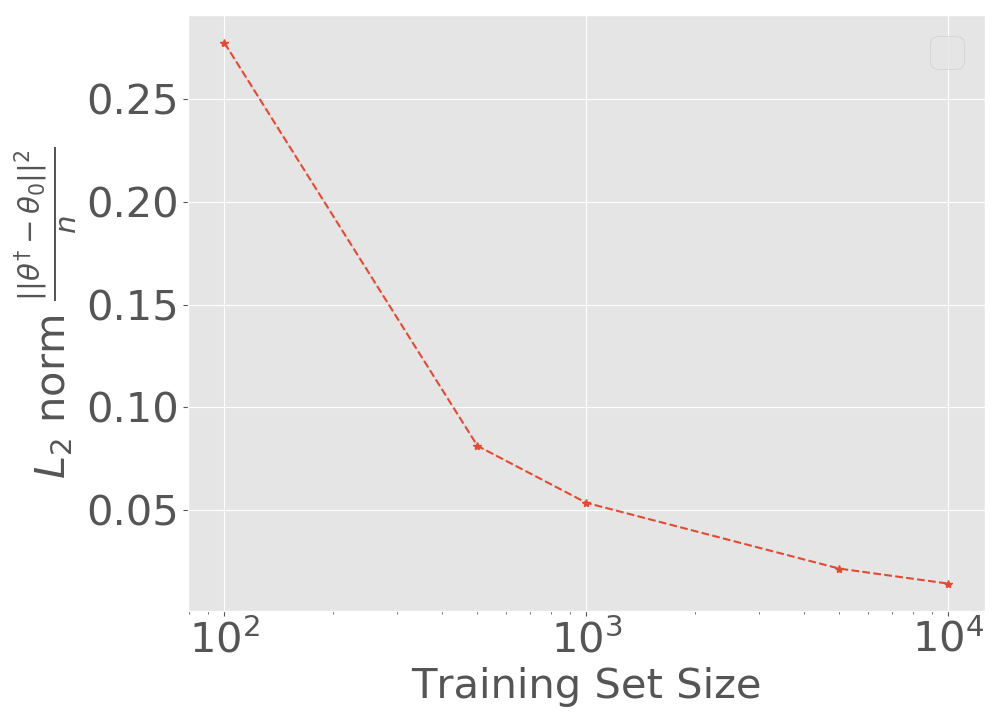}
 } 
  \subfigure[Margin Loss.]{
 \includegraphics[width = 0.28 \textwidth]{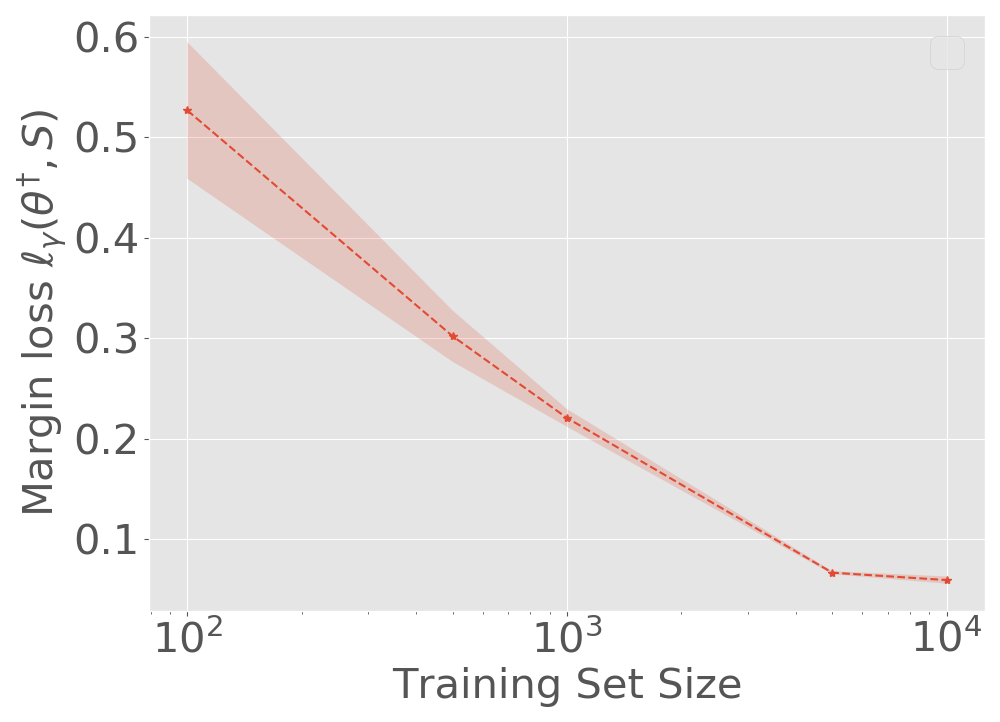}
 } 
  \subfigure[Generalization Bound.]{
 \includegraphics[width = 0.28 \textwidth]{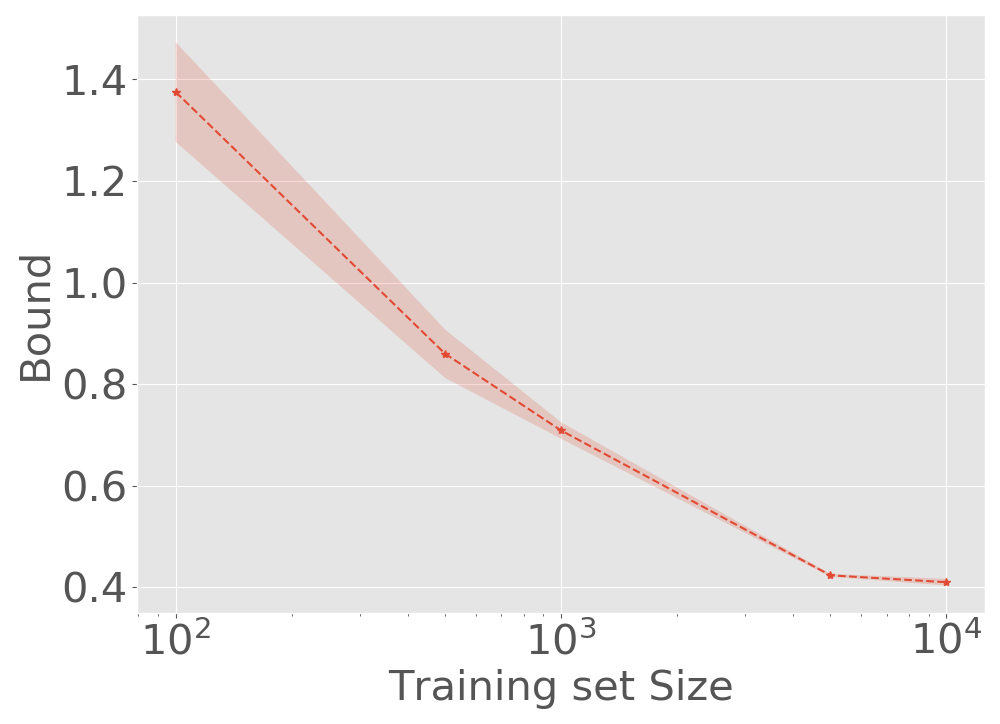}
 } 
\caption[]{Results for ReLU-nets with depth = 4, width =256 total 983,040 parameters, trained on CIFAR-10 with batch size = 128 and with increase in training set size $n$ (5 runs for each) from 100 to 10,000. 
 (a) test set error rate; (b) diagonal elements (mean) of $\tilde \cH_{l,\phi}^{\theta^{\dagger}}$; (c) effective curvature; (d) $L_2$ norm of $\theta^{\dagger}$; (e) margin loss; (f)  generalization bound. 
The bound and all its components decrease with increase in $n$ from 100 to 10,000.
}
\label{fig:main_cifar_sample}
\end{figure*}

\subsection{Bounds with Changing Random Labels.}
\label{exp:main_rand}

In the first set of experiments, we validate the theoretical promise of our bound with different level of randomness in the label by reporting the key factors i.e., empirical margin loss  $\ell_\gamma(\theta^{\dagger}, S)$, $L_2$ norm of the weights $\frac{\left\|\theta^{\dagger}-\theta_{0}\right\|^{2}}{n \sigma^{2}}$, and effective curvature $\sum_{\ell=1}^{\tilde p} \ln \frac{\tilde{\nu}^{2}(\ell)}{1 / \sigma^{2}}$, in our generalization bound.

Figure \ref{fig:main_mnist_rand} plots the change in test set error rate, the bound, and different components of the bound as the percentage of random labels is increased. Figure \ref{fig:main_mnist_rand} considers ReLU-nets  with  depth  =  4,  width  =128, rained  on  1000 samples from MNIST with batch size = 128. Figure \ref{fig:main_mnist_rand}(a) shows the test set error rate which understandably increases with the increase in random labels.
Figure \ref{fig:main_mnist_rand}(b) plots the sorted diagonal elements of $\tilde \cH_{l,\phi}^{\theta^{\dagger}}$ and shows that $\tilde \cH_{l,\phi}^{\theta^{\dagger}}[j,j]$ increases with increase in random labels, i.e., the curvature of the loss surface increases with
increase in random labels.
Figure \ref{fig:main_mnist_rand}(c) shows that the effective curvature increases with random labels, in line with the observations in Figure \ref{fig:main_mnist_rand}(b). While $\tilde \cH_{l,\phi}^{\theta^{\dagger}}[j,j]$ can change based on $\alpha$-scaling~\citep{dipb17}, the effective curvature is scale-invariant.
Figure \ref{fig:main_mnist_rand}(d) plots the $L_2$ norm $\|\theta^{\dagger}\|_2/(n\sigma^2)$ (with $\theta_0=0$) and shows that $\theta^{\dagger}$ learned with more random labels has a larger $L_2$ norm. 
Figure \ref{fig:main_mnist_rand}(e) shows that the empirical margin loss distribution shifts to a higher value with increase in random labels. Figure \ref{fig:main_mnist_rand}(f) plots the proposed bound as $a_{\eta} \ell_{\gamma}(\theta^{\dagger}, S) + \frac{b _{\eta}}{2 n}(\sum_{\ell=1}^{\tilde{p}} \ln \frac{\tilde{\nu}_{\ell}^{2}}{1 / \sigma^{2}}+\frac{\|\theta^{\dagger}-\theta_{0}\|^{2}}{\sigma^{2}})$ with $\eta = 0.1$ and $\sigma^2 =100$. We omit the $d_{\eta} \exp (-\min (c_{2} \gamma^{2}, c_{1} \gamma))+b_{\eta} (\log (\frac{1}{\delta}))/n$
terms since they do not change with change in random labels.  Figure \ref{fig:main_mnist_rand}(f) shows that with the randomness in labels increasing from $0\%$ to $50\%$, the generalization error shifts to a higher value and is consistent with the change of the test set error rate in Figure \ref{fig:main_mnist_rand}(a).

\noindent \textbf{Additional Results.} To validate our bound for ReLU-nets with different depth, width and trained with different batch size, we present the results for ReLU-nets with depth = 2 and  width = 128 trained with batch size 128 in
Figure \ref{fig:main_mnist_d2_l128}, and ReLU-nets with  depth  =  4,  width  =128 and trained with batch size 16 in \ref{fig:main_mnist_d4_l128_bs16}.
Both figures show that increasing percentage of random labels, the generalization bound as well as the components (effective curvature, $L_2$ norm, margin loss) increase, and the bound in Figure \ref{fig:main_mnist_d2_l128} (f) and \ref{fig:main_mnist_d4_l128_bs16} (f) indicates the observed test error rate in Figure \ref{fig:main_mnist_d2_l128} (a) and \ref{fig:main_mnist_d4_l128_bs16} (a) respectively. We also consider CIFAR-10 dataset, i.e., the results for ReLU-nets with depth = 4, width = \{256, 512\}, trained on 1000 samples from CIFAR-10 with batch size = 128 are presented in Figure \ref{fig:main_cifar_d4_l256} and \ref{fig:main_cifar_d4_l512}. The ReLU-nets with depth = 4, width = 256, trained with batch size = 16 are presented in Figure \ref{fig:main_cifar_d4_l256_bs16}. Those results demonstrate that the observations from MNIST are also valid for CIFAR-10 dataset and our bounds stay valid and non-vacuous as they match the observed test error rate.

\subsection{Bounds with Changing Training Set Size.}
\label{exp:main_sample}
In this section, we evaluate how the generalization bound behaves when the training set size increases. We report the key factors i.e., empirical margin loss  $\ell_\gamma(\theta^{\dagger}, S)$, $L_2$ norm of the weights $\frac{\left\|\theta^{\dagger}-\theta_{0}\right\|^{2}}{n\sigma^{2}}$, and effective curvature $\sum_{\ell=1}^{\tilde p} \ln \frac{\tilde{\nu}_{\ell}^{2}(\ell)}{1 / \sigma^{2}}$ in the bound for different size $n \in \{ 100, 500, 1000, 5000, 10000\}$ of the training set in Figures \ref{fig:main_mnist_sample} and \ref{fig:main_cifar_sample} for MNIST and CIFAR-10 respectively with ReLU-nets of depth = 4,  width = 256 and trained with batch size 128. Figures \ref{fig:main_mnist_sample} and \ref{fig:main_cifar_sample} show the change in test set error rate, the bound, and different components of the bound with increase in training set size $n$ for MNIST, and CIFAR-10. Figure \ref{fig:main_mnist_sample}(a) shows that the test set error rate decreases with increase in the training set size $n$ \citep{nako19b}. Figure \ref{fig:main_mnist_sample}(b) shows that the sorted diagonal elements of $\tilde \cH_{l,\phi}^{\theta^{\dagger}}$ decrease with increase in $n$. As a consequence, the effective curvature decreases with increase in $n$  as shown in Figure \ref{fig:main_mnist_sample}(c). Recall that the  effective curvature is scale invariant and hence does not change based on $\alpha$-scaling. Figure \ref{fig:main_mnist_sample}(d) shows that the $L_2$ norm term $\|\theta\|_2/(n\sigma^2)$ also decreases with increase in $n$. The behavior of the $L_2$ norm has been studied closely in recent literature~\citep{nako19b} and we revisit this in the Appendix. Figure \ref{fig:main_mnist_sample}(e) shows that the empirical margin loss $\ell_{\gamma}\left(\theta^{\dagger}, S\right)$ also decreases with increase in $n$. 
Figure \ref{fig:main_mnist_sample}(f) plots the proposed generalization bound with $\eta = 0.1$ and $\sigma^2 =1000$, and shows that with $n$ increasing from 100 to 10000, the generalization error decreases, and is consistent with the test set error rate behavior in Figure \ref{fig:main_mnist_sample}(a) and unlike bounds from several other recent bounds \citep{nako19b}.  Figure \ref{fig:main_cifar_sample}(a-f) show that the above observations for MNIST also hold for CIFAR-10.

\begin{figure*}[t] 
\centering
 \subfigure[Test Error Rate]{
 \includegraphics[width = 0.28 \textwidth]{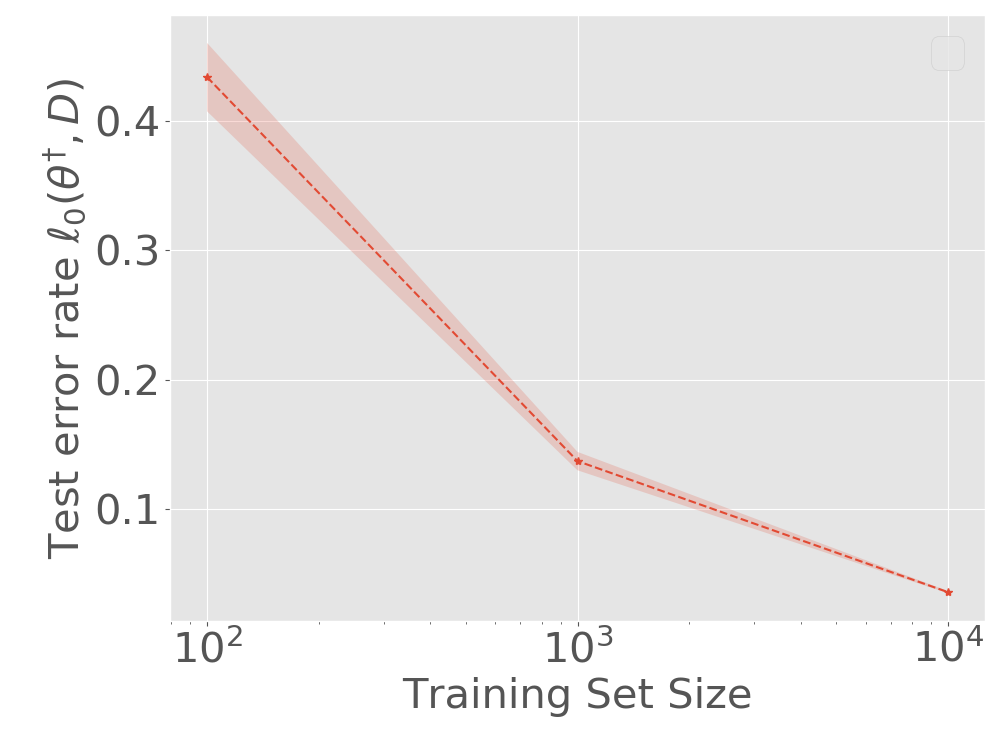}
 } 
 \subfigure[Diagonal Elements of  Hessian.]{
 \includegraphics[width = 0.28\textwidth,height =0.20\textwidth]{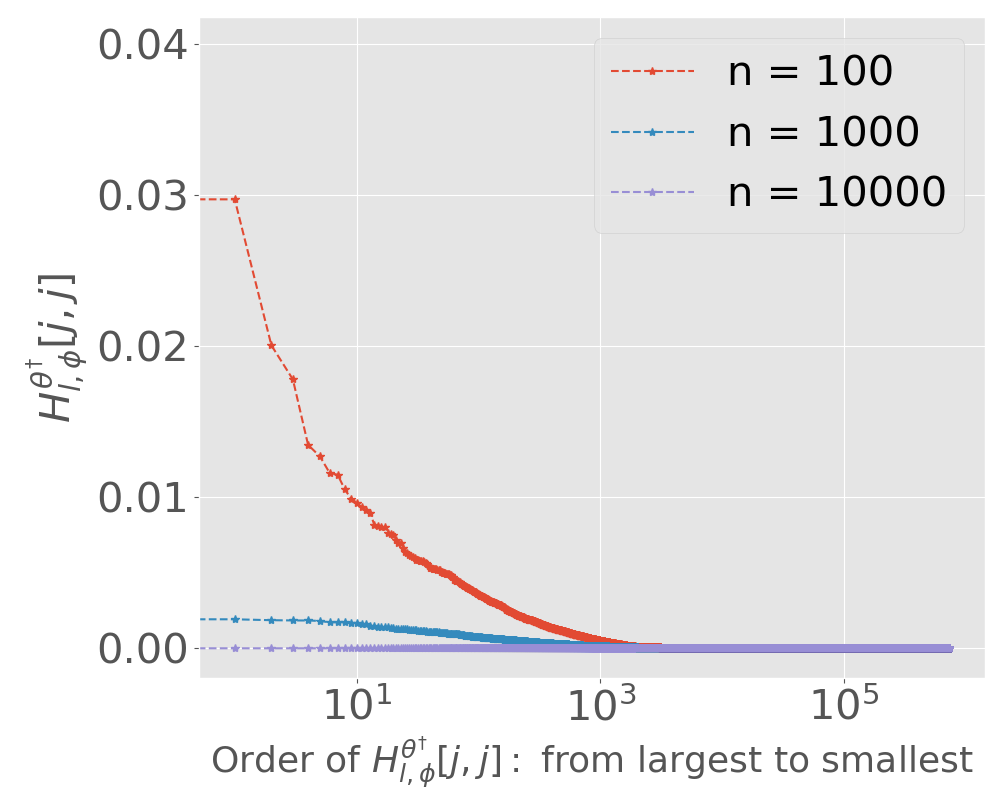}
 } 
 \subfigure[Effective Curvature.]{
 \includegraphics[width = 0.28 \textwidth,height =0.20\textwidth]{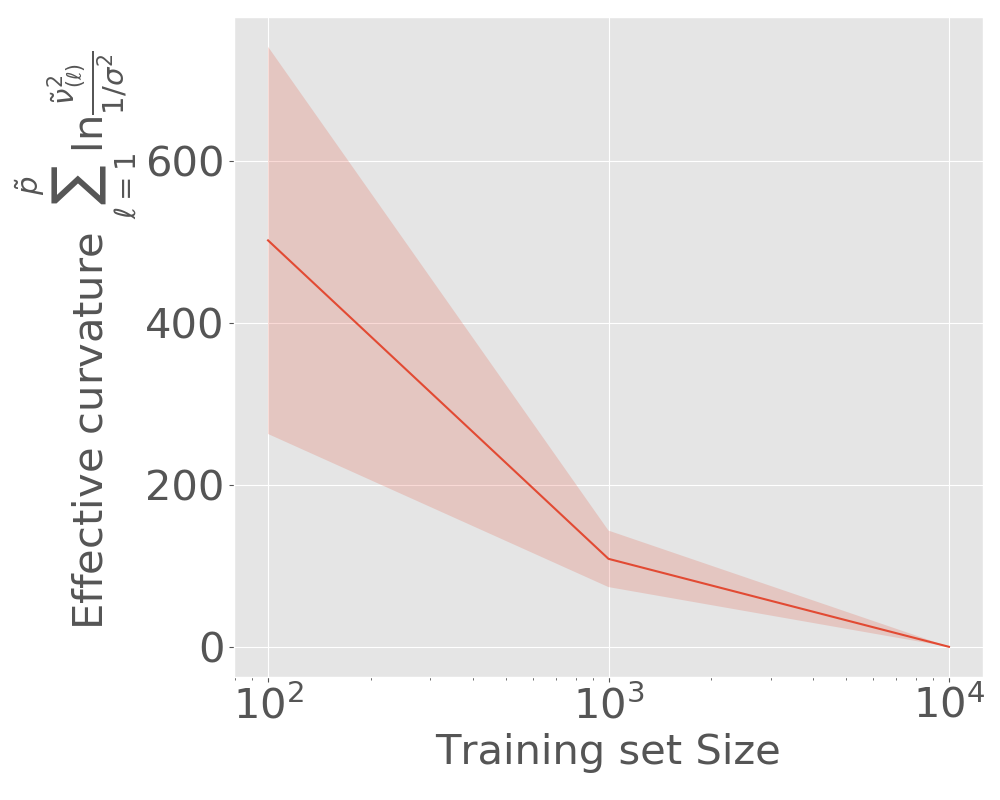}
 } 
 \subfigure[$L_2$ norm / no. sample.]{
 \includegraphics[width = 0.28 \textwidth]{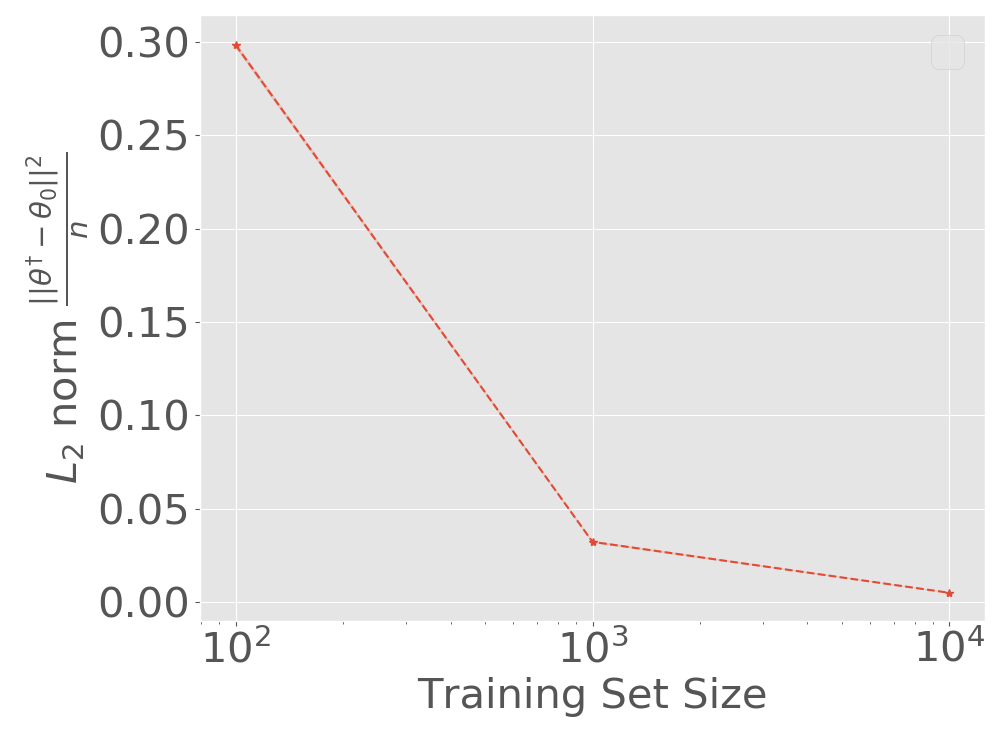}
 } 
  \subfigure[Margin Loss.]{
 \includegraphics[width = 0.28 \textwidth,height =0.20\textwidth]{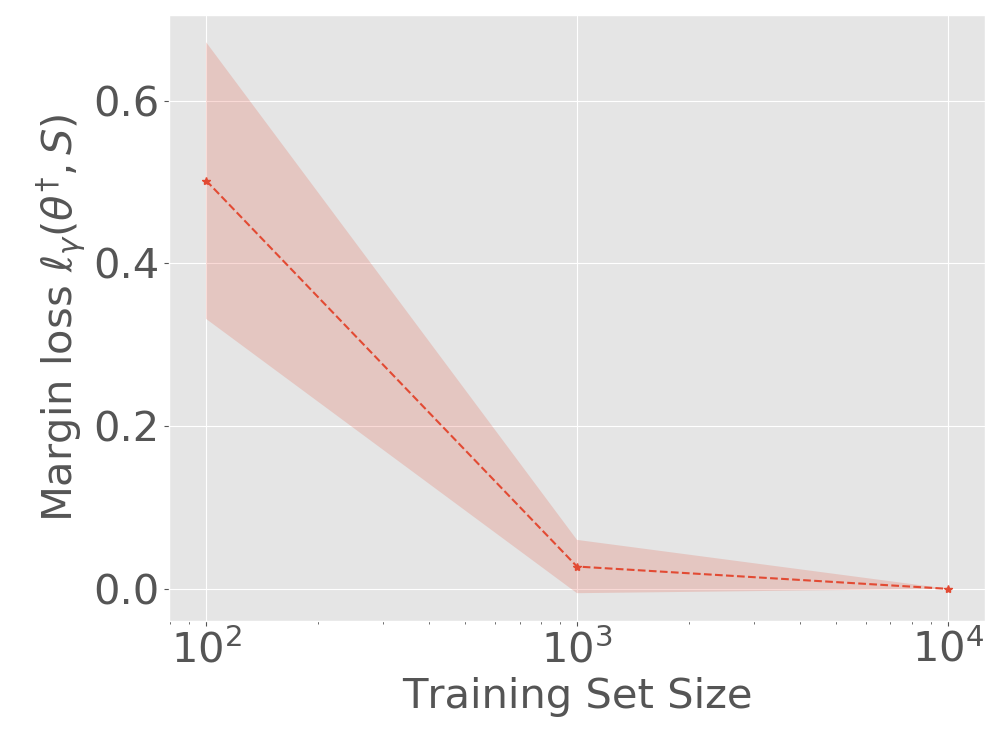}
 } 
  \subfigure[Generalization Bound.]{
 \includegraphics[width = 0.28 \textwidth,height =0.20\textwidth]{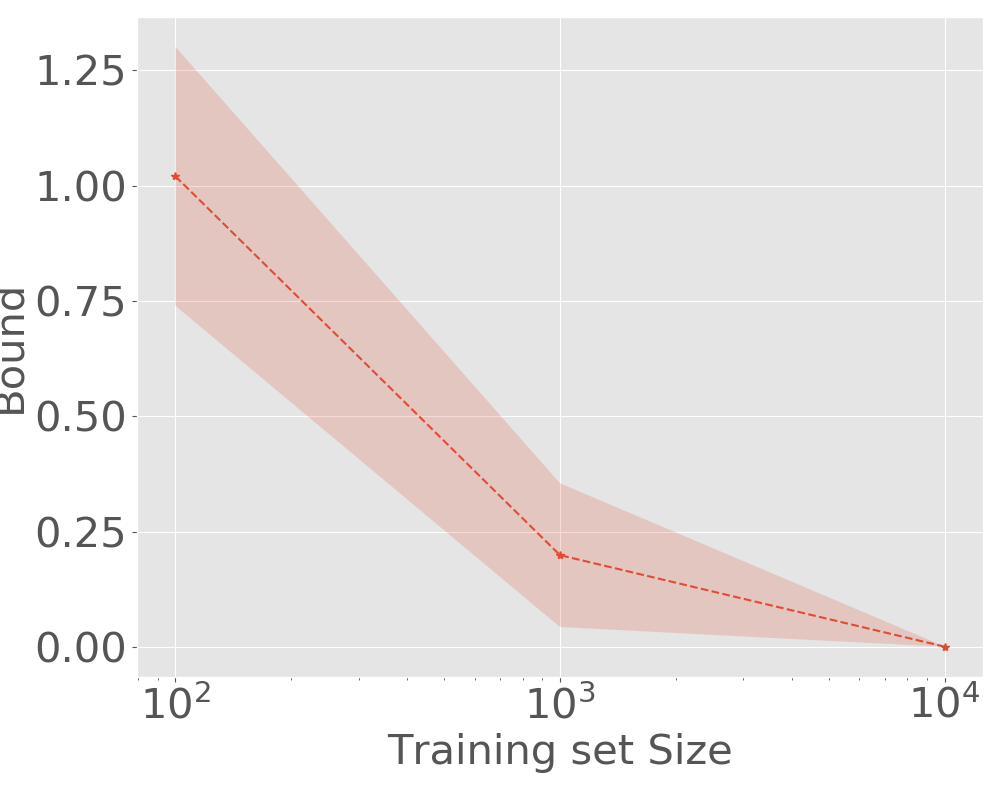}
 } 
\vspace{-2mm}
\caption[]{Results for ReLU-nets with depth = 8, width =256, total 727,552 parameters, trained on MNIST (batch size = 128) with increasing training set size $n$ (5 runs for each) from 100 to 10,000.
(a) test set error rate; (b) diagonal elements (mean) of $\tilde \cH_{l,\phi}^{\theta^{\dagger}}$; (c) effective curvature; (d) $L_2$ norm of $\theta^{\dagger}$; (e) margin loss; (f)  generalization bound. The bound and all its components decrease with increase in $n$ from 100 to 10,000.
}
\label{fig:main_mnist_sample_d8_l256}
\end{figure*}

\begin{figure*}[!t] 
\centering
 \subfigure[Test Error Rate]{
 \includegraphics[width = 0.28 \textwidth]{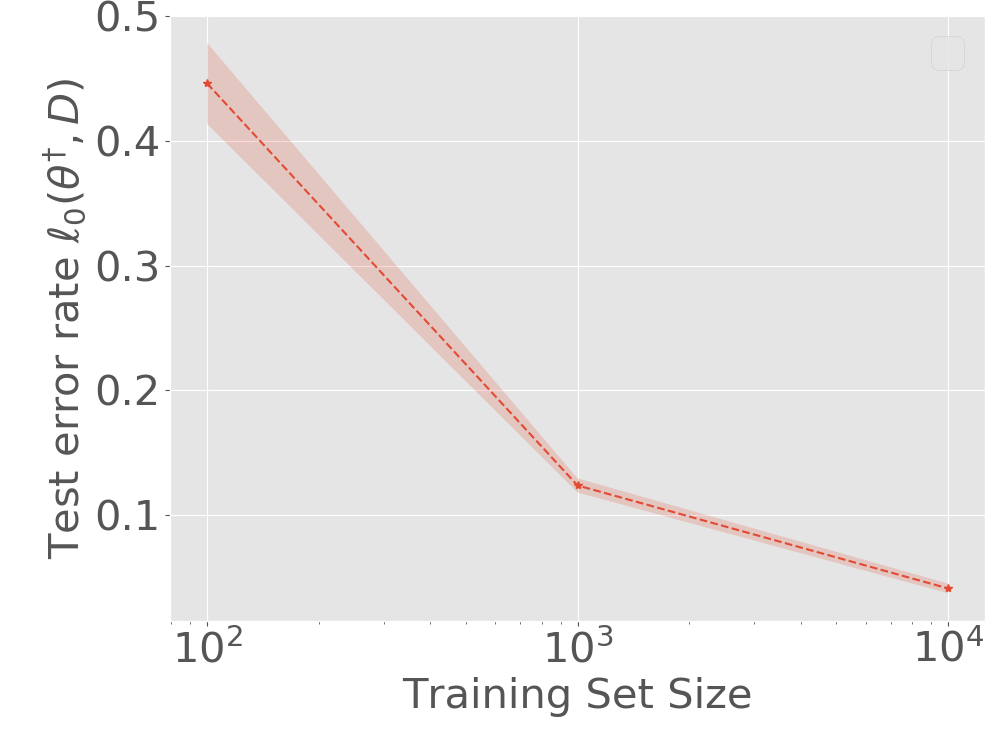}
 } 
 \subfigure[Diagonal Elements of  Hessian.]{
 \includegraphics[width = 0.28 \textwidth, height =0.20\textwidth]{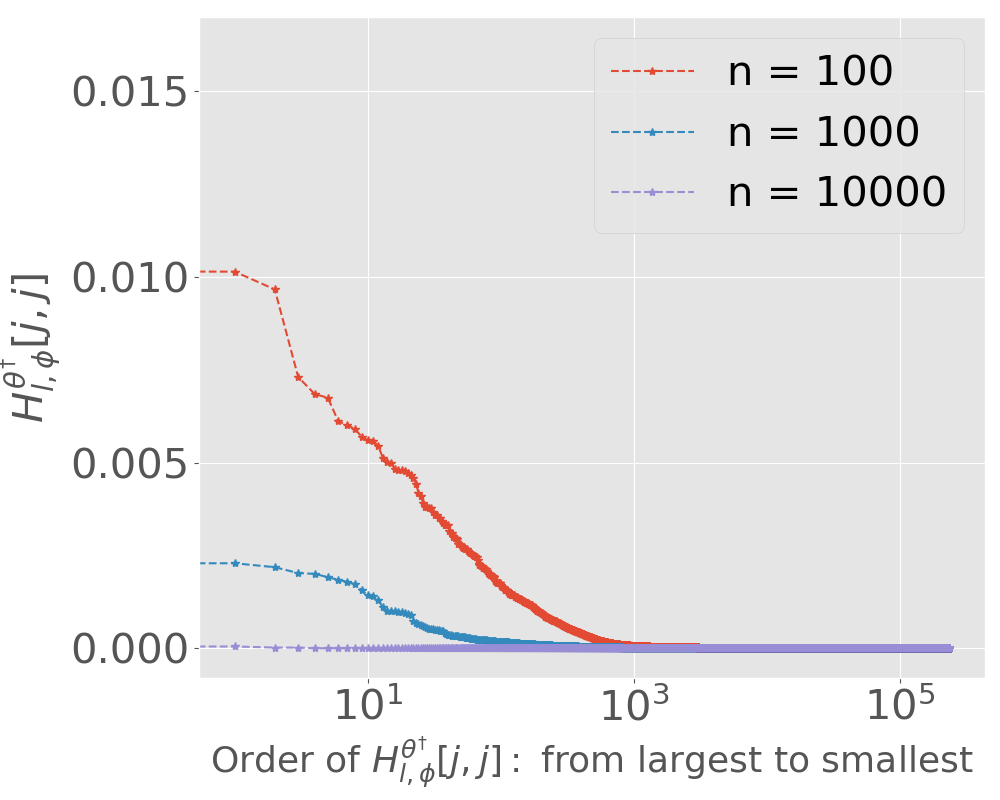}
 } 
 \subfigure[Effective Curvature.]{
 \includegraphics[width = 0.28 \textwidth, height =0.20\textwidth]{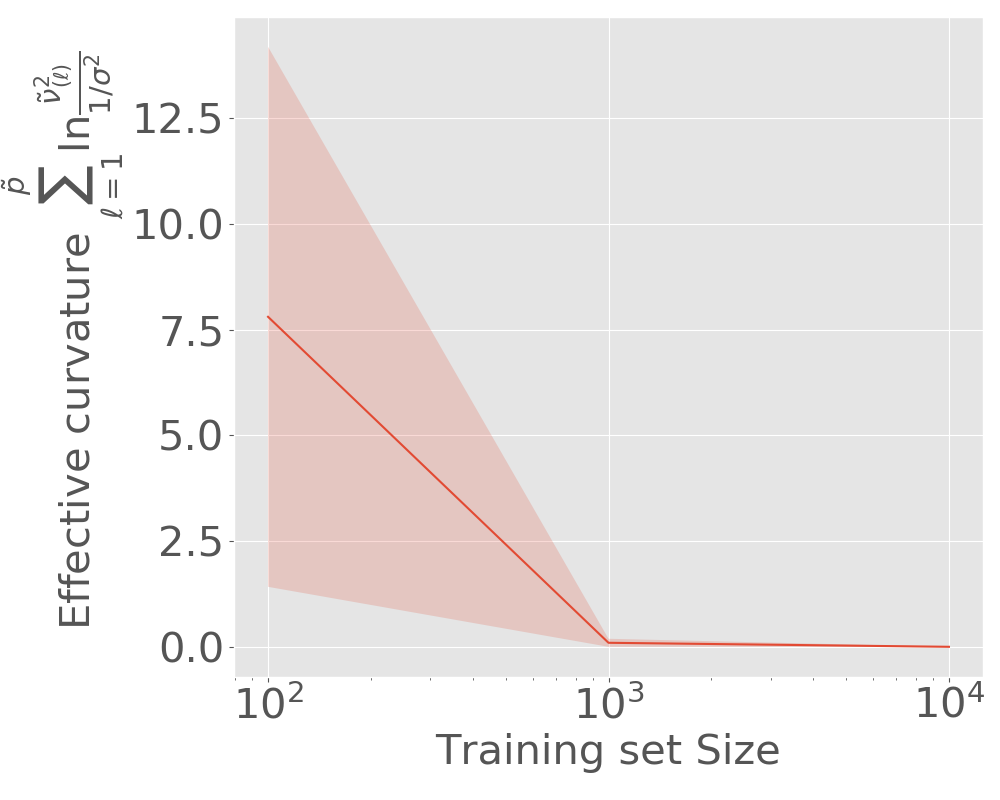}
 } 
 \subfigure[$L_2$ norm / no. sample.]{
 \includegraphics[width = 0.28 \textwidth]{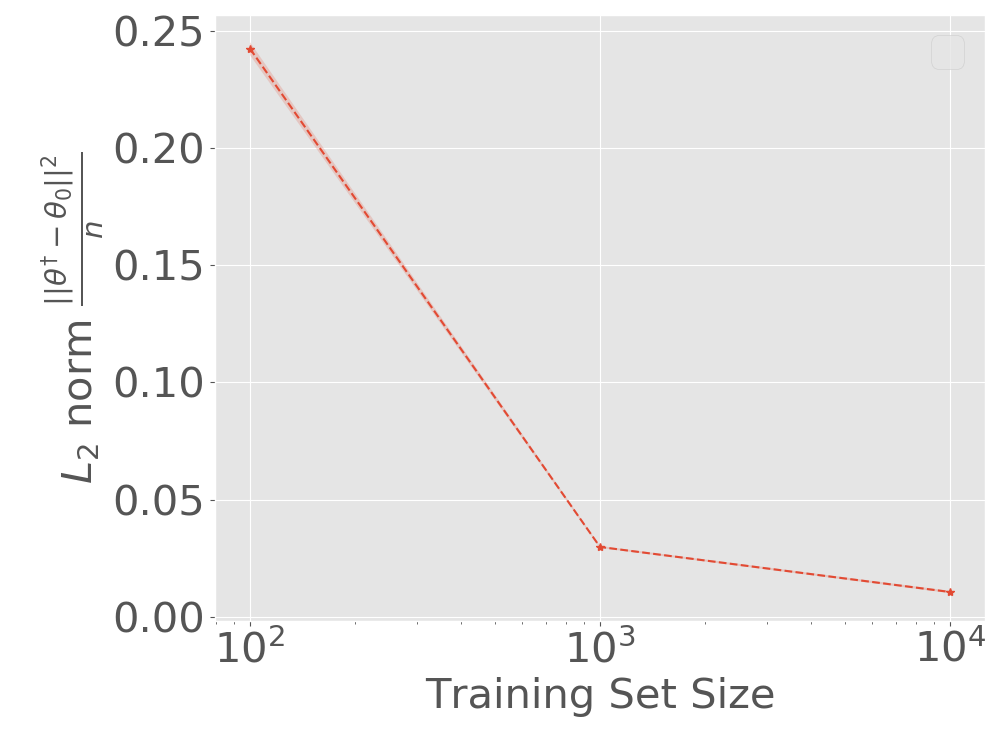}
 } 
  \subfigure[Margin Loss.]{
 \includegraphics[width = 0.28 \textwidth]{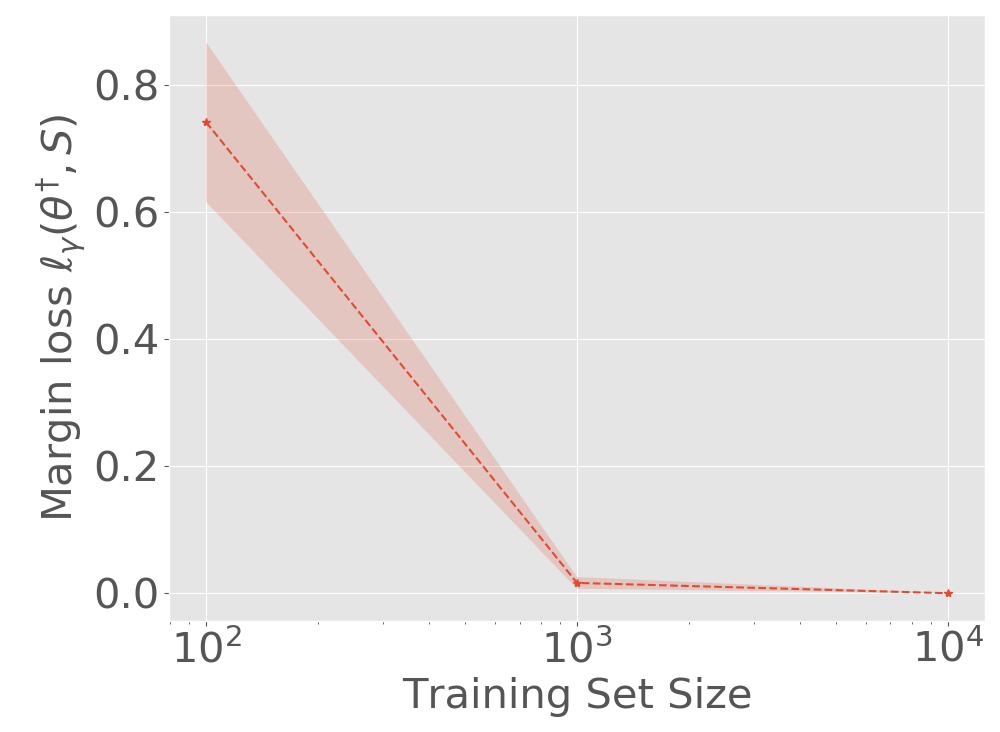}
 } 
  \subfigure[Generalization Bound.]{
 \includegraphics[width = 0.28 \textwidth, height =0.20\textwidth]{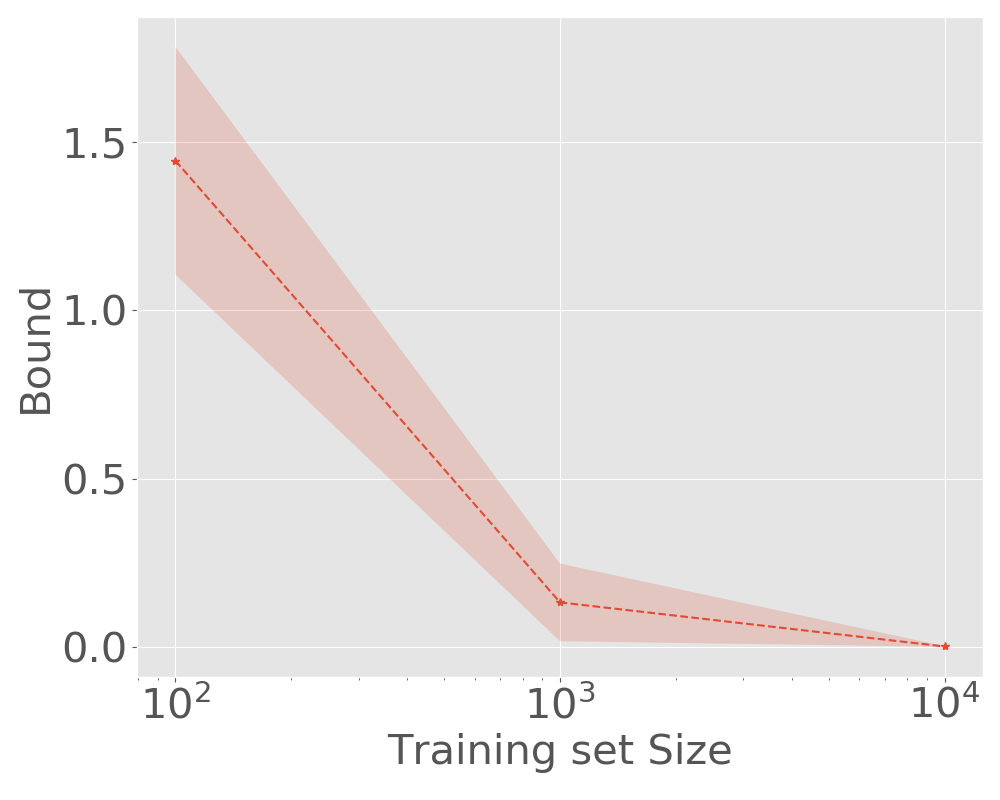}
 } 
\caption[]{Results for ReLU-nets with depth = 8, width =128, total 416,256 parameters, trained on MNIST (batch size = 16) with increasing training set size $n$ (5 runs for each) from 100 to 10,000.
(a) test set error rate; (b) diagonal elements (mean) of $\tilde \cH_{l,\phi}^{\theta^{\dagger}}$; (c) effective curvature; (d) $L_2$ norm of $\theta^{\dagger}$; (e) margin loss; (f)  generalization bound. The bound and all its components decrease with increase in $n$ from 100 to 10,000.
}
\label{fig:main_mnist_sample_d8_l128_bs16}
\end{figure*}

\begin{figure*}[t] 
\centering
 \subfigure[Test Error Rate]{
 \includegraphics[width = 0.28 \textwidth]{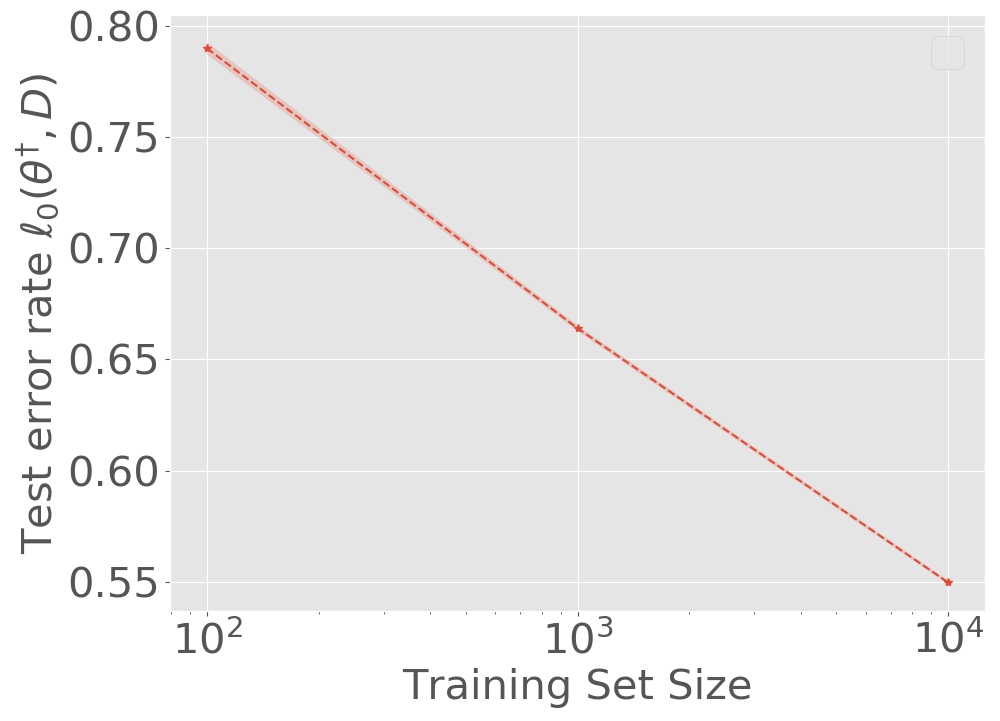}
 } 
 \subfigure[Diagonal Elements of  Hessian.]{
 \includegraphics[width = 0.28 \textwidth, height =0.20\textwidth]{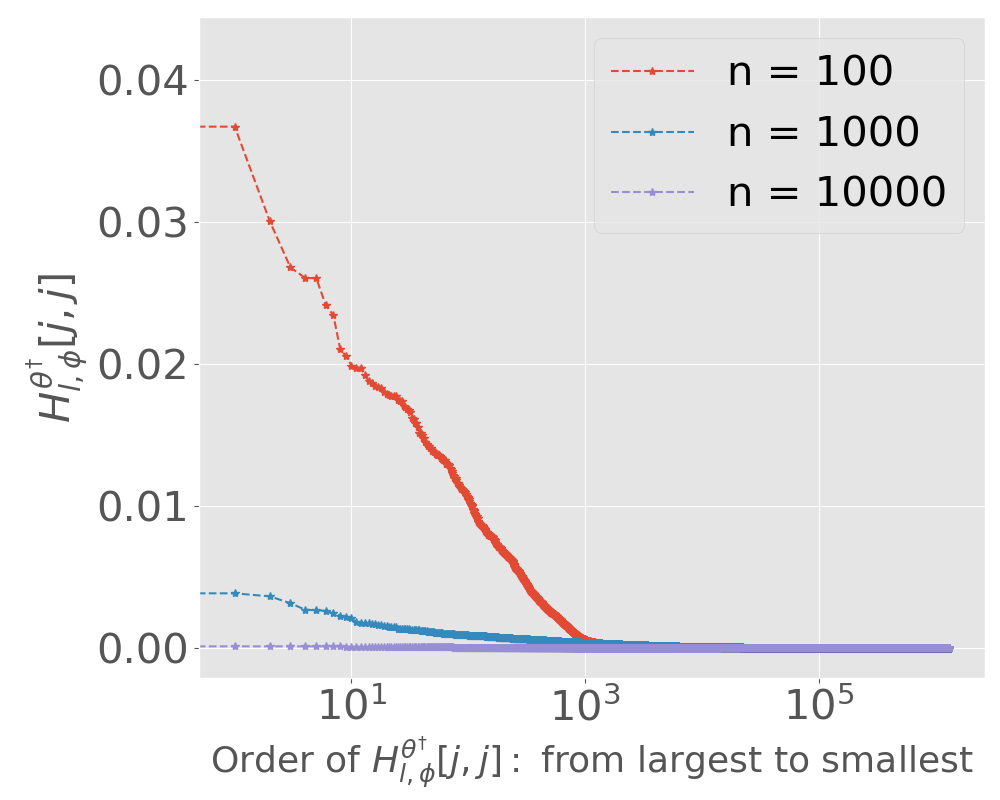}
 } 
 \subfigure[Effective Curvature.]{
 \includegraphics[width = 0.28 \textwidth, height =0.20\textwidth]{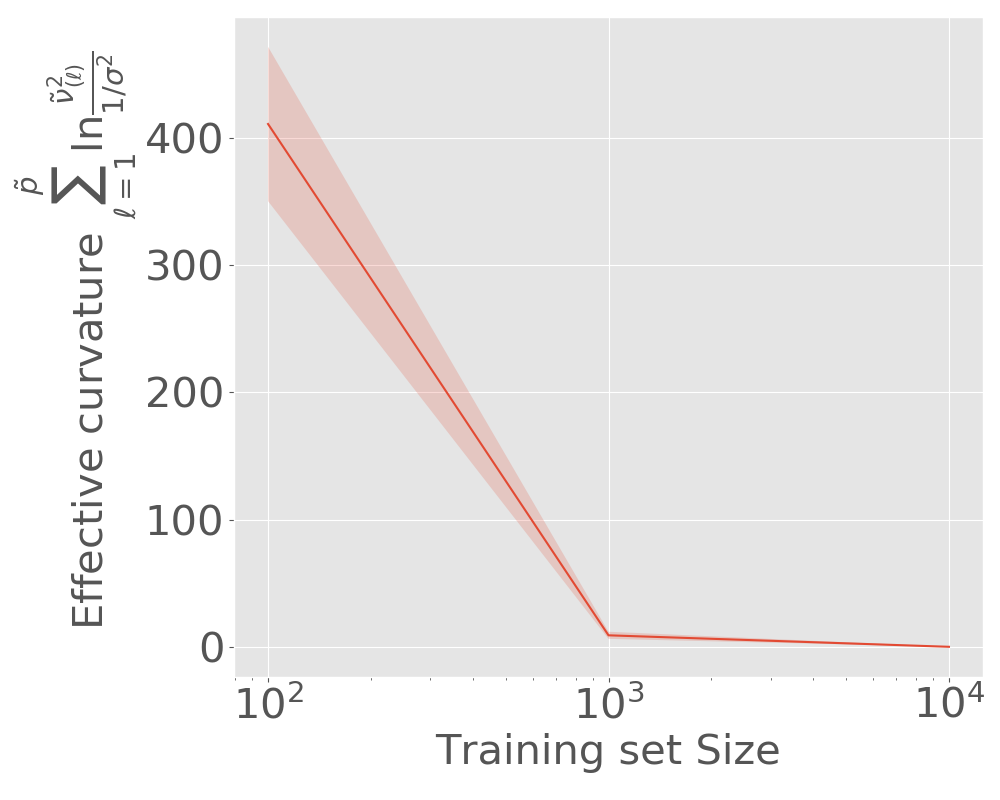}
 } 
 \subfigure[$L_2$ norm / no. sample.]{
 \includegraphics[width = 0.28 \textwidth]{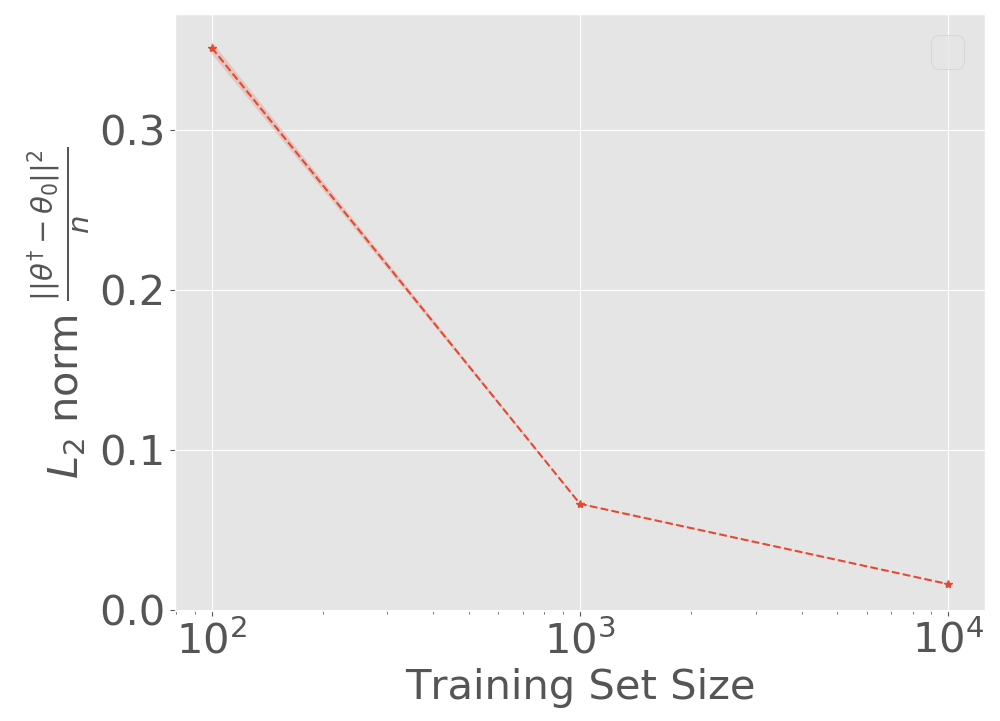}
 } 
  \subfigure[Margin Loss.]{
 \includegraphics[width = 0.28 \textwidth]{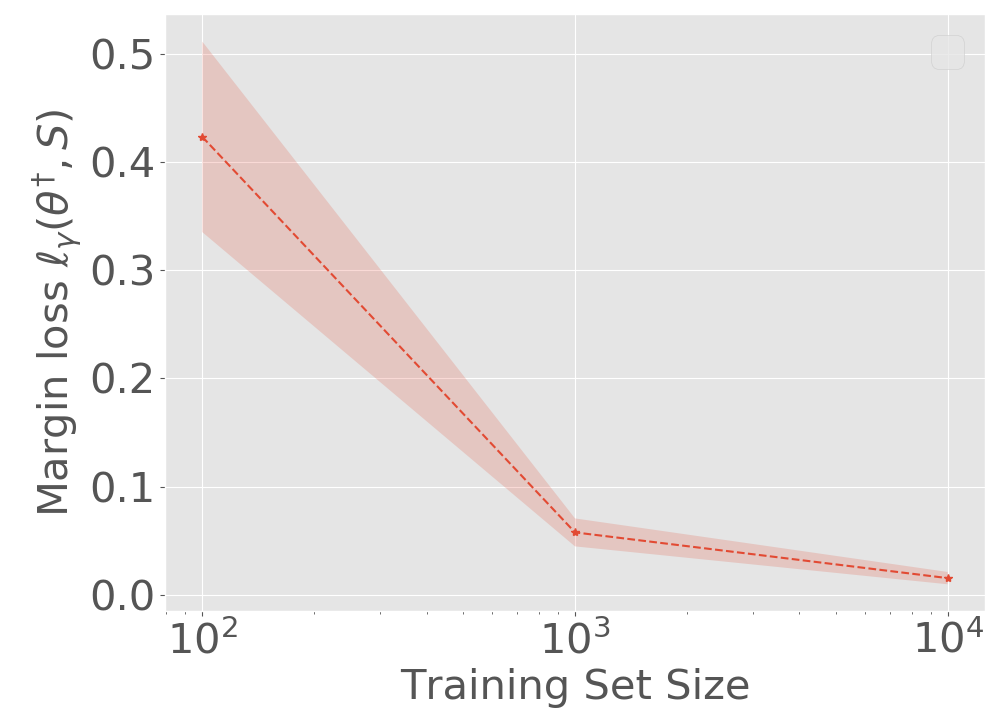}
 } 
  \subfigure[Generalization Bound.]{
 \includegraphics[width = 0.28 \textwidth, height =0.20\textwidth]{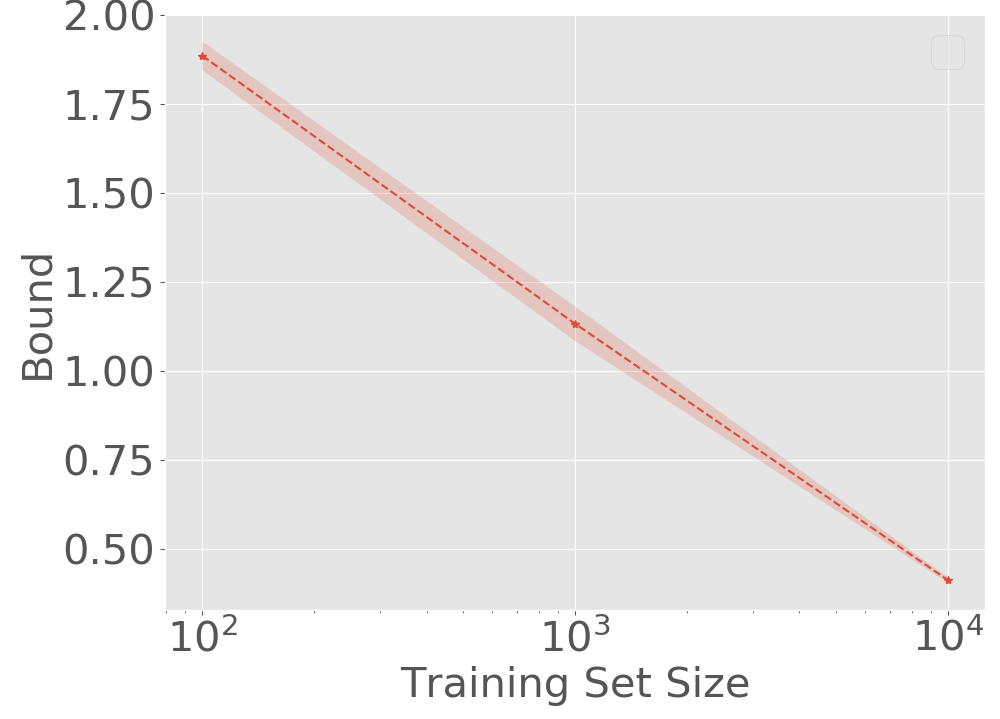}
 } 
\caption[]{Results for ReLU-nets with depth = 8, width =256, total 1,247,744 parameters, trained on CIFAR-10 (batch size = 128) with increasing training set size $n$ (5 runs for each) from 100 to 10,000. 
(a) test set error rate; (b) diagonal elements (mean) of $\tilde \cH_{l,\phi}^{\theta^{\dagger}}$; (c) effective curvature; (d) $L_2$ norm of $\theta^{\dagger}$; (e) margin loss; (f)  generalization bound. The bound and all its components decrease with increase in $n$ from 100 to 10,000.
}
\label{fig:main_cifar_sample_d8_l256}
\end{figure*}

\begin{figure*}[!t] 
\centering
 \subfigure[Test Error Rate]{
 \includegraphics[width = 0.28 \textwidth]{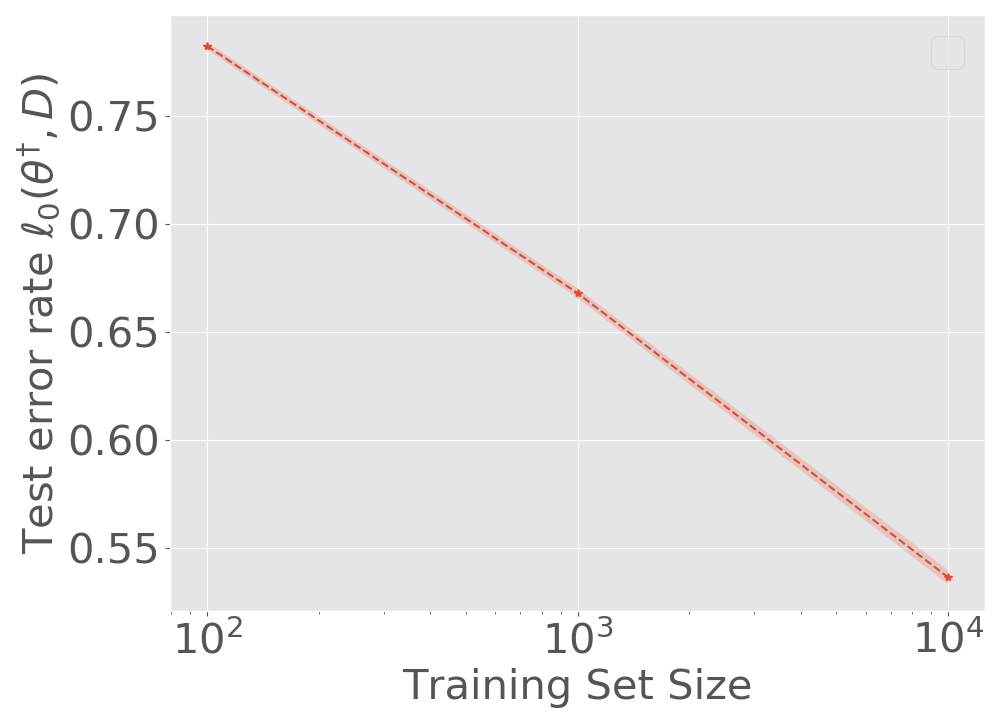}
 } 
 \subfigure[Diagonal Elements of  Hessian.]{
 \includegraphics[width = 0.28 \textwidth, height =0.20\textwidth]{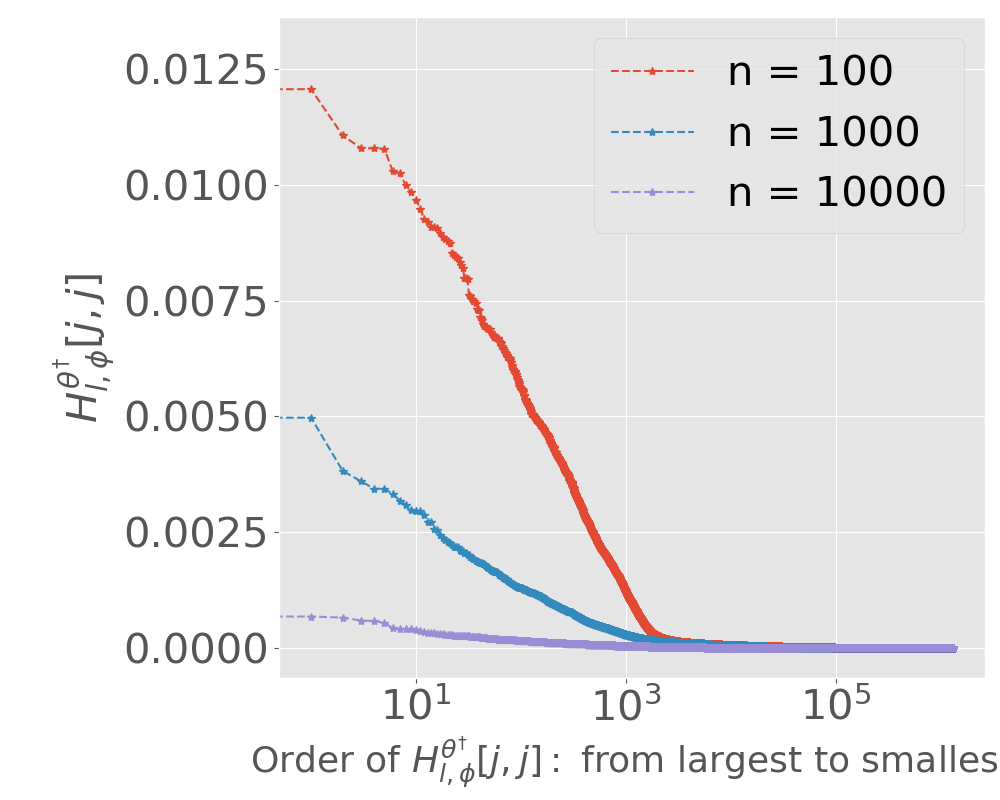}
 } 
 \subfigure[Effective Curvature.]{
 \includegraphics[width = 0.28 \textwidth, height =0.20\textwidth]{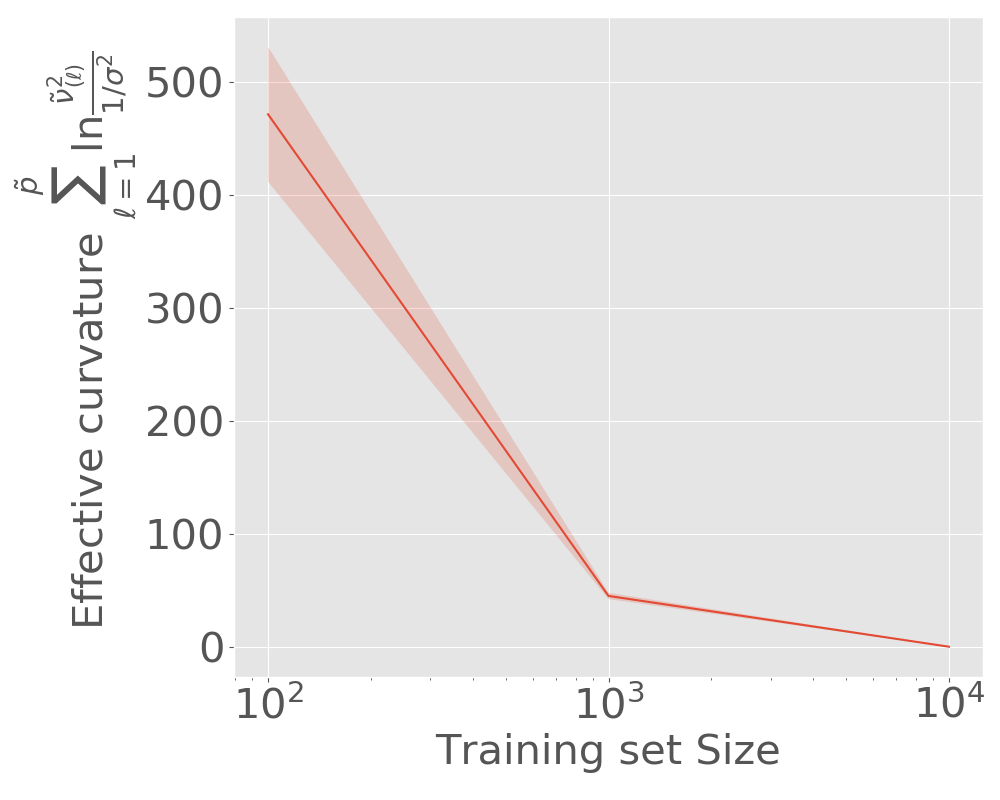}
 } 
 \subfigure[$L_2$ norm / no. sample.]{
 \includegraphics[width = 0.28 \textwidth]{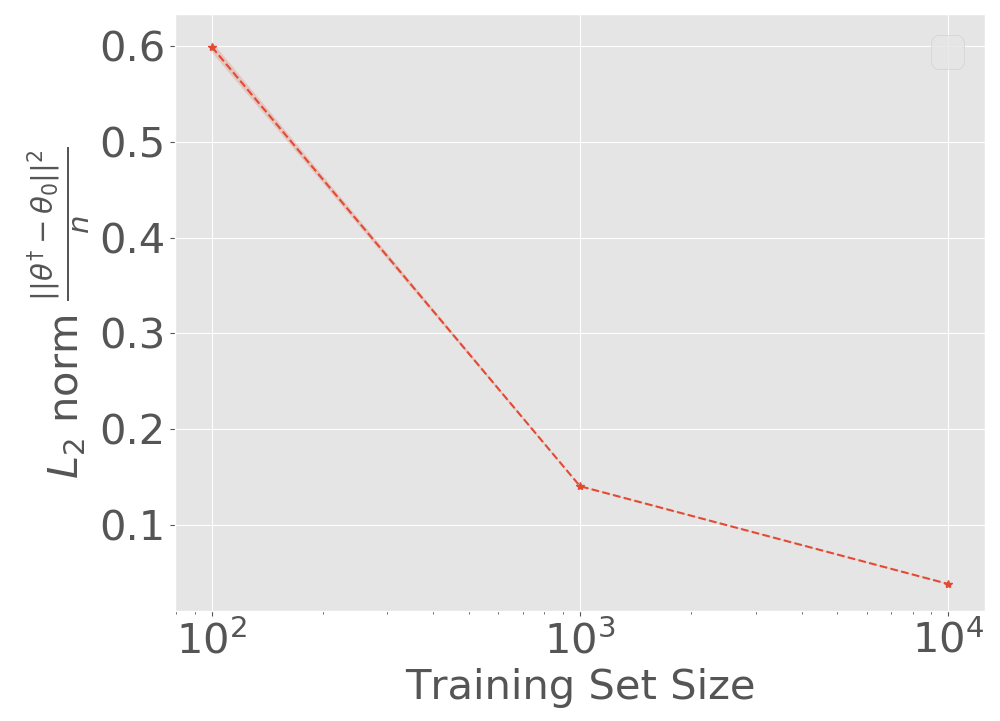}
 } 
  \subfigure[Margin Loss.]{
 \includegraphics[width = 0.28 \textwidth]{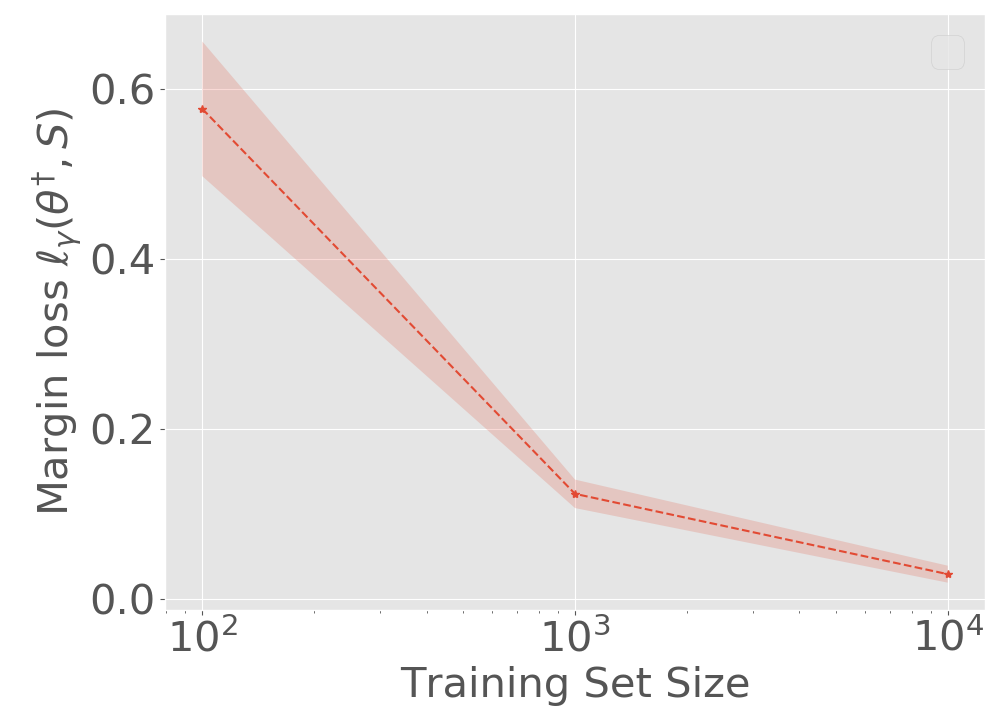}
 } 
  \subfigure[Generalization Bound.]{
 \includegraphics[width = 0.28 \textwidth, height =0.20\textwidth]{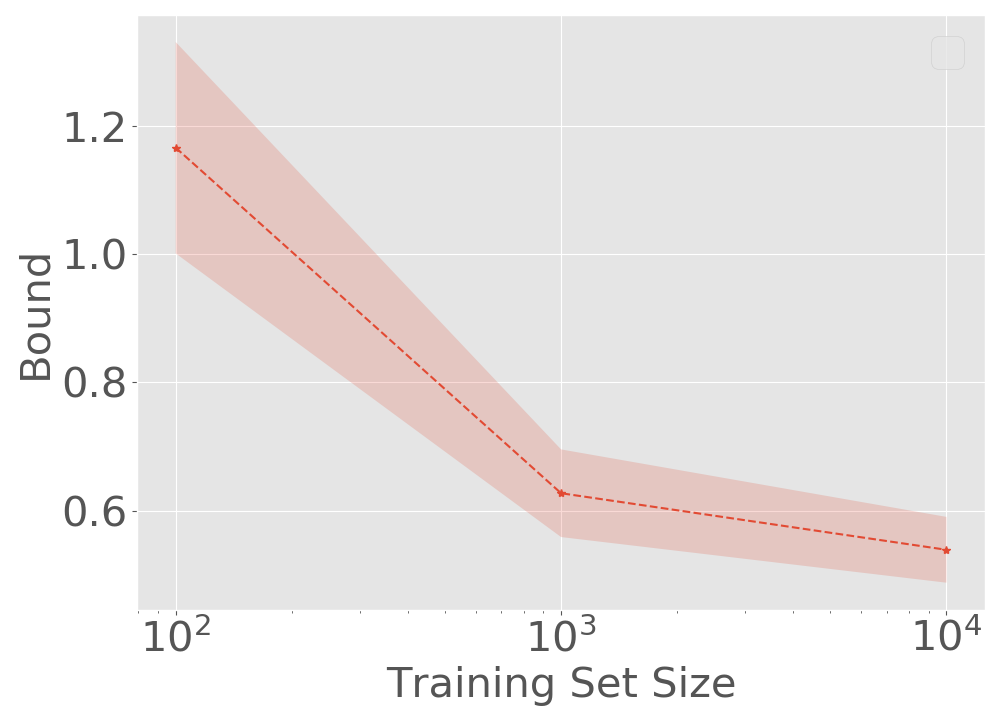}
 } 
\caption[]{Results for ReLU-nets with depth = 8, width =256, total 1,247,744 parameters, trained on CIFAR-10 (batch size = 16) with increasing training set size $n$ (5 runs for each) from 100 to 10,000.  
(a) test set error rate; (b) diagonal elements (mean) of $\tilde \cH_{l,\phi}^{\theta^{\dagger}}$; (c) effective curvature; (d) $L_2$ norm of $\theta^{\dagger}$; (e) margin loss; (f)  generalization bound. The bound and all its components decrease with increase in $n$ from 100 to 10,000.
}
\label{fig:main_cifar_sample_d8_l256_bs16}
\end{figure*}

\noindent \textbf{Additional Results.} Figure \ref{fig:main_mnist_sample_d8_l256} and \ref{fig:main_mnist_sample_d8_l128_bs16} presents additional results for ReLU-nets with depth 8 and batch size 16 for MNIST. Figure \ref{fig:main_mnist_sample_d8_l256} shows that the behavior observed in Figure \ref{fig:main_mnist_sample} also holds for different depths and widths (more results are presented in the Appendix). Figure  \ref{fig:main_mnist_sample_d8_l128_bs16} shows that the bound also holds for micro-batch training (batch size = 16), i.e., the generalization bound as well as the components (effective curvature, $L_2$ norm, margin loss) decreases as training set size increases. Figure \ref{fig:main_cifar_sample_d8_l256_bs16} and \ref{fig:main_cifar_sample_d8_l256} present the results for CIFAR-10 which considers ReLU-nets with depth 8 and batch size 16. They demonstrate that the bound also holds for CIFAR-10 with different depth and width as well as 
micro-batch training. 



\subsection{Additional Results}
\label{exp:add_result}

\begin{figure*}[t] 
\centering
\vspace*{-2mm}
 \subfigure[Norms]{
 \includegraphics[width = 0.35 \textwidth]{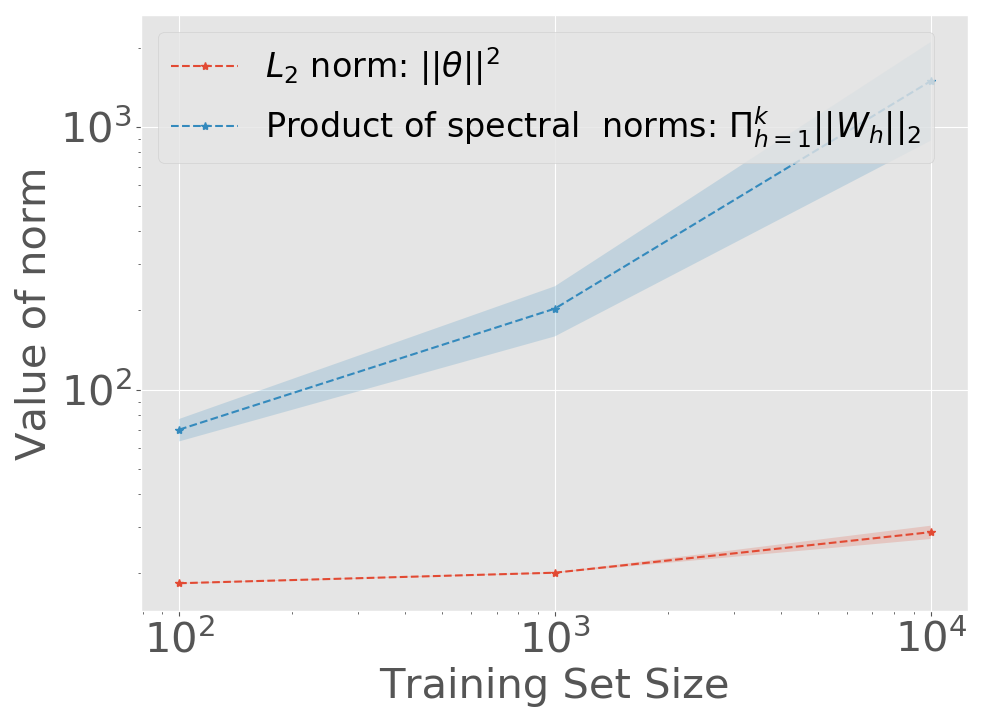}
 } 
  \subfigure[Norms scaled by sample]{
 \includegraphics[width = 0.35 \textwidth]{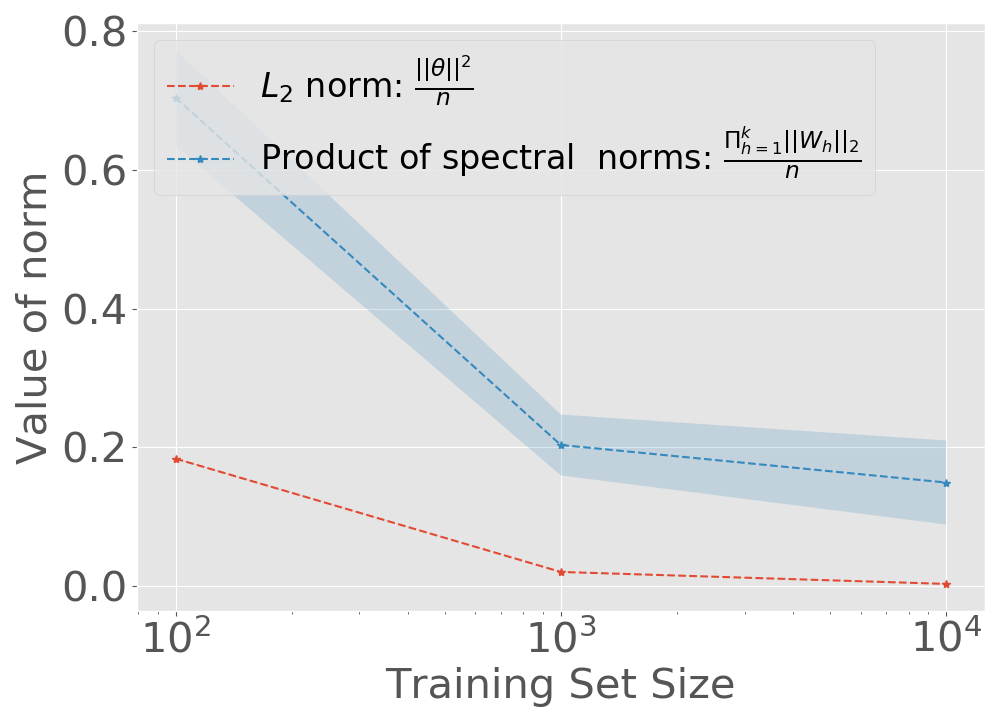}
 } 
 \subfigure[Norm ]{
 \includegraphics[width = 0.35 \textwidth]{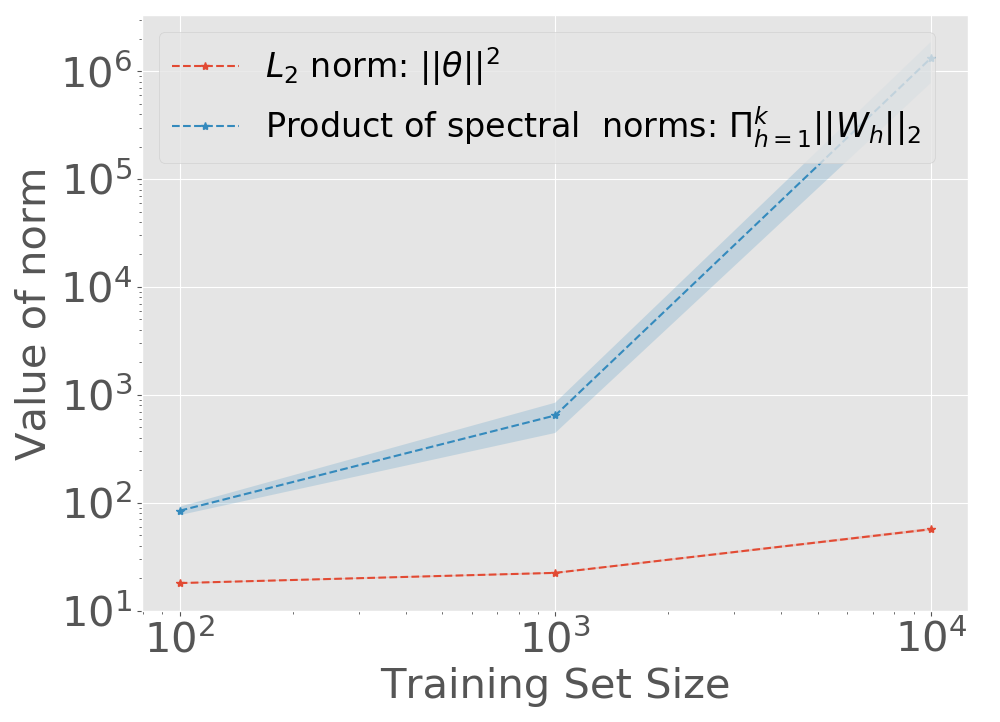}
 } 
 \subfigure[Norms scaled by sample]{
 \includegraphics[width = 0.35 \textwidth]{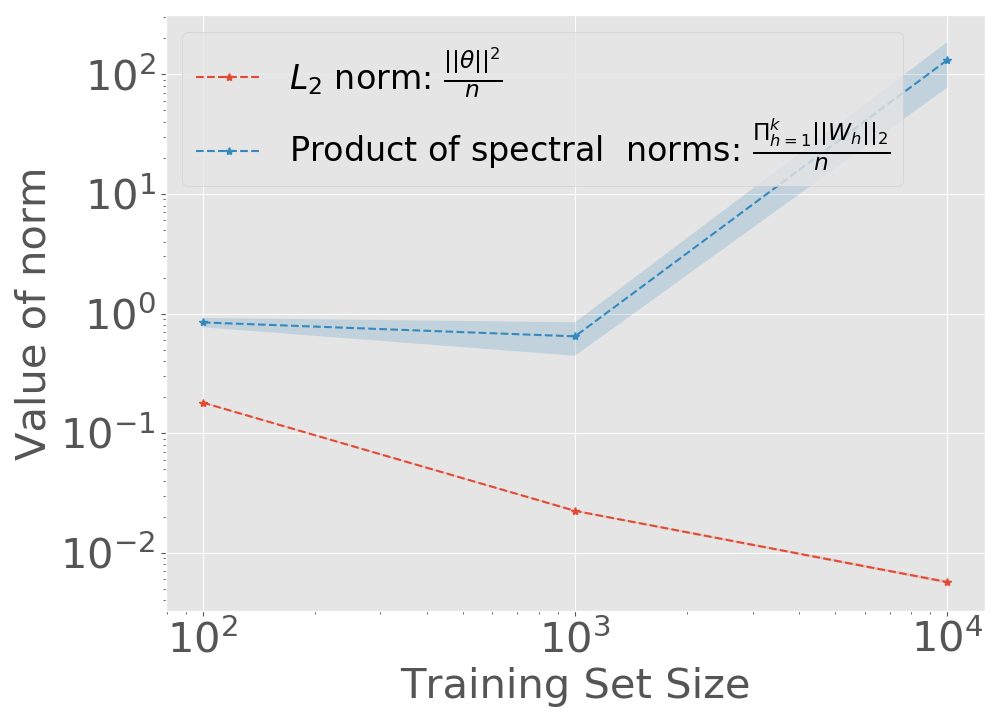}
 }
 \caption[]{$L_2$ norm and product of spectral norms for ReLU-nets with depth = 4, width =128, trained on MNIST with batch size 128 (a-b) and 16 (c-d). (a) and (c) present the $L_2$ norm and product of spectral norms; (b) and (d) present the $L_2$ norm and product of spectral norms averaged by sample size. Product of spectral norms grows much faster than $L_2$ norm with training sample increases.
}
\label{fig:main_mnist_norms}
\end{figure*}

\begin{figure*}[!t] 
\vspace*{-2mm}
\centering
 \subfigure[Norms]{
 \includegraphics[width = 0.35 \textwidth]{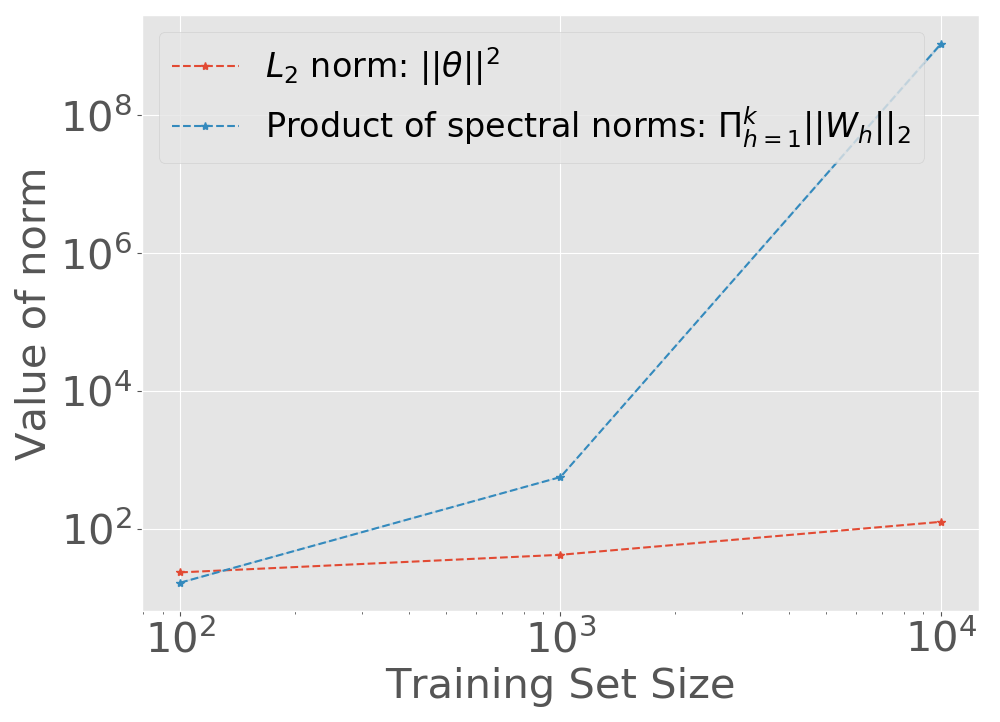}
 } 
  \subfigure[Norms scaled by sample]{
 \includegraphics[width = 0.35 \textwidth]{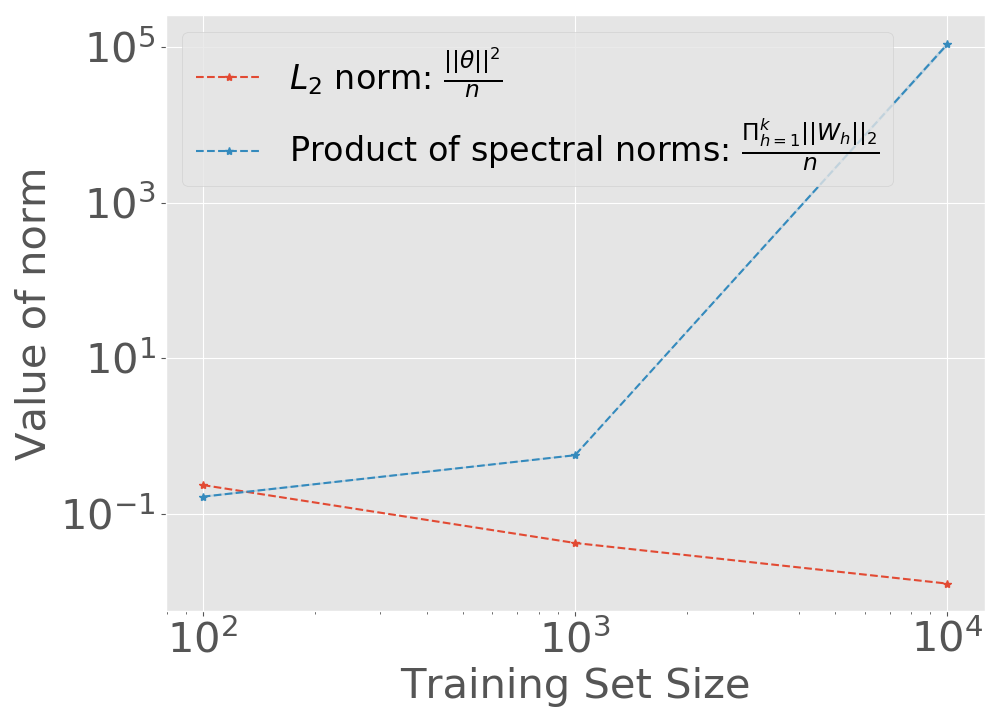}
 } 
 \subfigure[Norm ]{
 \includegraphics[width = 0.35 \textwidth]{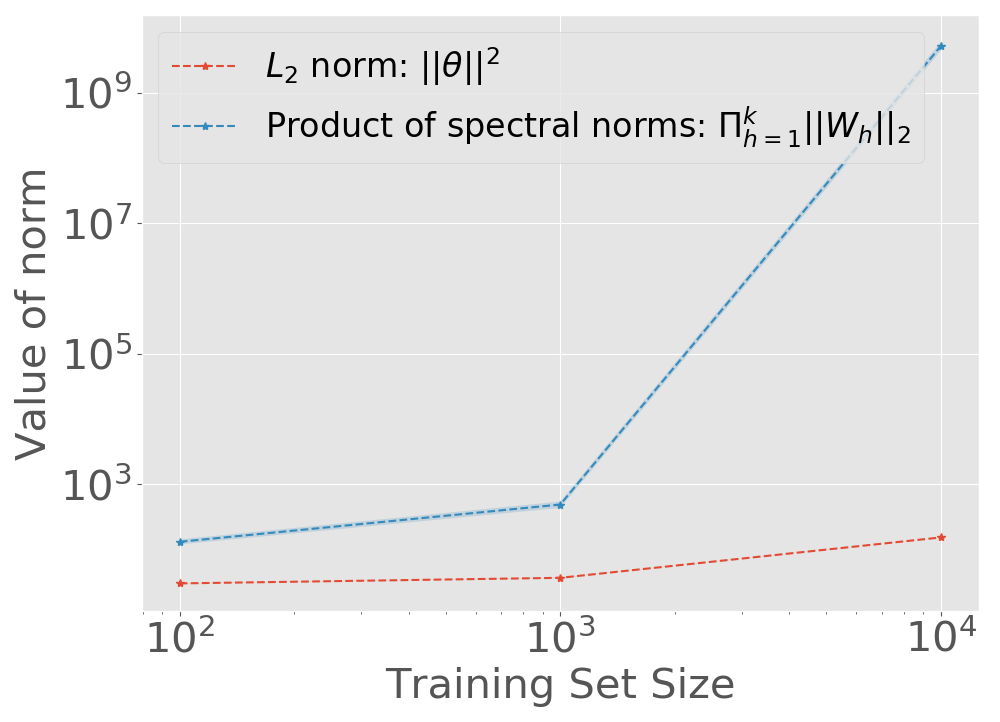}
 } 
 \subfigure[Norms scaled by sample]{
 \includegraphics[width = 0.35 \textwidth]{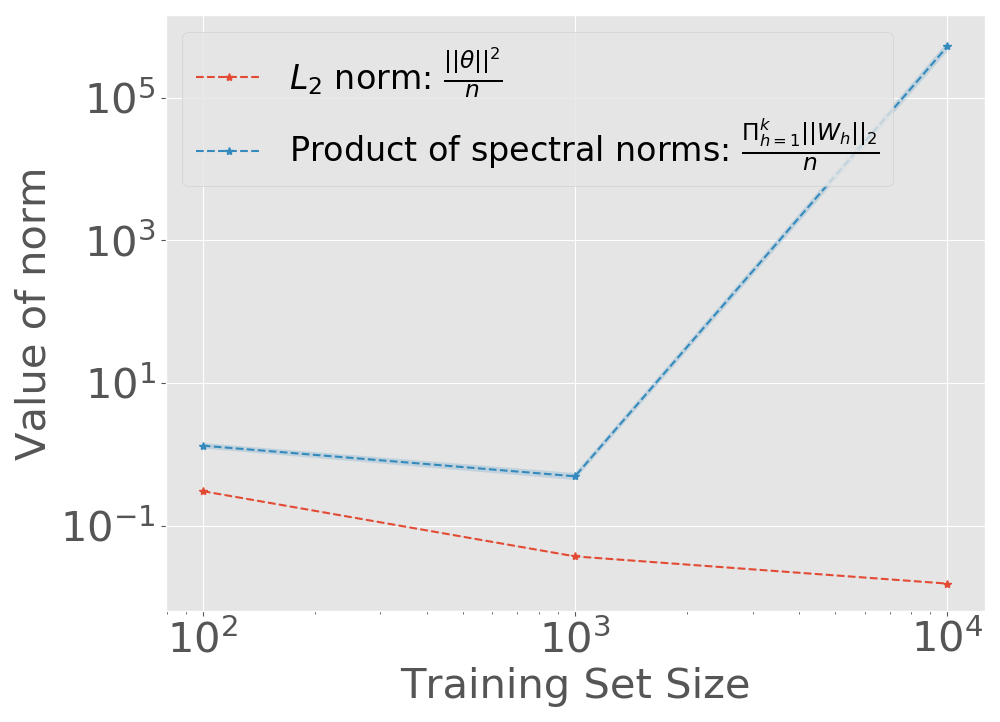}
 }
 \caption[]{$L_2$ norm and product of spectral norms for ReLU-nets with depth = 4, width =256, trained on CIFAR-10 with batch size 128 (a-b) and 16 (c-d). (a) and (c) present the $L_2$ norm and product of spectral norms; (b) and (d) present the $L_2$ norm and product of spectral norms averaged by sample size. Sample averaged  $L_2$ norm decreases as training sample increases. But sample averaged product of spectral norms increases as training sample increases. 
}
\label{fig:main_cifar_norms}
\end{figure*}

\noindent \textbf{Spectral Norm and $L_2$ Norm.} We now take a closer look at the relative behavior of the product of spectral norms often used in existing bounds and the $L_2$ norm in our bound. Figure \ref{fig:main_mnist_norms} (a-b) present the results for MNIST with mini-batch training, i.e., batch size = 128 and (c-d) present the results for MNIST with micro-batch training, i.e., batch size = 16. We observe in Figure \ref{fig:main_mnist_norms}(a) and (c) that both  quantities grow with training sample size $n$,  but the $L_2$ norm (red line) grows far slower than the product of the spectral norms (blue line).  Figure \ref{fig:main_mnist_norms}(b) and (d) shows the same quantities but divided by the number of samples. Note that in (a) both seem to decrease with increase in $n$, with $L_2$ norm having a tiny edge at higher $n$.  
Figure \ref{fig:main_cifar_norms} shows both the quantities for CIFAR-10 which also considers mini-batch training (a-b) and micro-batch training (c-d). Figure \ref{fig:main_cifar_norms} (a) and (d) shows that for both setting, product of spectral norm grows much faster than $L_2$ norm. Figure \ref{fig:main_cifar_norms}(d) shows the same quantities divided by the number of samples. The product of the spectral norms scaled by $n$ increases with sample increases whereas the scaled $L_2$ norm keeps decreasing. 

\begin{figure*}[t] 
\centering
 \subfigure[Bound with different $\sigma^2$]{
 \includegraphics[width = 0.35 \textwidth]{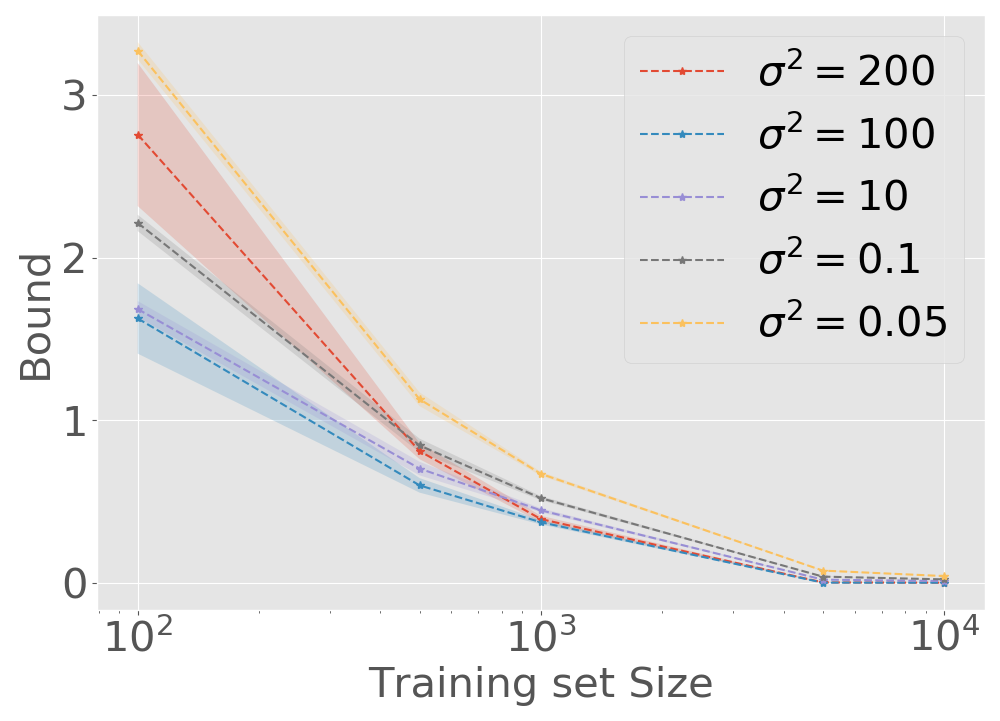}
 } 
\vspace{-2mm}
\subfigure[Bound with different $\sigma^2$]{
 \includegraphics[width = 0.35 \textwidth]{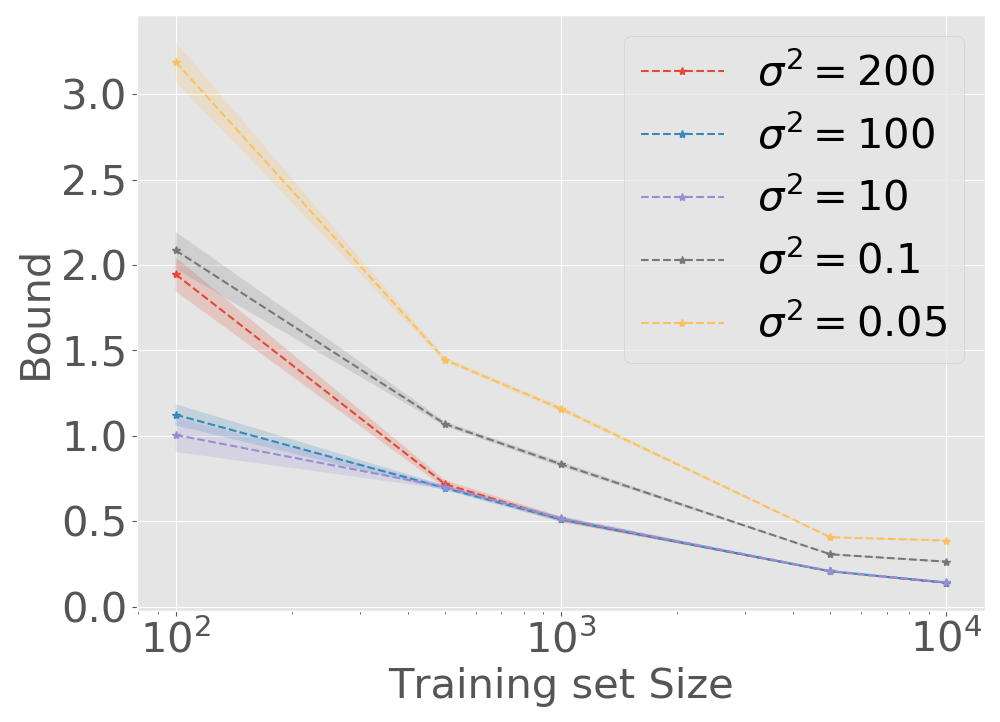}
 } 
\caption[]{Generalization bound with different $\sigma^2$. (a) Generalization bound for ReLU network with depth =4 and width = 256, trained on MNIST with batch size 128; (b) Generalization bound for ReLU network with depth =4 and width = 256, trained on CIFAR-10 with batch size 128. Optimizing the covariance $\sigma$ of the PAC-Bayes prior distribution can lead to a sharper bound.}
\label{fig:main_bound_sigma}
\end{figure*}

\noindent {\bf Optimal $\sigma$.} Note that the choice of variance $\sigma^2$ of the prior distribution also playa a role in the generalization bound: $\left(\sum_{\ell=1}^{\tilde p} \ln \frac{\tilde{\nu}_{\ell}^{2}(\ell)}{1 / \sigma^{2}}+\frac{\left\|\theta^{\dagger}-\theta_{0}\right\|^{2}}{\sigma^{2}}\right)$. The dependence on the prior covariance in the two terms  illustrates a trade-off, i.e., a large $\sigma$ diminishes the dependence on $\left\|\theta^{\dagger}-\theta_{0}\right\|^{2}$, but increases the dependence on the effective curvature , and vice versa. To illustrate how the value of $\sigma$ affects the bound, we choose $\sigma^2 \in \{0.05, 0.1, 10, 100, 200\}$, and present the corresponding bound for MNIST in Figure \ref{fig:main_bound_sigma}~(a) and bound for CIFAR-10 in Figure \ref{fig:main_bound_sigma}~(b). It shows that the optimal value of $\sigma^2$ may locate in $(100, 10)$. This observation suggests that optimizing the covariance $\sigma$ of the PAC-Bayes prior distribution, which is data-independent can lead to a sharper bound. We consider such analysis as our future work.

\begin{figure*}[!t] 
\centering
 \subfigure[ $\ell_\gamma(\theta^{\dagger}, S), \gamma = 14$]{
 \includegraphics[width = 0.28 \textwidth, height = 0.25\textwidth]{margin_dist_cifar10_d4_l256_bs128.png}
 }
 \subfigure[ $\ell_\gamma(\theta^{\dagger}, S), \gamma = 16$]{
 \includegraphics[width = 0.28 \textwidth, height = 0.25\textwidth]{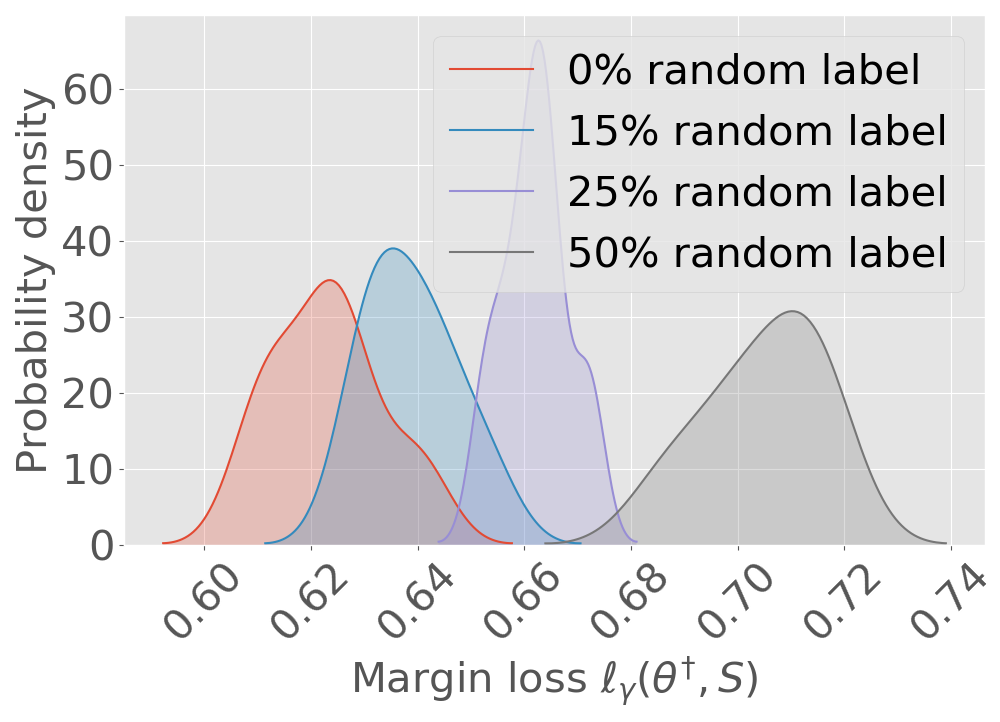}
 } 
 \subfigure[ $\ell_\gamma(\theta^{\dagger}, S), \gamma = 18$]{
 \includegraphics[width = 0.28 \textwidth, height = 0.25\textwidth]{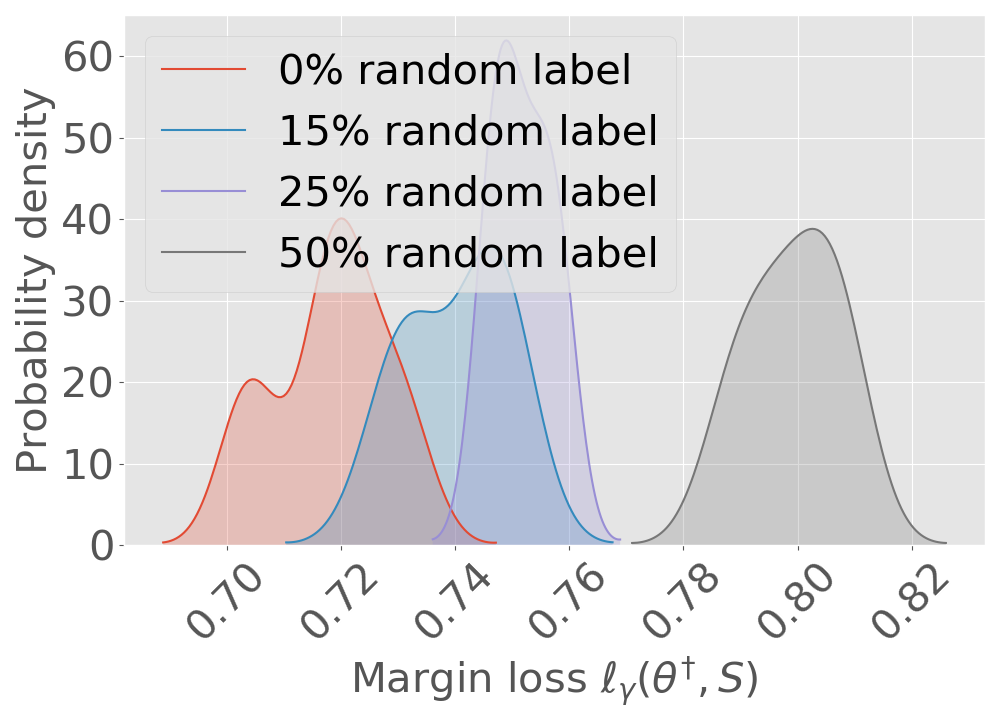}
 } 
\vspace{-2mm} 
\caption[]{Margin loss distribution with different $\gamma$. ReLU-nets with depth = 4, width =128, total 1,052,426 parameters, trained on CIFAR-10 with increase in number of random labels (10 runs each) from 0\% to 50\%.}
\label{fig:main_cifar_mar_gamma}
\end{figure*}

\begin{figure*}[!t] 
\centering
 \subfigure[ $\ell_\gamma(\theta^{\dagger}, S), \gamma = 7$]{
 \includegraphics[width = 0.28 \textwidth, height = 0.25\textwidth]{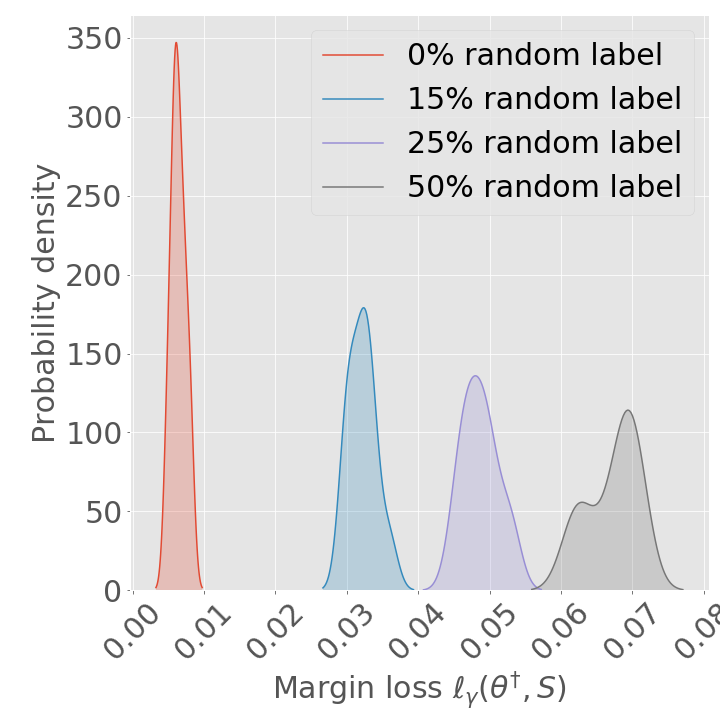}
 }
 \subfigure[ $\ell_\gamma(\theta^{\dagger}, S), \gamma = 10$]{
 \includegraphics[width = 0.28 \textwidth, height = 0.25\textwidth]{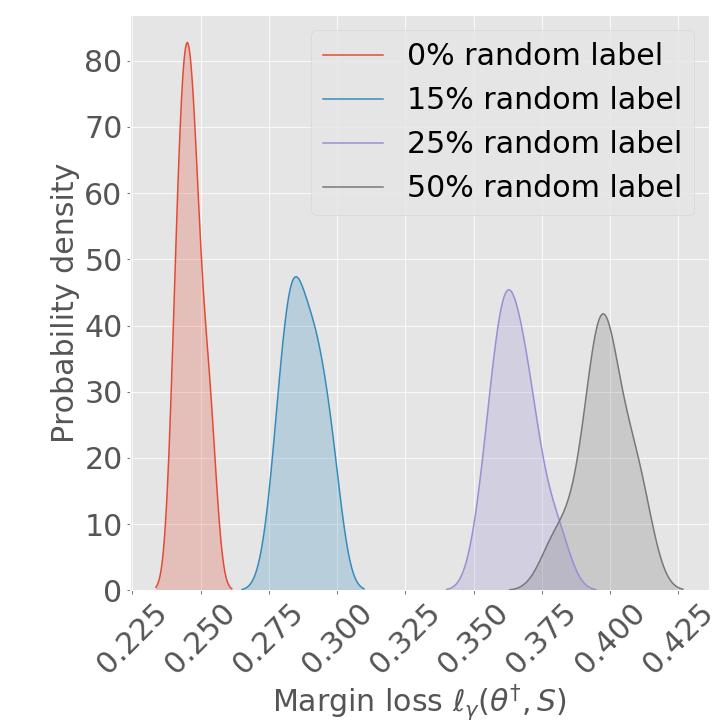}
 } 
 \subfigure[$\ell_\gamma(\theta^{\dagger}, S), \gamma = 15$]{
 \includegraphics[width = 0.28\textwidth, height = 0.25\textwidth]{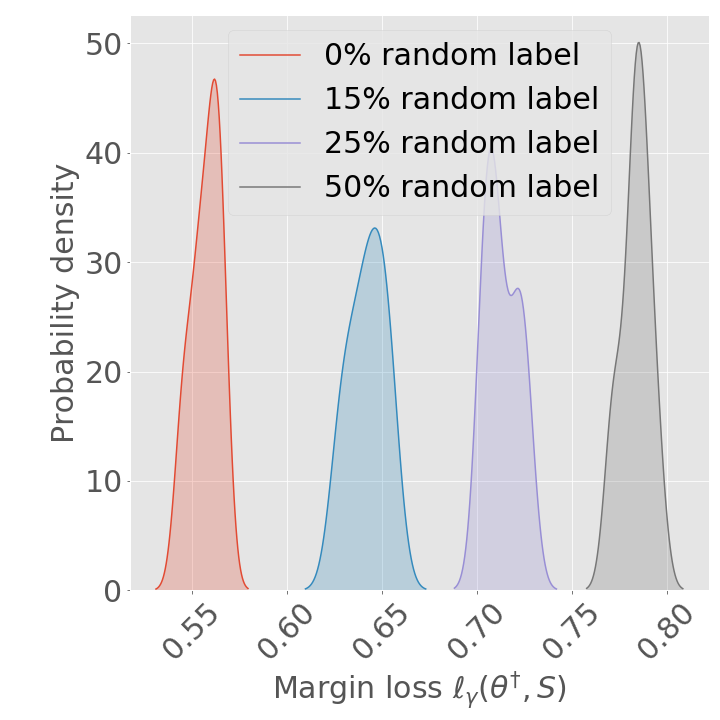}
 } 
\vspace{-2mm}
\caption[]{Margin loss distribution with different $\gamma$. ReLU-nets with depth = 4, width =128, total 167,818 parameters, trained on MNIST with increase in number of random labels (30 runs each) from 0\% to 50\%.
}
\label{fig:main_mnist_mar_gamma}
\end{figure*}

\noindent {\bf Margin Loss and  Margin $\gamma$.} Note that the empirical margin loss plays a role in the generalization bound. The choice of the margin $\gamma$ affects the empirical margin loss $\ell_\gamma(\theta^{\dagger}, S)$ and the terms $d_{\eta} \exp \left(-c \gamma\right)$ in our bound. Increasing the value of $\gamma$ will increase the empirical margin loss $\ell_\gamma(\theta^{\dagger}, S)$, but the term  $d_{\eta} \exp \left(-c\gamma\right)$ will decrease. Figure \ref{fig:main_mnist_mar_gamma} and \ref{fig:main_cifar_mar_gamma} illustrates how the margin loss changes $\ell_\gamma(\theta^{\dagger}, S)$  with different choice of $\gamma$
for CIFAR-10 and MNIST respectively. 
We can see that with the margin $\gamma$ increases, the margin loss distribution for MNIST (Figure \ref{fig:main_mnist_mar_gamma}) and CIFAR-10 (Figure \ref{fig:main_cifar_mar_gamma}) shifts to a higher value, implying the increases in the margin loss term.

\section{Conclusions}
\label{sec:conc}

Explaining the generalization of deterministic non-smooth deep nets  has remained challenging.
Recent work has shown that most existing bounds which relies on bounding the Lipschitz constant of such deep nets are not quantitatively tight, and often display unusual empirical behavior~\citep{nako19b}. 
In this paper, we have presented new bounds for non-smooth deep nets based on a de-randomization argument on PAC-Bayes. Our analysis uses the self-evident but tricky to use fact the ReLU-nets and related deep nets realize as linear deep nets for any given input. The bound demonstrates a trade-off between effective curvature (`flatness') of the predictor, and $L_2$ norm of the learned weights. The bounds display correct qualitative behavior with change in training set size and random labels. The empirical results look promising, are quantitatively meaningful and non-vacuous even without hyper-parameter tuning, and leaves room for future work on quantitative sharpening of the bounds.





\section*{Acknowledgement}
The research was supported by NSF grants IIS-1908104, OAC-1934634, and IIS-1563950. We would like to thank the Minnesota Super-computing Institute (MSI) for providing computational resources and support. 

\newpage
\bibliographystyle{abbrvnat}
\bibliography{reference}

\begin{thebibliography}{91}
\providecommand{\natexlab}[1]{#1}
\providecommand{\url}[1]{\texttt{#1}}
\expandafter\ifx\csname urlstyle\endcsname\relax
  \providecommand{\doi}[1]{doi: #1}\else
  \providecommand{\doi}{doi: \begingroup \urlstyle{rm}\Url}\fi

\bibitem[Alquier and Biau(2013)]{alquier2013sparse}
P.~Alquier and G.~Biau.
\newblock Sparse single-index model.
\newblock \emph{Journal of Machine Learning Research}, 14\penalty0
  (Jan):\penalty0 243--280, 2013.

\bibitem[Alquier et~al.(2016)Alquier, Ridgway, and
  Chopin]{alquier2016properties}
P.~Alquier, J.~Ridgway, and N.~Chopin.
\newblock On the properties of variational approximations of gibbs posteriors.
\newblock \emph{Journal of Machine Learning Research}, 17\penalty0
  (1):\penalty0 8374--8414, 2016.

\bibitem[Arora et~al.(2018)Arora, Ge, Neyshabur, and
  Zhang]{arora_stronger_2018}
S.~Arora, R.~Ge, B.~Neyshabur, and Y.~Zhang.
\newblock Stronger generalization bounds for deep nets via a compression
  approach.
\newblock In \emph{International Conference on Machine Learning}, pages
  254--263, 2018.

\bibitem[Arora et~al.(2019{\natexlab{a}})Arora, Cohen, Golowich, and
  Hu]{arco18}
S.~Arora, N.~Cohen, N.~Golowich, and W.~Hu.
\newblock A convergence analysis of gradient descent for deep linear neural
  networks.
\newblock In \emph{International Conference on Learning Representations},
  2019{\natexlab{a}}.

\bibitem[Arora et~al.(2019{\natexlab{b}})Arora, Du, Hu, Li, and Wang]{ardu19}
S.~Arora, S.~Du, W.~Hu, Z.~Li, and R.~Wang.
\newblock Fine-grained analysis of optimization and generalization for
  overparameterized two-layer neural networks.
\newblock In \emph{International Conference on Machine Learning}, pages
  322--332, 2019{\natexlab{b}}.

\bibitem[Banerjee(2006)]{bane06}
A.~Banerjee.
\newblock On {Bayesian} bounds.
\newblock In \emph{International Conference on Machine Learning}, pages 81--88,
  2006.

\bibitem[Bartlett and Mendelson(2002)]{bame02}
P.~L. Bartlett and S.~Mendelson.
\newblock Rademacher and {Gaussian} complexities: Risk bounds and structural
  results.
\newblock \emph{Journal of Machine Learning Research}, 3\penalty0
  (Nov):\penalty0 463--482, 2002.

\bibitem[Bartlett et~al.(1999)Bartlett, Maiorov, and Meir]{bama99}
P.~L. Bartlett, V.~Maiorov, and R.~Meir.
\newblock Almost linear {VC} dimension bounds for piecewise polynomial
  networks.
\newblock In \emph{Advances in Neural Information Processing Systems}, pages
  190--196, 1999.

\bibitem[Bartlett et~al.(2005)Bartlett, Bousquet, Mendelson,
  et~al.]{bartlett2005local}
P.~L. Bartlett, O.~Bousquet, S.~Mendelson, et~al.
\newblock Local {Rademacher} complexities.
\newblock \emph{Annals of Statistics}, 33\penalty0 (4):\penalty0 1497--1537,
  2005.

\bibitem[Bartlett et~al.(2017)Bartlett, Foster, and Telgarsky]{bafo17}
P.~L. Bartlett, D.~J. Foster, and M.~J. Telgarsky.
\newblock Spectrally-normalized margin bounds for neural networks.
\newblock In \emph{Advances in Neural Information Processing Systems}, pages
  6240--6249, 2017.

\bibitem[Bartlett et~al.(2019)Bartlett, Harvey, Liaw, and
  Mehrabian]{bartlett_nearly-tight_2019}
P.~L. Bartlett, N.~Harvey, C.~Liaw, and A.~Mehrabian.
\newblock Nearly-tight {VC}-dimension and {Pseudodimension} {Bounds} for
  {Piecewise} {Linear} {Neural} {Networks}.
\newblock page~17, 2019.

\bibitem[B{\'e}gin et~al.(2014)B{\'e}gin, Germain, Laviolette, and
  Roy]{begin2014pac}
L.~B{\'e}gin, P.~Germain, F.~Laviolette, and J.-F. Roy.
\newblock {PAC-Bayesian} theory for transductive learning.
\newblock In \emph{Artificial Intelligence and Statistics}, pages 105--113,
  2014.

\bibitem[B{\'e}gin et~al.(2016)B{\'e}gin, Germain, Laviolette, and
  Roy]{begin2016pac}
L.~B{\'e}gin, P.~Germain, F.~Laviolette, and J.-F. Roy.
\newblock {PAC-Bayesian} bounds based on the r{\'e}nyi divergence.
\newblock In \emph{Artificial Intelligence and Statistics}, pages 435--444,
  2016.

\bibitem[Benedek and Itai(1988)]{benedek_nonuniform_1988}
G.~Benedek and A.~Itai.
\newblock Nonuniform learnability.
\newblock In \emph{International Colloquium on Automata, Languages and
  Programming}, 1988.

\bibitem[Benedek and Itai(1994)]{benedek_nonuniform_1994}
G.~Benedek and A.~Itai.
\newblock Nonuniform learnability.
\newblock \emph{Journal of Computer and System Sciences}, 48\penalty0 (2),
  1994.

\bibitem[Blumer et~al.(1989)Blumer, Ehrenfeucht, Haussler, and
  Warmuth]{blumer_learnability_1989}
A.~Blumer, A.~Ehrenfeucht, D.~Haussler, and M.~K. Warmuth.
\newblock Learnability and the {Vapnik}-{Chervonenkis} dimension.
\newblock \emph{Journal of the ACM}, 36\penalty0 (4):\penalty0 929--965, Oct.
  1989.
\newblock ISSN 00045411.
\newblock \doi{10.1145/76359.76371}.
\newblock URL \url{http://portal.acm.org/citation.cfm?doid=76359.76371}.

\bibitem[Boucheron et~al.(2005)Boucheron, Bousquet, and
  Lugosi]{boucheron2005theory}
S.~Boucheron, O.~Bousquet, and G.~Lugosi.
\newblock Theory of classification: A survey of some recent advances.
\newblock \emph{ESAIM: probability and statistics}, 9:\penalty0 323--375, 2005.

\bibitem[Boucheron et~al.(2013)Boucheron, Lugosi, and
  Massart]{boucheron_concentration_2013}
S.~Boucheron, G.~Lugosi, and P.~Massart.
\newblock \emph{Concentration {Inequalities}: {A} {Nonasymptotic} {Theory} of
  {Independence}}.
\newblock Oxford University Press, Feb. 2013.

\bibitem[Bousquet(2002)]{bousquet2002concentration}
O.~Bousquet.
\newblock \emph{Concentration inequalities and empirical processes theory
  applied to the analysis of learning algorithms}.
\newblock PhD thesis, {\'E}cole Polytechnique: Department of Applied
  Mathematics Paris, France, 2002.

\bibitem[Bousquet and Elisseeff(2002)]{boel02}
O.~Bousquet and A.~Elisseeff.
\newblock Stability and generalization.
\newblock \emph{Journal of Machine Learning Research}, 2\penalty0
  (Mar):\penalty0 499--526, 2002.

\bibitem[Cantelli(1933)]{cantelli1933sulla}
F.~P. Cantelli.
\newblock Sulla determinazione empirica delle leggi di probabilita.
\newblock \emph{Giorn. Ist. Ital. Attuari}, 4\penalty0 (421-424), 1933.

\bibitem[Cao and Gu(2019)]{cagu19a}
Y.~Cao and Q.~Gu.
\newblock A generalization theory of gradient descent for learning
  over-parameterized deep {ReLU} networks.
\newblock \emph{arXiv preprint arXiv:1902.01384}, 2019.

\bibitem[Catoni(2007)]{cato07}
O.~Catoni.
\newblock {PAC-Bayesian} supervised classification: the thermodynamics of
  statistical learning.
\newblock \emph{Monograph Series of the Institute of Mathematical Statistics},
  2007.

\bibitem[Chaudhari et~al.(2019)Chaudhari, Choromanska, Soatto, LeCun, Baldassi,
  Borgs, Chayes, Sagun, and Zecchina]{chaudhari2019entropy}
P.~Chaudhari, A.~Choromanska, S.~Soatto, Y.~LeCun, C.~Baldassi, C.~Borgs,
  J.~Chayes, L.~Sagun, and R.~Zecchina.
\newblock Entropy-{SGD}: Biasing gradient descent into wide valleys.
\newblock \emph{Journal of Statistical Mechanics: Theory and Experiment},
  2019\penalty0 (12):\penalty0 124018, 2019.

\bibitem[Denker and LeCun(1991)]{denker1991transforming}
J.~S. Denker and Y.~LeCun.
\newblock Transforming neural-net output levels to probability distributions.
\newblock In \emph{Advances in neural information processing systems}, pages
  853--859, 1991.

\bibitem[Dinh et~al.(2017)Dinh, Pascanu, Bengio, and Bengio]{dipb17}
L.~Dinh, R.~Pascanu, S.~Bengio, and Y.~Bengio.
\newblock Sharp minima can generalize for deep nets.
\newblock In \emph{International Conference on Machine Learning}, pages
  1019--1028, 2017.

\bibitem[Du et~al.(2018)Du, Wang, Zhai, Balakrishnan, Salakhutdinov, and
  Singh]{du2018many}
S.~S. Du, Y.~Wang, X.~Zhai, S.~Balakrishnan, R.~R. Salakhutdinov, and A.~Singh.
\newblock How many samples are needed to estimate a convolutional neural
  network?
\newblock In \emph{Advances in Neural Information Processing Systems}, pages
  373--383, 2018.

\bibitem[Du et~al.(2019)Du, Zhai, Poczos, and Singh]{duzh2019}
S.~S. Du, X.~Zhai, B.~Poczos, and A.~Singh.
\newblock Gradient descent provably optimizes over-parameterized neural
  networks.
\newblock In \emph{International Conference on Learning Representations}, 2019.

\bibitem[Dziugaite and Roy(2018{\natexlab{a}})]{dziugaite2018entropy}
G.~K. Dziugaite and D.~Roy.
\newblock Entropy-{SGD} optimizes the prior of a {PAC-Bayes} bound:
  Generalization properties of {Entropy-SGD} and data-dependent priors.
\newblock In \emph{International Conference on Machine Learning}, pages
  1377--1386, 2018{\natexlab{a}}.

\bibitem[Dziugaite and Roy(2017)]{dzda17}
G.~K. Dziugaite and D.~M. Roy.
\newblock Computing nonvacuous generalization bounds for deep (stochastic)
  neural networks with many more parameters than training data.
\newblock In \emph{Uncertainty in Artificial Intelligence}, 2017.

\bibitem[Dziugaite and Roy(2018{\natexlab{b}})]{dzro18}
G.~K. Dziugaite and D.~M. Roy.
\newblock Data-dependent {PAC-Bayes} priors via differential privacy.
\newblock In \emph{Advances in Neural Information Processing Systems},
  2018{\natexlab{b}}.

\bibitem[Frei et~al.(2019)Frei, Cao, and Gu]{frei2019algorithm}
S.~Frei, Y.~Cao, and Q.~Gu.
\newblock Algorithm-dependent generalization bounds for overparameterized deep
  residual networks.
\newblock In \emph{Advances in Neural Information Processing Systems}, pages
  14797--14807, 2019.

\bibitem[Germain et~al.(2009)Germain, Lacasse, Laviolette, and
  Marchand]{germain2009pac}
P.~Germain, A.~Lacasse, F.~Laviolette, and M.~Marchand.
\newblock {PAC-Bayesian} learning of linear classifiers.
\newblock In \emph{International Conference on Machine Learning}, pages
  353--360, 2009.

\bibitem[Ghorbani et~al.(2019)Ghorbani, Krishnan, and
  Xiao]{ghorbani_investigation_2019}
B.~Ghorbani, S.~Krishnan, and Y.~Xiao.
\newblock An {Investigation} into {Neural} {Net} {Optimization} via {Hessian}
  {Eigenvalue} {Density}.
\newblock \emph{arXiv:1901.10159 [cs, stat]}, Jan. 2019.
\newblock URL \url{http://arxiv.org/abs/1901.10159}.
\newblock arXiv: 1901.10159.

\bibitem[Glivenko(1933)]{glivenko1933sulla}
V.~Glivenko.
\newblock Sulla determinazione empirica delle leggi di probabilita.
\newblock \emph{Gion. Ist. Ital. Attauri.}, 4:\penalty0 92--99, 1933.

\bibitem[Goldberg and Jerrum(1995)]{goldberg1995bounding}
P.~W. Goldberg and M.~R. Jerrum.
\newblock Bounding the vapnik-chervonenkis dimension of concept classes
  parameterized by real numbers.
\newblock \emph{Machine Learning}, 18\penalty0 (2-3):\penalty0 131--148, 1995.

\bibitem[Golowich et~al.(2018)Golowich, Rakhlin, and Shamir]{gora18}
N.~Golowich, A.~Rakhlin, and O.~Shamir.
\newblock Size-independent sample complexity of neural networks.
\newblock In \emph{Conference on Learning Theory}, pages 297--299, 2018.

\bibitem[Guedj and Alquier(2013)]{guedj2013pac}
B.~Guedj and P.~Alquier.
\newblock {PAC-Bayesian} estimation and prediction in sparse additive models.
\newblock \emph{Electronic Journal of Statistics}, 7:\penalty0 264--291, 2013.

\bibitem[Gunasekar et~al.(2018)Gunasekar, Lee, Soudry, and
  Srebro]{guls2018implicit}
S.~Gunasekar, J.~D. Lee, D.~Soudry, and N.~Srebro.
\newblock Implicit bias of gradient descent on linear convolutional networks.
\newblock In \emph{Advances in Neural Information Processing Systems}, pages
  9461--9471, 2018.

\bibitem[Hardt et~al.(2016)Hardt, Recht, and Singer]{hare16}
M.~Hardt, B.~Recht, and Y.~Singer.
\newblock Train faster, generalize better: Stability of stochastic gradient
  descent.
\newblock In \emph{International Conference on Machine Learning}, pages
  1225--1234, 2016.

\bibitem[Hochreiter and Schmidhuber(1997)]{hosc97b}
S.~Hochreiter and J.~Schmidhuber.
\newblock Flat minima.
\newblock \emph{Neural Computation}, 9\penalty0 (1):\penalty0 1--42, 1997.

\bibitem[Hsu et~al.(2012)Hsu, Kakade, Zhang, et~al.]{hska12}
D.~Hsu, S.~Kakade, T.~Zhang, et~al.
\newblock A tail inequality for quadratic forms of subgaussian random vectors.
\newblock \emph{Electronic Communications in Probability}, 17, 2012.

\bibitem[Keskar et~al.(2017)Keskar, Mudigere, Nocedal, Smelyanskiy, and
  Tang]{kemn17}
N.~S. Keskar, D.~Mudigere, J.~Nocedal, M.~Smelyanskiy, and P.~T.~P. Tang.
\newblock On large-batch training for deep learning: Generalization gap and
  sharp minima.
\newblock In \emph{International Conference on Learning Representations}, 2017.

\bibitem[Kleeman(2011)]{kl2011}
R.~Kleeman.
\newblock Information theory and dynamical system predictability.
\newblock \emph{Entropy}, 13\penalty0 (3):\penalty0 612--649, 2011.

\bibitem[Koltchinskii and Panchenko(2000)]{kopa00}
V.~Koltchinskii and D.~Panchenko.
\newblock Rademacher processes and bounding the risk of function learning.
\newblock \emph{High Dimensional Probability II}, pages 443--457, 2000.

\bibitem[Lacasse et~al.(2007)Lacasse, Laviolette, Marchand, Germain, and
  Usunier]{lacasse2007pac}
A.~Lacasse, F.~Laviolette, M.~Marchand, P.~Germain, and N.~Usunier.
\newblock {PAC-Bayes} bounds for the risk of the majority vote and the variance
  of the gibbs classifier.
\newblock In \emph{Advances in Neural Information Processing Systems}, pages
  769--776, 2007.

\bibitem[Langford(2005)]{langford2005tutorial}
J.~Langford.
\newblock Tutorial on practical prediction theory for classification.
\newblock \emph{Journal of Machine Learning Research}, 6\penalty0
  (Mar):\penalty0 273--306, 2005.

\bibitem[Langford and Caruana(2002)]{langford2002not}
J.~Langford and R.~Caruana.
\newblock {(Not)} bounding the true error.
\newblock In \emph{Advances in Neural Information Processing Systems}, pages
  809--816, 2002.

\bibitem[Langford and Seeger(2001)]{langford2001bounds}
J.~Langford and M.~Seeger.
\newblock Bounds for averaging classifiers.
\newblock \emph{Technical report, Carnegie Mellon, Department of Computer
  Science}, 2001.

\bibitem[Langford and Shawe-Taylor(2003)]{langford2003pac}
J.~Langford and J.~Shawe-Taylor.
\newblock {PAC-Bayes} \& margins.
\newblock In \emph{Advances in Neural Information Processing Systems}, pages
  439--446, 2003.

\bibitem[Laurent and Brecht(2018)]{labr18}
T.~Laurent and J.~Brecht.
\newblock Deep linear networks with arbitrary loss: All local minima are
  global.
\newblock In \emph{International Conference on Machine Learning}, pages
  2902--2907, 2018.

\bibitem[Lee et~al.(2019)Lee, Xiao, Schoenholz, Bahri, Novak, Sohl-Dickstein,
  and Pennington]{lee2019wide}
J.~Lee, L.~Xiao, S.~Schoenholz, Y.~Bahri, R.~Novak, J.~Sohl-Dickstein, and
  J.~Pennington.
\newblock Wide neural networks of any depth evolve as linear models under
  gradient descent.
\newblock In \emph{Advances in Neural Information Processing Systems}, pages
  8572--8583, 2019.

\bibitem[Li et~al.(2018)Li, Lu, Wang, Haupt, and Zhao]{lilu18}
X.~Li, J.~Lu, Z.~Wang, J.~Haupt, and T.~Zhao.
\newblock On tighter generalization bound for deep neural networks: Cnns,
  resnets, and beyond.
\newblock \emph{arXiv preprint arXiv:1806.05159}, 2018.

\bibitem[Li et~al.(2020)Li, Gu, Zhou, Chen, and Banerjee]{ligu19}
X.~Li, Q.~Gu, Y.~Zhou, T.~Chen, and A.~Banerjee.
\newblock Hessian based analysis of {SGD} for deep nets: Dynamics and
  generalization.
\newblock In \emph{Proceedings of the 2020 SIAM International Conference on
  Data Mining}, pages 190--198. SIAM, 2020.

\bibitem[Littlestone and Warmuth(1994)]{liwa1994}
N.~Littlestone and M.~Warmuth.
\newblock The weighted majority algorithm.
\newblock \emph{Information and Computation}, 108\penalty0 (2):\penalty0 212 --
  261, 1994.
\newblock ISSN 0890-5401.

\bibitem[London(2017)]{london2017pac}
B.~London.
\newblock A {PAC-Bayesian} analysis of randomized learning with application to
  stochastic gradient descent.
\newblock In \emph{Advances in Neural Information Processing Systems}, pages
  2931--2940, 2017.

\bibitem[London et~al.(2014)London, Huang, Taskar, and Getoor]{london2014pac}
B.~London, B.~Huang, B.~Taskar, and L.~Getoor.
\newblock {PAC-Bayesian} collective stability.
\newblock In \emph{Artificial Intelligence and Statistics}, pages 585--594,
  2014.

\bibitem[Long and Sedghi(2019)]{lose19}
P.~M. Long and H.~Sedghi.
\newblock Size-free generalization bounds for convolutional neural networks.
\newblock \emph{arXiv preprint arXiv:1905.12600}, 2019.

\bibitem[MacKay(1992)]{mackay1992practical}
D.~J. MacKay.
\newblock A practical bayesian framework for backpropagation networks.
\newblock \emph{Neural computation}, 4\penalty0 (3):\penalty0 448--472, 1992.

\bibitem[McAllester(1999{\natexlab{a}})]{mcda99}
D.~McAllester.
\newblock {PAC-Bayesian} model averaging.
\newblock In \emph{Conference on Learning Theory}, 1999{\natexlab{a}}.

\bibitem[McAllester(2003)]{mcda03}
D.~McAllester.
\newblock Simplified {PAC-Bayesian} margin bounds.
\newblock In \emph{Learning Theory and Kernel Machines}, pages 203--215.
  Springer, 2003.

\bibitem[McAllester(1999{\natexlab{b}})]{mcallester1999some}
D.~A. McAllester.
\newblock Some {PAC-Bayesian} theorems.
\newblock \emph{Machine Learning}, 37\penalty0 (3):\penalty0 355--363,
  1999{\natexlab{b}}.

\bibitem[Mohri et~al.(2018)Mohri, Rostamizadeh, and
  Talwalkar]{mohri_foundations_2018}
M.~Mohri, A.~Rostamizadeh, and A.~Talwalkar.
\newblock \emph{Foundations of machine learning}.
\newblock Adaptive computation and machine learning. The MIT Press, Cambridge,
  Massachusetts, second edition edition, 2018.
\newblock ISBN 978-0-262-03940-6.

\bibitem[Nagarajan and Kolter(2019{\natexlab{a}})]{nako19b}
V.~Nagarajan and J.~Z. Kolter.
\newblock Uniform convergence may be unable to explain generalization in deep
  learning.
\newblock In \emph{Advances in Neural Information Processing Systems}, pages
  11611--11622, 2019{\natexlab{a}}.

\bibitem[Nagarajan and Kolter(2019{\natexlab{b}})]{nako18}
V.~Nagarajan and Z.~Kolter.
\newblock Deterministic {PAC}-bayesian generalization bounds for deep networks
  via generalizing noise-resilience.
\newblock In \emph{International Conference on Learning Representations},
  2019{\natexlab{b}}.

\bibitem[Negrea et~al.(2020)Negrea, Dziugaite, and Roy]{negrea_defense_2019}
J.~Negrea, G.~K. Dziugaite, and D.~M. Roy.
\newblock In defense of uniform convergence: {Generalization} via
  derandomization with an application to interpolating predictors.
\newblock In \emph{International Conference on Machine Learning}, 2020.

\bibitem[Neyshabur et~al.(2015)Neyshabur, Tomioka, and Srebro]{nets15}
B.~Neyshabur, R.~Tomioka, and N.~Srebro.
\newblock Norm-based capacity control in neural networks.
\newblock In \emph{Conference on Learning Theory}, pages 1376--1401, 2015.

\bibitem[Neyshabur et~al.(2017)Neyshabur, Bhojanapalli, McAllester, and
  Srebro]{nebh17}
B.~Neyshabur, S.~Bhojanapalli, D.~McAllester, and N.~Srebro.
\newblock Exploring generalization in deep learning.
\newblock In \emph{Advances in Neural Information Processing Systems}, pages
  5947--5956, 2017.

\bibitem[Neyshabur et~al.(2018)Neyshabur, Bhojanapalli, and Srebro]{nebh18}
B.~Neyshabur, S.~Bhojanapalli, and N.~Srebro.
\newblock A {PAC}-{Bayesian} approach to spectrally-normalized margin bounds
  for neural networks.
\newblock In \emph{International Conference on Learning Representations}, 2018.

\bibitem[Papyan(2018)]{papyan_full_2018}
V.~Papyan.
\newblock The {Full} {Spectrum} of {Deepnet} {Hessians} at {Scale}: {Dynamics}
  with {SGD} {Training} and {Sample} {Size}.
\newblock \emph{arXiv:1811.07062 [cs, stat]}, Nov. 2018.
\newblock URL \url{http://arxiv.org/abs/1811.07062}.
\newblock arXiv: 1811.07062.

\bibitem[Papyan(2019)]{papyan2019measurements}
V.~Papyan.
\newblock Measurements of three-level hierarchical structure in the outliers in
  the spectrum of deepnet {Hessians}.
\newblock In \emph{International Conference on Machine Learning}, pages
  5012--5021, 2019.

\bibitem[Parrado-Hern{\'a}ndez et~al.(2012)Parrado-Hern{\'a}ndez, Ambroladze,
  Shawe-Taylor, and Sun]{parrado2012pac}
E.~Parrado-Hern{\'a}ndez, A.~Ambroladze, J.~Shawe-Taylor, and S.~Sun.
\newblock {PAC-Bayes} bounds with data dependent priors.
\newblock \emph{The Journal of Machine Learning Research}, 13\penalty0
  (1):\penalty0 3507--3531, 2012.

\bibitem[Rudelson and Vershynin(2013)]{ruve13}
M.~Rudelson and R.~Vershynin.
\newblock {Hanson-Wright} inequality and {sub-Gaussian} concentration.
\newblock \emph{Electronic Communications in Probability}, 18, 2013.

\bibitem[Sagun et~al.(2016)Sagun, Bottou, and LeCun]{sabo16}
L.~Sagun, L.~Bottou, and Y.~LeCun.
\newblock Eigenvalues of the {Hessian} in deep learning: Singularity and
  beyond.
\newblock \emph{arXiv preprint arXiv:1611.07476}, 2016.

\bibitem[Sagun et~al.(2017)Sagun, Evci, Guney, Dauphin, and
  Bottou]{sagun_empirical_2017}
L.~Sagun, U.~Evci, V.~U. Guney, Y.~Dauphin, and L.~Bottou.
\newblock Empirical analysis of the {Hessian} of over-parametrized neural
  networks.
\newblock \emph{arXiv:1706.04454 [cs]}, June 2017.
\newblock URL \url{http://arxiv.org/abs/1706.04454}.
\newblock arXiv: 1706.04454.

\bibitem[Sauer(1972)]{sauer1972density}
N.~Sauer.
\newblock On the density of families of sets.
\newblock \emph{Journal of Combinatorial Theory, Series A}, 13\penalty0
  (1):\penalty0 145--147, 1972.

\bibitem[Seeger(2002)]{seeger2002pac}
M.~Seeger.
\newblock {PAC-Bayesian} generalisation error bounds for {Gaussian} process
  classification.
\newblock \emph{Journal of machine learning research}, 3\penalty0
  (Oct):\penalty0 233--269, 2002.

\bibitem[Shalev-Shwartz and Ben-David(2014)]{ssbd14}
S.~Shalev-Shwartz and S.~Ben-David.
\newblock \emph{Understanding machine learning: From theory to algorithms}.
\newblock Cambridge university press, 2014.

\bibitem[Shalev-Shwartz et~al.(2010)Shalev-Shwartz, Shamir, Srebro, and
  Sridharan]{shalev-shwartz_learnability_2010}
S.~Shalev-Shwartz, O.~Shamir, N.~Srebro, and K.~Sridharan.
\newblock Learnability, {Stability} and {Uniform} {Convergence}.
\newblock \emph{Journal of Machine Learning Research}, 11\penalty0
  (Oct):\penalty0 2635--2670, 2010.
\newblock ISSN ISSN 1533-7928.
\newblock URL \url{http://www.jmlr.org/papers/v11/shalev-shwartz10a.html}.

\bibitem[Shawe-Taylor and Williamson(1997)]{shawe1997pac}
J.~Shawe-Taylor and R.~C. Williamson.
\newblock A {PAC} analysis of a {Bayesian} estimator.
\newblock In \emph{Proceedings of the tenth annual conference on Computational
  Learning Theory}, pages 2--9, 1997.

\bibitem[Smith and Le(2018)]{smle18}
S.~L. Smith and Q.~V. Le.
\newblock A {Bayesian} perspective on generalization and stochastic gradient
  descent.
\newblock In \emph{International Conference on Learning Representations}, 2018.

\bibitem[Soudry et~al.(2018)Soudry, Hoffer, Nacson, Gunasekar, and
  Srebro]{soho18}
D.~Soudry, E.~Hoffer, M.~S. Nacson, S.~Gunasekar, and N.~Srebro.
\newblock The implicit bias of gradient descent on separable data.
\newblock \emph{Journal of Machine Learning Research}, 19\penalty0
  (1):\penalty0 2822--2878, 2018.

\bibitem[Valiant(1984)]{vali84}
L.~G. Valiant.
\newblock A theory of the learnable.
\newblock \emph{Communications of the ACM}, 1984.

\bibitem[Vapnik(1968)]{vapnik1968uniform}
V.~Vapnik.
\newblock On the uniform convergence of relative frequencies of events to their
  probabilities.
\newblock In \emph{Doklady Akademii Nauk USSR}, volume 181, pages 781--787,
  1968.

\bibitem[Vapnik(1992)]{vapnik1992principles}
V.~Vapnik.
\newblock Principles of risk minimization for learning theory.
\newblock In \emph{Advances in Neural Information Processing Systems}, pages
  831--838, 1992.

\bibitem[Vapnik(2013)]{vapnik2013nature}
V.~Vapnik.
\newblock \emph{The nature of statistical learning theory}.
\newblock Springer science \& business media, 2013.

\bibitem[Vapnik(1999)]{vapnik1999overview}
V.~N. Vapnik.
\newblock An overview of statistical learning theory.
\newblock \emph{IEEE Transactions on Neural Networks}, 10\penalty0
  (5):\penalty0 988--999, 1999.

\bibitem[Vershynin(2018)]{vers18}
R.~Vershynin.
\newblock \emph{High-dimensional probability: An introduction with applications
  in data science}, volume~47.
\newblock Cambridge university press, 2018.

\bibitem[Yang et~al.(2019)Yang, Sun, and Roy]{yasu19}
J.~Yang, S.~Sun, and D.~M. Roy.
\newblock Fast-rate {PAC-Bayes} generalization bounds via shifted {Rademacher}
  processes.
\newblock In \emph{Advances in Neural Information Processing Systems}, 2019.

\bibitem[Zhang et~al.(2017)Zhang, Bengio, Hardt, Recht, and Vinyals]{zhbh17}
C.~Zhang, S.~Bengio, M.~Hardt, B.~Recht, and O.~Vinyals.
\newblock Understanding deep learning requires rethinking generalization.
\newblock In \emph{International Conference on Learning Representations}, 2017.

\bibitem[Zhou and Feng(2018)]{zhou2018understanding}
P.~Zhou and J.~Feng.
\newblock Understanding generalization and optimization performance of deep
  {CNNs}.
\newblock In \emph{International Conference on Machine Learning}, pages
  5960--5969, 2018.

\end{thebibliography}

\appendix

\newpage

\section{Gaussian Distributions: Technical Results}
\label{app:gauss}

\begin{lemm}
Let $\theta \sim N(0,\sigma^2 \I_{p \times p})$. Then, 
\begin{equation}
    \P(\exists j \in [p] ~ \mid ~ |\theta_j| > c\sigma + \sqrt{\log p + \log 1/\delta} ) \leq \delta~. 
\end{equation}
\end{lemm}

\begin{lemm} \label{lemm: up_lo_H}
For $\delta \sim \cN(0,\Sigma)$, where $\Sigma = \diag(\upsilon^2_j)$ with $\upsilon^2_j = \min\{\sigma^2, \sigma^2_j\}, \forall j \in [p]$, positive semi-definite matrices $H$ with $\|H\|_{2} \leq \zeta,$ and denote the ratio $\kappa:=\frac{\|H\|_{F}^{2}}{\|H\|_{2}^{2}}$ and the stable rank $\alpha=\frac{\operatorname{Tr}(H)}{\|H\|_{2}}$, we have the following upper bound and lower bound:
\begin{equation}
\begin{split}
    &\P \left[  \delta^T H \delta   >  \sigma^2 \zeta \alpha\tilde \gamma  \right]    \leq  \exp\left(- \frac{1}{2} \min \left[ \frac{  \alpha^2(\tilde \gamma-1)^2}{\kappa}, \alpha(\tilde \gamma-1) \right] \right)~,
    \label{eq:ub_h}
    \end{split}
\end{equation}
where $\tilde \gamma >1$.
\end{lemm}

\proof Since $H$ is positive semi-definite, the diagonals of $H$ must be non-negative, then we have,
\begin{align*}
\E\left[\delta^TH\delta\right]=\E[\tr(H\delta\delta^T)]=\tr( H\Sigma) &= \sum_i \min\{\sigma^2, \sigma^2_j\} H[i,i]\\
 &   \leq \sigma^2 \sum_i \min\left(1, \frac{\sigma_j^2}{\sigma^2 }\right) H[i,i]\\
    & \leq \sigma^2 \sum_i H[i,i] = \sigma^2\zeta \alpha~,
\end{align*}

From Hanson-Wright inequality and the fact that $\E\left[\delta^TH\delta\right]=\E[\tr(H\delta\delta^T)]=\sigma^2\tr(H)\leq\sigma^2\zeta \alpha$, and \begin{equation}
\label{eq:psi2_iso}
    \max_i \|\delta_i\|_{\psi_2}\leq \sqrt{2} \sigma~,
\end{equation} 
we have
\begin{align}
    \P\left[\delta^TH\delta-\sigma^2\zeta \alpha\geq t\right]
    &\leq \exp\left(-\frac{1}{2}\min\left[\frac{t^2}{\sigma^4\|H\|_F^2},\frac{t}{\sigma^2\|H\|_2}\right]\right)\\
    &\leq \exp\left(-\frac{1}{2}\min\left[\frac{t^2}{\sigma^4\kappa\zeta^2},\frac{t}{\sigma^2\zeta}\right]\right)~.
\end{align}
By taking $t=\tilde \gamma\sigma^2 \zeta \alpha(>0)$, we have:
\begin{equation}
\begin{split}
    &\P[\delta^TH\delta\geq (\tilde \gamma+1)\sigma^2\zeta \alpha]\leq \exp\left(-\frac{1}{2}\min\left[\frac{ \alpha^2\tilde \gamma^2}{\kappa},\alpha\tilde \gamma\right]\right)~.
\end{split}
\end{equation}
Denoting $\tilde \gamma+1$ as $\tilde\gamma$ completes the proof. \qed

\section{De-randomized Margin Bounds: Smooth Predictors and PAC-Bayes}
\label{app:margin2nc_updated}
 We consider the case where $\phi^{\theta}(x_i)$ is a smooth function of 
$\theta$, and provide detailed proofs of the technical results corresponding to PAC-Bayes 
with isotropic posteriors briefly mentioned in Section~\ref{sec:margin2nc_updated}.
We start by recalling the Assumption~\ref{asmp:gen} which will be used for the analysis:
\asmpgen*

\subsection{Bounds for Stochastic vs.~Deterministic Predictors}

For analyzing smooth predictors using PAC-Bayes with isotropic posterior, we first establish the following bound relating the performance of deterministic and stochastic predictors:

\theosmarg*


\proof
Since $\phi$ is twice differentiable, for some suitable (random) $\tilde{\theta} = (1-\tau) \theta^{\dagger} + \tau \theta = \theta^{\dagger} + \tau (\theta - \theta^{\dagger})$ where $\tau \in [0,1]$, we have
\begin{equation}
\phi^{\theta}(x_i) = \phi^{\theta^{\dagger}}(x_i) + \langle \theta - \theta^{\dagger}, \nabla \phi^{\theta^{\dagger}}(x_i) \rangle + \frac{1}{2} (\theta - \theta^{\dagger})^T H_{\phi}^{\tilde{\theta}}(x_i) (\theta - \theta^{\dagger})~. 
\end{equation}

Now consider the following set where $\theta^{\dagger}$ achieves a margin greater than $\left(\beta + \sigma^2\zeta \alpha\tilde \gamma\right)$:
\begin{equation}
\cZ^{(>)}_{\beta + \frac{1}{2}\sigma^2\zeta \alpha\tilde \gamma}(\theta^{\dagger} ) = \bigg\{ (x,y) \in \cX \times \cY ~\bigg|~  y \phi^{\theta^{\dagger}}(x) > \beta + \sigma^2\zeta \alpha\tilde \gamma ~ \bigg\}~.
\end{equation}

Let $P = N(0,\sigma^2 \I_p)$ be a multivariate distribution with mean 0 and covariance $\sigma^2 \I_p$. Now, for $z \in \cZ^{(>)}_{\beta + \sigma^2\zeta \alpha\tilde \gamma}(\theta^{\dagger} )$, we have

\begin{align*}
\P_{\theta \sim \cQ}&\bigg[ y \phi^{\theta}(x) ~\leq ~\beta  ~\big|~ z \in \cZ^{(>)}_{\beta + \sigma^2\zeta \alpha\tilde \gamma}(\theta^{\dagger} ) \bigg] = \P_{\delta \sim P}\bigg[ y \phi^{\theta^{\dagger}+\delta}(x) ~\leq ~\beta  ~\big|~ z \in \cZ^{(>)}_{\beta + \sigma^2\zeta \alpha\tilde \gamma}(\theta^{\dagger} ) \bigg] ~.
\end{align*}

Now we first present the proof of \eqref{eq:non_conv_marg_swd}, where we bound the loss on stochastic predictors with the the loss on deterministic predictor.

Conditioned on $y =+1$, we have 
	\begin{equation}\label{eq:con_cha}\begin{split}
&\P_{\delta \sim P}\bigg[  \phi^{\theta^{\dagger}+\delta}(x) ~\leq ~\beta  ~\big|~ z \in \cZ^{(>)}_{\beta + \sigma^2\zeta \alpha\tilde \gamma}(\theta^{\dagger} ), y = +1 \bigg] \\
& = \P_{\delta \sim P}\bigg[  \phi^{\theta^{\dagger}}(x) + \langle \delta, \nabla \phi^{\theta^{\dagger}}(x) \rangle +  \frac{1}{2}\delta^T H_{\phi}^{\tilde \theta}(x)  \delta \leq \beta ~\big|~ z \in \cZ^{(>)}_{\beta + \sigma^2\zeta \alpha\tilde \gamma}(\theta^{\dagger} ),  y = +1 \bigg] \\
& \leq  \P_{\delta \sim P}\bigg[   \langle \delta, \nabla \phi^{\theta^{\dagger}}(x) \rangle- \frac{1}{2} \delta^T H  \delta \leq \beta -\phi^{\theta^{\dagger}}(x)~\big|~ z \in \cZ^{(>)}_{\beta + \sigma^2\zeta \alpha\tilde \gamma}(\theta^{\dagger} ), y = +1 \bigg] \\
& \overset{(a)}{\leq}    \P_{\delta \sim P}\bigg[   \langle \delta, \nabla \phi^{\theta^{\dagger}}(x) \rangle - \frac{1}{2} \delta^T H \delta \leq - \sigma^2 \tilde \gamma \zeta \alpha~\big|~ z \in \cZ^{(>)}_{\beta + \sigma^2\zeta \alpha\tilde \gamma}(\theta^{\dagger} ), y = +1 \bigg]~ \\
& \overset{(b)}{\leq}    \P_{\delta \sim P}\bigg[   \langle \delta, \nabla \phi^{\theta^{\dagger}}(x) \rangle \leq -\frac{1}{2} \sigma^2 \zeta \alpha \tilde \gamma ~\big|~ z \in \cZ^{(>)}_{\beta + \sigma^2\zeta \alpha\tilde \gamma}(\theta^{\dagger} ), y = +1 \bigg] +\P_{\delta \sim P}\bigg[\delta^T H \delta \geq \sigma^2 \zeta \alpha\tilde \gamma\bigg]~\\
& \overset{(c)}{\leq} \exp \left( -\frac{\sigma^2 \zeta^2 \alpha^2\tilde \gamma^2}{2G^2} \right)+\exp\left(- \frac{1}{2}\min \left[ \frac{ \alpha^2(\tilde \gamma-1)^2}{\kappa},  \alpha(\tilde \gamma-1) \right] \right)~,
\end{split}\end{equation}
where (a) follows since for $(x,y) \in \cZ^{(>)}_{\beta +\sigma^2\zeta \alpha\tilde \gamma}(\theta^{\dagger})$ and $y=+1$ we have $\phi^{\theta^{\dagger}}(x) > \beta + \sigma^2\zeta \alpha\tilde \gamma$; (b) follows since $\P[x+y\leq a+b]\leq \P[x\leq a]+\P[y\leq b]$; (c) is from Hoeffding's inequality with $\| \nabla \phi^{\theta^{\dagger}}(x) \|_2^2 \leq G^2$ and Lemma \ref{lemm: up_lo_H}.

Similarly, conditioned in $y =-1$, we have 
\begin{align*}
&\P_{\delta \sim P}\bigg[  -\phi^{\theta^{\dagger}+\delta}(x) ~\leq ~\beta  ~\big|~ z \in \cZ^{(>)}_{\beta +  \sigma^2\zeta \alpha\tilde \gamma}(\theta^{\dagger} ), y = -1 \bigg] \\
& = \P_{\delta \sim P}\bigg[  -\phi^{\theta^{\dagger}}(x) -\langle \delta, \nabla \phi^{\theta^{\dagger}}(x) \rangle - \frac{1}{2} \delta^T H_{\phi}^{\tilde \theta}(x)  \delta \leq \beta  ~\big|~ z \in \cZ^{(>)}_{\beta +  \sigma^2\zeta \alpha\tilde \gamma}(\theta^{\dagger} ),  y = -1 \bigg] \\
& \leq  \P_{\delta \sim P}\bigg[   -\langle \delta, \nabla \phi^{\theta^{\dagger}}(x) \rangle - \frac{1}{2} \delta^T H \delta \leq \beta + \phi^{\theta^{\dagger}}(x)~\big|~ z \in \cZ^{(>)}_{\beta + \sigma^2\zeta \alpha\tilde \gamma}(\theta^{\dagger} ), y = -1 \bigg] \\
& \overset{(a)}{\leq}    \P_{\delta \sim P}\bigg[   -\langle \delta, \nabla \phi^{\theta^{\dagger}}(x) \rangle - \frac{1}{2}\delta^T H \delta \leq -  \sigma^2  \zeta \alpha\tilde \gamma  ~\big|~ z \in \cZ^{(>)}_{\beta +  \sigma^2\zeta \alpha\tilde \gamma}(\theta^{\dagger} ), y = -1 \bigg]~ \\
& \overset{(b)}{\leq}    \P_{\delta \sim P}\bigg[   -\langle \delta, \nabla \phi^{\theta^{\dagger}}(x) \rangle \leq -\frac{1}{2}\sigma^2 \zeta \alpha \tilde \gamma~\big|~ z \in \cZ^{(>)}_{\beta +  \sigma^2\zeta \alpha\tilde \gamma}(\theta^{\dagger} ), y = -1 \bigg]+\P_{\delta \sim P}\bigg[ -\delta^T H \delta\leq -\sigma^2\zeta \alpha \tilde \gamma\bigg]~\\
& \overset{(c)}{\leq} \exp \left( -\frac{\sigma^2 \zeta^2 \alpha^2\tilde \gamma^2}{2G^2} \right)+\exp\left(- \frac{1}{2}\min \left[ \frac{  \alpha^2(\tilde \gamma-1)^2}{ \kappa}, \alpha(\tilde \gamma-1)\right] \right)~,
\end{align*}
where (a) follows since for $(x,y) \in \cZ^{(>)}_{\beta + \sigma^2\zeta \alpha\tilde \gamma}(\theta^{\dagger})$ and $y=-1$ we have $-\phi^{\theta^{\dagger}}(x) > \beta + \sigma^2\zeta \alpha\tilde \gamma$; (b) follows since $\P[x+y\leq a+b]\leq \P[x\leq a]+\P[y\leq b]$; (c) is from Hoeffding's inequality with $\| \nabla \phi^{\theta^{\dagger}}(x) \|_2^2 \leq G^2$ and Lemma \ref{lemm: up_lo_H}.

Then, we have
\begin{equation}
\begin{split}
&\P_{\theta \sim \cQ}\bigg[ y \phi^{\theta}(x) ~\leq ~\beta  ~\big|~ z \in \cZ^{(>)}_{\beta +  \sigma^2\zeta \alpha\tilde \gamma}(\theta^{\dagger} ) \bigg]\\
&= \P_{\delta \sim P}\bigg[ y \phi^{\theta^{\dagger}+\delta}(x) ~\leq ~\beta  ~\big|~z \in \cZ^{(>)}_{\beta +  \sigma^2\zeta \alpha\tilde \gamma}(\theta^{\dagger} ) \bigg] \\
& \leq   \P \bigg[ y \phi^{\theta^{\dagger}+\delta}(x) ~\leq ~\beta  ~\big|~ z \in \cZ^{(>)}_{\beta + \sigma^2\zeta \alpha\tilde \gamma}(\theta^{\dagger} ),  y = +1 \bigg]  + \P \bigg[ y \phi^{\theta^{\dagger}+\delta}(x) ~\leq ~\beta  ~\big|~ z \in \cZ^{(>)}_{\beta +  \sigma^2\zeta \alpha\tilde \gamma}(\theta^{\dagger} ), y = -1 \bigg] \\
& \leq 2\exp \left( -\frac{\sigma^2 \zeta^2 \alpha^2\tilde \gamma^2}{2G^2} \right)+2\exp\left(- \frac{1}{2} \min \left[ \frac{  \alpha^2(\tilde \gamma-1)^2}{\kappa}, \alpha(\tilde \gamma-1) \right] \right)~.
\end{split}
\end{equation}
For $z \not \in \cZ^{(>)}_{\beta + \sigma^2\zeta \alpha\tilde \gamma}(\theta^{\dagger} )$, we have
\begin{align*}
\P_{\theta \sim Q} \bigg[ y \phi^{\theta^\dagger}(x) ~\leq ~\beta  ~\big|~ z \not \in \cZ^{(>)}_{\beta +  \sigma^2\zeta \alpha\tilde \gamma}(\theta^{\dagger} ) \bigg] \leq 1~.
\end{align*}
By definition, we have
\begin{align*}
&\ell_{\beta}(\cQ,W)  = \P_{\substack{\theta \sim \cQ\\ z \sim W}}\bigg[  y\phi^{\theta}(x) ~\leq ~\beta ~ \bigg] \\
& =   \P_{\substack{\theta \sim \cQ\\ z \sim W}}\bigg[ y \phi^{\theta}(x) ~\leq ~\beta  ~\big|~ z  \not \in \cZ^{(>)}_{\beta + \sigma^2\zeta \alpha\tilde \gamma}(\theta^{\dagger}) \bigg]\P_{z\sim W}\left[ y \phi^{\theta^{\dagger}}(x) \leq \beta + \sigma^2\zeta \alpha\tilde \gamma\right]\\
&\quad+  \P_{\substack{\theta \sim \cQ\\ z \sim W}} \bigg[ y \phi^{\theta}(x) ~\leq ~\beta  ~\big|~ z \in \cZ^{(>)}_{\beta + \sigma^2\zeta \alpha\tilde \gamma}(\theta^{\dagger}) \bigg]\P_{z\sim W}\left[ y \phi^{\theta^{\dagger}}(x) > \beta + \sigma^2\zeta \alpha\tilde \gamma\right] \\
& \leq  \P_{z \sim  W} \bigg[ y \phi^{\theta^{\dagger}}(x) \leq \beta + \sigma^2\zeta \alpha\tilde \gamma \bigg]+ 2\exp \left( -\frac{1}{2}\frac{\sigma^2 \zeta^2 \alpha^2\tilde \gamma^2}{G^2} \right)+2\exp\left(- \frac{1}{2} \min \left[ \frac{ \zeta^2 \alpha^2(\tilde \gamma-1)^2}{\zeta_F^2}, \frac{ \zeta \alpha(\tilde \gamma-1)}{\zeta} \right] \right)\\
& \leq  \P_{z \sim  W} \bigg[ y \phi^{\theta^{\dagger}}(x) \leq \beta + \sigma^2\zeta \alpha\tilde \gamma \bigg]+ 2\exp \left( -\frac{1}{2}\frac{\sigma^2 \zeta^2 \alpha^2\tilde \gamma^2}{G^2} \right)+2\exp\left(- \frac{1}{2} \min \left[ \frac{ \zeta^2 \alpha^2(\tilde \gamma-1)^2}{\zeta_F^2}, \frac{ \zeta \alpha(\tilde \gamma-1)}{\zeta} \right] \right) \\
& =  \ell_{\beta +  \sigma^2\zeta \alpha\tilde \gamma}(\phi^{\theta^{\dagger}},W)+ 2\exp \left( -\frac{\sigma^2 \zeta^2 \alpha^2\tilde \gamma^2}{2G^2} \right)+2\exp\left(- \frac{1}{2} \min \left[ \frac{ \alpha^2(\tilde \gamma-1)^2}{\kappa},  \alpha(\tilde \gamma-1) \right] \right) ~,
\end{align*}
which establishes \eqref{eq:non_conv_marg_swd}.

Now we first present the proof of \eqref{eq:non_conv_marg_dws}, where we bound the  the loss on deterministic predictor with the loss on stochastic predictors.
\label{sec: proof_dws}

Consider the following set where $\theta^{\dagger}$ achieves a margin of at most $\beta$:
\begin{equation}
\cZ^{(\leq)}_{\beta}(\theta^{\dagger} ) = \bigg\{ (x,y) \in \cX \times \cY ~\bigg|~  y \phi^{\theta^{\dagger}}(x) \leq \beta ~ \bigg\}~.
\end{equation}

Let $P = N(0,\sigma^2 \I_p)$ be a multivariate distribution with mean 0 and covariance $\sigma^2 \I_p$. Now, for $z \in \cZ^{(\leq)}_{\beta}(\theta^{\dagger})$, we have
\begin{align*}
\P_{\theta \sim \cQ}\bigg[ y \phi^{\theta}(x) ~> ~ \beta + \sigma^2\zeta \alpha\tilde \gamma ~\big|~ z \in \cZ^{(\leq)}_{\beta}(\theta^{\dagger}) \bigg]  = \P_{\delta \sim P}\bigg[ y \phi^{\theta^{\dagger}+\delta}(x) ~> ~\beta +  \sigma^2\zeta \alpha\tilde \gamma ~\big|~ z \in \cZ^{(\leq)}_{\beta}(\theta^{\dagger}) \bigg]~.
\end{align*}

Now, conditioned in $y =+1$, we have 
\begin{align*}
&\quad\P_{\delta \sim P}\bigg[  \phi^{\theta^{\dagger}+\delta}(x) ~> ~\beta +  \sigma^2\zeta \alpha\tilde \gamma~\big|~ z \in \cZ^{(\leq)}_{\beta}(\theta^{\dagger}), y = +1 \bigg] \\
& =~ \P_{\delta \sim P}\bigg[  \phi^{\theta^{\dagger}}(x) + \langle \delta, \nabla \phi^{\theta^{\dagger}}(x) \rangle +  \frac{1}{2} \delta^T H_{\phi}^{\tilde \theta}(x) \delta > \beta + \sigma^2\zeta \alpha\tilde \gamma~\big|~ z \in \cZ^{(\leq)}_{\beta}(\theta^{\dagger}),  y = +1 \bigg] \\
& \leq~  \P_{\delta \sim P}\bigg[   \langle \delta, \nabla \phi^{\theta^{\dagger}}(x) \rangle + \frac{1}{2}\delta^T H \delta > \beta +  \sigma^2\zeta \alpha\tilde \gamma -\phi^{\theta^{\dagger}}(x)~\big|~ z \in \cZ^{(\leq)}_{\beta}(\theta^{\dagger}),  y = +1 \bigg] \\
& \overset{(a)}{\leq}~    \P_{\delta \sim P}\bigg[   \langle \delta, \nabla \phi^{\theta^{\dagger}}(x) \rangle + \frac{1}{2} \delta^T H \delta >  \sigma^2\zeta \alpha\tilde \gamma~\big|~ z \in \cZ^{(\leq)}_{\beta}(\theta^{\dagger}), y = +1 \bigg] \\
& \overset{(b)}{\leq}~    \P_{\delta \sim P}\bigg[   \langle \delta, \nabla \phi^{\theta^{\dagger}}(x) \rangle  >  \frac{1}{2} \sigma^2\zeta \alpha\tilde \gamma ~\big|~ z \in \cZ^{(\leq)}_{\beta}(\theta^{\dagger}),  y = +1 \bigg]+\P_{\delta \sim P}\bigg[ \delta^T H \delta \geq \sigma^2 \zeta \alpha \tilde \gamma\bigg] \\
& \overset{(c)}{\leq} \exp \left( -\frac{1}{2}\frac{\sigma^2 \zeta^2 \alpha^2\tilde \gamma^2}{G^2} \right)+\exp\left(- \frac{1}{2} \min \left[ \frac{ \alpha^2(\tilde \gamma-1)^2}{ \kappa}, \alpha(\tilde \gamma-1) \right] \right)~,
\end{align*}
where (a) follows since for $(x,y) \in \cZ^{(\leq)}_{\beta}(\theta^{\dagger})$ and $y=+1$ we have $\phi^{\theta^{\dagger}}(x) \leq \beta \Rightarrow \beta -\phi^{\theta^{\dagger}}(x) \geq 0$;  (b) follows since $\P[x+y\leq a+b]\leq \P[x\leq a]+\P[y\leq b]$; (c) is from Hoeffding's inequality with $\| \nabla \phi^{\theta^{\dagger}}(x) \|_2^2 \leq G^2$ and Lemma \ref{lemm: up_lo_H}.

Similarly, conditioned on $y =-1$, we have 
\begin{align*}
&\quad \P_{\delta \sim P}\bigg[  - \phi^{\theta^{\dagger}+\delta}(x) ~> ~\beta + \sigma^2\zeta \alpha\tilde \gamma ~\big|~ z \in \cZ^{(\leq)}_{\beta}(\theta^{\dagger}), y = -1 \bigg] \\
& =~ \P_{\delta \sim P}\bigg[  - \phi^{\theta^{\dagger}}(x) - \langle \delta, \nabla \phi^{\theta^{\dagger}}(x) \rangle -  \frac{1}{2}\delta^T H_{\phi}^{\tilde \theta}(x)  \delta > \beta + \sigma^2\zeta \alpha\tilde \gamma~\big|~ z \in \cZ^{(\leq)}_{\beta}(\theta^{\dagger}),  y = -1 \bigg] \\
& \leq~  \P_{\delta \sim P}\bigg[ -  \langle \delta, \nabla \phi^{\theta^{\dagger}}(x) \rangle - \frac{1}{2} \delta^T H \delta > \beta + \sigma^2\zeta \alpha\tilde \gamma + \phi^{\theta^{\dagger}}(x)~\big|~ z \in \cZ^{(\leq)}_{\beta}(\theta^{\dagger}), y = -1 \bigg] \\
& \overset{(a)}{\leq}~    \P_{\delta \sim P}\bigg[ -  \langle \delta, \nabla \phi^{\theta^{\dagger}}(x) \rangle - \frac{1}{2}\delta^T H \delta >    \sigma^2\zeta \alpha\tilde \gamma ~\big|~ z \in \cZ^{(\leq)}_{\beta}(\theta^{\dagger}),  y = -1 \bigg] \\
& \overset{(b)}{\leq}~    \P_{\delta \sim P}\bigg[   - \langle \delta, \nabla \phi^{\theta^{\dagger}}(x) \rangle  > \frac{1}{2}\sigma^2\zeta \alpha\tilde \gamma ~\big|~ z \in \cZ^{(\leq)}_{\beta}(\theta^{\dagger}), y = -1 \bigg]+ \P_{\delta \sim P}\bigg[\delta^T H\delta \geq \sigma^2\zeta \alpha \tilde \gamma\bigg] \\
& \overset{(c)}{\leq} \exp \left( -\frac{1}{2}\frac{\sigma^2 \zeta^2 \alpha^2\tilde \gamma^2}{G^2} \right)+\exp\left(- \frac{1}{2} \min \left[ \frac{ \alpha^2(\tilde \gamma-1)^2}{\kappa}, \alpha(\tilde \gamma-1) \right] \right)
\end{align*}
where (a) follows since for $(x,y) \in \cZ^{(\leq)}_{\beta}(\theta^{\dagger}$ and $y=-1$ we have $-\phi(\theta^{\dagger}) \leq \beta \Rightarrow \beta + \phi(\theta^{\dagger}) \geq 0$; (b) follows since $\P[x+y\leq a+b]\leq \P[x\leq a]+\P[y\leq b]$; (c) is from Hoeffding's inequality with $\| \nabla \phi^{\theta^{\dagger}}(x) \|_2^2 \leq G^2,~\max_i \|\delta_i\|_{\psi_2}\leq \frac{1}{2} \sigma$, and Lemma \ref{lemm: up_lo_H}.

Then, we have
\begin{align*}
&\P_{\theta \sim \cQ}\bigg[ y \phi^{\theta}(x) ~> ~\beta + \sigma^2\zeta \alpha\tilde \gamma  ~\big|~ z \in \cZ^{(\leq)}_{\beta}(\theta^{\dagger}) \bigg]  \\
&= \P_{\delta \sim P}\bigg[ y \phi^{\theta^{\dagger}+\delta}(x) ~> ~\beta +\sigma^2\zeta \alpha\tilde \gamma ~\big|~ z \in \cZ^{(\leq)}_{\beta}(\theta^{\dagger})\bigg] \\
& \leq   \P \bigg[ y \phi^{\theta^{\dagger}+\delta}(x) ~> ~\beta + \sigma^2\zeta \alpha\tilde \gamma ~\big|~ z \in \cZ^{(\leq)}_{\beta}(\theta^{\dagger}), y = +1 \bigg]+ \P \bigg[ y \phi^{\theta^{\dagger}+\delta}(x) ~>~\beta +\sigma^2\zeta \alpha\tilde \gamma ~\big|~ z \in \cZ^{(\leq)}_{\beta }(\theta^{\dagger}),  y = -1 \bigg] \\
& \leq 2\exp \left( -\frac{1}{2}\frac{\sigma^2 \zeta^2 \alpha^2\tilde \gamma^2}{G^2} \right)+2\exp\left(- \frac{1}{2} \min \left[ \frac{ \alpha^2(\tilde \gamma-1)^2}{\kappa}, \alpha(\tilde \gamma-1) \right] \right)~ .
\end{align*}
For $z \not \in \cZ^{(\leq)}_{\beta}(\theta^{\dagger})$, we have
\begin{equation}
\begin{split}
\P_{\theta \sim Q}  \bigg[ y \phi^{\theta^\dagger}(x) ~> ~\beta + \sigma^2\zeta \alpha\tilde \gamma ~\big|~ z \not \in \cZ^{(\leq)}_{\beta }(\theta^{\dagger}) \bigg] \leq 1~.
\end{split}
\end{equation}
By definition, we have
\begin{equation}
\begin{split}
&~1-  \ell_{\beta +\sigma^2\zeta \alpha\tilde \gamma }(\cQ,W) \\ 
&= \P_{\substack{\theta \sim \cQ\\ z \sim W}}\bigg[  y\phi^{\theta}(x) ~> ~\beta + \sigma^2\zeta \alpha\tilde \gamma ~ \bigg] \\
& \leq \P_{z \sim  W} \bigg[ y \phi^{\theta^{\dagger}}(x) > \beta  \bigg] +  \P_{\substack{\theta \sim \cQ\\ z \sim W}} \bigg[ y \phi^{\theta}(x) ~> ~\beta + \sigma^2\zeta \alpha\tilde \gamma ~\big|~ z \in \cZ^{(\leq)}_{\beta}(\theta^{\dagger}) \bigg] \\
& \leq  \P_{z \sim  W} \bigg[ y \phi^{\theta^{\dagger}}(x) > \beta  \bigg]+ 2\exp \left( -\frac{1}{2}\frac{\sigma^2 \zeta^2 \alpha^2\tilde \gamma^2}{G^2} \right) +2\exp\left(- \frac{1}{2} \min \left[ \frac{  \alpha^2(\tilde \gamma-1)^2}{\kappa},  \alpha(\tilde \gamma-1) \right] \right) \\
& \leq \P_{z \sim  W} \bigg[ y \phi^{\theta^{\dagger}}(x) > \beta  \bigg] + 2\exp \left( -\frac{1}{2}\frac{\sigma^2 \zeta^2 \alpha^2\tilde \gamma^2}{G^2} \right)+2\exp\left(- \frac{1}{2} \min \left[ \frac{ \alpha^2(\tilde \gamma-1)^2}{\kappa}, \alpha(\tilde \gamma-1)\right] \right) \\
& = 1-\ell_{\beta}(\phi^{\theta^{\dagger}},W)  + 2\exp \left( -\frac{1}{2}\frac{\sigma^2 \zeta^2 \alpha^2\tilde \gamma^2}{G^2} \right) +2\exp\left(- \frac{1}{2} \min \left[ \frac{ \alpha^2(\tilde \gamma-1)^2}{\kappa},  \alpha(\tilde \gamma-1) \right] \right)~,
\end{split}
\end{equation}
which implies
\begin{equation}
\begin{split}
\ell_{\beta}&(\phi^{\theta^{\dagger}},W)  \leq  \ell_{\beta + \sigma^2\zeta \alpha\tilde \gamma }(\cQ,W) + 2\exp \left( -\frac{1}{2}\frac{\sigma^2 \zeta^2 \alpha^2\tilde \gamma^2}{G^2} \right) +2\exp\left(- \frac{1}{2} \min \left[ \frac{  \alpha^2(\tilde \gamma-1)^2}{\kappa},  \alpha(\tilde \gamma-1) \right] \right)~.
\end{split}
\end{equation}
By choosing $\tilde \gamma>2$, we have  $(\tilde\gamma-1)^2 >\frac{1}{4} \tilde\gamma^2$, $\tilde\gamma - 1>\frac{1}{2}\tilde\gamma$, which completes the proof.
\qed

\subsection{Deterministic Margin Bounds with Fast Rates}
With the above de-randomization, another piece we need to derive a deterministic margin bound is the fast rate Pac-Bayes bound in \cite{cato07,yasu19}, which is formally stated as below:
\begin{theo}
(Fast-Rate PAC-Bayes \citep{cato07}[Theorem 1.2.6] \citep{yasu19}) For any prior distribution $P$,  for any $\delta \in(0,1)$ and $\eta \in(0,1)$, with probability at least $1 - \delta$ over the draw of $n$ samples $S \sim D^n$, for any $Q$ we have 
\begin{equation}
\label{eq:fast_bayes}
    \ell(Q,D) \leq \frac{\log (1/\eta)}{1-\eta} \ell(Q,S) + \frac{1}{1 - \eta} \frac{ KL(Q \| P) + \log (\frac{1}{\delta}) }{n} ~,
\end{equation}
where $\ell(Q,D), \ell(Q,S)$ are true and empirical losses.
\end{theo}

Recall that these multiplicative factors are exactly the ones which appear in classical algorithms such as the Weighted Majority \citep{liwa1994} and the connections between online regret bounds and PAC-Bayes bounds are well known~\citep{bane06}. For settings where the empirical loss $\ell(Q,S)$ is small, e.g., (margin) loss with deep nets on the training set, one can choose relatively smaller values of $\eta$ to get quantitatively tighter bounds. Denoting $a_{\eta}= \frac{\log (1/\eta)}{1-\eta}$ and $b_{\eta} = \frac{1}{1 - \eta}$, Table~\ref{tab:my_label} illustrate the trade-off between the empirical loss and the KL-divergence terms.
\begin{table}[t]
    \centering
\begin{tabular}{|c||c|c|c|c|} \hline
$\eta$ & 0.5 & 0.25 & 0.1 & 0.05 \\ \hline 
$a_{\eta}$ & 1.39 & 1.85 & 2.56 & 3.15 \\
$b_{\eta}$ & 2 & 1.33 & 1.11 & 1.05 \\ \hline 
\end{tabular}
    \caption{Trade-off between the empirical loss and the KL-divergence terms for `fast-rate' PAC-Bayes bound.}
\label{tab:my_label}
\end{table}

Utilizing the fast rate PAC-Bayes bound, we have the following bound for the deterministic predictor:

\theosmooth*

\proof To get to a de-randomized margin bound, we utilize the results in Theorem~\ref{theo:s_marg}.
First, with $\beta = 0$,  $\sigma^2\tilde \gamma \zeta \alpha = \gamma$ and $W=D$ in Theorem~\ref{theo:s_marg}  we have
\begin{equation}
\begin{split}
\ell_0(\phi^{\theta^{\dagger}}, D)& \leq \ell_{\gamma/2}(\cQ,D) +  4\exp \left(-\min(c_2\frac{\gamma^2}{\sigma^4\zeta^2 \alpha^2},c_1\frac{\gamma}{\sigma^2\zeta \alpha})\right)~.
\end{split}
\label{eq:dwsc}
\end{equation}
Similarly, with $\beta = \gamma/2$ and $W=S$ in Theorem we have
\begin{equation}
\begin{split}
\ell_{\gamma/2}(\cQ, S) &\leq \ell_{\gamma}(\phi^{\theta^{\dagger}},S) +  4\exp \left(-\min(c_2\frac{\gamma^2}{\sigma^4\zeta^2 \alpha^2},c_1\frac{\gamma}{\sigma^2\zeta \alpha})\right)~.
\end{split}
\label{eq:swdc}
\end{equation}

Now, from the Fast Rate PAC-Bayesian bound \eqref{eq:fast_bayes}, with probability at least $(1-\delta)$ over the draw of $n$ samples $S \sim D^n$, for any $\beta \in (0,1)$ and for any $Q$ we have 
\begin{equation} \label{eq:pac-bound1}
    \ell_{\gamma/2}(Q,D) \leq \frac{\log (1/\eta)}{1-\eta} \ell_{\gamma/2}(Q,S) + \frac{1}{1 - \eta} \frac{ KL(Q \| P) + \log (\frac{1}{\delta}) }{n} ~,
\end{equation}

Using \eqref{eq:dwsc} and \eqref{eq:swdc}, and noting that $KL(Q \| P) = \frac{\| \theta^{\dagger}-\theta_0 \|_2^2}{2\sigma^2}$, we have 
\begin{equation}
\begin{split}
\ell_{0}(\phi^{\theta^{\dagger}},D)  &\leq \frac{\log (1/\eta)}{1-\eta} \ell_{\gamma}(\phi^{\theta^{\dagger}},S) + 
\frac{1}{1 - \eta} \frac{ \| \theta^{\dagger}-\theta_0 \|_2^2}{2\sigma^2 n} + \frac{1}{1 - \eta} \frac{\log (\frac{1}{\delta})}{n}\\
&+ 4\left(\frac{\log (1/\eta)}{1-\eta}+1\right)\exp \left( -\min(c_2 \gamma^2,c_1\gamma) \right)~.
\end{split}
\end{equation} 

To show that our bound is scale-invariant, we use the property of KL-divergence between any continuous distributions $Q$ and $P$ such that the KL-divergence between $Q$ and $P$ remains invariant under $\zeta \alpha$-scale transformation \citep{kl2011}, i.e., 
\begin{equation}
    KL(Q^\prime ||P^\prime) = KL(Q||P),
\end{equation}
where $Q^\prime$ and $P^\prime$ are the distributions after $\alpha$-scale transformation corresponding to $Q$ and $P$ respectively. Thus, the $KL(Q\|P)$ in \eqref{eq:pac-bound1} remains invariant under $\alpha$-scale transformation. Note that the other terms apart from the $KL(Q\|P)$ in \eqref{eq:pac-bound1} do not change by $\alpha$-scale transformation since the functions represented by  the networks are the same. Thus, our bound is scale-invariant. That completes the proof. \qed

Recent work \cite{dipb17} show that 
$\alpha$-scale transformation can arbitrarily change the flatness of the loss landscape for deep networks
with positively homogeneous activation without changing the functions represented by the networks, which invalid many flatness-based generalization bound. Our generalization bound  remains invariant under $\alpha$-scale transformation since the KL-divergence between two continuous distributions remains invariant under invertible transformations, such as $\alpha$-scale transformation.

\section{De-randomized Margin Bounds: Multi-class Classification with Smooth Predictors}
\label{app:margink}

In this section, we focus on constructing the fast-rate deterministic bound using the anisotropic posterior case for multi-class problem, with  a similar de-randomization. Results for isotropic case shall be essentially the same.

Let $\phi: \R^p \times \R^d \mapsto \R^k$ be the output $\phi^{\theta}( x_i) \in \R^k$ of a deep net with parameter $\theta$ and input $x_i$. For a sample point $(x_i,y_i) \in \cX\times \cY $, where $\cX,\cY$ denotes the input and output space respectively, $\phi^{\theta}(x_i)[y_i]$ denotes the score corresponding to class $y_i$, and in general, $\phi^{\theta}(x_i)[h]$ denotes the score corresponding to class $h$. Note that the classification is correct when
\begin{equation}
    \phi^{\theta}(x_i)[y_i] > \phi^{\theta}(x_i)[h]~, ~~~\forall h \neq y_i~. 
\end{equation}
We define margin loss for a specific $z=(x,y)$ as
\begin{equation}
        \ell_{\beta}(\theta,z) \triangleq \1 \big[  \phi^{\theta}(x)[y] ~\leq ~\beta + \phi^{\theta}(x)[h]~, \forall h \neq y \big]~,
\end{equation}
where $\1[a] = 1$ if $a$ is true, and 0 otherwise. For a Bayesian predictor, we maintain a distribution $Q$ over the parameters $\theta$, and the corresponding margin loss
\begin{equation}
        \ell_{\beta}(Q,z) \triangleq \P_{\theta \sim Q}\big[  \phi^{\theta}(x)[y] ~\leq ~\beta + \phi^{\theta}(x)[h]~, \forall h \neq y \big]~.
\end{equation}

For any distribution $W$ on $\cX \times \cY$, and parameter $\theta$, we define the margin loss as
\begin{equation}
    \ell_{\beta}(\theta,W) \triangleq \P_{(x,y) \sim W}\bigg[  \phi^{\theta}(x)[y] ~\leq ~\beta + \phi^{\theta}(x)[h]~, \forall h \neq y \bigg]~.
\end{equation}
Further, for any distribution $W$ on $\cX \times \cY$, and any distribution over parameter $\theta$, we define the margin loss as
\begin{equation}
\begin{split}
    \ell_{\beta}(Q,W) \triangleq \P_{\theta \sim Q}[\ell_{\beta}(\theta,W)]  = \P_{\substack{\theta \sim Q\\ (x,y) \sim W}}\bigg[  \phi^{\theta}(x)[y] ~\leq ~\beta + \phi^{\theta}(x)[h]~, \forall h \neq y \bigg]~.
\end{split}
\end{equation}
We assume the Assumption~\ref{asmp:gen} holds for functions on each class $h\in[k]$, the proof idea is essentially the same as in the 2-class case, with constants changed.
\subsection{Bounds for Stochastic vs. Deterministic Predictors for Multi-class}

We first establishes the relationship between stochastic and deterministic predictors:

\begin{theo}
\label{theo:s_marg_mul}
For k-class classification problem, let $\sigma^2 > 0$ be chosen before seeing the training data. Let $W$ be any distribution on pairs $(\x,y)$ with $\x \in \R^d,\y \in \R^k$. For any $\theta^{\dagger} \in \R^d$, let $\cQ$ be a multivariate anisotropic Gaussian distribution with mean $\theta^{\dagger}$ and covariance $\Sigma_{\theta^{\dagger}}$, where $\Sigma_{\theta^{\dagger}} = \diag(\nu^2_j)$ with $\nu^2_j = \min\{\sigma^2, \sigma^2_j\}, \forall j \in [p]$,
Suppose Assumption~\ref{asmp:gen} holds for prediction function on each class $h\in[k]$, for any $\tilde \gamma > 2$ and any $\beta \in R$, we have 

\begin{equation}
\begin{split}
\ell_{\beta}\left(\cQ, W\right) &\leq \ell_{\beta + \sigma^2\zeta \alpha\tilde \gamma}(\phi^{\theta^{\dagger}},W)  + k\exp\left(- \min(c_2\tilde \gamma^2,c_1\tilde \gamma)\right)~,
\label{eq:non_conv_marg_swd}
\end{split}
\end{equation}
and,
\begin{equation}
\begin{split}
\ell_{\beta}(\phi^{\theta^{\dagger}},W)  &\leq  \ell_{\beta +\sigma^2\zeta \alpha\tilde \gamma }(\cQ,W) + k\exp\left(- \min(c_2\tilde \gamma^2,c_1\tilde \gamma)\right)~,
\label{eq:non_conv_marg_dws}
\end{split}
\end{equation} 
where constant $c_2 =  \min\left[\frac{\sigma^2\zeta^2 \alpha^2}{2G^2},\frac{ \alpha^2}{8\kappa}\right]$, $c_1= \frac{ \alpha}{4}$ and $G$, $\kappa$, $\alpha$, $\zeta$ are as in Assumption \ref{asmp:gen}.

\end{theo}

\proof The proof follows the proof of Theorem \ref{theo:s_marg}. 
Since $\phi^{\theta}(x)[h]$ is twice differentiable, for some suitable (random) $\tilde{\theta} = (1-\tau) \theta^{\dagger} + \tau \theta = \theta^{\dagger} + \tau (\theta - \theta^{\dagger})$ where $\tau \in [0,1]$, we have  

\begin{equation}
\phi^{\theta}(x)[h] = \phi^{\theta^{\dagger}}(x)[h] + \langle \theta - \theta^{\dagger}, \nabla \phi^{\theta^{\dagger}}(x)[h] \rangle +  \frac{1}{2}(\theta - \theta^{\dagger})^T H_{\phi}^{\tilde{\theta}}(x)[h] (\theta - \theta^{\dagger})~. 
\end{equation}

Consider the following set where $\theta^{\dagger}$ achieves a margin greater than $\left(\beta + \sigma^2\zeta \alpha\tilde \gamma\right)$:
\begin{equation}
\label{eq:hoe_hw_swd_mul}
\cZ^{(>)}_{\beta + \sigma^2\zeta \alpha\tilde \gamma}(\theta^{\dagger} ) = \bigg\{ (x,y) \in \cX \times \cY ~\bigg|~  \phi^{\theta^{\dagger}}(x)[y]-\phi^{\theta^{\dagger}}(x)[h] > \beta + \sigma^2\zeta \alpha\tilde \gamma, \forall h\neq y ~ \bigg\}~.
\end{equation} 
Let $P = \cN(0,\Sigma_{\theta^{\dagger}})$ be a multivariate distribution with mean 0 and covariance $\Sigma_{\theta^{\dagger}}$.  Now, for $z \in \cZ^{(>)}_{\beta + \sigma^2\zeta \alpha\tilde \gamma}(\theta^{\dagger} )$, we have
\begin{equation} \label{eq:repeat}
\begin{split}
&\P_{\delta \sim P}\bigg[  \phi^{\theta^{\dagger}+\delta}(x)[y]-\phi^{\theta^{\dagger}+\delta}(x)[h] ~\leq ~\beta,\forall h\neq y  ~\big|~ z \in \cZ^{(>)}_{\beta + \sigma^2\zeta \alpha\tilde \gamma}(\theta^{\dagger} ) \bigg] \\
& = \P_{\delta \sim P}\bigg[  \phi^{\theta^{\dagger}}(x)[y] + \langle \delta, \nabla \phi^{\theta^{\dagger}}(x)[y] \rangle + \frac{1}{2} \delta^T H_{\phi}^{\tilde \theta_1}(x)[y]  \delta-\left(\phi^{\theta^{\dagger}}(x)[h] + \langle \delta, \nabla \phi^{\theta^{\dagger}}(x)[h] \rangle +  \frac{1}{2}\delta^T H_{\phi}^{\tilde \theta_2}(x)[h]  \delta\right)\leq \beta,\\
&\qquad\forall h\neq y~\big|~ z \in \cZ^{(>)}_{\beta + \sigma^2\zeta \alpha\tilde \gamma}(\theta^{\dagger} ) \bigg] \\
& \leq  \P_{\delta \sim P}\bigg[   \langle \delta, \nabla \phi^{\theta^{\dagger}}(x)[y]-\nabla \phi^{\theta^{\dagger}}(x)[h] \rangle + \frac{1}{2} \delta^T 2H  \delta \leq \beta -\phi^{\theta^{\dagger}}(x)[y]+\phi^{\theta^{\dagger}}(x)[h],\forall h\neq y\big|~ z \in \cZ^{(>)}_{\beta + \sigma^2\zeta \alpha\tilde \gamma}(\theta^{\dagger} ) \bigg] \\
& \leq    \P_{\delta \sim P}\bigg[  \langle \delta, \nabla \phi^{\theta^{\dagger}}(x)[y]-\nabla \phi^{\theta^{\dagger}}(x)[h] \rangle + \frac{1}{2} \delta^T 2H \delta \leq -\sigma^2\alpha \tilde \gamma ,\forall h\neq y~\big|~ z \in \cZ^{(>)}_{\beta + \sigma^2\zeta \alpha\tilde \gamma}(\theta^{\dagger} ) \bigg] \\
& \leq    \P_{\delta \sim P}\bigg[  \langle \delta, \nabla \phi^{\theta^{\dagger}}(x)[y]-\nabla \phi^{\theta^{\dagger}}(x)[h] \rangle \leq -\frac{1}{2}\sigma^2 \alpha \tilde \gamma ,\forall h\neq y~\big|~ z \in \cZ^{(>)}_{\beta + \sigma^2\zeta \alpha\tilde \gamma}(\theta^{\dagger} ) \bigg] +\P_{\delta \sim P}\bigg[\delta^T 2H \delta \leq -\sigma^2 \alpha\tilde \gamma\bigg]\\
&\leq (k-1)\exp \left( -\frac{\sigma^2 \zeta^2 \alpha^2\tilde \gamma^2}{2G^2} \right)+\exp\left(- \frac{1}{2}\min \left[ \frac{  \alpha^2(\tilde \gamma-1)^2}{ \kappa}, \alpha(\tilde \gamma-1)\right] \right)
\end{split}
\end{equation}
where the last step is from Hoeffding's inequality with $\| \nabla \phi^{\theta^{\dagger}}(x)[y]-\phi^{\theta^{\dagger}}(x)[h] \|_2^2\leq 2\|\nabla \phi^{\theta^{\dagger}}(x)[y]\|_2^2 +2\|\phi^{\theta^{\dagger}}(x)[h]\|_2^2 \leq 4G^2$, $\max_i \|\delta_i\|_{\psi_2}\leq c_0 \sigma$, and taking union bound over all class $h\neq y$; and Lemma~\ref{lemm: up_lo_H}.

Therefore, based on Bayes rule, we have the following bound:
\begin{align*}
\ell_{\beta}(\cQ,W) &= \P_{\substack{\theta \sim \cQ\\ z \sim W}}\bigg[ \phi^{\theta^{\dagger}+\delta}(x)[y]-\phi^{\theta^{\dagger}+\delta}(x)[h] ~\leq ~\beta ~ ,\forall h\neq y\bigg] \\
& \leq   \P_{z \sim  W} \bigg[ \phi^{\theta^{\dagger}}(x)[y]-\phi^{\theta^{\dagger}}(x)[h] \leq \beta+\sigma^2\zeta \alpha\tilde \gamma,\forall h\neq y \bigg]\\
&\quad+  \P_{\substack{\theta \sim \cQ\\ z \sim W}} \bigg[ \phi^{\theta^{\dagger}+\delta}(x)[y]-\phi^{\theta^{\dagger}+\delta}(x)[h] ~\leq ~\beta ~ ,\forall h\neq y  ~\big|~ z \in \cZ^{(>)}_{\beta + \sigma^2\zeta \alpha\tilde \gamma}(\theta^{\dagger}) \bigg] \\
& =  \ell_{\beta + \sigma^2\zeta \alpha\tilde \gamma}(\phi^{\theta^{\dagger}},W) +(k-1)\exp \left( -\frac{\sigma^2 \zeta^2 \alpha^2\tilde \gamma^2}{2G^2} \right)+\exp\left(- \frac{1}{2}\min \left[ \frac{  \alpha^2(\tilde \gamma-1)^2}{ \kappa}, \alpha(\tilde \gamma-1)\right] \right)~.
\end{align*}
Similarly, consider the following set:
\begin{equation}
\cZ^{(\leq)}_{\beta}(\theta^{\dagger} ) = \bigg\{ (x,y) \in \cX \times \cY ~\bigg|~  \phi^{\theta^{\dagger}}(x)[y]-\phi^{\theta^{\dagger}}(x)[h] \leq \beta, \forall h\neq y ~ \bigg\}~.
\end{equation} 
Now, for $z \in \cZ^{(\leq)}_{\beta}(\theta^{\dagger} )$, 
and follow a similar argument as \eqref{eq:repeat}, we have
\begin{equation}
\label{eq:hoe_hw_dws_mul}
\begin{split}
&\P_{\delta \sim P}\bigg[  \phi^{\theta^{\dagger}+\delta}(x)[y]-\phi^{\theta^{\dagger}+\delta}(x)[h] >\beta+\sigma^2\alpha\tilde\gamma,\forall h\neq y  ~\big|~ z \in \cZ^{(\leq)}_{\beta}(\theta^{\dagger} ) \bigg] \\
& = \P_{\delta \sim P}\bigg[  \phi^{\theta^{\dagger}}(x)[y] + \langle \delta, \nabla \phi^{\theta^{\dagger}}(x)[y] \rangle +  \frac{1}{2}\delta^T H_{\phi}^{\tilde \theta_1}(x)[y]  \delta-\left(\phi^{\theta^{\dagger}}(x)[h] + \langle \delta, \nabla \phi^{\theta^{\dagger}}(x)[h] \rangle +  \frac{1}{2}\delta^T H_{\phi}^{\tilde \theta_2}(x)[h]  \delta\right)\\
&\qquad>\beta+\sigma^2\alpha\tilde\gamma,\forall h\neq y~\big|~ z \in \cZ^{(\leq)}_{\beta}(\theta^{\dagger} ) \bigg] \\
&\leq (k-1)\exp \left( -\frac{\sigma^2 \zeta^2 \alpha^2\tilde \gamma^2}{2G^2} \right)+\exp\left(- \frac{1}{2}\min \left[ \frac{  \alpha^2(\tilde \gamma-1)^2}{ \kappa}, \alpha(\tilde \gamma-1)\right] \right)~.
\end{split}
\end{equation}
where the last step is from Hoeffding's inequality with $\| \nabla \phi^{\theta^{\dagger}}(x)[y]-\phi^{\theta^{\dagger}}(x)[h] \|_2^2\leq 2\|\nabla \phi^{\theta^{\dagger}}(x)[y]\|_2^2 +2\|\phi^{\theta^{\dagger}}(x)[h]\|_2^2 \leq 4G^2$, $\max_i \|\delta_i\|_{\psi_2}\leq c_0 \sigma$, and taking union bound over all classes $h\neq y$; and Lemma~\ref{lemm: up_lo_H}. 

Similarly,
\begin{equation}
\begin{split}
1-  \ell_{\beta + \sigma^2\zeta \alpha\tilde \gamma }(\cQ,W)  &= \P_{\substack{\theta \sim \cQ\\ z \sim W}}\bigg[  \phi^{\theta^{\dagger}+\delta}(x)[y]-\phi^{\theta^{\dagger}+\delta}(x)[h] ~> ~\beta + \sigma^2\zeta \alpha\tilde \gamma ,\forall h\neq y~ \bigg] \\
& \leq \P_{\substack{\theta \sim \cQ\\ z \sim W}}\bigg[ \phi^{\theta^{\dagger}}(x)[y]-\phi^{\theta^{\dagger}}(x)[h] ~> ~\beta ,\forall h\neq y  ~\big|~ z \not \in \cZ^{(\leq)}_{\beta}(\theta^{\dagger}) \bigg]\\
&\qquad+  \P_{\substack{\theta \sim \cQ\\ z \sim W}}\bigg[\phi^{\theta^{\dagger}+\delta}(x)[y]-\phi^{\theta^{\dagger}+\delta}(x)[h] ~> ~\beta + \sigma^2\zeta \alpha\tilde \gamma ,\forall h\neq y~\big|~ z \in \cZ^{(\leq)}_{\beta}(\theta^{\dagger}) \bigg] \\
& = 1-\ell_{\beta}(\phi^{\theta^{\dagger}},W)+(k-1)\exp \left( -c_0\frac{\sigma^2 \alpha^2\tilde \gamma^2}{G^2} \right)+\exp\left(- c_0 \min \left[ \frac{ \alpha^2(\tilde \gamma-1)^2}{\eta_F^2}, \frac{ \alpha(\tilde \gamma-1)}{\eta_2} \right] \right)~,
\end{split}
\end{equation}
which implies
\begin{equation}
\ell_{\beta}(\phi^{\theta^{\dagger}},W)  \leq  \ell_{\beta + \sigma^2\zeta \alpha\tilde \gamma }(\cQ,W)+ (k-1)\exp \left( -\frac{\sigma^2 \zeta^2 \alpha^2\tilde \gamma^2}{2G^2} \right)+\exp\left(- \frac{1}{2}\min \left[ \frac{  \alpha^2(\tilde \gamma-1)^2}{ \kappa}, \alpha(\tilde \gamma-1)\right] \right)~.
\end{equation}
Choose $\tilde \gamma>2$, we have  $(\tilde\gamma-1)^2 >\frac{1}{4} \tilde\gamma^2$, $\tilde\gamma - 1>\frac{1}{2}\tilde\gamma$. 
That completes the proof.
\qed

\subsection{Deterministic Margin Bound for Multi-class}
With Theorem~\ref{theo:s_marg_mul}, we can construct the deterministic margin bound for multi-class problem.

\begin{theo}
\label{theo:smooth_mul}
For k-class classification problem, consider any $\theta_0 \in \R^p, \sigma^2 > 0$ chosen before training, and let $\theta^{\dagger}$ be the parameters of the model after training. 
Let  $\nu^2_j = \min\{\sigma^2, \sigma^2_j\}, \forall j \in [p]$.
Suppose Assumption~\ref{asmp:gen} holds for prediction function on every classes $h\in[k]$, then with probability at least $1-\delta$, for any $\theta^{\dagger}$, $\eta \in (0,1), \gamma > 2 \sigma^2 \alpha$, we have the following scale-invariant bound:
\begin{align}
\ell_{0}(\phi^{\theta^{\dagger}},D)  &\leq a_{\eta} \ell_{\gamma}(\phi^{\theta^{\dagger}},S) + 
\frac{b_{\eta}}{2n}\left( \sum_{j=1}^{p} \ln \frac{\nu_j^2}{1 / \sigma^{2}}+ \frac{ \| \theta^{\dagger}-\theta_0 \|_2^2}{\sigma^2 } \right) + d_{\eta}\exp \left( -\min(c_2 \gamma^2,c_1\gamma) \right)+ b_{\eta} \frac{\log (\frac{1}{\delta})}{n}~,
\end{align}
where $a_{\eta} = \frac{\log (1/\eta)}{1-\eta}, b_{\eta} = \frac{1}{1-\eta}$, $d_{\eta} = k(a_{\eta}+1)$, $c_2 =  \min\left[\frac{1}{2\sigma^2 G^2},\frac{1}{8\sigma^4r\zeta^2}\right],~c_1=\frac{1}{4\sigma^2\zeta}$ and $G$, $\zeta$, $\alpha$, $r$ are as in Assumption \ref{asmp:gen}.
\end{theo}

\proof The proof follows the proof of Theorem \ref{theo:smooth} with the use of Theorem \ref{theo:s_marg}
changed to Theorem \ref{theo:s_marg_mul}. 

\qed

\section{De-randomized Margin Bounds: Non-Smooth Predictor}
\label{app:margin2ns}

\subsection{Deterministic Bound for Non-smooth Predictors}

\theonsmarg*

We start with the following extension of the Hoeffding bound where the coefficients can depend on the randomness of prior random variables.
\begin{lemm}
Let $\{Z_t\}$ be a sub-Gaussian martingale difference sequence (MDS) and let $z_{1:t}$ denote a realization of $Z_{1:t}$. Let $\{a_t\}$ be a 
sequence of random variables such that $a_t = f_t(z_{1:(t-1)})$ for some sequence of functions $\{f_t\}$ with $|a_t| \leq \alpha_t$ a.s.~for suitable constants $\alpha_t, t=1,\ldots,T$. Then, for any $\tau > 0$, we have
\beq
\P \left( \left|\sum_{t=1}^T a_t z_t \right| \geq \tau \right) \leq 2 \exp \left\{ - \frac{\tau^2}{4c\kappa^2 \sum_{t=1}^T \alpha_t^2}  \right\}~,
\label{eq:dephef1}
\eeq
for some absolute constant $c > 0$ and where $\kappa$ is the $\psi_2$-norm of the conditional sub-Gaussian random variables.
\label{lem:dephef1}
\end{lemm}
\proof For any realization $z_{1:(t-1)}$ since $Z_t|z_{1:(t-1)}$ is a sub-Gaussian random variable with zero mean, then the conditional moment-generating function (MGF) satisfies: for all $s > 0$
\beq
\E[\exp(sZ_t) \mid z_{1:(t-1)}] \leq \exp(c s^2 \kappa^2)~,
\label{eq:mgf1}
\eeq
where $\kappa$ is $\psi_2$-norm of $Z_t$ conditioned on any realization $z_{1:(t-1)}$. Further, for $a_t = f(z_{1:(t-1)})$ with $|a_t| \leq \alpha_t$, we have
\begin{equation}
\E[\exp(sa_t Z_t) \mid z_{1:(t-1)}] \leq \exp(\frac{1}{2} a_t^2 s^2 \kappa^2)~,
\label{eq:mgf2}
\end{equation}
where the last inequality holds for all realiztions $z_{1:(t-1)}$. 

For any $s > 0$, note that
\begin{align}
\P \left( \sum_{t=1}^T a_t Z_t \geq \tau \right) &= \P\left( \exp\left( s\sum_{t=1}^T a_t Z_t \right) \geq \exp(s\tau) \right) \nonumber  \\ 
&\leq \exp(-s \tau) \E\left[ \exp\left( s\sum_{t=1}^T a_t Z_t \right) \right]~.
\label{eq:bnd1}
\end{align}
Now, using~\eqref{eq:mgf1}, we have 
\begin{align*}
\E\left[ \exp\left( s\sum_{t=1}^T a_t Z_t \right) \right] &= \E_{(Z_1,\ldots,Z_T)}\left[ \prod_{t=1}^T \exp (s a_t Z_t) \right] ~\\
&= \E_{(Z_1,\ldots,Z_{T-1})}\left[ E_{Z_T|Z_1,\ldots,Z_{T-1}}\left[ \exp( s a_T Z_T )\right] \prod_{t=1}^{T-1} \exp (s a_t Z_t) \right]~\\
& \leq \exp( \frac{1}{2} s^2 \alpha_T^2 \kappa^2) E_{(Z_1,\ldots,Z_{T-1})}\left[ \prod_{t=1}^{T-1} \exp (s a_t Z_t) \right]\\
& \leq  \exp( \frac{1}{2} s^2 \alpha_T^2 \kappa^2) \exp(\frac{1}{2} s^2 \alpha_{T-1}^2 \kappa^2) E_{(Z_1,\ldots,Z_{T-2})}\left[ \prod_{t=1}^{T-2} \exp (s a_t Z_t) \right] \\
& \leq \exp\left( \frac{1}{2} s^2 \kappa^2 \sum_{t=1}^T \alpha_t^2 \right)~.
\end{align*}
Plugging this back to \myref{eq:bnd1}, we have
\begin{align}
\P & \left( \sum_{t=1}^T a_t Z_t \geq \tau \right) \leq \exp \left(-s \tau + \frac{1}{2} s^2 \kappa^2 \sum_{t=1}^T \alpha_t^2 \right)~.
\end{align}
Choosing $s = \frac{\tau}{ \kappa^2 \sum_{t=1}^T \alpha_t^2}$, we obtain 
\begin{equation}
\P \left( \sum_{t=1}^T a_t Z_t \geq \tau \right) \leq \exp \left\{ -  \frac{\tau^2}{2\kappa^2 \sum_{t=1}^T \alpha_t^2} \right\}~.
\end{equation}

Repeating the same argument with $-Z_t$ instead of $X_t$, we obtain the same bound for $\P( - \sum_t a_t Z_t \geq \tau)$. Combining the two results gives us \myref{eq:dephef1}. \qed

We now consider a variant of the above result to a special case where the dependent random coefficients are binary and depends on the historical variables. 
\begin{corr}
\label{corr:mds_subg}
Let $\{Z_t\}$ be a sub-Gaussian martingale difference sequence (MDS) of length $T$ and let $z_{1:t}$ denote a realization of $Z_{1:t}$. Let $\{b_t\}$ be a sequence of binary random variables, i.e., $b_t \in \{0,1\}$ such that $b_t = f_t(z_{1:(t-1)}, b_{1:(1-t)})$ for some sequence of functions $\{f_t\}$. Then, for any $\tau > 0$ and any given $u = u_{1:T} \in \R^T$, we have
\beq
\P \left( \left|\sum_{t=1}^T b_t u_t z_t \right| \geq \tau \right) \leq 2 \exp \left\{ - \frac{\tau^2}{2\kappa^2 \| u \|_2^2}  \right\}~,
\label{eq:dephef2}
\eeq
where $\kappa$ is the $\psi_2$-norm of the conditional sub-Gaussian random variables.
\end{corr}
\proof The result follows by a direct application of Lemma~\ref{lem:dephef1} by a suitable choice of $\{a_t\}$ and $\{\alpha_t\}$. In particular, note that with $a_t = b_t u_t$ we have $|a_t| = |b_t u_t| \leq |u_t|$ so that with $\alpha_t = |u_t|$, $\sum_{t=1}^T \alpha_t^2 = \sum_{t=1}^T |u_t|^2 = \| u \|_2^2$. \qed

We now apply the result to deep nets. Let $\delta \in \R^p$ be a (sub)-Gaussian random vector corresponding to the parameters of a deep net. The indices of the components of $\delta$ are ordered by layers, so that if there are $k$ layers and $p_h, h=1,\ldots,k$ parameters in each of the layers,
\begin{itemize}
    \item $\delta_{1:p_1}$ correspond to parameters in the first layer, 
    \item $\delta_{(p_1+1):(p_1+p_2)}$ correspond to parameters in the second layer, and so on till 
    \item $\delta_{\left(\sum_{h=1}^{k-1} p_h+1\right):p}$ correspond to parameters in the last layer.
\end{itemize}
Consider a deep net $\psi^\delta$ with parameters $\delta$. For any input $x$, $\psi^\delta(x)$ can be equivalently written as a linear deep net with parameters $\phi^{\delta \odot \xi_x^\delta}(x)$ where $\xi_x^\delta \in \{0,1\}^p$ is a binary vector which is 0 for edges (connections) which are inactive and 1 for edges which are active. Our analysis will be primarily for a fixed $x$, and we will denote $\xi_x^\delta$ as $\xi^\delta$ for convenience, and clearly utilize the subscript when extending the analysis over all $x$.

For a ReLU-net with a fixed input $x$ and any parameters $\delta$, we have
\begin{equation}
    \xi_i^\delta = f_i(\delta_{1:(i-1)}, \xi_{1:(i-1)},x)~,
    \label{eq:recdn}
\end{equation}
for some suitable function $f_i$. In other words, whether an edge will be active or inactive for a given input $x$ depends on the earlier parameters 
$\delta_{1:(i-1)}$ and their active/inactive status $\xi_{1:(i-1)}$. In fact, for a ReLU-net, if $\xi_i$ is in layer $h, h=1,\ldots,k$, then $\xi_i$ only depends on parameters $\delta_{i'}$ and status $\xi_{i'}$ for edges (connections) in the earlier layers of the ReLU-net, i.e., layers $h'=1,\ldots,(h-1)$. In particular, such $\xi_i$ do not depend on parameters $\delta_{i'}$ and status $\xi_{i'}$ for edges in the same layer or subsequent layers. 

The above seemingly simple observation is a direct consequence of the structure of feed-forward networks, but implies the desired MDS structure we need to proceed with the analysis. In particular, an immediate consequence of the observation is that $\delta \odot \xi$ is sub-Gaussian for deep nets.  
\begin{corr}
\label{corr:subg_ns}
Let $\delta$ be an independent (Sub)-Gaussian random vector with $\max_i\|\delta_i\|_{\psi_2}=\kappa$, then for any fixed input $x$ and any $\theta^\dagger, u\in \R^p$, we have
\begin{equation}
    \P\bigg(\bigg|\langle \delta \odot \xi^{\theta^\dagger+\delta}, u\rangle\bigg|\geq \tau\bigg)\leq 2 \exp\bigg\{-\frac{\tau^2}{2\kappa^2\|u\|_2^2}\bigg\}~,
\end{equation}
which implies $\delta \odot \xi^{\theta^\dagger+\delta}$ is sub-Gaussian random vector with $\psi_2$-norm equals to $\kappa/\sqrt{2}$.
\end{corr}
\proof Note that $\langle \delta \odot \xi^{\theta^\dagger+\delta}, u\rangle = \sum_{i=1}^p  \xi_i^{\theta^\dagger + \delta} u_i \delta_i$. Comparing with Corollary \ref{corr:mds_subg}, using index $i$ instead of $t$, with $z_t = \delta_i$, $u_t = u_i$, and $b_t = \xi^{\theta^\dagger+\delta}_i$, which by \eqref{eq:recdn} follows the condition in Corollary \ref{corr:mds_subg}, the result follows by an application of Corollary \ref{corr:mds_subg}.\qed 

The corollary shows that for fixed $x$, while the components of $\delta \odot \xi^{\theta^\dagger + \delta}$ are not independent, the vector is sub-Gaussian and satisfies a Hoeffding-type inequality. To pull off something similar to our smooth case analysis, we now investigate if in addition to the Hoeffding-type inequality for linear forms of $\delta \odot \xi^{\theta^\dagger + \delta}$, can we establish Hanson-Wright type inequality for quadratic forms of $\delta \odot \xi^{\theta^\dagger + \delta}$. In the next result, we show that while the components of $\delta \odot \xi^{\theta^\dagger + \delta}$ are not independent, we can in fact establish a Hanson-Wright type inequality on quadratic forms of $\delta \odot \xi^{\theta^\dagger + \delta}$:


\begin{lemm} \label{lemm: up_lo_H_ani_ns}
For $\delta \sim \cN(0,\Sigma_{\theta^{\dagger}})$,where 
\begin{equation}
\Sigma_{\theta^{\dagger}}^{-1} = \diag(\nu_1^2,\ldots,\nu_p^2)~,\qquad \nu_j^2 \triangleq \max \left\{ \tilde \cH_{l,\phi}^{\theta^{\dagger}}[j,j], \frac{1}{\sigma^2} \right\}~,
\label{eq: cov_pos}
\end{equation}
where $\tilde \cH_{l,\phi}^{\theta^{\dagger}}$ is as in Example \ref{exa:ios_ns} and Assumption \ref{asmp:gen} holds, then we have the following bound
\begin{equation}
\begin{split}
    \P \left[  (\delta \odot \xi_x^{\theta^\dagger+\delta})^T H (\delta\odot \xi_x^{\theta^\dagger+\delta})   >  \sigma^2 \zeta\alpha\tilde \gamma  \right]   \leq  \exp\left(- \frac{1}{2} \min \left[ \frac{ \alpha^2(\tilde \gamma-1)^2}{\kappa}, \alpha(\tilde \gamma-1) \right] \right)~,
    \label{eq:ub_h}
    \end{split}
\end{equation}
where $\tilde \gamma >1$.
\end{lemm}
The traditional form of Hanson-Wright inequality which we use in the smooth case Lemma \ref{lemm: up_lo_H} does not hold for random vectors with dependent coordinates. Our way of dealing with this issue is bounding the moment generating function of quadratic form of dependent random vector by the MGF of quadratic form of independent random vector, so we can apply the Hanson-Wright inequality as usual. The sequential nautre of the parameters is again essential for our analysis.

{\em Proof of Lemma \ref{lemm: up_lo_H_ani_ns}.} For the quadratic form of $(\delta \odot \xi_x^{\theta^\dagger+\delta})^T H (\delta\odot \xi_x^{\theta^\dagger+\delta})$, i.e., 
\begin{equation}
\label{eq:bound_diag_detoindep}
    \sum_{i\neq j}h_{ij}\delta_i\delta_j \xi_i\xi_j+ \sum_{i}h_{ii}\delta_i^2 \xi_i^2~, 
\end{equation}
where $h_{ij}$ is the $i$-th row, $j$-th column component of $H$, and $\xi_i$ is short for $i$-th component of $\xi_x^{\theta^\dagger+\delta}$, the diagonal part can be easily bounded since $\xi_i\in \{0,1\}$ and $h_{ii}\geq 0$ ($H$ is positive semi-definite so the diagonal is non-negative.) :
\begin{equation}
    \sum_{i}h_{ii}\delta_i^2 \xi_i^2 \leq \sum_{i}h_{ii}\delta_i^2 ~,
\end{equation}
and the expectation can be bounded as
\begin{equation}
\label{eq:bound_ex_diag}
     \E \sum_{i}h_{ii}\delta_i^2 =  \sum_{i}h_{ii} \min\left(\sigma^2, \frac{1}{\tilde \cH_{l,\phi}^{\theta^{\dagger}}[i,i]}\right) \leq \sigma^2\tr(H)\leq \sigma^2\zeta\alpha~.
\end{equation}
Since $\delta_i$ are independent sub-Gaussian random variables, $\delta_i^2$ are independent sub-Exponential random variables. Thus,
\begin{equation}
    \|\delta_i^2 - \E\delta_i^2\|_{\psi_1} \leq \|\delta_i^2\|_{\psi_1} \leq \|\delta_i\|_{\psi_2}^2~, 
\end{equation}
then from Bernstein's inequality, we have
\begin{equation}
\label{eq:bound_diag}
    \P\left(\sum_i h_{ii}(\delta_{i}^2 - \E\delta_{i}^2 )\geq t/2\right) \leq \exp\left(-\frac{1}{2} \min \left(\frac{t^2}{\sum_i h_{ii}^2},\frac{t}{\max_i|h_{ii}|}\right)\right)\leq \exp\left(-\frac{1}{2} \min \left(\frac{t^2}{\|H\|_F^2},\frac{t}{\|H\|_2}\right)\right).
\end{equation}
For the off-diagonal part, without loss of generality, we assume $i<j$. Note that $\xi_j$ is a constant when $\delta_1,\cdots,\delta_{j-1}$ is fixed, we can prove each of the off-diagonal terms are a centered random variable by considering the conditional expectation:
\begin{align*}
    \E_{\delta}[h_{ij}\delta_i\delta_j \xi_i\xi_j] &= \E_{\delta_1,\cdots,\delta_j} [h_{ij}\delta_i\delta_j \xi_i\xi_j]\\
    &= \E_{\delta_1,\cdots,\delta_{j-1}}\E_{\delta_j|\delta_1,\cdots,\delta_{j-1}}[h_{ij}\delta_i\delta_j \xi_i\xi_j]\\
    &= \E_{\delta_1,\cdots,\delta_{j-1}} \left[h_{ij}\delta_i\xi_i \xi_j \E_{\delta_j|\delta_1,\cdots,\delta_{j-1}} [\delta_j] \right]\\
    &=0~.
\end{align*}
Next, we consider the moment generating function of off-diagonal part, i.e. $\E \exp(\lambda \sum_{1\leq i,j\leq p, i\neq j}h_{ij}\delta_i\delta_j \xi_i\xi_j)$, where $\lambda>0$. Conditioned on $\delta_1,\cdots,\delta_{p-1}$, $\xi_p\in \{0,1\}$ is a constant, and the randomness only comes from $\delta_p$ term. Noting the fact that exponential of linear function of $\delta_p$ is convex, denoting $\phi(a) = \exp(\lambda a)$, from Jensen's inequality on $\delta_p$ conditioned on $\delta_i, i<p$ and using the fact that $H$ is symmetric, we have
\begin{align*}
\hspace*{-5mm}
\phi\left( \sum_{1\leq i,j\leq p-1,i\neq j}h_{ij}\delta_i\delta_j \xi_i\xi_j\right) 
& = \phi\left( \sum_{1\leq i,j\leq p-1,i\neq j}h_{ij}\delta_i\delta_j \xi_i\xi_j +  \E_{\delta_p|\delta_i, i<p} \left[ 2 \sum_{1\leq i \leq (p-1)} h_{ip} \delta_i \delta_p \xi_i \xi_p \right] \right)\\
& \leq \E_{\delta_p|\delta_i, i<p} \phi\left( \sum_{1\leq i,j\leq p-1,i\neq j}h_{ij}\delta_i\delta_j \xi_i\xi_j +  2 \sum_{1\leq i\leq p-1}h_{ip}\delta_i \delta_p\xi_i\right) \\
 &\leq  \phi\left( \sum_{1\leq i,j\leq p-1,i\neq j}h_{ij}\delta_i\delta_j \xi_i\xi_j \right) \E_{\delta_p|\delta_i, i<p} \phi \left(  2\sum_{1\leq i\leq p-1}h_{ip}\delta_i \delta_p\xi_i\right)~,
\end{align*}
where we have only kept the terms corresponding to $\xi_p=1$ in the last expression since the terms corresponding to $\xi_p=0$ drop out. Now, we extend the sum on the left to include all terms and consider two cases, based on whether $\xi_p=0$ or 1. When $\xi_p=0$, we have
\begin{align*}
\phi\left( \sum_{1\leq i,j\leq p,i\neq j}h_{ij}\delta_i\delta_j \xi_i\xi_j\right) & = \phi\left( \sum_{1\leq i,j\leq p-1,i\neq j}h_{ij}\delta_i\delta_j \xi_i\xi_j\right) \\
& \leq \phi\left( \sum_{1\leq i,j\leq p-1,i\neq j}h_{ij}\delta_i\delta_j \xi_i\xi_j \right) \E_{\delta_p|\delta_i, i<p} \phi \left( 2 \sum_{1\leq i\leq p-1}h_{ip}\delta_i \delta_p\xi_i\right)~,
\end{align*}
and taking conditional expectation w.r.t.~$\delta_p|\delta_i, i<p$ does not change anything since 
\begin{align}
\E_{\delta_p|\delta_i, i<p}  \left[\phi\left( \sum_{1\leq i,j\leq p,i\neq j}h_{ij}\delta_i\delta_j \xi_i\xi_j\right) \right] 
\leq \phi\left( \sum_{1\leq i,j\leq p-1,i\neq j}h_{ij}\delta_i\delta_j \xi_i\xi_j \right) \E_{\delta_p|\delta_i, i<p} \phi \left( 2 \sum_{1\leq i\leq p-1}h_{ip}\delta_i \delta_p\xi_i\right)~.
\label{eq:cep1}
\end{align}
When $\xi_p=1$, we have 
\begin{align*}
\phi& \left( \sum_{1\leq i,j\leq p,i\neq j}h_{ij}\delta_i\delta_j \xi_i\xi_j\right) = \phi\left( \sum_{1\leq i,j\leq p-1,i\neq j}h_{ij}\delta_i\delta_j \xi_i\xi_j \right)
\phi \left( 2 \sum_{1\leq i\leq p-1}h_{ip}\delta_i \delta_p\xi_i \right) 
\end{align*}
so that by taking conditional expectation w.r.t.~$\delta_p|\delta_i, i<p$, we have
\begin{align}
\E_{\delta_p|\delta_i, i<p}&  \left[ \phi \left( \sum_{1\leq i,j\leq p,i\neq j}h_{ij}\delta_i\delta_j \xi_i\xi_j\right)  \right]
 = \phi\left( \sum_{1\leq i,j\leq p-1,i\neq j}h_{ij}\delta_i\delta_j \xi_i\xi_j \right) \left[ \E_{\delta_p|\delta_i, i<p} \phi \left( 2 \sum_{1\leq i\leq p-1}h_{ip}\delta_i \delta_p\xi_i\right) \right]~. 
\label{eq:cep2}
\end{align}
From \eqref{eq:cep1} and \eqref{eq:cep2}, we have
\begin{align}
\E_{\delta_{1:p}}&\left[ \phi\left( \sum_{1\leq i,j \leq p, i\neq j} h_{ij} \delta_i \delta_j \xi_i \xi_j  \right) \right] 
 \leq \E_{\delta_{1:p}}\left[ \phi\left( \sum_{1\leq i,j \leq (p-1), i\neq j} h_{ij} \delta_i \delta_j \xi_i \xi_j  +  2 \sum_{1\leq i\leq p-1}h_{ip}\delta_i \delta_p\xi_i\right)  \right] ~. 
\label{eq:step1}
\end{align}

%
%

Next we focus on $\xi_{p-1}$, and condition on $\delta_1,\cdots,\delta_{p-2}, \delta_p$, so that $\xi_{p-1}$ is fixed, so we have  
\begin{align*}
\hspace*{-10mm}
\phi & \left( \sum_{1\leq i,j\leq p-2,i\neq j}h_{ij}\delta_i\delta_j \xi_i\xi_j + 2 \sum_{1 \leq i \leq p-2} h_{ip} \delta_i \delta_p \xi_i \right)\\
=& \phi \left(\sum_{1\leq i,j\leq p-2,i\neq j}h_{ij}\delta_i\delta_j \xi_i\xi_j 
+ \E_{\delta_{p-1}|(\delta_i,i < p-1, \delta_p)} \left[2 \sum_{1\leq i\leq p-2}h_{i,p-1}\delta_i \delta_{p-1}\xi_i \xi_{p-1} + 2 h_{p-1,p} \delta_{p-1}\delta_p \xi_{p-1} \right] \right. \\
& \qquad \qquad \qquad \qquad \qquad  \left. +  2 \sum_{1\leq i\leq p-2}h_{ip}\delta_i \delta_p\xi_i  \right) \\
\leq~ & \E_{\delta_{p-1}|(d_{i, i < (p-1)},\delta_p)} \phi \left( \sum_{1\leq i,j\leq p-2,i\neq j}h_{ij}\delta_i\delta_j \xi_i\xi_j + 2 \sum_{1\leq i\leq p-2}h_{i,p-1}\delta_i \delta_{p-1}\xi_i + 2 h_{p-1,p} \delta_{p-1}\delta_p + 2 \sum_{1\leq i\leq p-2}h_{ip}\delta_i \delta_p\xi_i  \right) \\
= &  \phi \left( \sum_{1\leq i,j\leq p-2,i\neq j}h_{ij}\delta_i\delta_j \xi_i\xi_j +  2 \sum_{1\leq i\leq p-2}h_{ip}\delta_i \delta_p\xi_i  \right) \\
& \qquad
\times \E_{\delta_{p-1}|(d_{i, i < (p-1)},\delta_p)} \phi \left( 2 \sum_{1\leq i\leq p-2}h_{i,p-1}\delta_i \delta_{p-1}\xi_i + 2 h_{p-1,p} \delta_{p-1}\delta_p \right) ~
\end{align*}
where we again only keep the terms corresponding to $\xi_{p-1}=1$ since the terms with $\xi_{p-1}=0$ drop out. 
As before, we multiply both sides by the missing terms on the left hand side (LHS), and consider the cases $\xi_{p-1}=0$ and $\xi_{p-1}=1$. For $\xi_{p-1}=0$, we have 
\begin{align*}
\hspace*{-10mm}
\phi & \left( \sum_{1\leq i,j\leq p-1,i\neq j}h_{ij}\delta_i\delta_j \xi_i\xi_j + 2 \sum_{1 \leq i \leq p-1} h_{ip} \delta_i \delta_p \xi_i \right)\\
& = \phi  \left( \sum_{1\leq i,j\leq p-2,i\neq j}h_{ij}\delta_i\delta_j \xi_i\xi_j + 2 \sum_{1 \leq i \leq p-2} h_{ip} \delta_i \delta_p \xi_i \right)\\
& \leq \phi \left( \sum_{1\leq i,j\leq p-2,i\neq j}h_{ij}\delta_i\delta_j \xi_i\xi_j +  2 \sum_{1\leq i\leq p-2}h_{ip}\delta_i \delta_p\xi_i  \right) \\
& \qquad \qquad \qquad \times \E_{\delta_{p-1}|(d_{i, i < (p-1)},\delta_p)} \phi \left( 2 \sum_{1\leq i\leq p-2}h_{i,p-1}\delta_i \delta_{p-1}\xi_i + 2 h_{p-1,p} \delta_{p-1}\delta_p \right) ~,
\end{align*}
and taking conditional expectation w.r.t.~$\delta_{p-1}|(\delta_i, i < p-1,\delta_p)$ does not change anything since 
\begin{align}
&\E_{\delta_{p-1}|(d_{i, i < (p-1)},\delta_p)}  \phi  \left( \sum_{1\leq i,j\leq p-1,i\neq j}h_{ij}\delta_i\delta_j \xi_i\xi_j + 2 \sum_{1 \leq i \leq p-1} h_{ip} \delta_i \delta_p \xi_i \right) \nonumber \\
& \leq \phi \left( \sum_{1\leq i,j\leq p-2,i\neq j}h_{ij}\delta_i\delta_j \xi_i\xi_j +  2 \sum_{1\leq i\leq p-2}h_{ip}\delta_i \delta_p\xi_i  \right) \label{eq:cep3} \\
& \qquad \qquad \qquad \times \E_{\delta_{p-1}|(d_{i, i < (p-1)},\delta_p)} \phi \left( 2 \sum_{1\leq i\leq p-2}h_{i,p-1}\delta_i \delta_{p-1}\xi_i + 2 h_{p-1,p} \delta_{p-1}\delta_p \right) ~. \nonumber
\end{align}
For $\xi_{p-1}=1$, we have 
\begin{align*}
\hspace*{-5mm}
\phi & \left( \sum_{1\leq i,j\leq p-1,i\neq j}h_{ij}\delta_i\delta_j \xi_i\xi_j + 2 \sum_{1 \leq i \leq p-1} h_{ip} \delta_i \delta_p \xi_i \right)\\
& = \phi  \left( \sum_{1\leq i,j\leq p-2,i\neq j}h_{ij}\delta_i\delta_j \xi_i\xi_j + 2 \sum_{1\leq i\leq p-2}h_{i,p-1}\delta_i \delta_{p-1}\xi_i + 2 h_{p-1,p} \delta_{p-1}\delta_p + 2 \sum_{1 \leq i \leq p-2} h_{ip} \delta_i \delta_p \xi_i \right)\\
& = \phi \left( \sum_{1\leq i,j\leq p-2,i\neq j}h_{ij}\delta_i\delta_j \xi_i\xi_j 
+  2 \sum_{1\leq i\leq p-2}h_{ip}\delta_i \delta_p\xi_i  \right) 
\phi \left( 2 \sum_{1\leq i\leq p-2}h_{i,p-1}\delta_i \delta_{p-1}\xi_i + 2 h_{p-1,p} \delta_{p-1}\delta_p \right) ~.
\end{align*}
Taking expectations w.r.t.~$\delta_{p-1}|(\delta_i, i < p-1,\delta_p)$, we have
\begin{align}
\hspace*{-5mm}
& \E_{\delta_{p-1}|(d_{i, i < (p-1)},\delta_p)}  \phi \left( \sum_{1\leq i,j\leq p-1,i\neq j}h_{ij}\delta_i\delta_j \xi_i\xi_j + 2 \sum_{1 \leq i \leq p-1} h_{ip} \delta_i \delta_p \xi_i \right)\\
& = \phi \left( \sum_{1\leq i,j\leq p-2,i\neq j}h_{ij}\delta_i\delta_j \xi_i\xi_j +  2 \sum_{1\leq i\leq p-2}h_{ip}\delta_i \delta_p\xi_i  \right) \\
& \qquad \qquad \qquad \times \left[ \E_{\delta_{p-1}|(d_{i, i < (p-1)},\delta_p)} \phi \left( 2 \sum_{1\leq i\leq p-2}h_{i,p-1}\delta_i \delta_{p-1}\xi_i + 2 h_{p-1,p} \delta_{p-1}\delta_p \right) \right]~.
\label{eq:cep4}
\end{align}
From~\eqref{eq:step1}, using \eqref{eq:cep3} and \eqref{eq:cep4}, we have 
\begin{align}
\hspace*{-10mm}
& \E_{\delta_{1:p}}\left[ \phi\left( \sum_{1\leq i,j \leq p, i\neq j} h_{ij} \delta_i \delta_j \xi_i \xi_j  \right) \right]  \\
& \leq \E_{\delta_{1:p}}\left[ \phi\left( \sum_{1\leq i,j \leq (p-1), i\neq j} h_{ij} \delta_i \delta_j \xi_i \xi_j  +  2 \sum_{1\leq i\leq p-1}h_{ip}\delta_i \delta_p\xi_i\right)  \right] \\
& \leq \E_{\delta_{1:p}}\left[ \phi\left( \sum_{1\leq i,j \leq (p-2), i\neq j} h_{ij} \delta_i \delta_j \xi_i \xi_j + 2 \sum_{1\leq i\leq p-2}h_{i,p-1}\delta_i \delta_{p-1}\xi_i + 2 h_{p-1,p} \delta_{p-1}\delta_p  \right)  \right]~.
\label{eq:step2}
\end{align}
%
%
%

Continuing in the same manner all the way and considering the same analysis for $\delta_{p-2},\ldots,\delta_1$, using $\phi(a) = \exp(\lambda a)$, we have
\begin{equation}
\E_{\delta_{1:p}} \exp \left(\lambda \sum_{1\leq i,j\leq p, i\neq j}h_{ij}\delta_i\delta_j \xi_i\xi_j \right) 
\leq \E_{\delta_{1:p}} \exp\left( \lambda \sum_{1\leq i,j\leq p, i\neq j}h_{ij}\delta_i\delta_j \right)~.
\end{equation}


Then we can do decoupling on the independent random vector $\delta$ as in classical Hanson-Wright inequality,  see Theorem 6.2.1 in \citep{vers18}, we have 
\begin{align*}
    \E \exp(\lambda \sum_{1\leq i,j\leq p, i\neq j}h_{ij}\delta_i\delta_j) &\leq \E \exp(4\lambda \delta^T H \delta^{\prime}) \quad \text{(by decoupling - Remark 6.1.3 in \citep{vers18})}\\
    & \leq \E \exp(\lambda g^T H g^{\prime}) \quad \text{(by comparison lemma - Lemma 6.2.3 in \citep{vers18})}\\
    & \leq \exp(\frac{1}{2}\lambda^2 \|H\|_F^2) \quad \text{(by bound on Gaussian chaos - Lemma 6.2.2 in \citep{vers18})},
\end{align*}
provided that $|\lambda|\leq 1/2\|H\|_2$, where $g$ is standard Gaussian random vector and $\delta^\prime, g^\prime$ are independent copy of $\delta,g$. Therefore, we have
\begin{align*}
    \P\left(\sum_{1\leq i,j\leq p, i\neq j}h_{ij}\delta_i\delta_j \xi_i\xi_j \geq t/2\right) &= \P \left(\lambda \sum_{1\leq i,j\leq p, i\neq j}h_{ij}\delta_i\delta_j \xi_i\xi_j \geq \lambda t/2\right) \\
    &\leq \exp(-\lambda t/2)\E \exp\left(\lambda \sum_{1\leq i,j\leq p, i\neq j}h_{ij}\delta_i\delta_j \xi_i\xi_j\right)\\
    & \leq \exp(-\lambda t/2)\E \exp\left(\lambda \sum_{1\leq i,j\leq p, i\neq j}h_{ij}\delta_i\delta_j\right)\\
    &\leq \exp(-\lambda t/2) \exp(\frac{1}{2} \lambda^2 \|H\|_F^2)~.
\end{align*}
Optimizing over $0\leq \lambda \leq c/\|H\|_2$, we conclude that
\begin{equation}
\label{eq:bound_offd}
    \P\left(\sum_{1\leq i,j\leq p, i\neq j}h_{ij}\delta_i\delta_j \xi_i\xi_j \geq t/2\right)\leq \exp\left(-\frac{1}{2} \min (\frac{t^2}{\|H\|_F^2},\frac{t}{\|H\|_2})\right)~.
\end{equation}
Therefore, gathering deviation bound on diagonal term, bound on expectation of diagonal term, and deviation bound on off-diagonal term, we have deviation bound on the quadratic form:
\begin{align*}
    &\P\left[(\delta\odot \xi_x^{\theta^\dagger+\delta})^TH(\delta\odot \xi_x^{\theta^\dagger+\delta})-\sigma^2\zeta\alpha\geq t\right]\\
    \overset{(a)}{\leq}~&\P\left[(\delta\odot \xi_x^{\theta^\dagger+\delta})^TH(\delta\odot \xi_x^{\theta^\dagger+\delta})-\E \sum_{i}h_{ii}\delta_i^2\geq t\right]\\
    \overset{(b)}{\leq}~& \P\left(\sum_i h_{ii}(\delta_{i}^2 - \E\delta_{i}^2 )\geq t/2\right) + \P\left(\sum_{1\leq i,j\leq p, i\neq j}h_{ij}\delta_i\delta_j \xi_i\xi_j \geq t/2\right) \\
    \overset{(c)}{\leq}~&\exp\left(-\frac{1}{2} \min \left(\frac{t^2}{\|H\|_F^2},\frac{t}{\|H\|_2}\right)\right) + \exp\left(-\frac{1}{2} \min \left(\frac{t^2}{\|H\|_F^2},\frac{t}{\|H\|_2}\right)\right)\\
    =~& 2\exp\left(-\frac{1}{2} \min \left(\frac{t^2}{\|H\|_F^2},\frac{t}{\|H\|_2}\right)\right)~,
\end{align*}
where (a) is from Equation \eqref{eq:bound_ex_diag}, (b) is from Equation \eqref{eq:bound_diag_detoindep}, (c) is from Equation \eqref{eq:bound_diag}\eqref{eq:bound_offd}.

By taking $t=\tilde \gamma\sigma^2 \zeta\alpha(>0)$, we have
\begin{equation}
\begin{split}
    \P[(\delta\odot \xi_x^{\theta^\dagger+\delta})^TH(\delta\odot \xi_x^{\theta^\dagger+\delta})\geq (\tilde \gamma+1)\sigma^2\zeta\alpha]\leq\exp\left(-\frac{1}{2}\min\left[\frac{ \alpha^2\tilde \gamma^2}{\kappa},\alpha\tilde \gamma\right]\right)~.
\end{split}
\end{equation}
Denote $\tilde \gamma+1$ as $\tilde \gamma$ completes the proof. \qed

{\em Proof of Theorem~\ref{theo:ns_marg}:} Since $\phi$ is twice differentiable, we have  
\begin{equation}
	\phi^{\vartheta_x}(x) = \phi^{\vartheta_x^{\dagger}}(x) + \langle \vartheta_x  - \vartheta_x^{\dagger}, \nabla \phi^{\vartheta_x^{\dagger}}(x) \rangle +  \frac{1}{2}(\vartheta_x - \vartheta_x^{\dagger})^T H_{\phi}^{\tilde{\vartheta}_x}(x) (\vartheta_x - \vartheta_x^\dagger)~. 
\end{equation}
where
\begin{equation}
\vartheta_x \triangleq \theta \odot \xi_{x}^\theta \qquad \text{and} \qquad \vartheta_x^{\dagger} \triangleq \theta^{\dagger} \odot \xi_x^{\theta^{\dagger}}~,
\end{equation}
and $\tilde{\vartheta}_x = (1-\tau) \vartheta_x^{\dagger} + \tau \vartheta_x = \vartheta_x^{\dagger} + \tau (\vartheta_x - \vartheta_x^{\dagger})$ for some $\tau \in [0,1]$.

Now consider the following set where $\theta^{\dagger}$ achieves a margin greater than $\left(\beta + \frac{3}{2}\sigma^2\zeta\alpha\tilde \gamma +G\|\theta^\dagger\|_2+\frac{1}{2}\zeta \|\theta^\dagger\|_2^2 \right)$:
\begin{equation}
\cZ^{(>)}_{\beta + \frac{3}{2}\sigma^2\zeta\alpha\tilde \gamma+G\|\theta^\dagger\|_2+\frac{1}{2}\zeta \|\theta^\dagger\|_2^2}(\theta^{\dagger} ) = \bigg\{ (x,y) \in \cX \times \cY ~\bigg|~  y \phi^{\vartheta^{\dagger}_x}(x) > \beta + \frac{3}{2}\sigma^2\zeta\alpha\tilde \gamma +G\|\theta^\dagger\|_2+\frac{1}{2}\zeta \|\theta^\dagger\|_2^2 ~ \bigg\}~.
\end{equation}

Let $P = N(0,\Sigma_{\theta^\dagger})$ be a multivariate Gaussian distribution with mean 0 and covariance $\Sigma_{\theta^\dagger}$. Let $Q = N(\theta^\dagger, \Sigma_{\theta^\dagger})$ be the posterior for the PAC-Bayes analysis. Note that $\theta \sim Q$ is same as $\theta = \theta^\dagger + \delta$, where $\delta \sim P$. Now we have
\begin{align*}
	&\P_{\theta \sim \cQ}\bigg[ y \phi^{\theta \odot \xi_{x}^\theta}(x) ~\leq ~\beta  ~\big|~ z \in \cZ^{(>)}_{\beta + \frac{3}{2}\sigma^2\zeta\alpha\tilde \gamma +G\|\theta^\dagger\|_2+\frac{1}{2}\zeta \|\theta^\dagger\|_2^2}(\theta^{\dagger} ) \bigg] \\
	= &\P_{\delta \sim P}\bigg[ y \phi^{(\theta^{\dagger}+\delta) \odot (\xi_{x}^{\theta^\dagger+\delta})}(x) ~\leq ~\beta  ~\big|~ z \in \cZ^{(>)}_{\beta + \frac{3}{2}\sigma^2\zeta\alpha\tilde \gamma+G\|\theta^\dagger\|_2+\frac{1}{2}\zeta \|\theta^\dagger\|_2^2}(\theta^{\dagger} ) \bigg] ~,
\end{align*}
where, by definition,
\begin{equation}
\phi^{(\theta^{\dagger}+\delta) \odot (\xi_{x}^{\theta^\dagger+\delta})}(x) = \psi^{\theta^\dagger + \delta}(x)~.
\end{equation} 
Further, with $\theta = \theta^\dagger + \delta$, note that we have
\begin{align}
&\phi^{\vartheta_x}(x) 
\nr = \phi^{\vartheta_x^{\dagger}}(x) + \langle \vartheta_x  - \vartheta_x^{\dagger}, \nabla \phi^{\vartheta_x^{\dagger}}(x) \rangle + \frac{1}{2} (\vartheta_x - \vartheta_x^{\dagger})^T H_{\phi}^{\tilde{\vartheta}_x}(x) (\vartheta_x - \vartheta_x^\dagger)\\
\nr & \geq \phi^{\vartheta_x^{\dagger}}(x) + \langle \vartheta_x  - \vartheta_x^{\dagger}, \nabla \phi^{\vartheta_x^{\dagger}}(x) \rangle - \frac{1}{2} (\vartheta_x - \vartheta_x^{\dagger})^T H (\vartheta_x - \vartheta_x^\dagger)\\
\nr & = \phi^{\vartheta_x^{\dagger}}(x) + \langle (\theta^{\dagger}+\delta) \odot \xi_{x}^{\theta^\dagger+\delta}  - \theta^{\dagger} \odot \xi_{x}^{\theta^\dagger}, \nabla \phi^{\vartheta_x^{\dagger}}(x) \rangle\\
\nr &\quad-  \frac{1}{2}((\theta^{\dagger}+\delta) \odot \xi_{x}^{\theta^\dagger+\delta}  - \theta^{\dagger} \odot \xi_{x}^{\theta^\dagger})^T H ((\theta^{\dagger}+\delta) \odot \xi_{x}^{\theta^\dagger+\delta}  - \theta^{\dagger} \odot \xi_{x}^{\theta^\dagger})\\
\nr & = \phi^{\vartheta_x^{\dagger}}(x) + \langle \delta \odot \xi_{x}^{\theta^\dagger+\delta}, \nabla \phi^{\vartheta_x^{\dagger}}(x) \rangle +\langle \theta^\dagger \odot (\xi_{x}^{\theta^\dagger+\delta}-\xi_x^{\theta^\dagger}), \nabla \phi^{\vartheta_x^{\dagger}}(x) \rangle -\frac{1}{2} (\delta \odot \xi_{x}^{\theta^\dagger+\delta})^T H (\delta \odot \xi_{x}^{\theta^\dagger+\delta} )\\
\nr &\quad-  \frac{1}{2}( \theta^{\dagger} \odot (\xi_{x}^{\theta^\dagger+\delta} - \xi_{x}^{\theta^\dagger}))^T H ( \theta^{\dagger} \odot (\xi_{x}^{\theta^\dagger+\delta} - \xi_{x}^{\theta^\dagger})) -\langle \delta\odot \xi_{x}^{\theta^\dagger+\delta},H( \theta^{\dagger} \odot (\xi_{x}^{\theta^\dagger+\delta} - \xi_{x}^{\theta^\dagger}))\rangle \\
\nr & = \phi^{\vartheta_x^{\dagger}}(x) + \langle \delta \odot \xi_{x}^{\theta^\dagger+\delta}, \nabla \phi^{\vartheta_x^{\dagger}}(x) \rangle +\langle \theta^\dagger \odot (\xi_{x}^{\theta^\dagger+\delta}-\xi_x^{\theta^\dagger}), \nabla \phi^{\vartheta_x^{\dagger}}(x) \rangle - \frac{1}{2}(\delta \odot \xi_{x}^{\theta^\dagger+\delta})^T H (\delta \odot \xi_{x}^{\theta^\dagger+\delta} )\\
&\quad- \frac{1}{2} ( \theta^{\dagger} \odot (\xi_{x}^{\theta^\dagger+\delta} - \xi_{x}^{\theta^\dagger}))^T H ( \theta^{\dagger} \odot (\xi_{x}^{\theta^\dagger+\delta} - \xi_{x}^{\theta^\dagger})) -\langle \delta\odot \xi_{x}^{\theta^\dagger+\delta}\odot (\xi_{x}^{\theta^\dagger+\delta} - \xi_{x}^{\theta^\dagger}),H \theta^{\dagger} \rangle~, 
\end{align}
where the last step is from the property of Hadamard product. Note that $\theta^\dagger \odot (\xi_{x}^{\theta^\dagger+\delta}-\xi_x^{\theta^\dagger})$ is a random vector with $\theta^\dagger_i$ be multiplied by $\{-1,0,1\}$ so that $\|\theta^\dagger \odot (\xi_{x}^{\theta^\dagger+\delta}-\xi_x^{\theta^\dagger})\|_2\leq \|\theta^\dagger\|_2$, almost surely. Therefore, from Cauchy-Schwarz inequality, we have almost surely
\begin{align*}
    &\quad\phi^{\vartheta_x}(x)\\
    &\geq \phi^{\vartheta_x^{\dagger}}(x) + \langle \delta \odot \xi_{x}^{\theta^\dagger+\delta}, \nabla \phi^{\vartheta_x^{\dagger}}(x) \rangle - \frac{1}{2}(\delta \odot \xi_{x}^{\theta^\dagger+\delta})^T H (\delta \odot \xi_{x}^{\theta^\dagger+\delta} )-\langle \delta\odot \xi_{x}^{\theta^\dagger+\delta}\odot (\xi_{x}^{\theta^\dagger+\delta} - \xi_{x}^{\theta^\dagger}),H \theta^{\dagger} \rangle\\
    &\quad- \|\theta^\dagger \odot (\xi_{x}^{\theta^\dagger+\delta}-\xi_x^{\theta^\dagger})\|_2\| \nabla \phi^{\vartheta_x^{\dagger}}(x)\|_2-\frac{1}{2}\|H\|_2\|\theta^\dagger \odot (\xi_{x}^{\theta^\dagger+\delta}-\xi_x^{\theta^\dagger})\|_2^2\\
    &\geq \phi^{\vartheta_x^{\dagger}}(x) + \langle \delta \odot \xi_{x}^{\theta^\dagger+\delta}, \nabla \phi^{\vartheta_x^{\dagger}}(x) \rangle - \frac{1}{2} (\delta \odot \xi_{x}^{\theta^\dagger+\delta})^T H (\delta \odot \xi_{x}^{\theta^\dagger+\delta} )-\langle \delta\odot \xi_{x}^{\theta^\dagger+\delta}\odot (\xi_{x}^{\theta^\dagger+\delta} - \xi_{x}^{\theta^\dagger}),H \theta^{\dagger} \rangle\\& - G\|\theta^\dagger\|_2 -\frac{1}{2}\zeta \|\theta^\dagger\|_2^2~.
\end{align*}

Notice that $\xi_{x}^{\theta^\dagger+\delta}\odot (\xi_{x}^{\theta^\dagger+\delta} - \xi_{x}^{\theta^\dagger})$ is also a MDS similar to $\xi_{x}^{\theta^\dagger+\delta}$. Therefore, For a fixed $x$, conditioned on $y=+1$, we have 
\begin{equation}\label{eq:dont_know_what_to_call}
\begin{split}
&\P_{\delta \sim P}\bigg[  \phi^{\vartheta_x}(x) ~\leq ~\beta  ~\big|~ z \in \cZ^{(>)}_{\beta + \frac{3}{2}\sigma^2\zeta\alpha\tilde \gamma + G\|\theta^\dagger\|_2+\frac{1}{2}\zeta \|\theta^\dagger\|_2^2}(\theta^{\dagger} ), y = +1 \bigg] \\
& = \P_{\delta \sim P}\bigg[  \phi^{\vartheta_x^{\dagger}}(x) + \langle \delta \odot \xi_{x}^{\theta^\dagger+\delta}, \nabla \phi^{\vartheta_x^{\dagger}}(x) \rangle - \langle \delta\odot \xi_{x}^{\theta^\dagger+\delta}\odot (\xi_{x}^{\theta^\dagger+\delta} - \xi_{x}^{\theta^\dagger}),H \theta^{\dagger} \rangle\\
&- \frac{1}{2}(\delta \odot \xi_{x}^{\theta^\dagger+\delta})^T H (\delta \odot \xi_{x}^{\theta^\dagger+\delta} ) - G\|\theta^\dagger\|_2-\frac{1}{2}\zeta \|\theta^\dagger\|_2^2 \leq \beta ~\big|~ z \in \cZ^{(>)}_{\beta + \frac{3}{2}\sigma^2\zeta\alpha\tilde \gamma + G\|\theta^\dagger\|_2+\frac{1}{2}\zeta \|\theta^\dagger\|_2^2}(\theta^{\dagger} ),  y = +1 \bigg] \\
& \leq  \P_{\delta \sim P}\bigg[   \langle \delta \odot \xi_{x}^{\theta^\dagger+\delta}, \nabla \phi^{\vartheta_x^{\dagger}}(x) \rangle - \langle \delta\odot \xi_{x}^{\theta^\dagger+\delta}\odot (\xi_{x}^{\theta^\dagger+\delta} - \xi_{x}^{\theta^\dagger}),H \theta^{\dagger} \rangle - \frac{1}{2}(\delta \odot \xi_{x}^{\theta^\dagger+\delta})^T H (\delta \odot \xi_{x}^{\theta^\dagger+\delta} )\leq \\
&-\sigma^2 \tilde \gamma \zeta\alpha ~\big|~ z \in \cZ^{(>)}_{\beta + \frac{3}{2}\sigma^2\zeta\alpha\tilde \gamma+G\|\theta^\dagger\|_2+\frac{1}{2}\zeta \|\theta^\dagger\|_2^2}(\theta^{\dagger} ), y = +1 \bigg] \\
& \leq    \P_{\delta \sim P}\bigg[   \langle \delta \odot \xi_{x}^{\theta^\dagger+\delta}, \nabla \phi^{\vartheta_x^{\dagger}}(x) \rangle \leq -\frac{1}{2}\sigma^2 \tilde \gamma \zeta\alpha~\big|~ z \in \cZ^{(>)}_{\beta + \frac{3}{2}\sigma^2\zeta\alpha\tilde \gamma +G\|\theta^\dagger\|_2+\frac{1}{2}\zeta \|\theta^\dagger\|_2^2}(\theta^{\dagger} ), y = +1 \bigg]~ \\
& \quad +   \P_{\delta \sim P}\bigg[    -\langle \delta\odot \xi_{x}^{\theta^\dagger+\delta}\odot (\xi_{x}^{\theta^\dagger+\delta} - \xi_{x}^{\theta^\dagger}),H \theta^{\dagger} \rangle \leq -\frac{1}{2}\sigma^2 \tilde \gamma \zeta\alpha~\big|~ z \in \cZ^{(>)}_{\beta + \frac{3}{2}\sigma^2\zeta\alpha\tilde \gamma +G\|\theta^\dagger\|_2+\frac{1}{2}\zeta \|\theta^\dagger\|_2^2}(\theta^{\dagger} ), y = +1 \bigg]~ \\
&\quad +   \P_{\delta \sim P}\bigg[  -\frac{1}{2}(\delta \odot \xi_{x}^{\theta^\dagger+\delta})^T H (\delta \odot \xi_{x}^{\theta^\dagger+\delta} ) \leq -\frac{1}{2}\sigma^2 \zeta\alpha \tilde \gamma ~\big|~ z \in \cZ^{(>)}_{\beta + \frac{3}{2}\sigma^2\zeta\alpha\tilde \gamma+G\|\theta^\dagger\|_2+\frac{1}{2}\zeta \|\theta^\dagger\|_2^2}(\theta^{\dagger} ), y = +1 \bigg]\\
& \overset{(a)}{\leq} \exp \left( -\frac{1}{2}\frac{\sigma^2 \zeta^2\alpha^2\tilde \gamma^2}{\|\nabla \phi^{\vartheta_x^{\dagger}}(x)\|_2^2} \right) +\exp \left( -\frac{1}{2}\frac{\sigma^2 \zeta^2\alpha^2\tilde \gamma^2}{\|H\theta^\dagger\|_2^2} \right) +\exp\left(- \frac{1}{2} \min \left[ \frac{ \alpha^2(\tilde \gamma-1)^2}{\kappa}, \alpha(\tilde \gamma-1) \right] \right)~\\
& \leq \exp \left( -\frac{1}{2}\frac{\sigma^2 \zeta^2\alpha^2\tilde \gamma^2}{G^2} \right)+\exp \left( -\frac{1}{2}\frac{\sigma^2 \zeta^2\alpha^2\tilde \gamma^2}{\|\theta^\dagger\|_2^2} \right)+ \exp\left(- \frac{1}{2} \min \left[ \frac{ \alpha^2(\tilde \gamma-1)^2}{\kappa},  \alpha(\tilde \gamma-1) \right] \right)~,
\end{split}\end{equation}
where step (a) is from Corollary \ref{corr:subg_ns} with Cauchy-Schwarz inequality, and Lemma \ref{lemm: up_lo_H_ani_ns}.

Similarly, conditioned on $y=-1$, we have
\begin{align*}
&\P_{\delta \sim P}\bigg[  -\phi^{\vartheta_x}(x) ~\leq ~\beta  ~\big|~ z \in \cZ^{(>)}_{\beta + \frac{3}{2}\sigma^2\zeta\alpha\tilde \gamma+ G\|\theta^\dagger\|_2+\frac{1}{2}\zeta \|\theta^\dagger\|_2^2}(\theta^{\dagger} ), y = -1 \bigg] \\
&\leq \exp \left( -\frac{1}{2}\frac{\sigma^2 \zeta^2\alpha^2\tilde \gamma^2}{G^2} \right)+\exp \left( -\frac{1}{2}\frac{\sigma^2 \alpha^2\tilde \gamma^2}{\|\theta^\dagger\|_2^2} \right)+ \exp\left(- \frac{1}{2} \min \left[ \frac{ \alpha^2(\tilde \gamma-1)^2}{\kappa},  \alpha(\tilde \gamma-1) \right] \right)~.
\end{align*}

The following analysis is similar as in smooth case, we have
\begin{align*}
&\ell_{\beta}(\cQ,W)\\
& = \P_{\substack{\theta \sim \cQ\\ z \sim W}}\bigg[  y\phi^{\vartheta_x}(x) ~\leq ~\beta ~ \bigg] \\
& \leq   \P_{z \sim  W} \bigg[ y \phi^{\vartheta^{\dagger}_x}(x) \leq \beta + \frac{3}{2}\sigma^2\zeta\alpha\tilde \gamma+G\|\theta^\dagger\|_2+\frac{1}{2}\zeta \|\theta^\dagger\|_2^2 \bigg]+  \P_{\substack{\theta \sim \cQ\\ z \sim W}} \bigg[ y \phi^{\vartheta_x}(x) ~\leq ~\beta  ~\big|~ z \in \cZ^{(>)}_{\beta + \frac{3}{2}\sigma^2\zeta\alpha\tilde \gamma+G\|\theta^\dagger\|_2+\zeta \|\theta^\dagger\|_2^2}(\theta^{\dagger}) \bigg] \\
& =  \ell_{\beta + \frac{3}{2}\sigma^2\zeta\alpha\tilde \gamma+G\|\theta^\dagger\|_2+\frac{1}{2}\zeta \|\theta^\dagger\|_2^2}(\phi^{\vartheta_x^{\dagger}},W)+2\exp \left( -\frac{\sigma^2 \zeta^2\alpha^2\tilde \gamma^2}{2G^2} \right)+2\exp \left( -\frac{\sigma^2 \alpha^2\tilde \gamma^2}{2\|\theta^\dagger\|_2^2} \right)+ 2\exp\left(-\frac{1}{2}  \min \left[ \frac{ \alpha^2(\tilde \gamma-1)^2}{\kappa},  \alpha(\tilde \gamma-1) \right] \right).
\end{align*}
and recalling that $\phi^{\vartheta_x}(x) = \psi^{\theta}(x)$ establishes Equation \eqref{eq:ns_aniso_swd}, the proof of Equation \eqref{eq:ns_aniso_dws} is similar.

For 2-layer neural network, we have the network is equivalent to a quadratic form of parameters (with de-activation), which means for any $\vartheta$, we have 
\begin{equation}
\label{eq:2_layer_property}
  \nabla \phi^{\vartheta}(x) =  H_{\phi}^{\vartheta} \vartheta = H \vartheta~, 
\end{equation}
where $H$ is the deterministic hessian matrix. Therefore, replace the $H\vartheta$ by $\nabla \phi^{\vartheta}(x)$ (bounding hessian by deterministic matrix is not needed) in the proof of layer $k\geq 3$ cases, we have
\begin{align*}
	\phi^{\vartheta_x}(x)
	& =  \phi^{\vartheta_x^{\dagger}}(x) + \langle \delta \odot \xi_{x}^{\theta^\dagger+\delta}, \nabla \phi^{\vartheta_x^{\dagger}}(x) \rangle +\langle \theta^\dagger \odot (\xi_{x}^{\theta^\dagger+\delta}-\xi_x^{\theta^\dagger}), \nabla \phi^{\vartheta_x^{\dagger}}(x) \rangle+ \frac{1}{2}(\delta \odot \xi_{x}^{\theta^\dagger+\delta})^T H (\delta \odot \xi_{x}^{\theta^\dagger+\delta} )\\
& \quad +  \frac{1}{2}( \theta^{\dagger} \odot (\xi_{x}^{\theta^\dagger+\delta} - \xi_{x}^{\theta^\dagger}))^T H ( \theta^{\dagger} \odot (\xi_{x}^{\theta^\dagger+\delta} - \xi_{x}^{\theta^\dagger})) -\langle \delta\odot \xi_{x}^{\theta^\dagger+\delta}\odot (\xi_{x}^{\theta^\dagger+\delta} - \xi_{x}^{\theta^\dagger}),H \theta^{\dagger} \rangle.
\end{align*}
Then from equation \eqref{eq:2_layer_property}, we have almost surely
\begin{align*}
    ( \theta^{\dagger} \odot (\xi_{x}^{\theta^\dagger+\delta} - \xi_{x}^{\theta^\dagger}))^T H ( \theta^{\dagger} \odot (\xi_{x}^{\theta^\dagger+\delta} - \xi_{x}^{\theta^\dagger})) &=\langle \theta^{\dagger} \odot (\xi_{x}^{\theta^\dagger+\delta} - \xi_{x}^{\theta^\dagger}), \nabla \phi^{\theta^{\dagger} \odot (\xi_{x}^{\theta^\dagger+\delta} - \xi_{x}^{\theta^\dagger})}\rangle \\
    &\leq G \|\theta^\dagger\|_2~,
\end{align*}
where we basically replace $\frac{1}{2}\zeta\|\theta^\dagger\|_2^2$ in the $k>2$ cases, by a lower order term $\frac{1}{2}G \|\theta^\dagger\|_2$. A same property applies to equation \eqref{eq:dont_know_what_to_call}, i.e. we can upper bound $\|H\theta^\dagger\|_2^2$ by $G^2$ instead of $\zeta\|\theta^\dagger \|_2^2$, therefore, for $k=2$, we have 
\begin{equation}
\begin{split}
\ell_{\beta}\left(\cQ, W\right) \leq \ell_{\beta + \frac{3}{2}\sigma^2\zeta\alpha\tilde \gamma+\frac{3}{2}G\|\theta^\dagger\|_2}(\psi^{\theta^{\dagger}},W) + 6\exp\left(- c\tilde \gamma\right)~,
\end{split}
\end{equation}
and, 
\begin{equation}
\begin{split}
\ell_{\beta}(\psi^{\theta^{\dagger}},W)  &\leq  \ell_{\beta + \frac{3}{2}\sigma^2\zeta\alpha\tilde \gamma+\frac{3}{2}G\|\theta^\dagger\|_2}(\cQ,W)+ 6\exp\left(- c\tilde \gamma\right)~,
\end{split}
\end{equation} 
which completes the proof.
\qed

With careful de-randomization on the Pac-Bayes theorem, we get the deterministic bound for the non-smooth predictors:
\nonsmooth*

\proof For $k>2$, to get to a de-randomized margin bound, we utilize the results in Theorem~\ref{theo:ns_marg}.
First, with $\beta = 0$,  $3\sigma^2 \zeta\alpha \tilde\gamma= \gamma$ and $W=D$ in Theorem~\ref{theo:ns_marg}  we have
\begin{equation}
\begin{split}
\ell_0(\psi^{\theta^{\dagger}}, D)& \leq \ell_{\gamma/2+G\|\theta^\dagger\|_2+\frac{1}{2}\zeta \|\theta^\dagger\|_2^2}(\cQ,D) +    6\exp \left( -\min(c_2 \frac{\gamma^2}{9\sigma^4\zeta^2\alpha^2}, c_1\frac{\gamma}{3\sigma^2\zeta\alpha}) \right)~.
\end{split}
\label{eq:dwsc_ani}
\end{equation}
Similarly, with $\beta = \gamma/2+G\|\theta^\dagger\|_2+\frac{1}{2}\zeta \|\theta^\dagger\|_2^2$ and $W=S$ in Theorem~\ref{theo:ns_marg} we have
\begin{equation}
\begin{split}
\ell_{\gamma/2+G\|\theta^\dagger\|_2+\frac{1}{2}\zeta \|\theta^\dagger\|_2^2}(\cQ, S) &\leq \ell_{\gamma+2G\|\theta^\dagger\|_2+\zeta \|\theta^\dagger\|_2^2}(\psi^{\theta^{\dagger}},S) +   6\exp \left( -\min(c_2 \frac{\gamma^2}{9\sigma^4\zeta^2\alpha^2}, c_1\frac{\gamma}{3\sigma^2\zeta\alpha}) \right)~.
\end{split}
\label{eq:swdc_ani}
\end{equation}
From the Fast Rate PAC-Bayesian bound \eqref{eq:fast_bayes}, with probability at least $(1-\delta)$ over the draw of $n$ samples $S \sim D^n$, for any $\beta \in (0,1)$ and for any $Q$ we have 
\begin{equation} \label{eq:pac_bound2}
    \ell_{\gamma/2+G\|\theta^\dagger\|_2+\frac{1}{2}\zeta \|\theta^\dagger\|_2^2}(Q,D) \leq \frac{\log (1/\beta)}{1-\beta} \ell_{\gamma/2+G\|\theta^\dagger\|_2+\frac{1}{2}\zeta \|\theta^\dagger\|_2^2}(Q,S) + \frac{1}{1 - \beta} \frac{ KL(Q \| P) + \log (\frac{1}{\delta}) }{n} ~,
\end{equation}

Noting that
\begin{align}
	2KL(\cQ|| P) &= \sum_{j = 1}^{p} (\frac{1}{\sigma^2 \nu_j} -1)+ \sum_{j =1}^{p}\frac{|\theta[j]-\theta_0[j]|^2}{\sigma^2} + \sum_{i =1}^{p} \ln \frac{\nu_j}{1/\sigma^2} \nonumber\\
	& \leq \sum_{l =1}^{\tilde p} \ln \frac{\tilde \nu_{(l)}^2}{1/\sigma^2} +\frac{\|\theta^{\dagger}-\theta_0\|_2^2}{\sigma^2}, 
\end{align}
we have 
\begin{equation}
\begin{split}
\ell_{0}(\psi^{\theta^{\dagger}},D)  \leq \frac{\log (1/\beta)}{1-\beta} &\ell_{\gamma+2G\|\theta^\dagger\|_2+\zeta \|\theta^\dagger\|_2^2}(\psi^{\theta^{\dagger}},S) + 
\frac{1}{2n(1 - \beta)} \left(  \sum_{\ell =1}^{\tilde p} \ln \frac{\tilde{\nu}^2_{(\ell)}}{1/\sigma^2}  +  \frac{\|\theta^{\dagger}-\theta_0\|_2^2}{\sigma^2} \right)\\
&+6\left(\frac{\log (1/\beta)}{1-\beta}+1\right)\exp \left( -\min(c_2 \frac{\gamma^2}{\sigma^4\zeta^2\alpha^2}, c_1\frac{\gamma}{\sigma^2\zeta\alpha}) \right)+ \frac{1}{1 - \beta} \frac{\log (\frac{1}{\delta})}{n}.
\end{split}
\end{equation} 

For 2-layer neural network, we have the deterministic bound formed as
\begin{align}
\ell_{0}(\psi^{\theta^{\dagger}},D)  &\leq a_{\beta} \ell_{\gamma+3G\|\theta^\dagger\|_2}(\psi^{\theta^{\dagger}},S) + 
\frac{b_{\beta}}{2n} \left(  \underbrace{\sum_{\ell =1}^{\tilde p} \ln \frac{\tilde{\nu}^2_{(\ell)}}{1/\sigma^2}}_{\textup{effective curvature}}  +  \underbrace{\frac{\|\theta^{\dagger}-\theta_0\|_2^2}{\sigma^2}}_{\textup{$L_2$ norm}} \right) +d_{\beta}\exp\left( -\min(c_2 \gamma^2,c_1\gamma) \right) +  b_{\beta} \frac{\log (\frac{1}{\delta})}{n},
\end{align}
which completes the proof.
\qed

\subsection{Sample Complexity for Non-uniform Bounds}

\sampcomp*

\proof We start with the slow rate version of Theorem~\ref{theo:non_smooth} which states that with probability at least $(1-\delta)$, for any $\psi^{\theta^{\dagger}}, \gamma > 6\sigma^2\zeta \alpha$, we have:
\begin{align}
\ell_{0}(\psi^{\theta^{\dagger}},D) &\leq   \ell_{\gamma + 2\varrho_k}(\psi^{\theta^{\dagger}},S)     + \frac{1}{2\sqrt{n}} \left(  \sum_{\ell =1}^{\tilde p_{1/\sigma^2}} \ln \frac{\tilde{\nu}^2_{(\ell)}}{1/\sigma^2}  +  \frac{\|\theta^{\dagger}-\theta_0\|_2^2}{\sigma^2} \right) + 6 \exp(-\min(c_2\gamma^2,c_1 \gamma)) +  \frac{\log (\frac{1}{\delta})}{\sqrt{n}}~.
\label{eq:slow}
\end{align}
Our specific goal is the following: for any $\psi^{\theta^\dagger}$, we want to find a scaling $\lambda = \lambda(\sigma,\epsilon,\|\theta^\dagger\|_2) \geq 1$ and the sample complexity $n_0 = n_0(\psi^{\theta^\dagger},\sigma,\epsilon,\delta,\lambda)$ such that for any $n \geq n_0$, the scaled predictor $\psi^{\lambda \theta^\dagger}$ satisfies:
\begin{equation}
\ell_{0}(\psi^{\lambda \theta^{\dagger}},D) \leq  \ell_0(\psi^{\lambda \theta^\dagger},S) + \epsilon~, 
\label{eq:samcomp2}
\end{equation}
which yields the desired main result by noting that for deep nets, the 0-1 loss does not change by parameter scaling so that $\ell_{0}(\psi^{\theta^{\dagger}},D) = \ell_{0}(\psi^{\lambda \theta^{\dagger}},D)$ and $\ell_{0}(\psi^{\theta^{\dagger}},S) = \ell_{0}(\psi^{\lambda \theta^{\dagger}},S)$. For the depth $k$ deep net with parameter $\lambda \theta^\dagger$, we choose the posterior $\cQ$ as an anisotropic Gaussian with mean $\lambda \theta^\dagger$ and diagonal covariance $\Sigma_{\lambda \theta^\dagger}$ such that
\begin{equation}
\Sigma_{\lambda \theta^{\dagger}}^{-1} = \diag(\rho_1^2,\ldots,\rho_p^2)~,\qquad \rho_j^2 \triangleq \max \left\{ \tilde \cH_{l,\phi}^{\lambda \theta^{\dagger}}[j,j], \frac{\lambda^{2k}}{\sigma^2} \right\}~,
\label{eq:scov_pos}
\end{equation}
where
\begin{equation}
    \tilde{\cH}_{l,\phi}^{\lambda \theta^{\dagger}} \triangleq \frac{1}{n} \sum_{i=1}^n \nabla^2 l(y_i,\phi^{\lambda \theta^\dagger \odot \xi_{x_i}^{\lambda \theta^\dagger}}(x_i))~.
\label{eq:ns_loss_shess}
\end{equation}
Now, considering \eqref{eq:slow} for the predictor $\psi^{\lambda \theta^{\dagger}}$, we get: with probability at least $(1-\delta)$, for any $\psi^{\lambda \theta^{\dagger}}, \gamma > 2\sigma^2 \zeta_{\lambda}\alpha$, we have:
\begin{align}
\hspace*{-5mm}
\ell_{0}(\psi^{\theta^{\lambda \dagger}},D) &\leq   \ell_{\gamma + 2\varrho_k}(\psi^{\lambda \theta^{\dagger}},S) + \frac{1}{2\sqrt{n}} \left(  \sum_{\ell =1}^{\tilde p_{\lambda^{2k}/\sigma^2}} \ln \frac{\tilde{\rho}^2_{(\ell)}}{\lambda^{2k}/\sigma^2}  +  \frac{\|\theta^{\dagger}-\theta_0\|_2^2}{\sigma^2} \right) + 6 \exp(-\min(c_{2,\lambda} \gamma^2, c_{1,\lambda}\gamma)) +  \frac{\log (\frac{1}{\delta})}{\sqrt{n}}~.
\label{eq:slow2}
\end{align}
where $\varrho_k = G_{\lambda} \lambda \| \theta^\dagger\|_2 + \frac{1}{2}\zeta_{\lambda} \lambda^2 \| \theta^\dagger \|_2^2$, $\tilde p_{\lambda^{2k}/\sigma^2} = | \{ j : \tilde \cH_{l,\phi}^{\lambda \theta^{\dagger}}[j,j] > \lambda^{2k}/ \sigma^2 \}|$, and $\tilde \rho^2_{(\ell)}$ are the diagonal elements which strictly satisfy the inequality. The constants $\zeta_{\lambda}$ and $c_{\lambda}$ are the ones corresponding to gradients and Hessians of the loss on scaled LDNs $\phi^{\lambda \theta^{\dagger}} \triangleq \phi^{\lambda \theta^\dagger \odot \xi_{x_i}^{\lambda \theta^\dagger}}(x_i)$, which in turn will determine $c_{1,\lambda},c_{2,\lambda}$ in \eqref{eq:slow2} corresponding to $c_1,c_2$ in Theorem~\ref{theo:non_smooth}. First, note that 
\begin{align}
    \nabla \phi^{\lambda \theta^{\dagger}} = \lambda^{k-1} \nabla \phi^{\theta^{\dagger}} ~~~\text{and}~~~  \nabla^2 \phi^{\lambda \theta^{\dagger}} = \lambda^{k-2} \nabla \phi^{\theta^{\dagger}}~,
\end{align}
where the gradients are w.r.t.~the scaled parameters $\lambda \theta^\dagger$. As a result, in Assumption~\ref{asmp:gen}, the constant related to the first order gradient $G_{\lambda} = \lambda^{k-1} G$, and the constants related to the Hessian, viz.~$\zeta_{\lambda}$, scaled by $\lambda^{k-2}$. Then, we have
\begin{align*}
c_{1,\lambda} &= \frac{1}{\lambda^{k-2}} c_1~,
\end{align*}
and 
\begin{align*}
c_{2,\lambda} & = c_2 =  \min\left[\frac{1}{2\sigma^2G^2_\lambda},\frac{1}{2\sigma^2\zeta_\lambda^2 \|\lambda\theta^\dagger\|_2^2},\frac{1}{8\sigma^4\kappa\zeta_\lambda^2}\right] \geq \frac{1}{\lambda^{2k-2}} c_2~.
\end{align*}


To go from \eqref{eq:slow2} to \eqref{eq:samcomp2}, choosing $\theta_0 = 0$, it is sufficient to choose $\lambda, n_0$ such that for $n \geq n_0$
\begin{align}
\ell_{\gamma + 2G_\lambda \lambda \| \theta^\dagger\|_2 + \zeta_\lambda \lambda^2 \| \theta^\dagger \|_2^2}(\psi^{\lambda \theta^{\dagger}},S) & \leq \ell_{0}(\psi^{\lambda\theta^{\dagger}},S) + \frac{\epsilon}{4}~, \label{eq:t1}\\
\frac{1}{2\sqrt{n}} \left(  \sum_{\ell =1}^{\tilde p_{\lambda^{2k}/\sigma^2}} \ln \frac{\tilde{\rho}^2_{(\ell)}}{\lambda^{2k}/\sigma^2}  +  \frac{\lambda^2\|\theta^{\dagger}\|_2^2}{\sigma^2} \right) & \leq \frac{\epsilon}{4}~, \label{eq:t2}\\
  6 \exp(-\min[c_{2,\lambda}\gamma^2,c_{1,\lambda}\gamma]) & \leq \frac{\epsilon}{4}~, \label{eq:t3} \\
 \frac{\log (\frac{1}{\delta})}{\sqrt{n}} & \leq \frac{\epsilon}{4}~. \label{eq:t4}
\end{align}
%
We first analyze the conditions \eqref{eq:t3} and then \eqref{eq:t1}, which are independent of $n$. Subsequently, we analyze the conditions \eqref{eq:t4} and \eqref{eq:t2}, which yield the sample complexity.

For \eqref{eq:t3},  with our analysis on $c_{1,\lambda}, c_{2,\lambda}$ it is sufficient to have
\begin{align*}
 6 \exp(-\min\{c_2\gamma^2/\lambda^{2k-2},c_1\gamma/\lambda^{k-2}\})  \leq \frac{\epsilon}{4}~,\\
\Rightarrow \quad \gamma  \geq \max \left\{ \frac{\lambda^{k-1}}{\sqrt{c_2}} \sqrt{\log \frac{24}{\epsilon}}~, \frac{\lambda^{k-2}}{c_1} \log \frac{24}{\epsilon}   \right\}
 \end{align*}
so that
 \begin{align}
    \gamma \geq \max \left\{  \frac{\lambda^{k-1}}{\sqrt{c_2}} \sqrt{\log \frac{24}{\epsilon}}~, \frac{\lambda^{k-2}}{c_1} \log \frac{24}{\epsilon}~, 2\lambda^{k-2} \sigma^2 \alpha \right\}~,
    \label{eq:t3cond3}
\end{align}
where we have also included the constraint $\gamma > 2\sigma^2 \alpha_\lambda$ needed for the bound in Theorem~\ref{theo:non_smooth} with proper scaling, i.e., $\alpha_{\lambda}$. 

We now focus on \eqref{eq:t1}. The margin loss $\ell_{\gamma}(\psi^{\theta^\dagger},S)$ for any predictor is a non-decreasing function of the margin $\gamma \geq 0$, and we assume:
\begin{equation}
    \ell_{\gamma}(\psi^{\theta^\dagger},S) - \ell_{0}(\psi^{\theta^\dagger},S) = g(\gamma)~,
    \label{eq:margf}
\end{equation}
where the margin function $g(\gamma), \gamma \geq 0$ is a non-decreasing function. We define the inverse function $g^{-1} : \R_+ \mapsto \R_+$ as $g^{-1}(a) =  \min_{\gamma :  g(\gamma) \geq a} \gamma$. Based on the margin function in \eqref{eq:margf}, for a depth $k > 2$ deep net, we have
\begin{align*}
\ell_{\gamma + 2G \| \theta^\dagger\|_2 + \zeta \| \theta^\dagger \|_2^2}(\psi^{ \theta^{\dagger}},S)  -  \ell_{0}(\psi^{\theta^{\dagger}},S) \leq g\left(\gamma + 2G \| \theta^\dagger\|_2 + \zeta \| \theta^\dagger \|_2^2 \right)~.
\end{align*}
Now, with the scaled parameters $\lambda \theta^\dagger$, we have
\begin{align*}
\ell_{\gamma + 2 \lambda G \| \theta^\dagger\|_2 +  \lambda^2 \zeta \| \theta^\dagger \|_2^2}(\psi^{\lambda \theta^{\dagger}},S)  -  \ell_{0}(\psi^{\lambda\theta^{\dagger}},S)  
& \overset{(a)}{=} \ell_{\lambda^{-k}\gamma + 2\lambda^{1-k} G \| \theta^\dagger\|_2 + \lambda^{2-k} \zeta  \| \theta^\dagger \|_2^2}(\psi^{\theta^{\dagger}},S)  -  \ell_{0}(\psi^{\theta^{\dagger}},S) \\
& \leq g\left(\lambda^{-k}\gamma + 2 \lambda^{1-k}G \| \theta^\dagger\|_2 +  \lambda^{2-k} \zeta \| \theta^\dagger \|_2^2 \right)~,
\end{align*}
where (a) follows since for a depth $k > 2$ deep net
\begin{align*}
\ell_{\gamma + 2\lambda G \| \theta^\dagger\|_2 +  \lambda^2 \zeta \| \theta^\dagger \|_2^2}(\psi^{\lambda \theta^{\dagger}},S) & = \P_S( y \psi^{\lambda \theta^{\dagger}}(x) 
\leq \gamma +  2\lambda G \| \theta^\dagger\|_2 + \lambda^2 \zeta \| \theta^\dagger \|_2^2 ) \\
& = \P_S( y \lambda^k \psi^{\theta^{\dagger}}(x) 
\leq \gamma +  2\lambda G \| \theta^\dagger\|_2 + \lambda^2 \zeta \| \theta^\dagger \|_2^2 ) \\
& = \P_S( y \psi^{\theta^{\dagger}}(x) 
\leq \lambda^{-k} \gamma +  2\lambda^{1-k} G \| \theta^\dagger\|_2 + \lambda^{2-k} \zeta \| \theta^\dagger \|_2^2 )\\
& = \ell_{\lambda^{-k}\gamma + 2\lambda^{1-k} G \| \theta^\dagger\|_2 + \lambda^{2-k} \zeta  \| \theta^\dagger \|_2^2}(\psi^{\theta^{\dagger}},S)~,
\end{align*}
and $\ell_{0}(\psi^{\theta^{\dagger}},S) = \ell_{0}(\psi^{\lambda \theta^{\dagger}},S)$. Then, for \eqref{eq:t1}, it is sufficient to have 
\begin{align*}
    \lambda^{-k}\gamma +  2\lambda^{1-k}G \| \theta^\dagger\|_2 + \lambda^{2-k} \zeta \| \theta^\dagger \|_2^2 \leq g^{-1}(\epsilon/4)~,
\end{align*}
which is satisfied if
\begin{align}
\lambda & \geq \max\left\{ \left(\frac{3\gamma}{g^{-1}(\epsilon/4)}\right)^{\frac{1}{k}}, ~~\left(\frac{6G \|\theta^\dagger\|_2}{g^{-1}(\epsilon/4)}\right)^{\frac{1}{k-1}},~~ \left(\frac{3\zeta \|\theta^\dagger\|_2^2}{g^{-1}(\epsilon/4)} \right)^{\frac{1}{k-2}} \right\}~
    \label{eq:t1cond1}
\end{align}
Just focusing on the first term, by taking the constraints on $\gamma$ in \eqref{eq:t3cond3} into account, with $\zeta = \frac{3}{g^{-1}(\epsilon/4)}$, we get
\begin{align*}
\lambda \geq \zeta^{1/k} \max\left\{  
\lambda^{1-1/k} \left(\frac{1}{c_2} \log \frac{24}{\epsilon}\right)^{1/2k},~~ \lambda^{1-2/k} \left(\frac{1}{c_1} \log \frac{24}{\epsilon} \right)^{1/k},~~ \lambda^{1-2/k} (2\sigma^2 \alpha)^{1/k}
\right\}~,
\end{align*}
which is satisfied if
\begin{align}
\lambda \geq  \max\left\{  
\left(\frac{\zeta^2}{c_2} \log \frac{24}{\epsilon}\right)^{1/2},~ ~ \left(\frac{\zeta}{c_1} \log \frac{24}{\epsilon} \right)^{1/2},~~ \sqrt{2\zeta \sigma^2 \alpha} \right\}~.
\label{eq:t1cond1part1}
\end{align}
Combining \eqref{eq:t1cond1} and \eqref{eq:t1cond1part1}, with $\zeta = \frac{3}{g^{-1}(\epsilon/4)}$, we have 
\begin{equation}
\hspace*{-5mm} \lambda \geq \max\left\{  
\left(\frac{\zeta^2}{c_2} \log \frac{24}{\epsilon}\right)^{1/2},~\left(\frac{\zeta}{c_1} \log \frac{24}{\epsilon} \right)^{1/2},~ \sqrt{2\zeta \sigma^2 \alpha},~\left(\frac{6G \|\theta^\dagger\|_2}{g^{-1}(\epsilon/4)}\right)^{\frac{1}{k-1}}, ~ \left(\frac{3\zeta \|\theta^\dagger\|_2^2}{g^{-1}(\epsilon/4)} \right)^{\frac{1}{k-2}} \right\}~.
\label{eq:t1cond1long}
\end{equation}

Now we focus on the sample dependent terms \eqref{eq:t2} and \eqref{eq:t4}. For \eqref{eq:t2}, first recall that the effective curvature for the unscaled and scaled models are respectively
\begin{align}
     \sum_{\ell =1}^{\tilde p_{1/\sigma^2}} \ln \frac{\tilde{\nu}^2_{(\ell)}}{1/\sigma^2}  \qquad \text{and} \qquad \sum_{\ell =1}^{\tilde p_{\lambda^{2k}/\sigma^2}} \ln \frac{\tilde{\rho}^2_{(\ell)}}{\lambda^{2k}/\sigma^2} ~,
\end{align}
where $\tilde \nu^2_j$ are diagonal elements of $\tilde \cH_{l,\phi}^{\theta^{\dagger}}$ which are strictly larger than $\frac{1}{\sigma^2}$ with a total of $p_{1/\sigma^2}$ such elements and
$\tilde \rho^2_j$ are diagonal elements of $\tilde \cH_{l,\phi}^{\lambda \theta^{\dagger}}$ which are strictly larger than $\frac{\lambda^{2k}}{\sigma^2}$ with a total of $p_{\lambda^{2k}/\sigma^2}$ such elements. Next we show that $p_{\lambda^{2k}/\sigma^2} \leq p_{1/\sigma^2}$ since for any $j$ the diagonal element $\rho^2_j \leq \lambda^{2k} \nu^2_j$. To see this, first note that being a depth $k$ LDN, $\phi^{\lambda \theta^\dagger \odot \xi_{x_i}^{\lambda \theta^\dagger}} = \lambda^k \phi^{\theta^\dagger \odot \xi_{x_i}^{\theta^\dagger}}$, i.e., we just get a scaled prediction. 
Then, the $i$-th term in the definition of $\tilde \cH_{l,\phi}^{\lambda \theta^{\dagger}}$  in \eqref{eq:ns_loss_shess}, we have
\begin{align*}
    \nabla^2 l(y_i,\phi^{\lambda \theta^\dagger \odot \xi_{x_i}^{\lambda \theta^\dagger}}(x_i))) = \nabla^2 l(y_i,\lambda^k \phi^{\theta^\dagger \odot \xi_{x_i}^{\theta^\dagger}}(x_i))) = 
    \nabla^2 l(y_i,\lambda^k \hat{y}_i(\theta^\dagger))~,
\end{align*}
where $\hat{y}_i(\theta^\dagger) = \phi^{\theta^\dagger \odot \xi_{x_i}^{\theta^\dagger}}(x_i)$ is the prediction from the unscaled predictor. Now, writing the gradients w.r.t.~$\theta^\dagger_j$ explicitly, by chain rule using intermediate variable $\hat{y}_i^{(\lambda)} = \lambda^k \hat{y}_i(\theta^\dagger)$, we have
\begin{align*}
    \frac{\partial l(y_i,\lambda^k \hat{y}_i(\theta^\dagger)}{\partial \theta_j^{\dagger}} = \frac{\partial l(y_i,\hat{y}_i^{(\lambda)})}{\partial \hat{y}_i^{(\lambda)}} \frac{\partial (\lambda^k \hat{y}_i(\theta))}{\partial \theta_j^{\dagger}} = \lambda^k  \frac{\partial l(y_i,\hat{y}_i^{(\lambda)})}{\partial \hat{y}_i^{(\lambda)}} \frac{\partial \hat{y}_i(\theta)}{\partial \theta_j^{\dagger}}~,
\end{align*}
so that
\begin{align}
  \frac{\partial^2 l(y_i,\lambda^k \hat{y}_i(\theta^\dagger)}{\partial {\theta_j^{\dagger}}^2} = \lambda^{2k}  \frac{\partial^2 l(y_i,\hat{y}_i^{(\lambda)})}{\partial (\hat{y}_i^{(\lambda)})^2} \left(\frac{\partial \hat{y}_i(\theta)}{\partial \theta_j^{\dagger}} \right)^2~,
 \label{eq:2d1}
\end{align}
since $\hat{y}_i(\theta^\dagger)$ being a LDN, $\frac{\partial^2 \hat{y}_i(\theta^{\dagger})}{\partial {\theta_j^{\dagger}}^2}=0$. With $\sigma(a) = 1/(1+\exp(-a))$, the cross-entropy loss 
\begin{align*}
l(y_i,\hat{z}_i) =  -(y_i \log(\sigma(\hat{z}_i)) + (1-y_i) \log(1-\sigma(\hat{z}_i))~.
\end{align*}
Then, the second derivative is 
\begin{align*}
\frac{\partial^2 l(y_i,\hat{z}_i)}{\partial z_i^2} = \sigma(\hat{z}_i)(1-\sigma(\hat{z}_i))~,
\label{eq:2d2}
\end{align*}
the variance of the prediction, which is maximum when $\hat{z}_i =0$. Comparing the variances for $\hat{z}^{(\lambda)}_i = \lambda^k \hat{y}_i(\theta^\dagger)$ and $\hat{z}_i = \hat{y}_i(\theta^\dagger)$, we get
\begin{align*}
\sigma( \lambda^k \hat{y}_i(\theta^\dagger) )(1-\sigma( \lambda^k \hat{y}_i(\theta^\dagger) )) \leq \sigma( \hat{y}_i(\theta^\dagger) )(1-\sigma( \hat{y}_i(\theta^\dagger) ))~
\end{align*}
since scaling of the prediction reduces the variance. As a result, 
\begin{align}
  \frac{\partial^2 l(y_i,\lambda^k \hat{y}_i(\theta^\dagger)}{\partial {\theta_j^{\dagger}}^2} = \lambda^{2k}  \frac{\partial^2 l(y_i,\hat{y}_i^{(\lambda)})}{\partial (\hat{y}_i^{(1)})^2} \left(\frac{\partial \hat{y}_i(\theta)}{\partial \theta_j^{\dagger}} \right)^2 \leq \lambda^{2k}  \frac{\partial^2 l(y_i,\hat{y}_i^{(1)})}{\partial (\hat{y}_i^{(1)})^2} \left(\frac{\partial \hat{y}_i(\theta)}{\partial \theta_j^{\dagger}} \right)^2 = \lambda^{2k}  \frac{\partial^2 l(y_i, \hat{y}_i(\theta^\dagger))}{\partial {\theta_j^{\dagger}}^2}~,
\end{align}
a $\lambda^k$-scaled version of the $j$-th component of the Hessian of the unscaled LDN $\phi^{\theta^\dagger \odot \xi_{x_i}^{\theta^\dagger}}(x_i)$. Averaging over all samples $i$, we get $\rho_j^2 \leq \lambda^{2k} \nu_j^2$ so that $\frac{\rho_j^2}{\lambda^{2k}\sigma^2} \leq \frac{\nu_j^2}{\sigma^2}$. As a result, $\tilde{p}_{\lambda^{2k}/\sigma^2}$, the number of diagonal elements of $\tilde \cH_{l,\phi}^{\lambda\theta^\dagger}$ which cross the threshold $\lambda^{2k}/\sigma^2$, can be at most $\tilde{p}_{1/\sigma^2}$, the number of diagonal elements of $\tilde \cH_{l,\phi}^{\theta^\dagger}$ which cross the threshold $1/\sigma^2$. Hence,
\begin{align}
 \sum_{\ell =1}^{\tilde p_{\lambda^{2k}/\sigma^2}} \ln \frac{\tilde{\rho}^2_{(\ell)}}{\lambda^{2k}/\sigma^2}  
 & \leq \sum_{\ell =1}^{\tilde p_{1/\sigma^2}} \ln \frac{\tilde{\nu}^2_{(\ell)}}{1/\sigma^2} ~.
\end{align}
As a result, for \eqref{eq:t2}, it is sufficient to have
\begin{align}
\frac{1}{2\sqrt{n}} \left( \sum_{\ell =1}^{\tilde p_{1/\sigma^2}} \ln \frac{\tilde{\nu}^2_{(\ell)}}{1/\sigma^2}   +  \frac{\lambda^2\|\theta^{\dagger}\|_2^2}{\sigma^2} \right)  \leq \frac{\epsilon}{4} \\
\Rightarrow \quad 
n  \geq \frac{4}{\epsilon^2} \left( \sum_{\ell =1}^{\tilde p_{1/\sigma^2}} \ln \frac{\tilde{\nu}^2_{(\ell)}}{1/\sigma^2}   +  \frac{\lambda^2\|\theta^{\dagger}\|_2^2}{2\sigma^2} \right)^2~.
\label{eq:t2cond2}
\end{align}
For \eqref{eq:t4}, it is sufficient to have 
\begin{equation}
    n \geq \frac{16}{\epsilon^2} \log \left( \frac{1}{\delta} \right)~,
    \label{eq:t4cond4}
\end{equation}
so that it suffices to have
\begin{equation}
n  \geq \max\left\{ \frac{4}{\epsilon^2} \left( \sum_{\ell =1}^{\tilde p_{1/\sigma^2}} \ln \frac{\tilde{\nu}^2_{(\ell)}}{1/\sigma^2}   +  \frac{\lambda^2\|\theta^{\dagger}\|_2^2}{2\sigma^2} \right)^2~, ~\frac{16}{\epsilon^2} \log \left( \frac{1}{\delta} \right) \right\}~,
\label{eq:sampcomp2}
\end{equation}
where $\lambda$ satisfies \eqref{eq:t1cond1long}. 


Finally, to establish the polynomial nature of the sample complexity $n_0$, i.e., the lower bound in \eqref{eq:sampcomp2}, it suffices to focus on \eqref{eq:sampcomp2} and also \eqref{eq:t1cond1long}. First, note that $\lambda$ in \eqref{eq:t1cond1long} has no dependency on $1/\delta$ and hence $n_0$ only has a polynomial dependence on $\log(1/\delta)$ from \eqref{eq:sampcomp2}. Next, for $\sigma$, the lower bound on $\lambda$ in \eqref{eq:t1cond1long} has a polynomial dependence on $\sigma$, and hence $n_0$ in \eqref{eq:sampcomp2}, after taking its dependence on $\lambda$ into account, also has a polynomial dependence on $\sigma$. Under the assumption that the margin function $g(\gamma) \geq c_1 \gamma^{c_2/q}$ for some finite integer $q$, we have $g^{-1}(a) \leq c_3 a^{c_4 q}$ for suitable constants $c_3,c_4$. In particular, $g^{-1}(\epsilon/4)$ in \eqref{eq:t1cond1long} is a polynomial in $\epsilon$, implying the lower bound to $\lambda$ in \eqref{eq:t1cond1long} has a polynomial dependency on $1/\epsilon$, which in turn implies $n_0$ has a polynomial dependence on $1/\epsilon$ from \eqref{eq:sampcomp2}. That completes the proof. \qed

\section{Experimental Results}
\label{app:expt}

In this section, we present the experimental setup 
and additional results. We divide our experiments into two sets to address questions: (i) How does our bound behave as we increase the number of random labels? (ii) How does our bound behave with an increase in the number of training samples? We presented the setup for these two sets of experiment
in Section \ref{exp:set_up}. Then we present additional results in Section \ref{app:exp_more_random} and \ref{app:exp_more_sample} for question (i) and (ii), respectively.


\subsection{Experimental Setup}
\label{exp:set_up}
{\bf Network Architecture and Datasets}: We focus on fully connected networks with ReLU activation. 
We consider two tasks: the MNIST image classification and the CIFAR-10 image
classification. The MNIST dataset contains 60,000 black and white training images, representing handwritten digits 0 to 9. Each image of size $28\times28$  is normalized by subtracting the mean and dividing the standard deviation of the training set and converted into a vector of size 784.  The CIFAR-10 dataset consists of 60,000 color images including 10 categories. 50,000 of them are for training, and the rest 10,000 are for validation/testing purpose. Every image is of size 32x32 and has 3 color channels. For the random label experiment, We use a subsample of size $1000$ from the original MNIST training set and CIFAR-10 training set, with an equal number of samples from each class.  Then we introduce different levels of randomness $r$ in labels \cite{zhbh17}. In our context, $r$ is the portion of labels for each class that has been replaced by random labels uniformly chosen from $k$ classes. $r = 0$ denotes the original dataset with no corruption, and $r = 1$ means a dataset with completely random labels. For the sample size experiment, i.e., question (ii), we train the ReLU network with sample size $n \in \{ 100, 500, 1000, 5000, 10000\}$ for MNIST and CIFAR-10, with an equal number of samples from each class.


\noindent {\bf Training and Evaluation}:
We use Adam with learning rate $0.001$ to minimize cross-entropy loss until convergence. To evaluate the proposed generalization bound, we first report the test error rate for each set of experiments. Note that our bound is composed of several key terms: empirical margin loss, $L_2$ norm of the weights, and effective curvature. We report these metrics and the value of the bound to check if the behaviors of our bound are aligned with those of the test error rate. For random label experiment for question (i), we repeat each experiment for $30$ times for MNIST and  $10$ times  for CIFAR-10. We report the distribution of the above measurement, and the mean and standard deviation. For the sample size experiment for question (ii), we repeat $5$ times for both MNIST and CIFAR-10 datasets. The details of the setting are presented in Table \ref{table:rand_exp} and Table \ref{table:sample_exp} for MNIST corresponding to two sets of experiments. The setting for CIFAR-10 are presented in Table \ref{table:sample_exp_cifar} and Table \ref{table:rand_exp_cifar} respectively for these two questions.

\begin{table}[t]
\caption{Summary of the setting for MNIST with different level of randomness. 
}
\label{table:rand_exp}
\centering
\begin{tabular}{lllll}
\hline
\multicolumn{5}{c}{Data} \\
\hline 
Input Dimension:     &784     & 784           & 784  & 784  \\
No. of Classes k:      &10      & 10           & 10  & 10  \\
Sample size: & 1000 & 1000 & 1000& 1000 \\
Random Labels:              & 0   & 15\% & 25\% & 50\% \\
\hline
\multicolumn{4}{c}{Training Parameters} \\
\hline 
Learning Rate $\eta$:       & 0.001   & 0.001     & 0.001   & 0.001 \\
Batch Size:               & 128  & 128   & 128  & 128  \\
Epochs:             & 100    & 120  & 130   & 150 \\
\hline        
\end{tabular}
\end{table}

\begin{table}[t]
\caption{Summary of the setting for MNIST with different number of training samples.}
\label{table:sample_exp}
\centering
\begin{tabular}{llllll}
\hline
\multicolumn{5}{c}{Data} \\
\hline 
Input Dimension:      & 784  & 784  & 784  & 784 & 784  \\
No. of Classes k:         & 10    & 10  & 10  & 10 & 10 \\
Training set size: & 100 & 500& 1000 & 5000& 10000\\
Random Labels:               & 0\% & 0\% & 0\% & 0\% & 0\% \\
\hline
\multicolumn{4}{c}{Training Parameters} \\
\hline 
Learning Rate $\eta$:         & 0.001     & 0.001   & 0.001 & 0.001 & 0.001 \\
Batch Size:    &  128  &  128 &  128 & 128 & 128 \\
Epochs:             & 100    & 150  & 250   & 300 & 350 \\
\hline        
\end{tabular}
\end{table}

\begin{table}[t]
\caption{Summary of the setting for CIFAR-10 with different level of randomness. 
}
\label{table:rand_exp_cifar}
\centering
\begin{tabular}{lllll}
\hline
\multicolumn{5}{c}{Data} \\
\hline 
Input Dimension:     &3072     & 3072           & 3072  & 3072  \\
No. of Classes k:      &10      & 10           & 10  & 10  \\
Sample size: & 1000 & 1000 & 1000& 1000 \\
Random Labels:              & 0   & 15\% & 25\% & 50\% \\
\hline
\multicolumn{4}{c}{Training Parameters} \\
\hline 
Learning Rate $\eta$:       & 0.001   & 0.001     & 0.001   & 0.001 \\
Batch Size:               & 128  & 128   & 128  & 128  \\
Epochs:             & 100    & 120  & 130   & 150 \\
\hline        
\end{tabular}
\end{table}

\begin{table}[t]
\caption{Summary of the setting for CIFAR-10 with different number of training samples.}
\label{table:sample_exp_cifar}
\centering
\begin{tabular}{llllll}
\hline
\multicolumn{5}{c}{Data} \\
\hline 
Input Dimension:      & 3072  & 3072  & 3072  & 3072 & 3072  \\
No. of Classes k:         & 10    & 10  & 10  & 10 & 10 \\
Training set size: & 100 & 500& 1000 & 5000& 10000\\
Random Labels:               & 0\% & 0\% & 0\% & 0\% & 0\% \\
\hline
\multicolumn{4}{c}{Training Parameters} \\
\hline 
Learning Rate $\eta$:         & 0.001     & 0.001   & 0.001 & 0.001 & 0.001 \\
Batch Size:    &  128  &  128 &  128 & 128 & 128 \\
Epochs:             & 100    & 200  & 500   & 1000 & 1500 \\
\hline        
\end{tabular}
\end{table}

 \subsection{Additional Results for Randomness and Generalization Bound}
 \label{app:exp_more_random}
 We also evaluate the proposed generalization bound for ReLU network with various depth and various width for both MNIST and CIFAR-10 datasets. For MNIST, we present the results for depth = 4 and width = $\{128, 256, 512\}$ to illustrate how the bound behaves for different widths for a fixed depth.  We  also present the results for width = 128 and depth = $\{2, 4, 6\}$ to illustrate how the bound behaves for different depths for a fixed width. For CIFAR-10, we present the results for depth = 4 and width = $\{256, 384, 512\}$ to illustrate how the bound behaves for different widths for a fixed depth.  We  also present the results for width = 256 and depth = $\{4, 6\}$ to illustrate how the bound behaves for different depths for a fixed width. We also evaluate our bound for micro-batch training, i.e., batch size = 16. The detailed setup is in Table \ref{table:more_rand}.

\begin{table}[h]
\caption{Summary of the setting for each specific experiment with MNIST dataset and CIFAR-10 dataset.}
\label{table:randomness}
\centering
\begin{tabular}{l|ccc|ccc}
\hline 
& \multicolumn{3}{c|}{MNIST} & \multicolumn{3}{c}{CIFAR-10}\\
\hline
No. of Classes k: & \multicolumn{3}{c|}{10} & \multicolumn{3}{c}{10}\\
Input Dimension:  & \multicolumn{3}{c|}{784}& \multicolumn{3}{c}{3072}\\
Level of Randomness: & \multicolumn{3}{c|}{[0\%, 15\%, 25\%,50\%]} & \multicolumn{3}{c}{[0\%, 15\%, 25\%,50\%]}\\
\hline
&\multicolumn{6}{c}{Network Structure}\\
\hline 
No. of Layers: & \multicolumn{3}{c|}{[2, 4, 6]} & \multicolumn{3}{c}{[4, 6, 8]}\\
No. of Nodes per Layer:& \multicolumn{3}{c|}{[128, 256, 512]} & \multicolumn{3}{c}{[256, 384, 512]}\\
\hline
Batch size:& \multicolumn{3}{c|}{[16, 128]} & \multicolumn{3}{c}{[16, 128]}\\
\hline
\end{tabular}
\label{table:more_rand}
\end{table}

\begin{figure*}[t] 
\centering
 \subfigure[Test Error Rate]{
 \includegraphics[width =  0.31\textwidth, height = 0.24\textwidth]{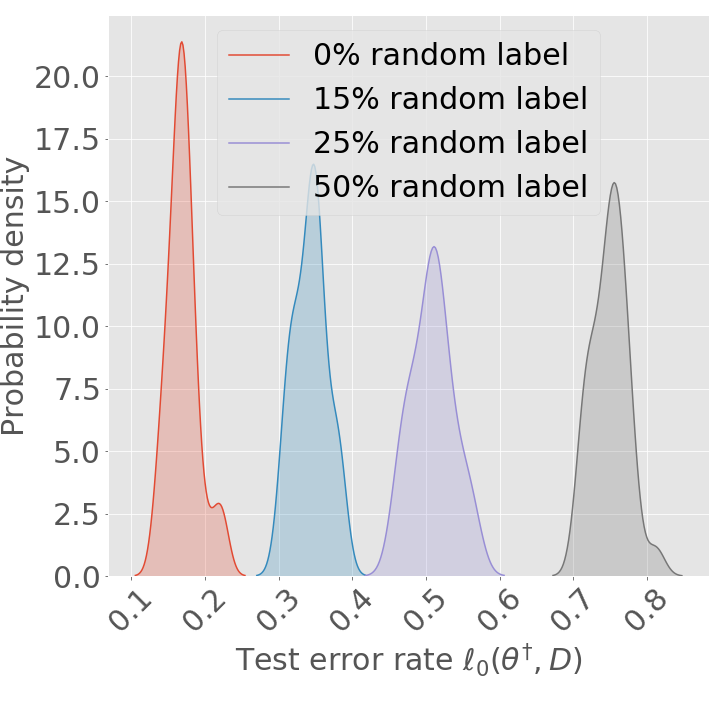}
 } 
 \subfigure[Diagonal Elements of  Hessian.]{
 \includegraphics[width =  0.31\textwidth, height = 0.24\textwidth]{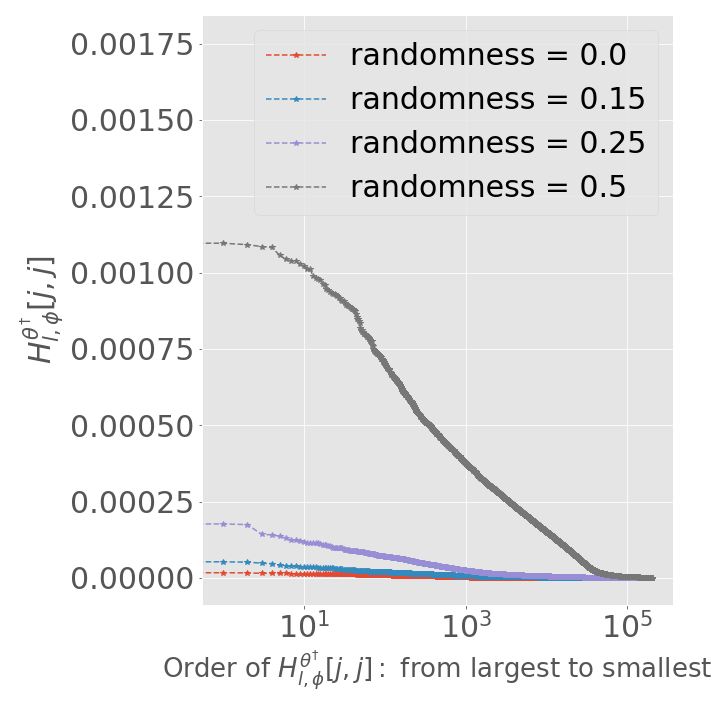}
 } 
 \subfigure[Effective Curvature.]{
 \includegraphics[width =  0.31\textwidth, height = 0.24\textwidth]{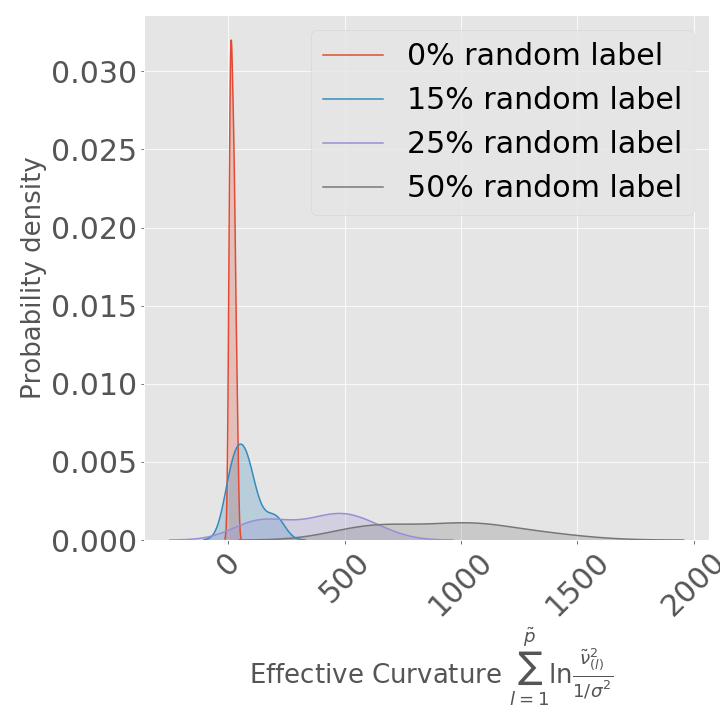}
 } 
 \subfigure[$L_2$ norm / no. sample.]{
 \includegraphics[width =  0.31\textwidth, height = 0.24\textwidth]{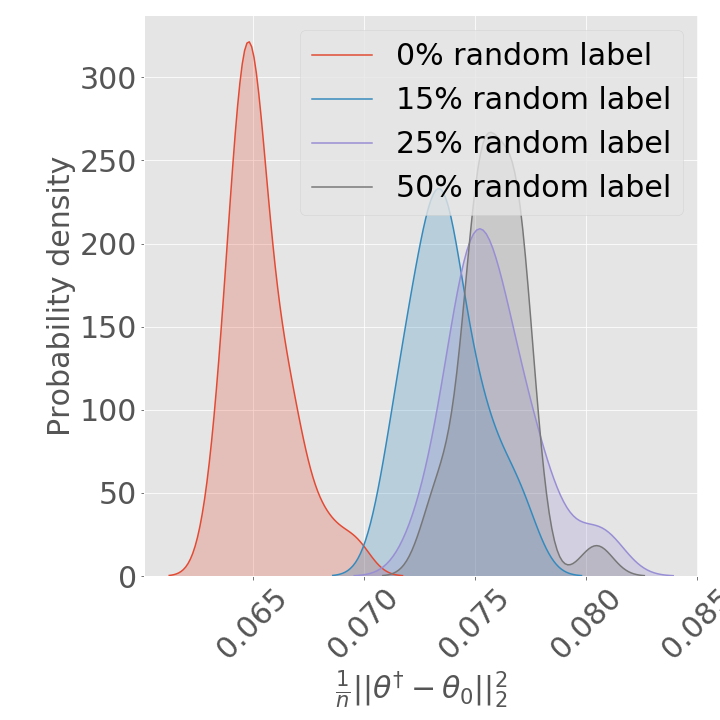}
 } 
  \subfigure[Margin Loss.]{
 \includegraphics[width =  0.31\textwidth, height = 0.24\textwidth]{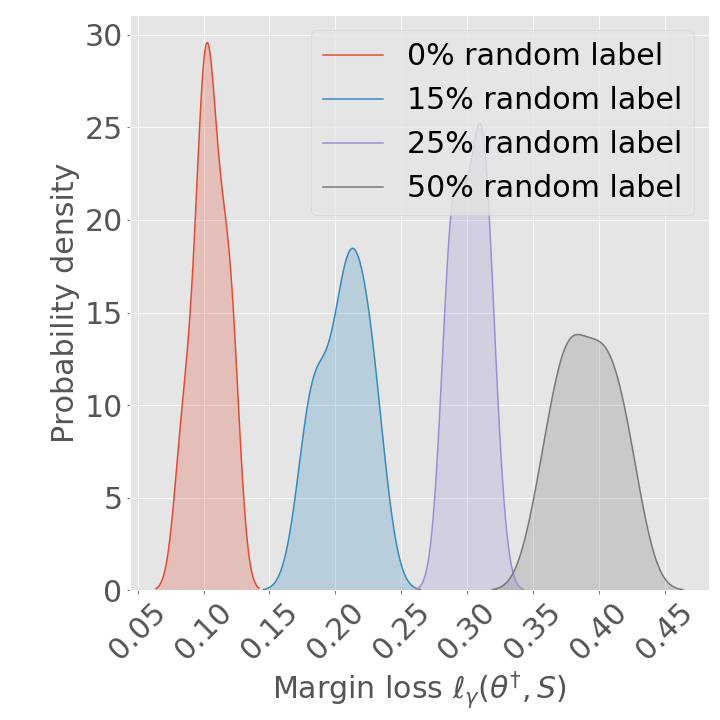}
 } 
   \subfigure[Scale-invariant Generalization Bound.]{
  \includegraphics[width =  0.31\textwidth, height = 0.24\textwidth]{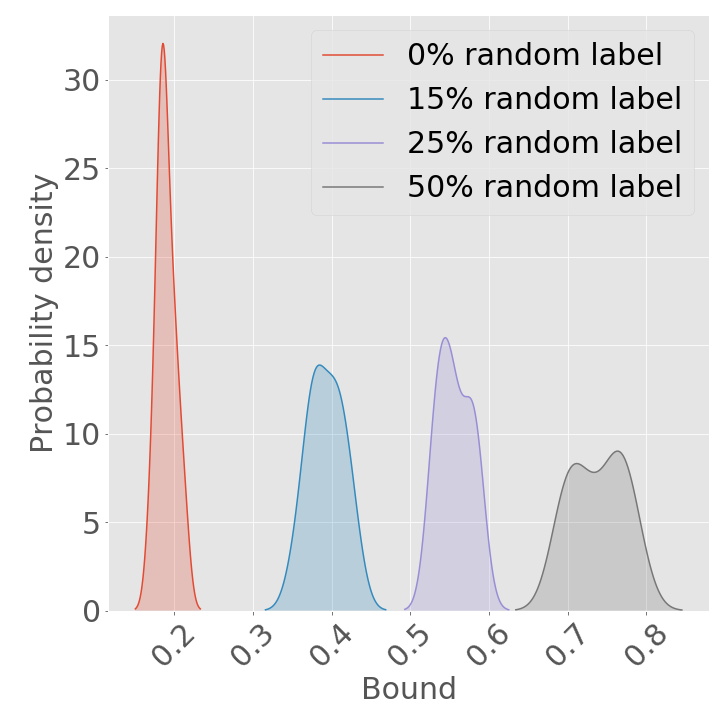}
  } 
\vspace{-2mm}
\caption[]{
Results for ReLU-nets with depth = 6, width =128, total 200,842 parameters, trained on 1000 samples from MNIST with batch size = 128 and a increase in number of random labels (30 runs each) from 0\% to 50\%. (a) test set error rate; (b) diagonal elements (mean) of $\tilde H_{l,\phi}^{\theta^\dagger}$; (c) effective curvature; d) $L_2$ norm of $\theta^{\dagger}$; (e) margin loss; (f) generalization bound.
Increasing percentage of random labels, the generalization bound as well as the components (effective curvature, $L_2$ norm, margin loss) increase, and the bound in (f) stays valid for the test error rate in (a).}
\vspace*{-4mm}
\label{fig:mnist_d6_l128}
\end{figure*}

\begin{figure*}[t] 
\centering
 \subfigure[Test Error Rate]{
 \includegraphics[width =  0.31\textwidth, height = 0.24\textwidth]{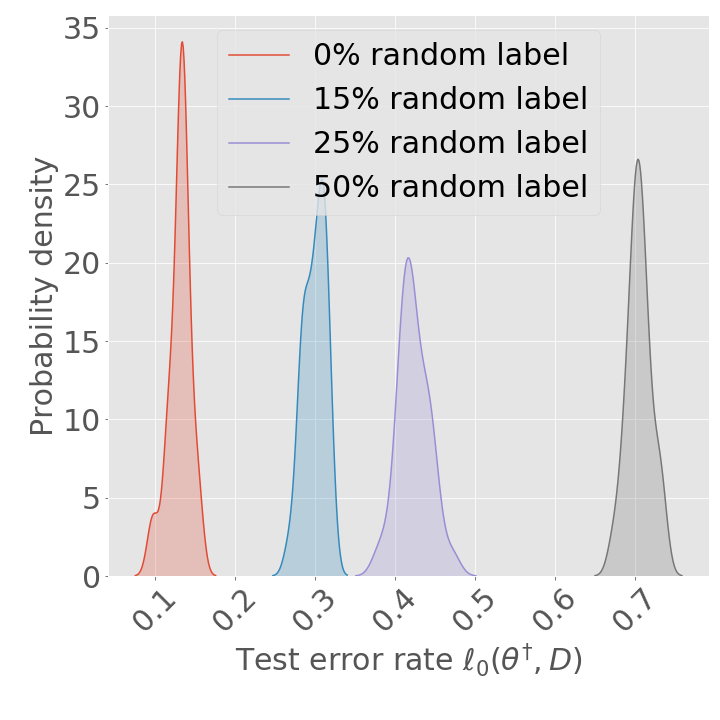}
 }  \vspace{-2mm}
 \subfigure[Diagonal Elements of  Hessian.]{
 \includegraphics[width =  0.31\textwidth, height = 0.24\textwidth]{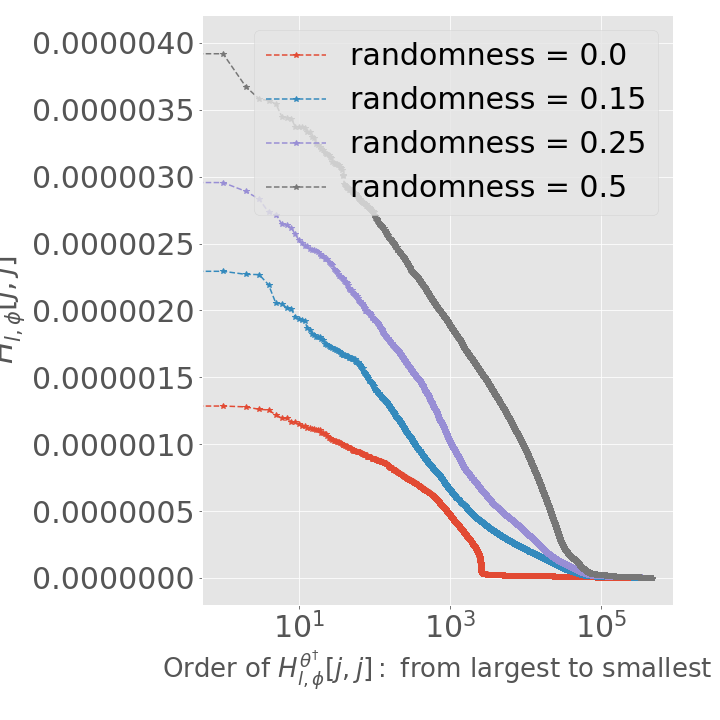}
 }  \vspace{-2mm}
 \subfigure[Effective Curvature.]{
 \includegraphics[width =  0.31\textwidth, height = 0.24\textwidth]{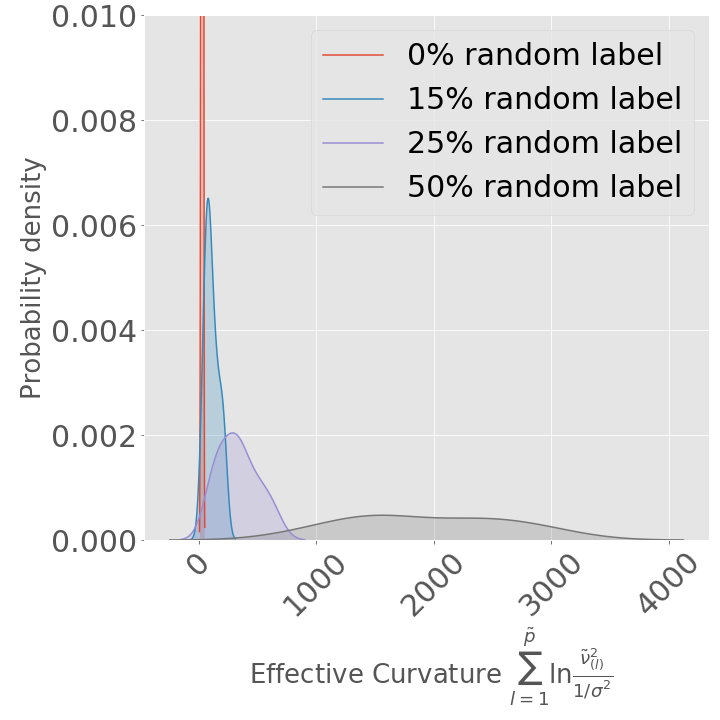}
 }  \vspace{-2mm}
 \subfigure[$L_2$ norm / no. sample.]{
 \includegraphics[width =  0.31\textwidth, height = 0.24\textwidth]{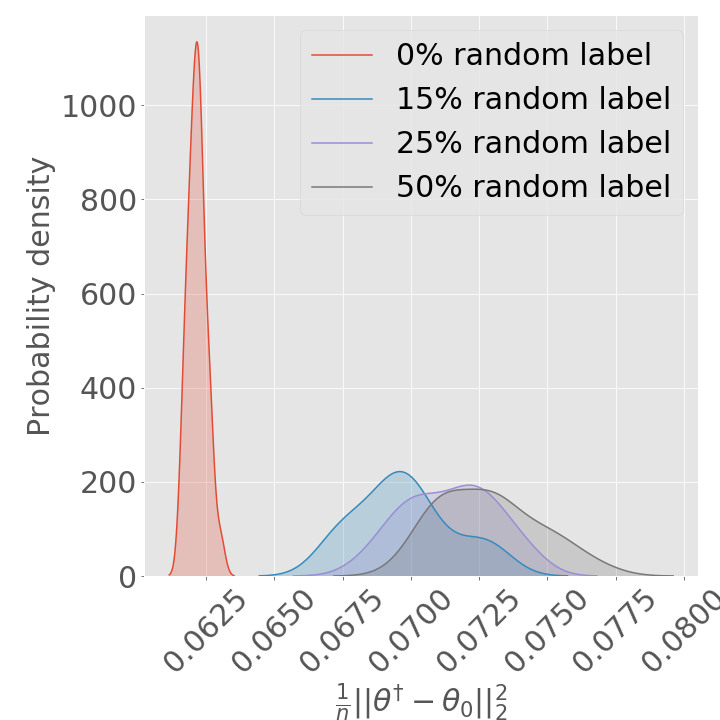}
 } 
  \subfigure[Margin Loss.]{
 \includegraphics[width =  0.31\textwidth, height = 0.24\textwidth]{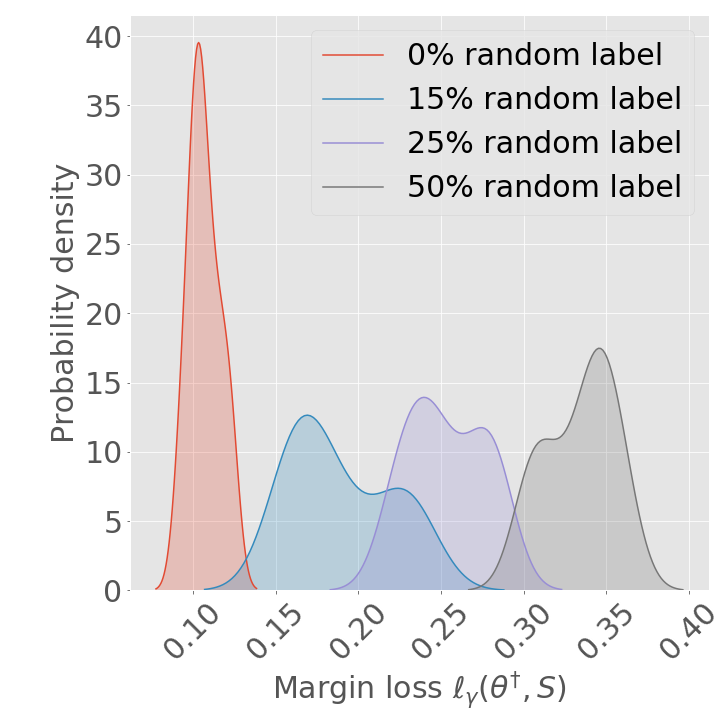}
 } 
   \subfigure[Scale-invariant Generalization Bound.]{
  \includegraphics[width =  0.31\textwidth, height = 0.24\textwidth]{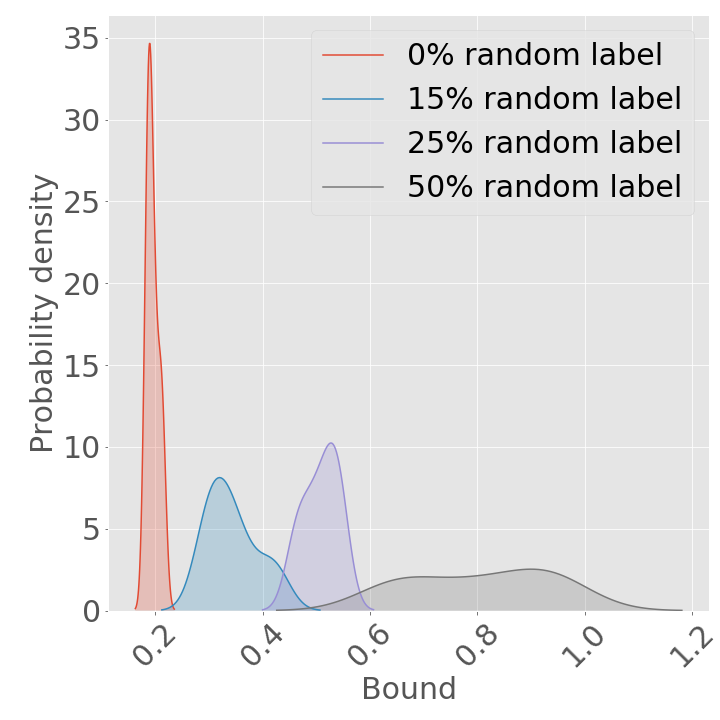}
  } 
\vspace{-2mm}
\caption[]{
Results for ReLU-nets with depth = 4, width =256, total 466,698 parameters, trained on 1000 samples from MNIST with batch size = 128.
(a-f) refer to Figure \ref{fig:mnist_d6_l128}.
Increasing percentage of random labels, the generalization bound as well as the components 
increase, and the bound in (f) stays valid for the test error rate in (a).}
\vspace*{-4mm}
\label{fig:mnist_d4_l256}
\end{figure*}

\begin{figure*}[t] 
\centering
 \subfigure[Test Error Rate]{
 \includegraphics[width =  0.31\textwidth, height = 0.24\textwidth]{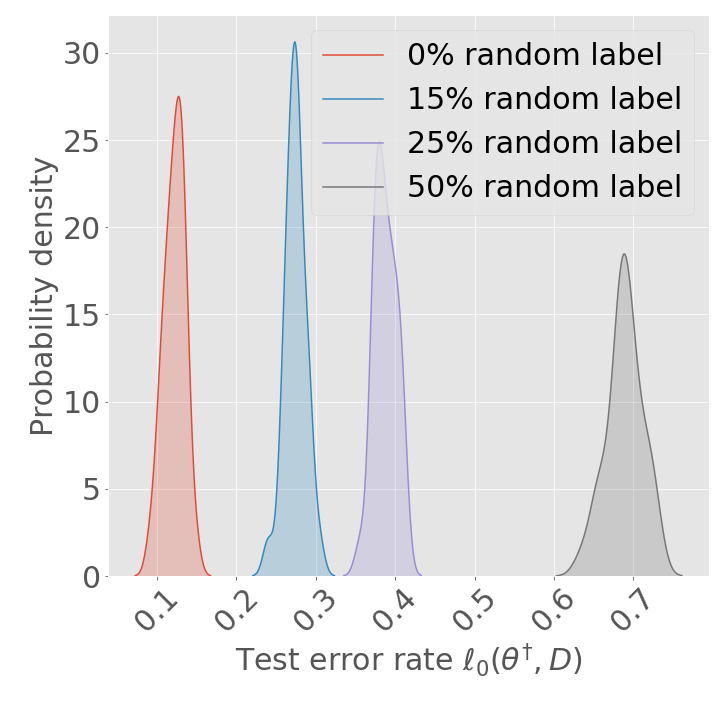}
 }  \vspace{-2mm}
 \subfigure[Diagonal Elements of  Hessian.]{
 \includegraphics[width =  0.31\textwidth, height = 0.24\textwidth]{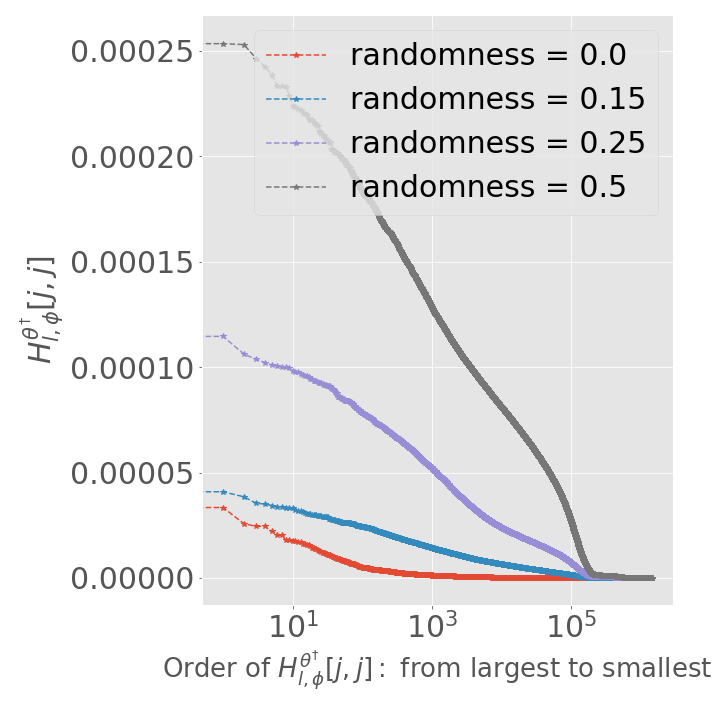}
 }  \vspace{-2mm}
 \subfigure[Effective Curvature.]{
 \includegraphics[width =  0.31\textwidth, height = 0.24\textwidth]{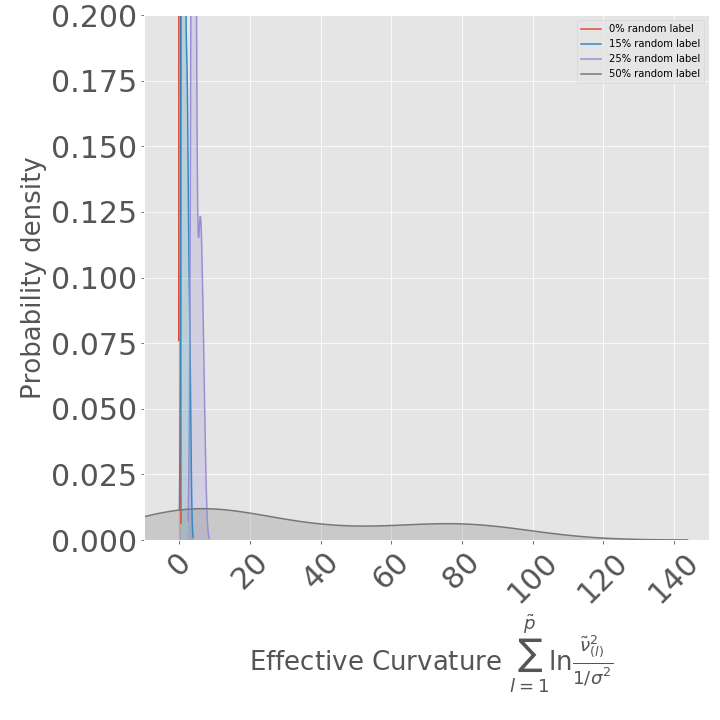}
 }  \vspace{-2mm}
 \subfigure[$L_2$ norm / no. sample.]{
 \includegraphics[width =  0.31\textwidth, height = 0.24\textwidth]{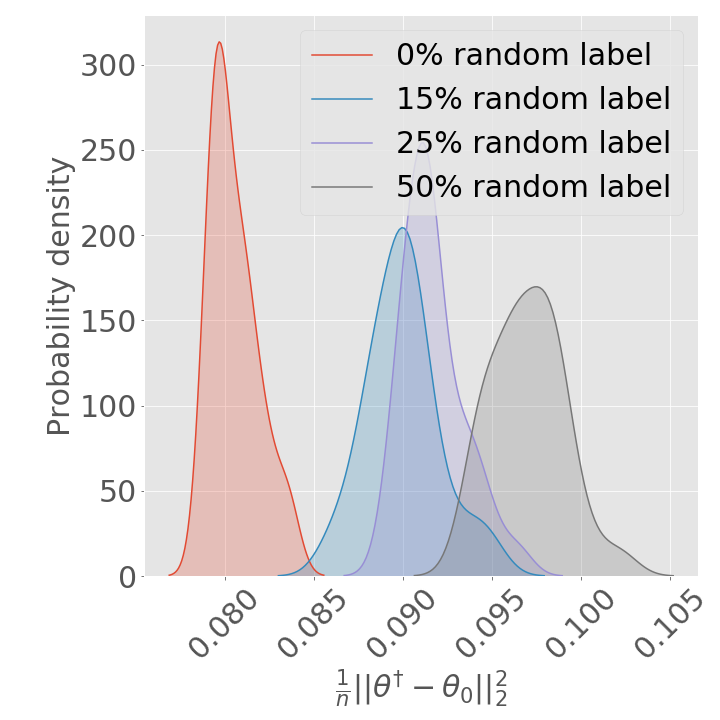}
 } 
  \subfigure[Margin Loss.]{
 \includegraphics[width =  0.31\textwidth, height = 0.24\textwidth]{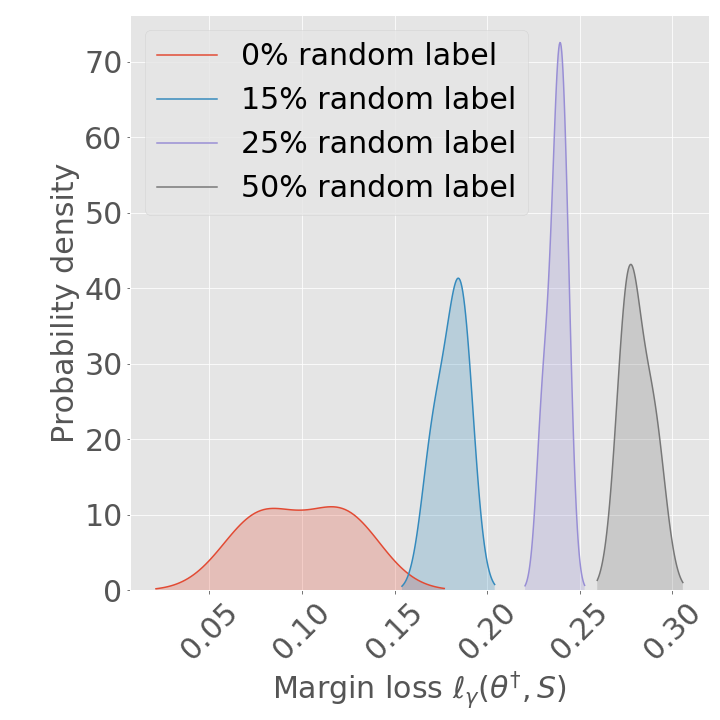}
 } 
   \subfigure[Scale-invariant Generalization Bound.]{
  \includegraphics[width =  0.31\textwidth, height = 0.24\textwidth]{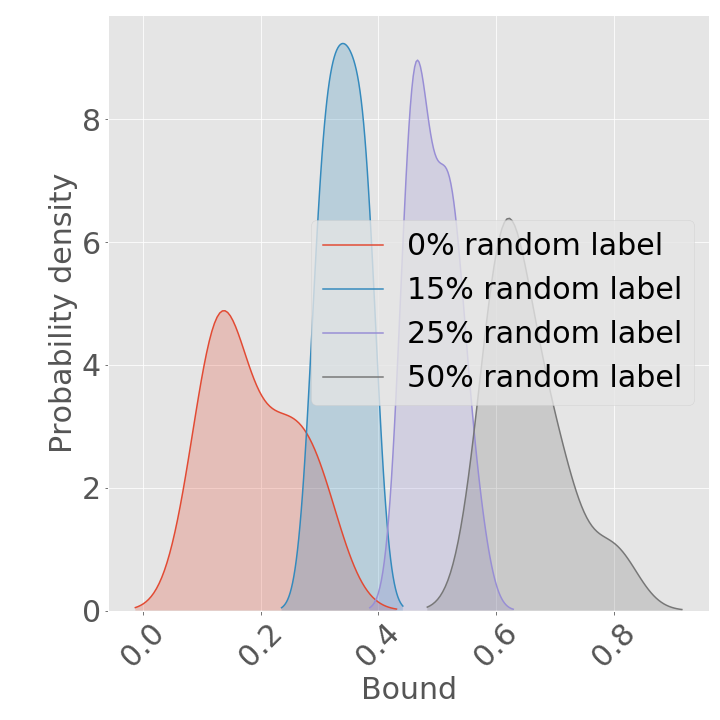}
  } 
\vspace{-2mm}
\caption[]{
Results for ReLU-nets with depth = 4, width =512, total 1,457,674 parameters, trained on 1000 samples from MNIST with batch size = 128.
(a-f) refer to Figure \ref{fig:mnist_d6_l128}.
Increasing percentage of random labels, the generalization bound as well as the components 
increase, and the bound in (f) stays valid for the test error rate in (a).
}
\vspace*{-4mm}
\label{fig:mnist_d4_l512}
\end{figure*}

\noindent {\bf MNIST.} The results for MNIST with ReLU network with width = 128 and depth = $\{2, 4, 6\}$ are presented in Figure \ref{fig:main_mnist_d2_l128}, \ref{fig:main_mnist_rand}, and \ref{fig:mnist_d6_l128}. The results suggests that our bound is also valid for ReLU network with different depth. The results for MNIST with ReLU network with depth = 4 and depth = $\{128, 256, 512\}$ are presented in Figure \ref{fig:main_mnist_rand}, \ref{fig:mnist_d4_l256}, and \ref{fig:mnist_d4_l512}, suggesting that our bound holds for ReLU network with different width. We also presented the result when training the ReLU network with micro-batch 16 in Figure \ref{fig:main_mnist_d4_l128_bs16}, which shows that our bound holds for micro-batch training as well.

\begin{figure*}[h] 
\vspace{-3mm}
\centering
 \subfigure[Test Error Rate]{
 \includegraphics[width =  0.31\textwidth, height = 0.24\textwidth]{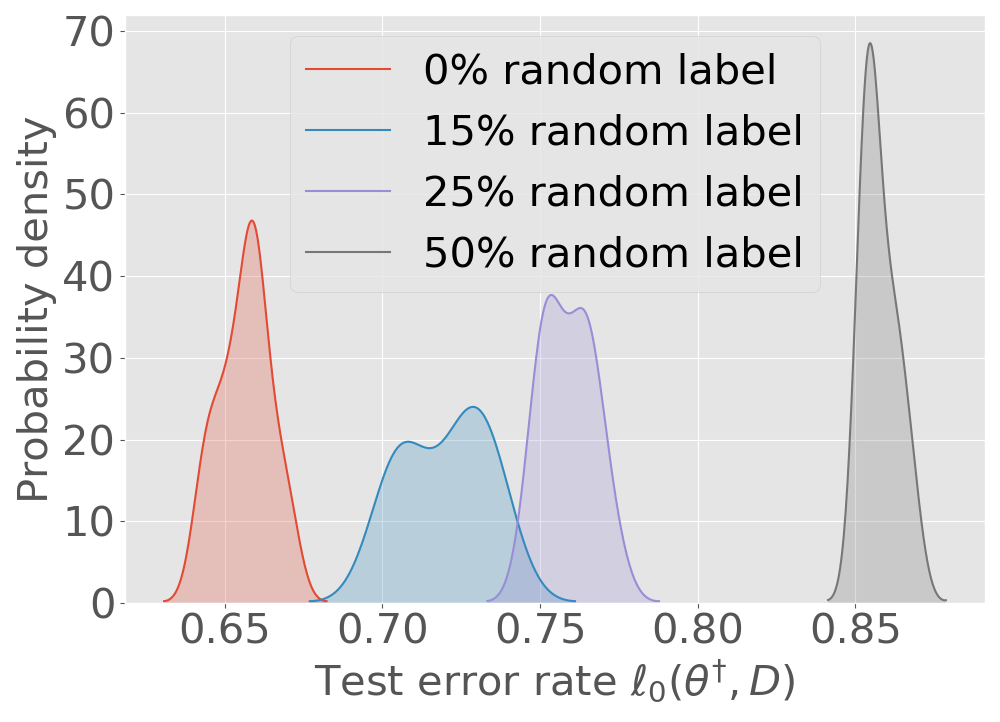}
 }  \vspace{-2mm}
 \subfigure[Diagonal Elements of  Hessian.]{
 \includegraphics[width =  0.31\textwidth, height = 0.24\textwidth]{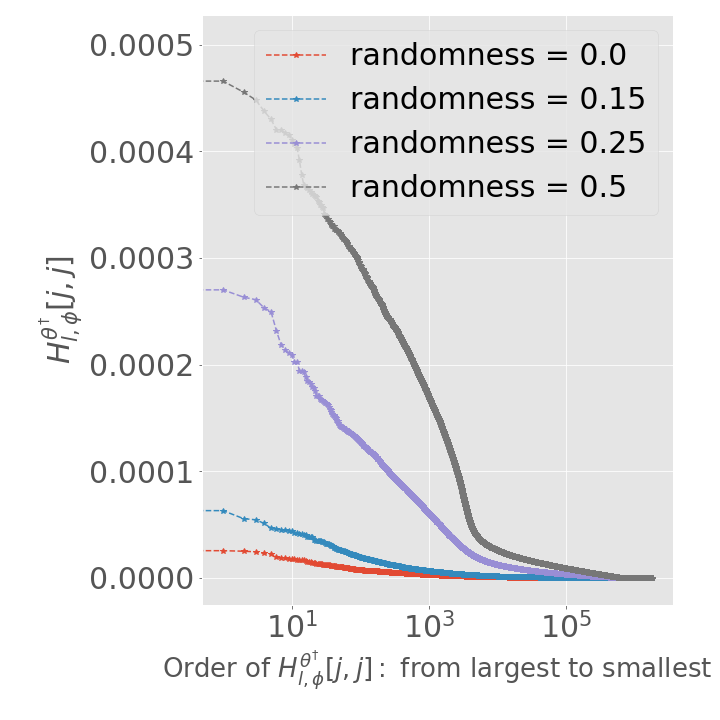}
 } 
 \subfigure[Effective Curvature.]{
 \includegraphics[width =  0.31\textwidth, height = 0.24\textwidth]{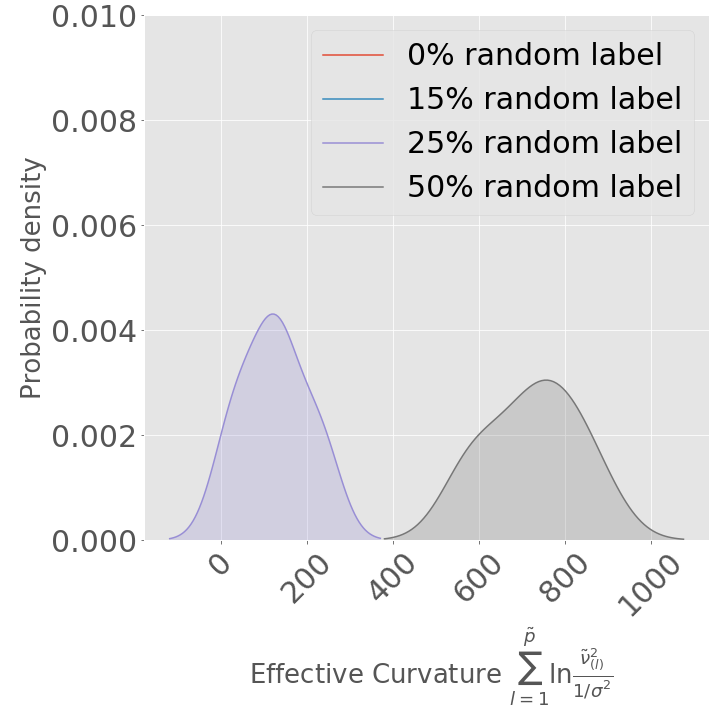}
 } 
 \subfigure[$L_2$ norm / no. sample.]{
 \includegraphics[width =  0.31\textwidth, height = 0.24\textwidth]{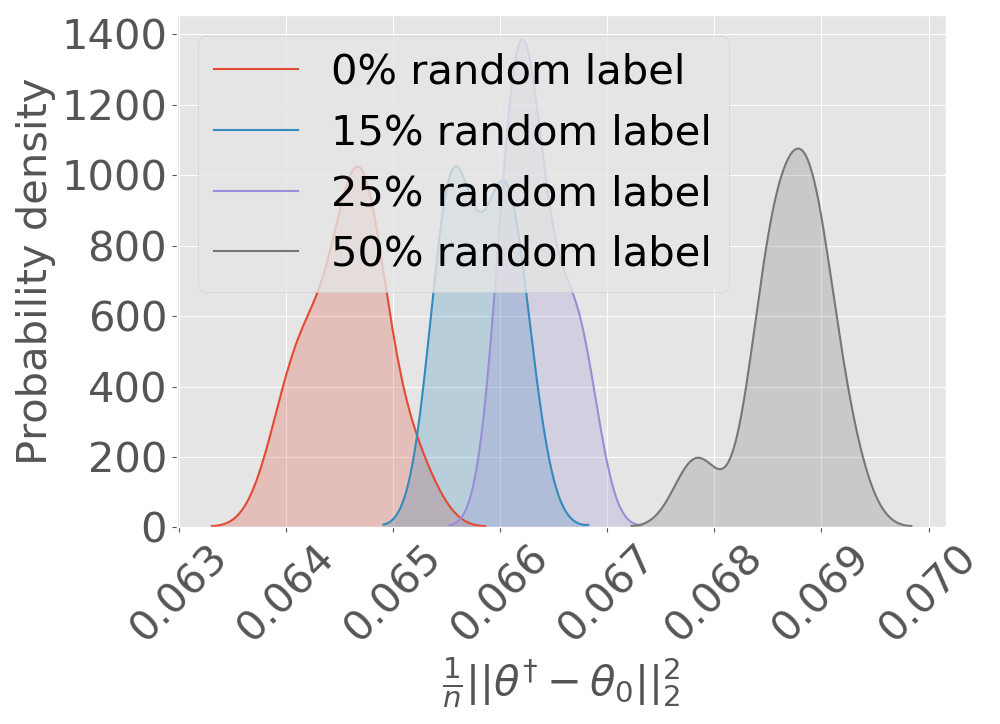}
 } 
  \subfigure[Margin Loss.]{
 \includegraphics[width =  0.31\textwidth, height = 0.24\textwidth]{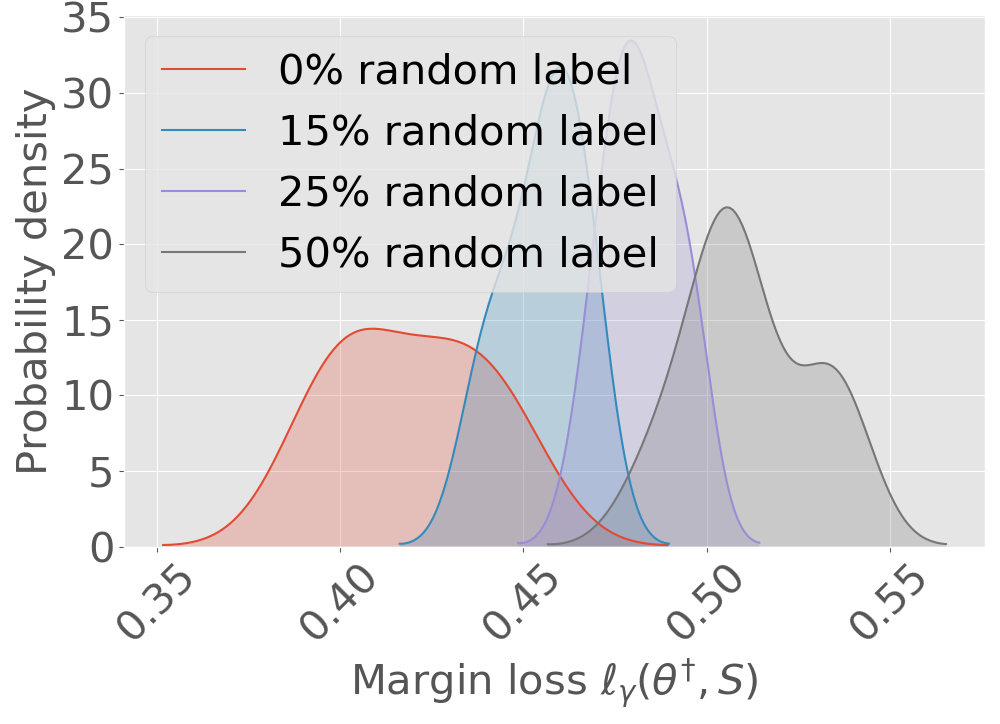}
 } 
   \subfigure[Scale-invariant Generalization Bound.]{
  \includegraphics[width =  0.31\textwidth, height = 0.24\textwidth]{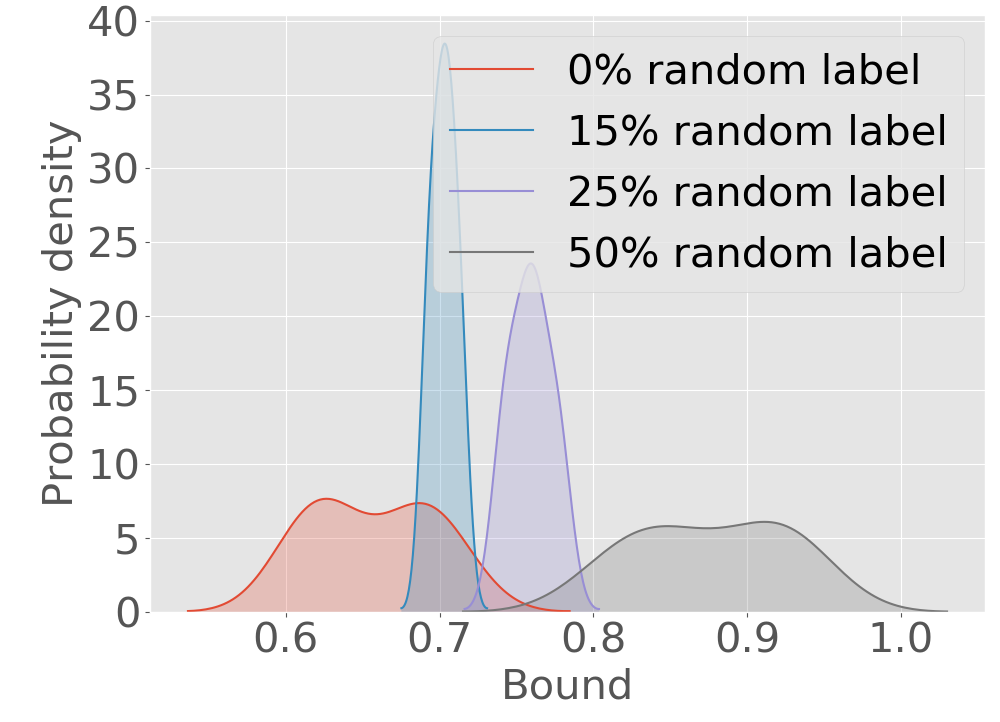}
  } 
\vspace{-4mm}
\caption[]{
Results for ReLU-nets with depth = 4, width =384, total 1,775,242 parameters, trained on 1000 samples from CIFAR-10 with batch size = 128. 
(a-f) refer to Figure \ref{fig:mnist_d6_l128}. In (c), the effective curvature for 0 \% and 15\% random label is zero.
Increasing percentage of random labels, the generalization bound as well as the components 
increase, and the bound in (f) stays valid for the test error rate in (a).
}
\vspace*{-4mm}
\label{fig:cifar_d4_l384}
\end{figure*}

\begin{figure*}[h] 
\vspace{-5mm}
\centering
 \subfigure[Test Error Rate]{
 \includegraphics[width =  0.31\textwidth, height = 0.24\textwidth]{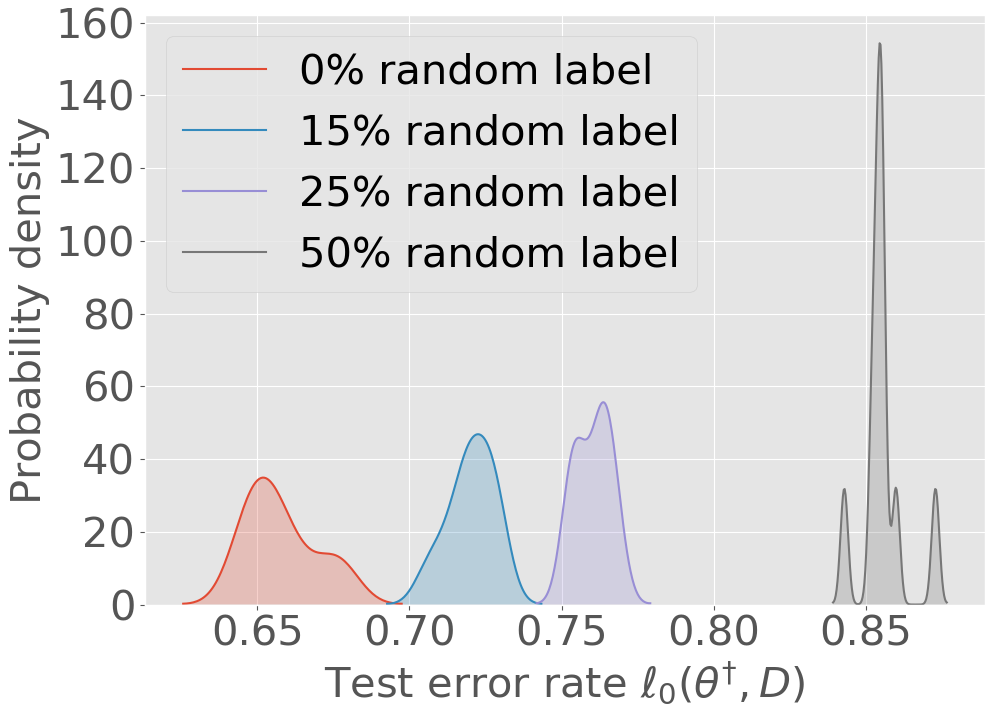}
 } 
 \subfigure[Diagonal Elements of  Hessian.]{
 \includegraphics[width =  0.31\textwidth, height = 0.24\textwidth]{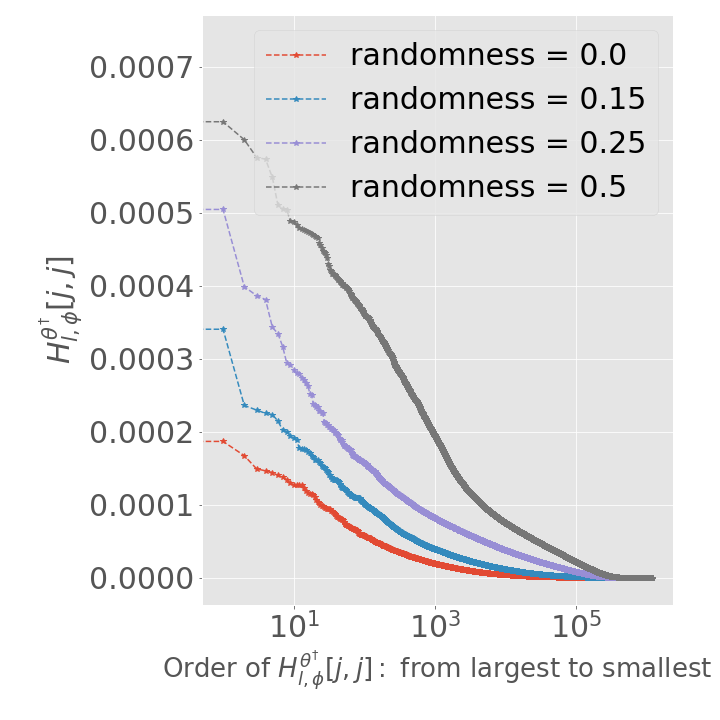}
 } 
 \subfigure[Effective Curvature.]{
 \includegraphics[width =  0.31\textwidth, height = 0.24\textwidth]{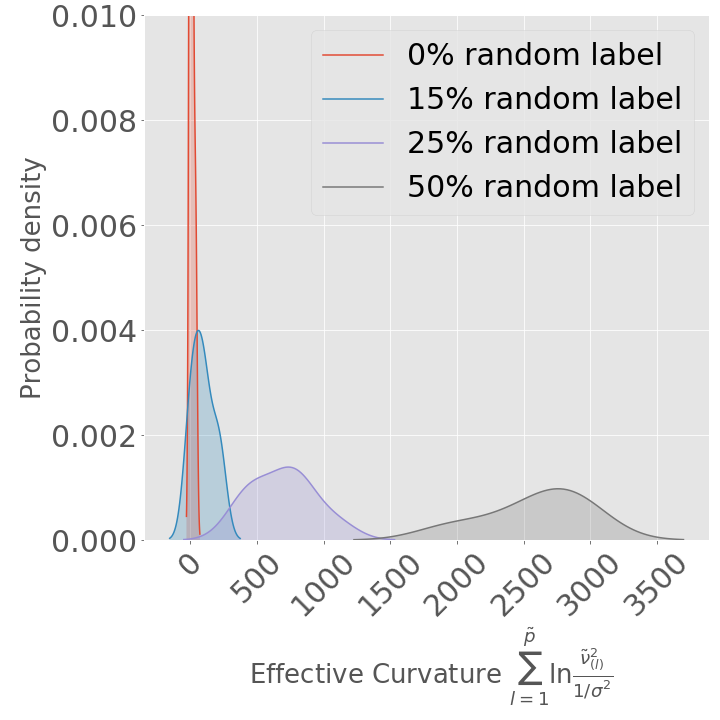}
 } 
 \subfigure[$L_2$ norm / no. sample.]{
 \includegraphics[width =  0.31\textwidth, height = 0.24\textwidth]{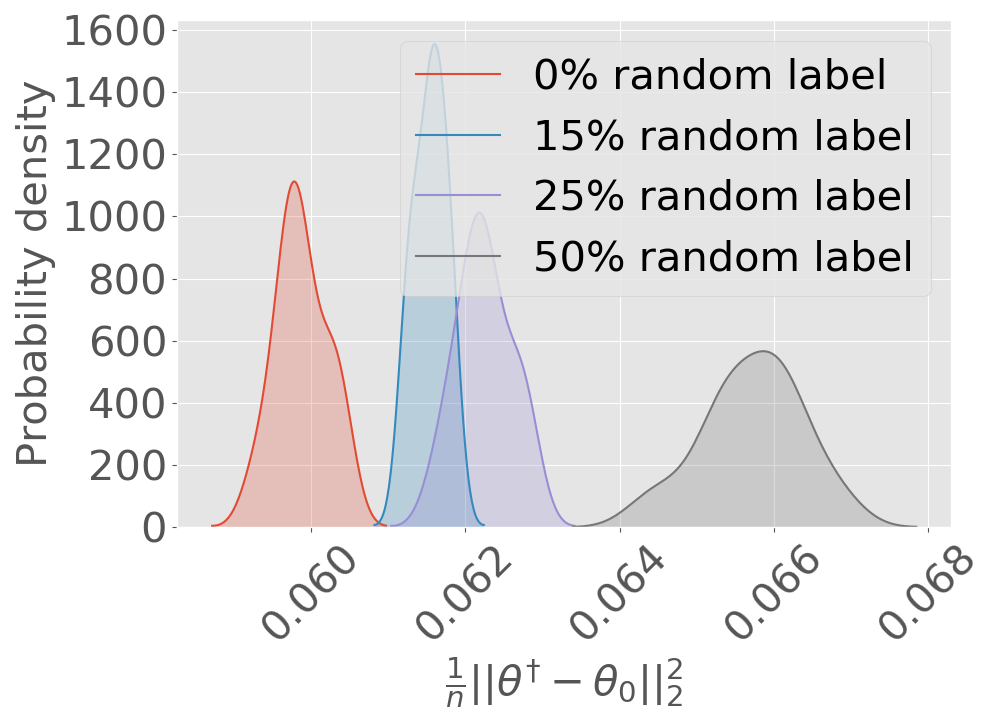}
 } 
  \subfigure[Margin Loss.]{
 \includegraphics[width =  0.31\textwidth, height = 0.24\textwidth]{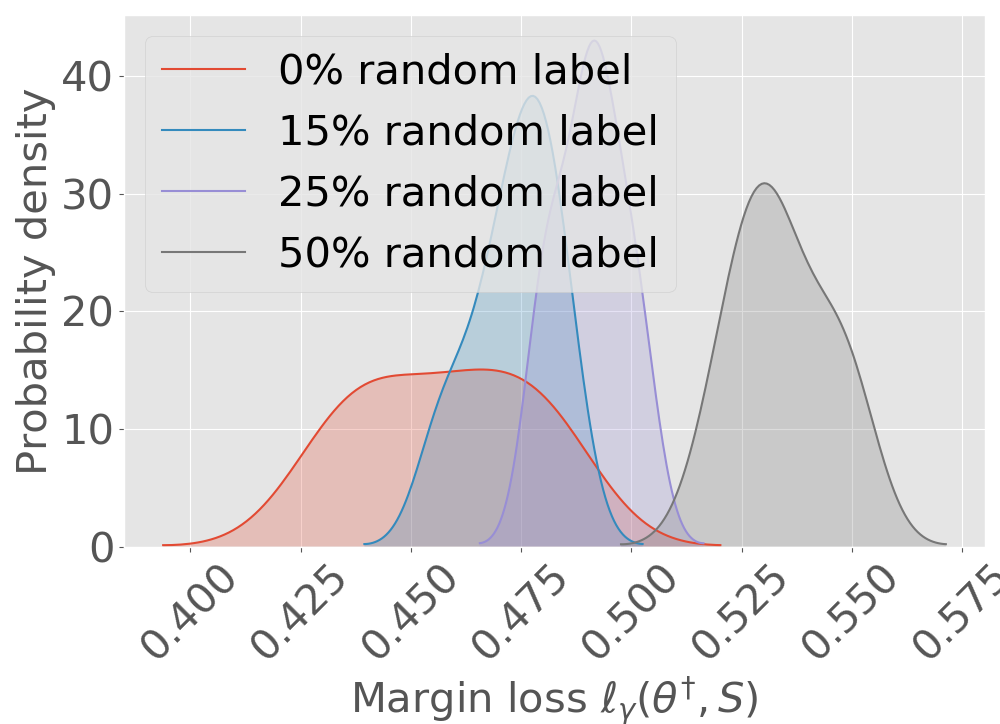}
 } 
   \subfigure[Scale-invariant Generalization Bound.]{
  \includegraphics[width =  0.31\textwidth, height = 0.24\textwidth]{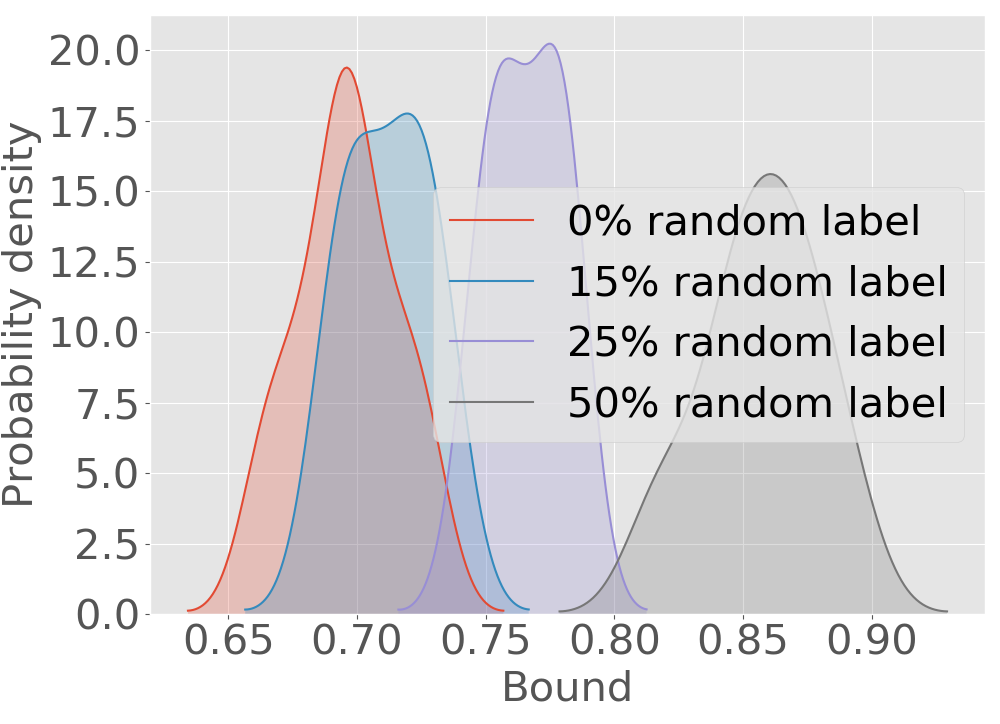}
  } 
\vspace{-4mm}
\caption[]{
Results for ReLU-nets with depth = 6, width =256, total 1,184,010 parameters, trained on 1000 samples from CIFAR-10 with batch size = 128. 
(a-f) refer to Figure \ref{fig:mnist_d6_l128}. 
Increasing percentage of random labels, the generalization bound as well as the components 
increase, and the bound in (f) stays valid for the test error rate in (a).
}
\vspace*{-4mm}
\label{fig:cifar_d6_l256}
\end{figure*}

\noindent {\bf CIFAR-10.} The results for CIFAR-10 with ReLU network with depth = 4 and width = $\{256, 384, 512\}$ are presented in Figure \ref{fig:main_cifar_d4_l256}, \ref{fig:cifar_d4_l384}, and \ref{fig:main_cifar_d4_l512}, suggesting that our bound holds for ReLU network with different width. The results for CIFAR-10 with ReLU network with width = 256 and depth = $\{ 4, 6\}$ are presented in Figure \ref{fig:main_cifar_d4_l256} and \ref{fig:cifar_d6_l256}, showing that the proposed bound is also valid for different depths. Figure \ref{fig:main_cifar_d4_l256_bs16} and \ref{fig:main_cifar_d4_l256} give the results for batch size 16 and batch size 128 for the same ReLU network structure i.e, depth = 4 and width = 256, showing that our bound is also valid for mirco-batch training and mini-batch training.



 \subsection{Additional Results for Sample Size and Generalization Bound}
 \label{app:exp_more_sample}
 \begin{table}[t]
\caption{Summary of the setting for each specific experiment with MNIST dataset and CIFAR-10 dataset.}
\label{table:real}
\centering
\begin{tabular}{l|ccc|ccc}
\hline 
& \multicolumn{3}{c|}{MNIST} & \multicolumn{3}{c}{CIFAR-10}\\
\hline
No. of Classes k: & \multicolumn{3}{c|}{10} & \multicolumn{3}{c}{10}\\
Input Dimension:  & \multicolumn{3}{c|}{784}& \multicolumn{3}{c}{3072}\\
Training set size: & \multicolumn{3}{c|}{[100, 1000, 10000]} & \multicolumn{3}{c}{[100, 1000, 10000]}\\
\hline
&\multicolumn{6}{c}{Network Structure}\\
\hline 
No. of Layers: & \multicolumn{3}{c|}{[2, 4, 8]} & \multicolumn{3}{c}{[4, 6, 8]}\\
No. of Nodes per Layer:& \multicolumn{3}{c|}{[128, 256, 512]} & \multicolumn{3}{c}{[256, 384, 512]}\\
\hline
Batch size:& \multicolumn{3}{c|}{[16, 128]} & \multicolumn{3}{c}{[16, 128]}\\
\hline
\end{tabular}
\label{table:more_sample}
\end{table}

We also evaluate the proposed generalization bound for ReLU network with various depth and various width for both MNIST and CIFAR-10 datasets. For MNIST, we present the results for depth = 4 and width = $\{128, 256, 512\}$ to illustrate how the bound behaves for different widths for a fixed depth.  We  also present the results for width = 128 and depth = $\{2, 4, 6\}$ to illustrate how the bound behaves for different depths for a fixed width. For CIFAR-10, we present the results for depth = 4 and width = $\{256, 384, 512\}$ to illustrate how the bound behaves for different widths for a fixed depth.  We  also present the results for width = 256 and depth = $\{4, 6\}$ to illustrate how the bound behaves for different depths for a fixed width. We also evaluate our bound for micro-batch training, i.e., batch size = 16. The detailed setup is in Table \ref{table:more_sample}.

\begin{figure*}[h] 
\centering
 \subfigure[Test Error Rate]{
 \includegraphics[width = 0.31 \textwidth]{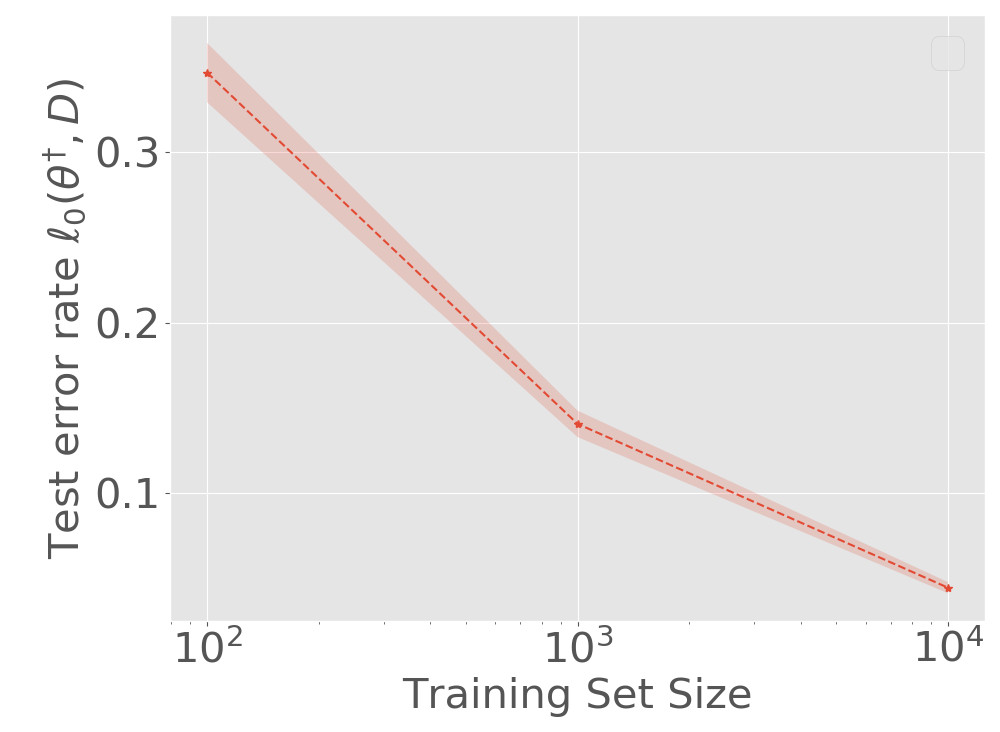}
 } 
 \subfigure[Diagonal Elements of  Hessian.]{
 \includegraphics[width = 0.31 \textwidth, height =0.23\textwidth]{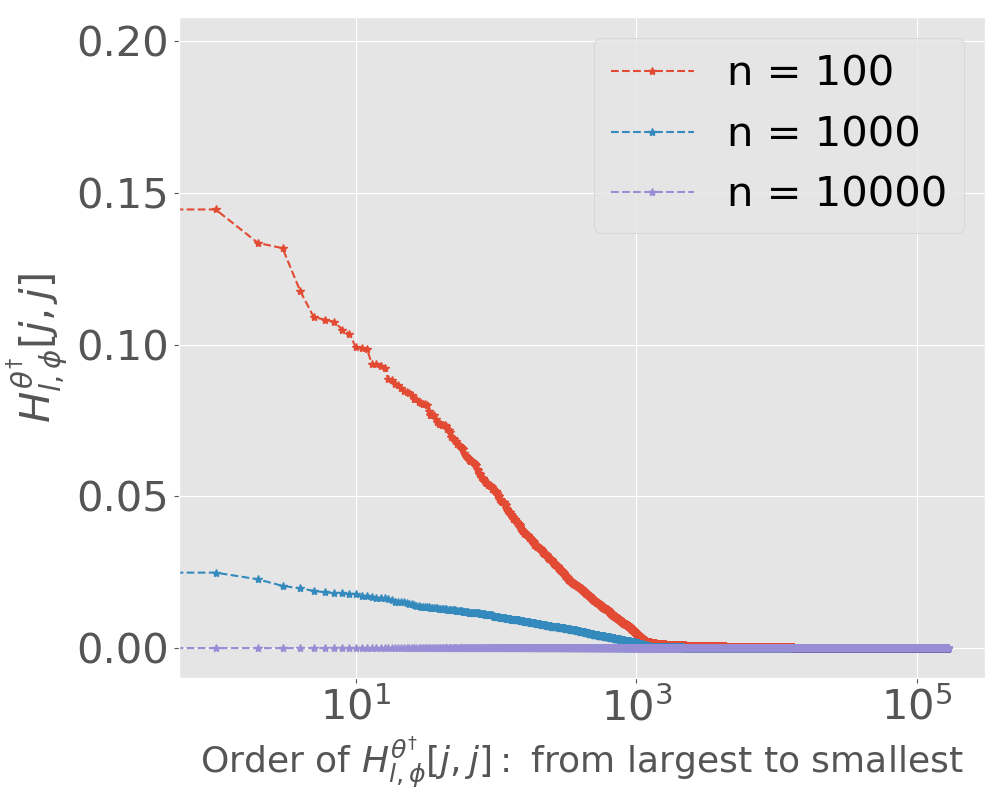}
 } 
 \subfigure[Effective Curvature.]{
 \includegraphics[width = 0.31 \textwidth, height =0.23\textwidth]{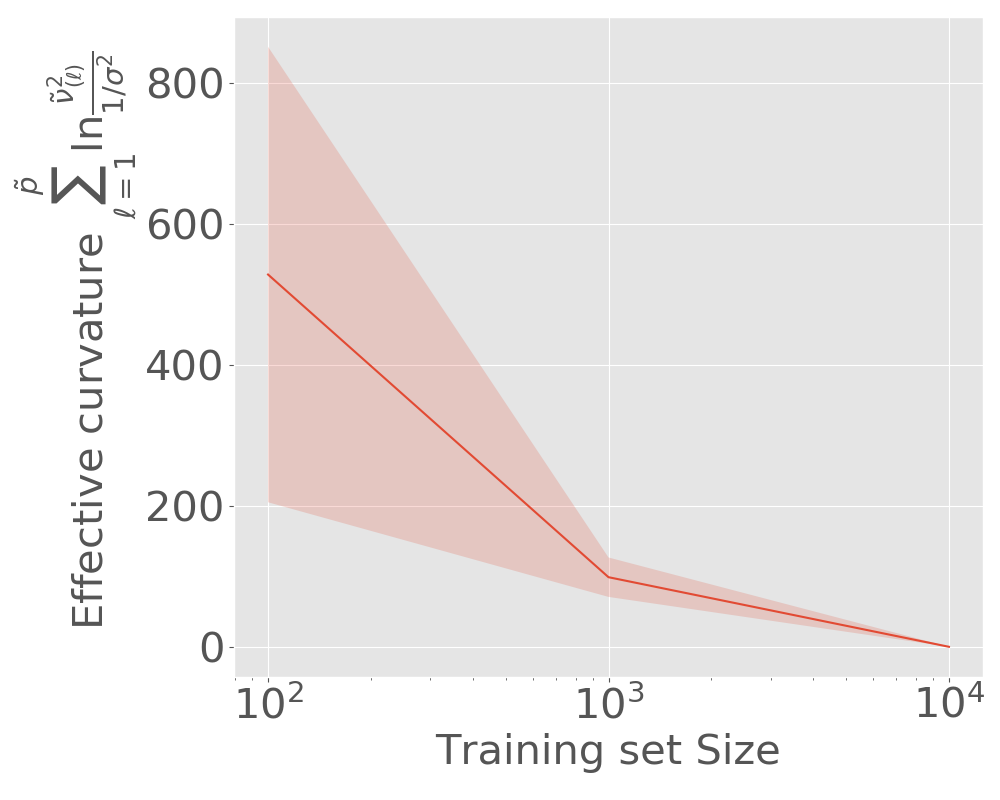}
 } 
 \subfigure[$L_2$ norm / no. sample.]{
 \includegraphics[width = 0.31 \textwidth]{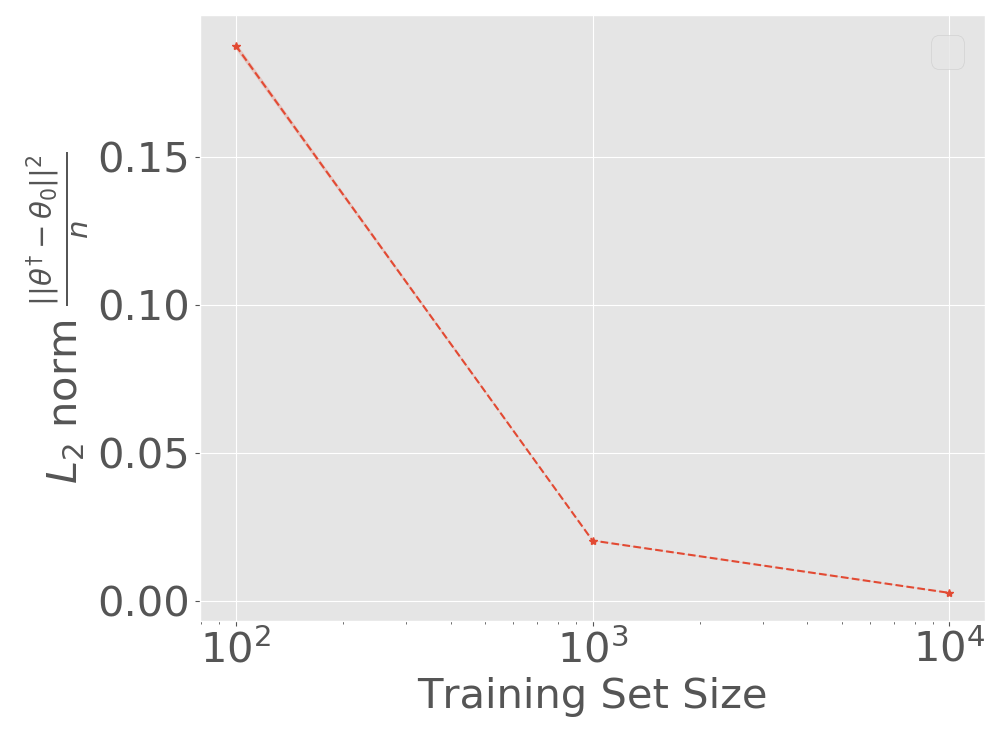}
 } 
  \subfigure[Margin Loss.]{
 \includegraphics[width = 0.31 \textwidth]{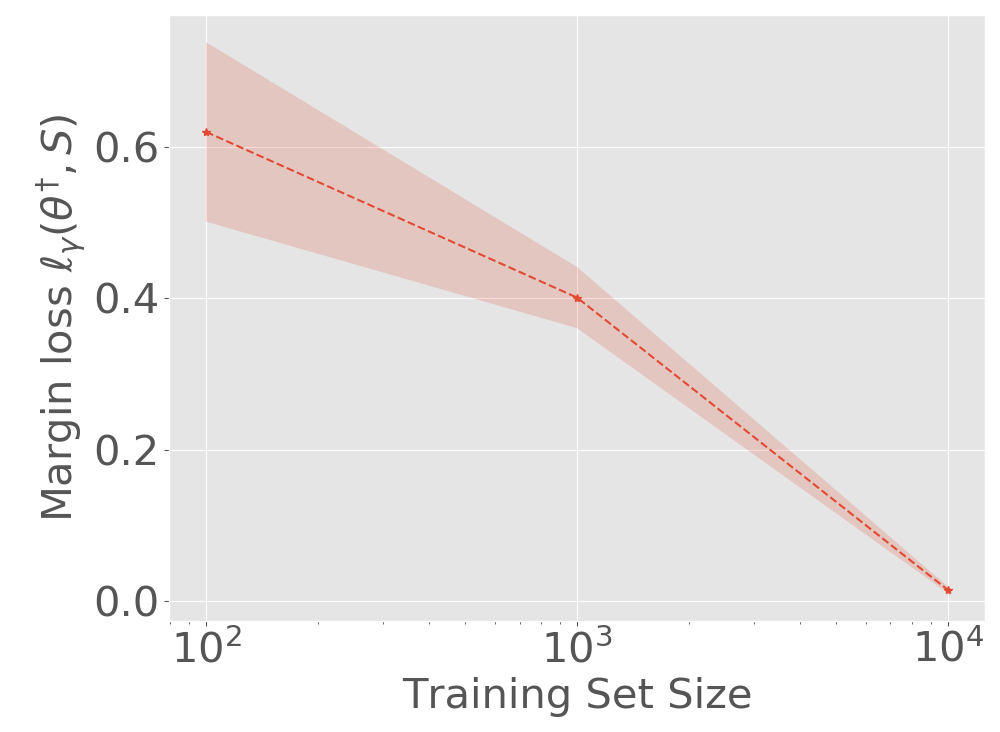}
 } 
  \subfigure[Scale-invariant Generalization Bound.]{
 \includegraphics[width = 0.31 \textwidth, height =0.23\textwidth]{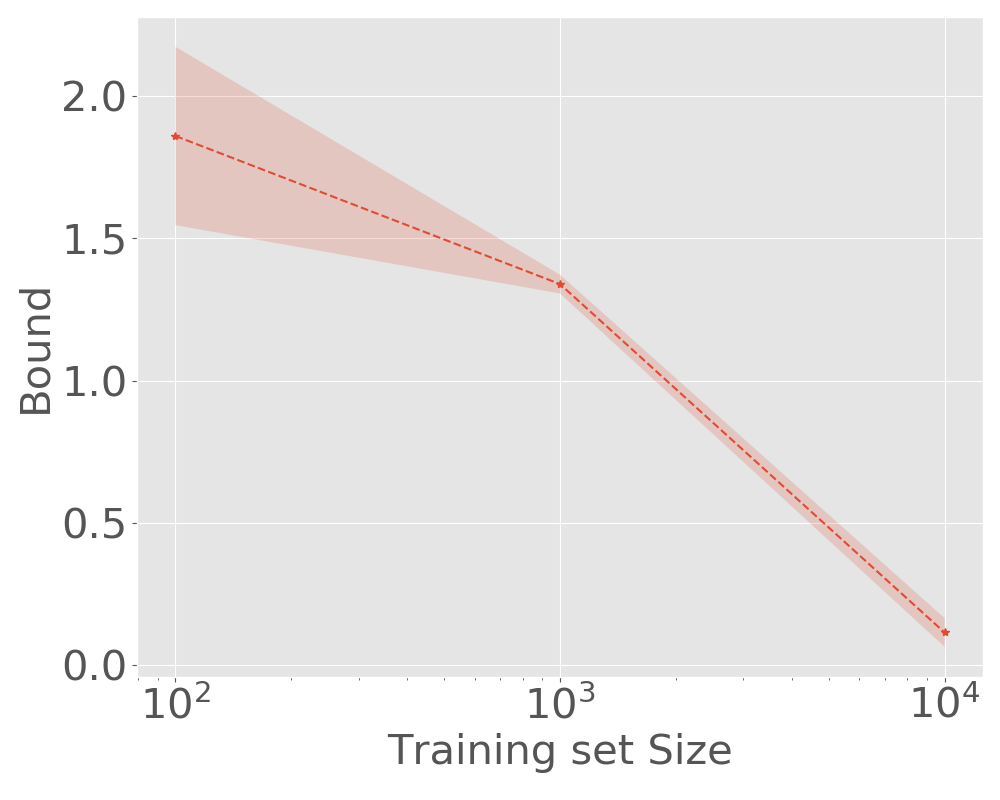}
 } 
\vspace{-2mm}
\caption[]{
Results for ReLU-nets with depth = 4, width =128 total 167,818 parameters, trained on MNIST with batch size = 128.
 (a) test set error rate; (b) diagonal elements (mean) of $\tilde \cH_{l,\phi}^{\theta^{\dagger}}$; (c) effective curvature; (d) $L_2$ norm of $\theta^{\dagger}$; (e) margin loss; (f)  generalization bound. 
The bound and all its components decrease with increase in $n$ from 100 to 10,000.
}
\label{fig:mnist_sample_d4_l128}
\vspace*{-4mm}
\end{figure*}

\begin{figure*}[h] 
\centering
 \subfigure[Test Error Rate]{
 \includegraphics[width = 0.31 \textwidth]{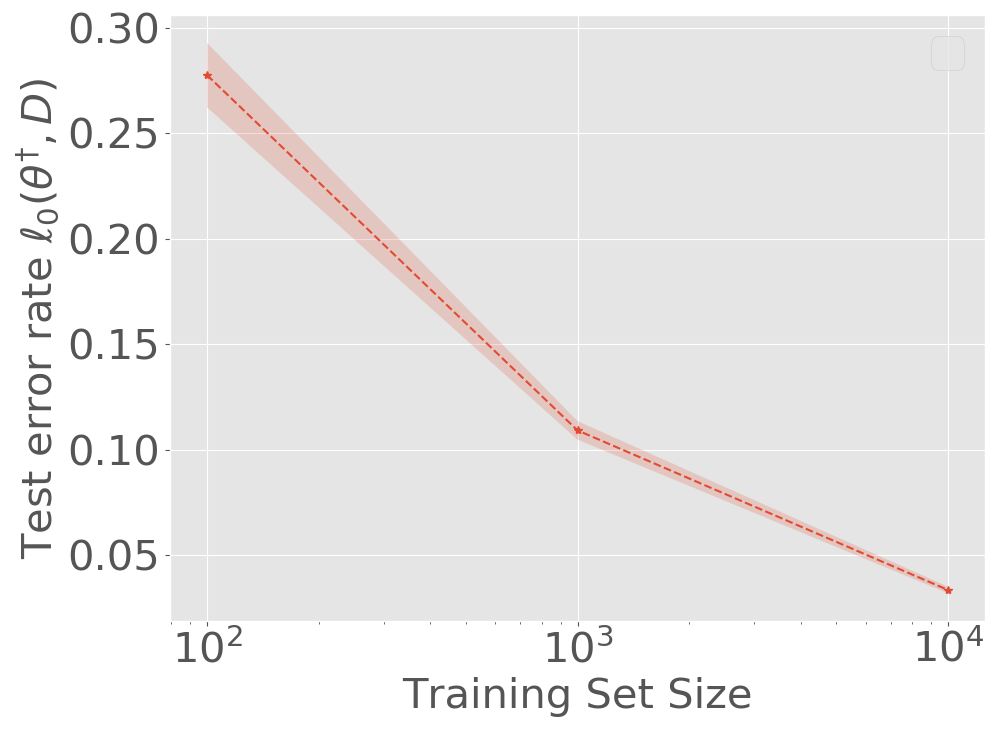}
 } 
 \subfigure[Diagonal Elements of  Hessian.]{
 \includegraphics[width = 0.31 \textwidth, height =0.23\textwidth]{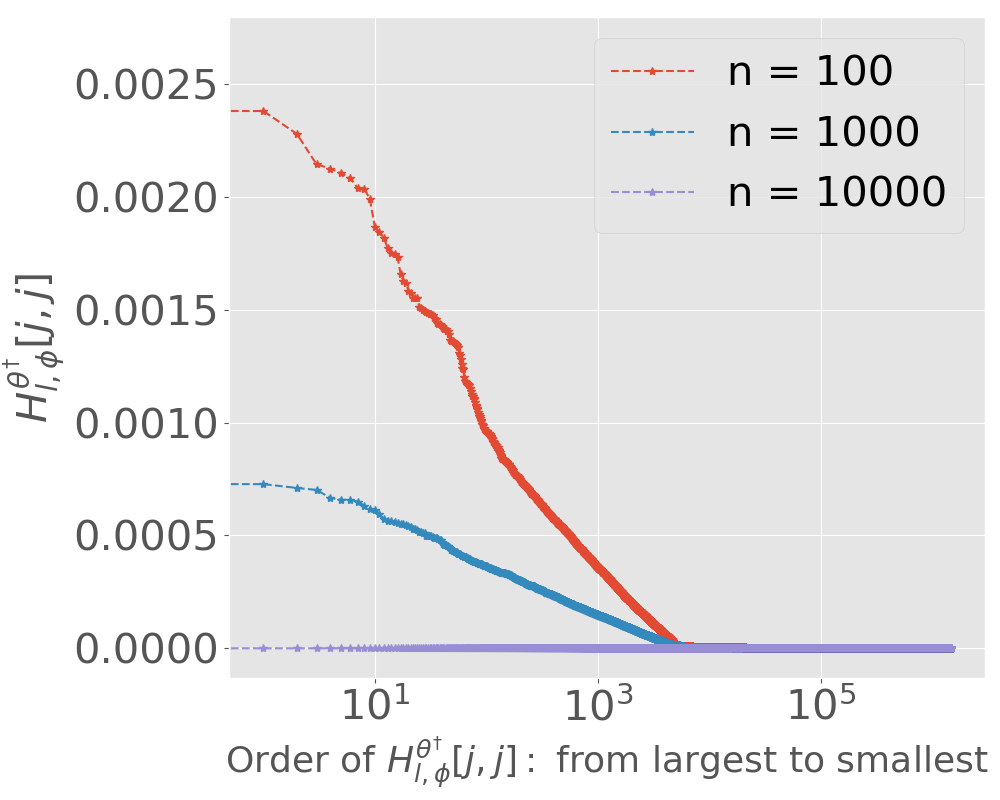}
 } 
 \subfigure[Effective Curvature.]{
 \includegraphics[width = 0.31 \textwidth, height =0.23\textwidth]{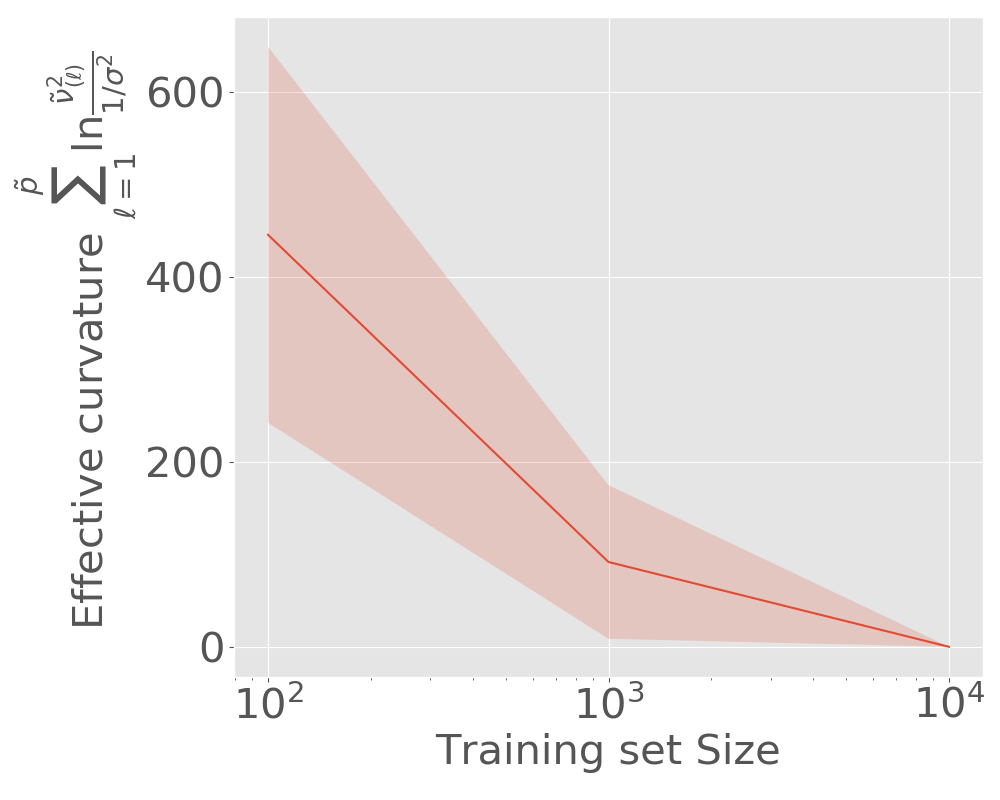}
 } 
 \subfigure[$L_2$ norm / no. sample.]{
 \includegraphics[width = 0.31 \textwidth]{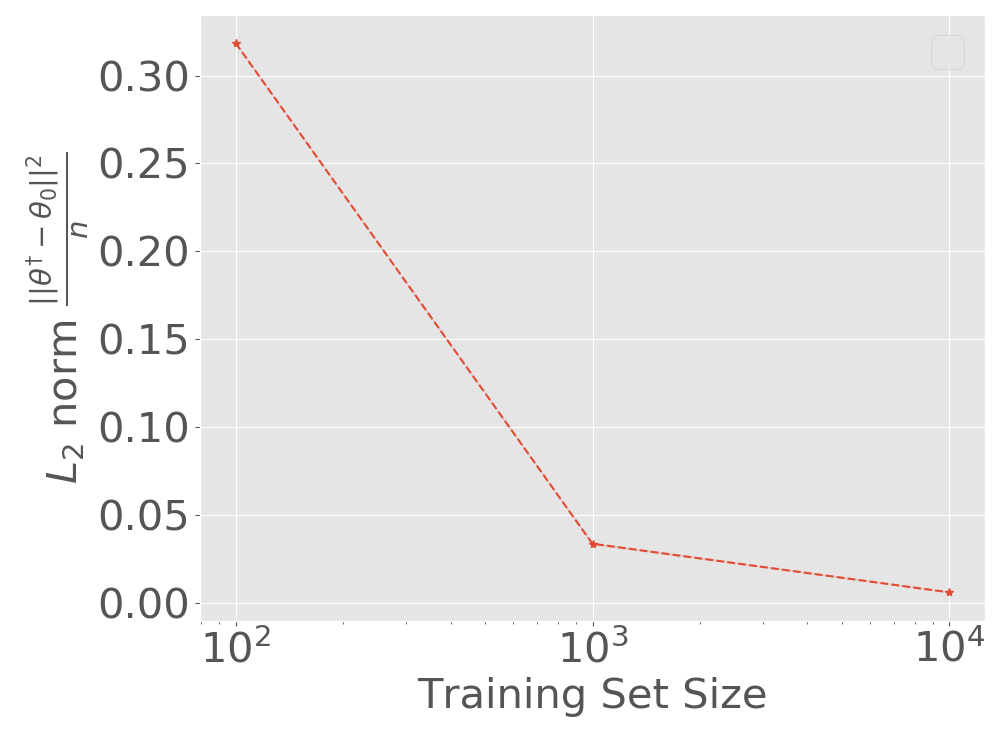}
 } 
  \subfigure[Margin Loss.]{
 \includegraphics[width = 0.31 \textwidth]{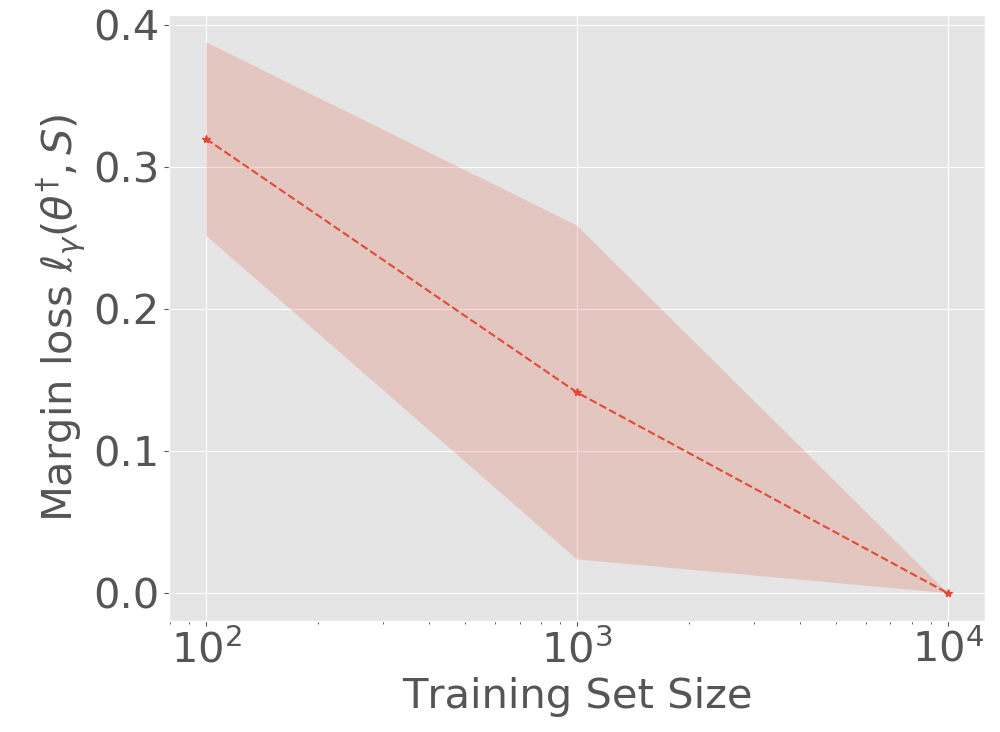}
 } 
  \subfigure[Scale-invariant Generalization Bound.]{
 \includegraphics[width = 0.31 \textwidth, height =0.23\textwidth]{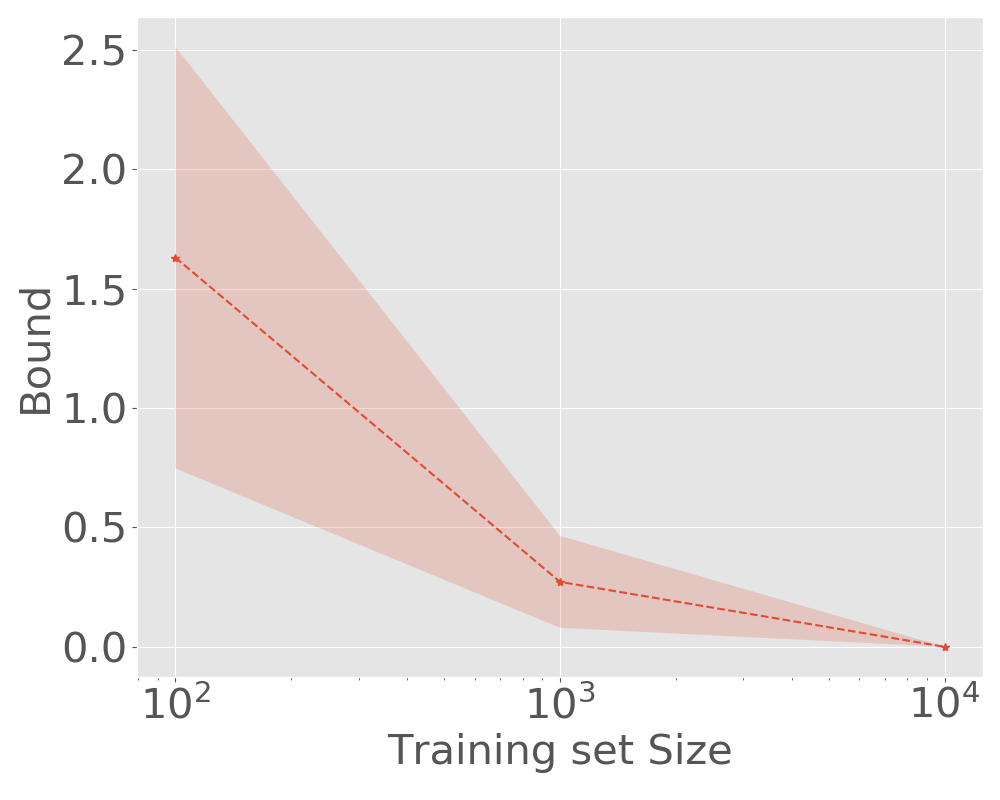}
 } 
\vspace{-2mm}
\caption[]{
Results for ReLU-nets with depth = 4, width =512 total 1,457,674 parameters, trained on MNIST with batch size = 128.
(a-f) refer to Figure \ref{fig:mnist_sample_d4_l128}.
The bound and all its components decrease with increase in $n$ from 100 to 10,000.
}
\label{fig:mnist_sample_d4_l512}
\vspace*{-4mm}
\end{figure*}

\begin{figure*}[h] 
\centering
 \subfigure[Test Error Rate]{
 \includegraphics[width = 0.31 \textwidth]{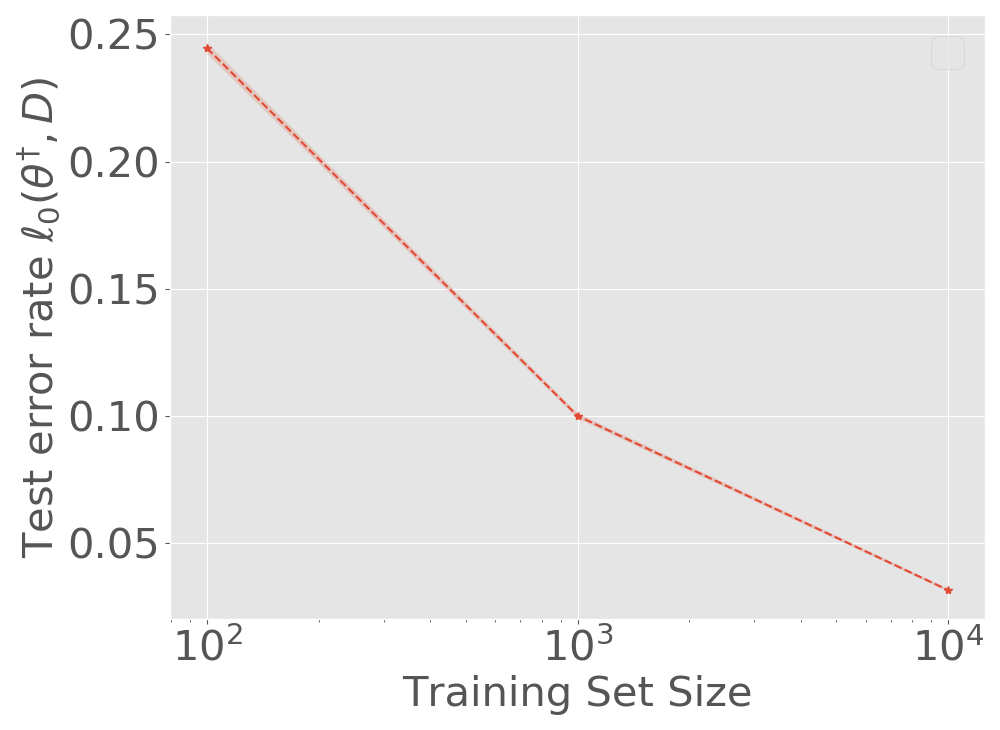}
 } 
 \subfigure[Diagonal Elements of  Hessian.]{
 \includegraphics[width = 0.31 \textwidth, height =0.23\textwidth]{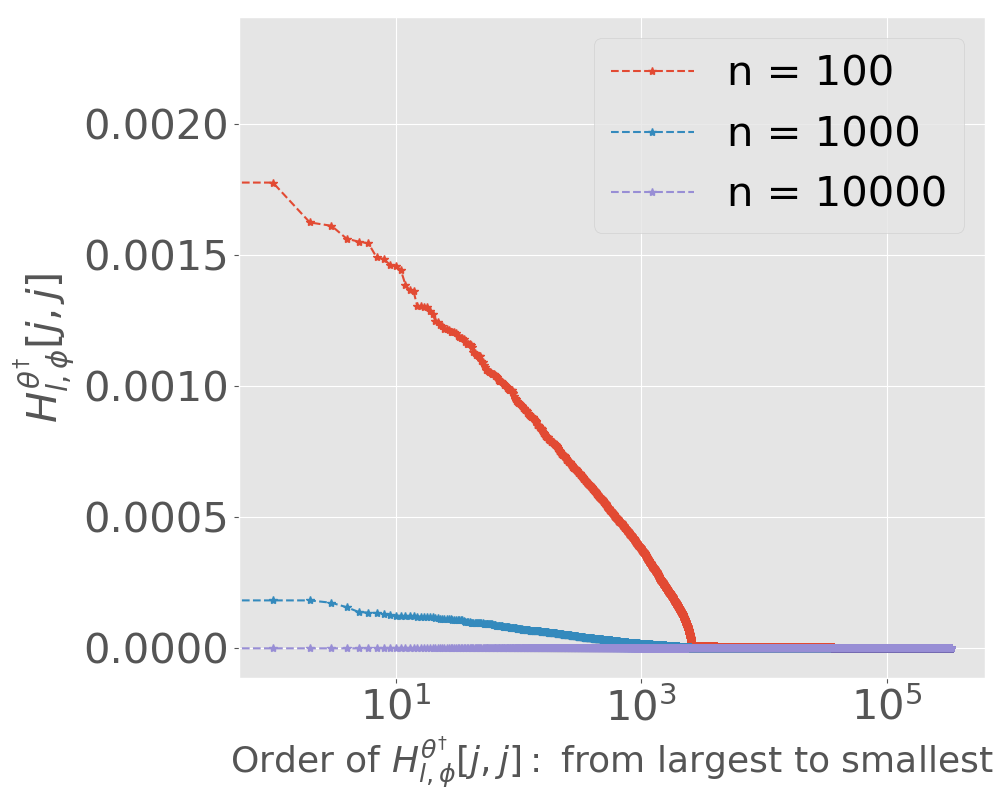}
 } 
 \subfigure[Effective Curvature.]{
 \includegraphics[width = 0.31 \textwidth, height =0.23\textwidth]{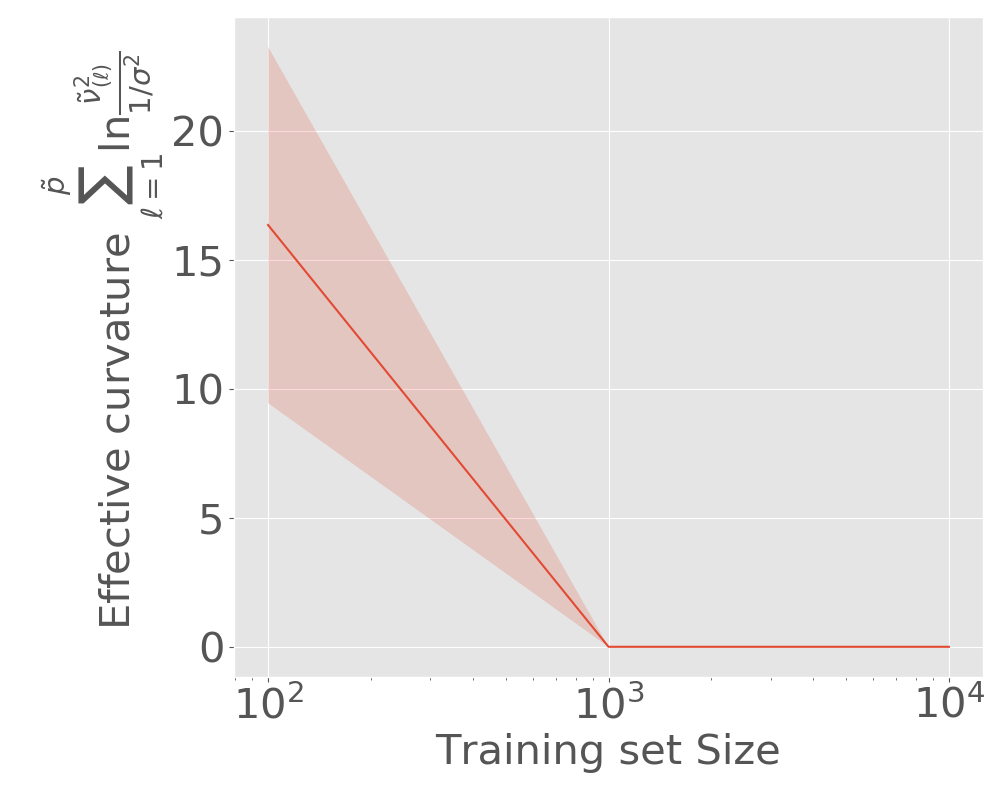}
 } 
 \subfigure[$L_2$ norm / no. sample.]{
 \includegraphics[width = 0.31 \textwidth]{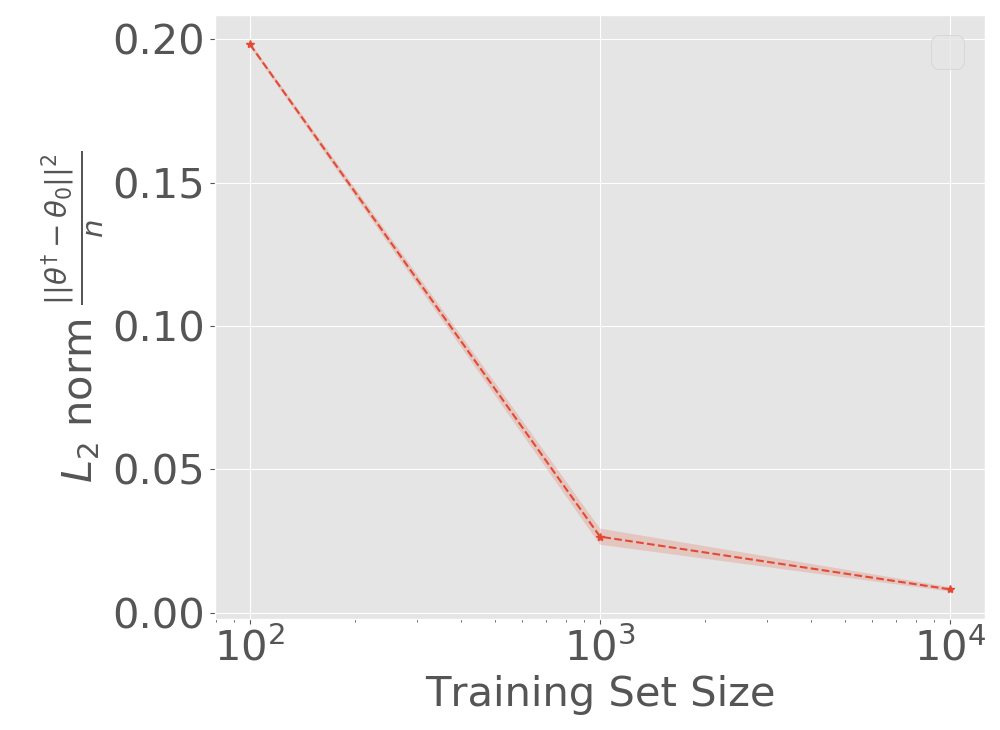}
 } 
  \subfigure[Margin Loss.]{
 \includegraphics[width = 0.31 \textwidth]{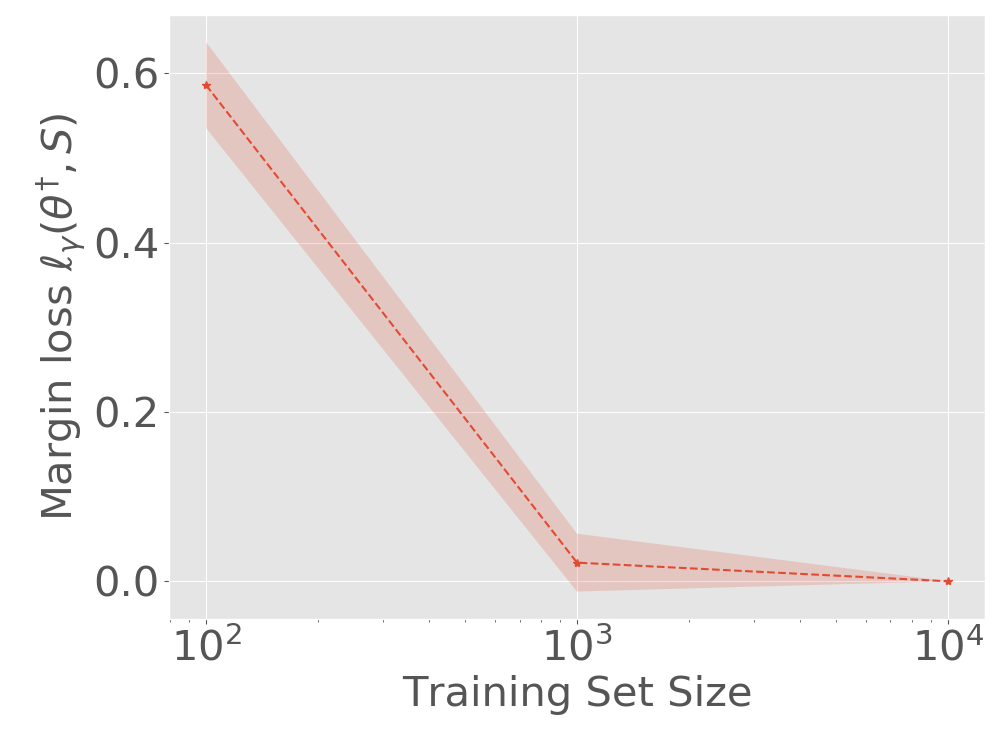}
 } 
  \subfigure[Scale-invariant Generalization Bound.]{
 \includegraphics[width = 0.31 \textwidth, height =0.23\textwidth]{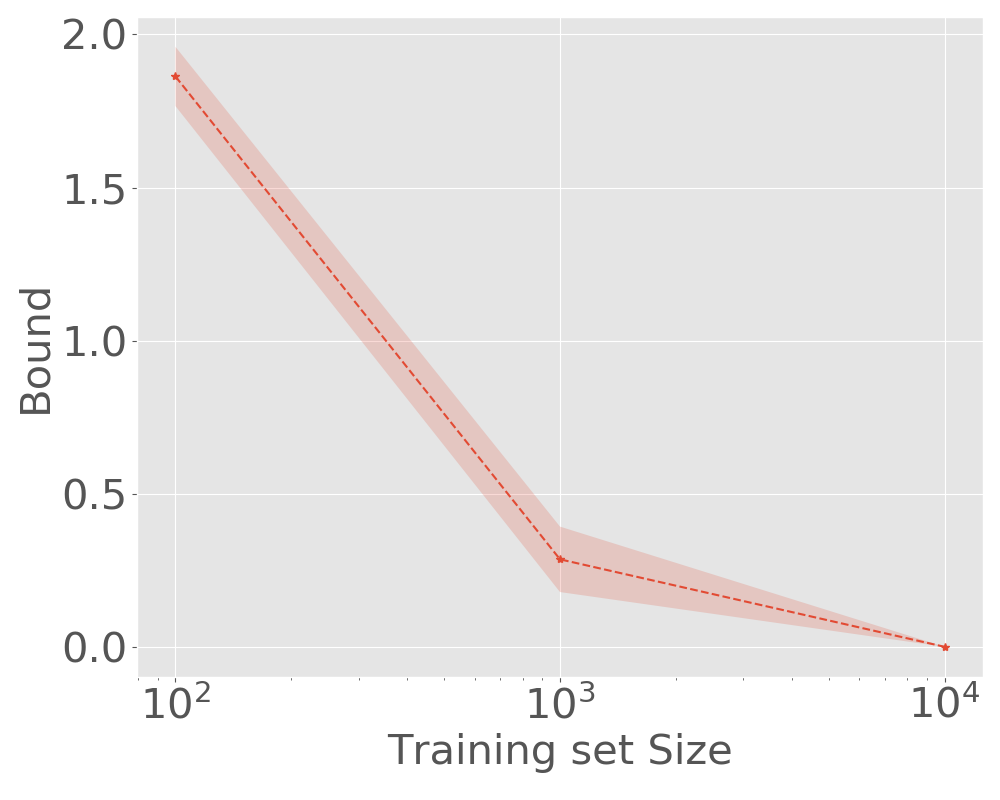}
 } 
\vspace{-2mm}
\caption[]{
Results for ReLU-nets with depth = 2, width =256 total 334,336 parameters, trained on MNIST with batch size = 128.
(a-f) refer to Figure \ref{fig:mnist_sample_d4_l128}.
The bound and all its components decrease with increase in $n$ from 100 to 10,000.
}
\label{fig:mnist_sample_d2_l256}
\vspace*{-4mm}
\end{figure*}

\begin{figure*}[t] 
\centering
 \subfigure[Test Error Rate]{
 \includegraphics[width = 0.31 \textwidth]{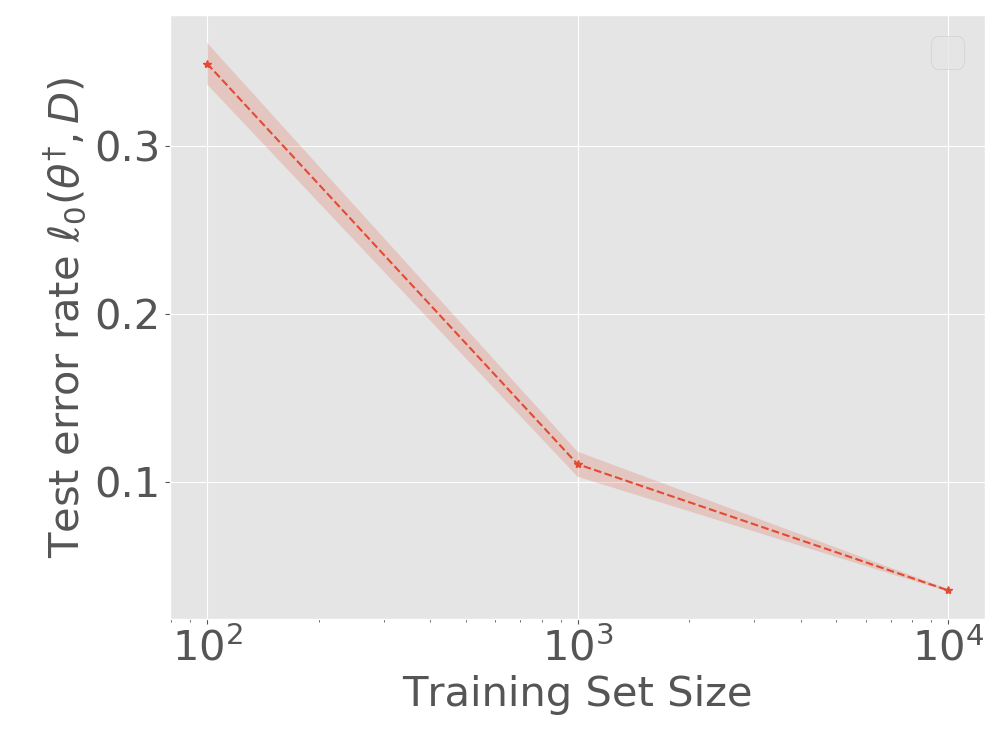}
 } 
 \subfigure[Diagonal Elements of  Hessian.]{
 \includegraphics[width = 0.31 \textwidth, height =0.23\textwidth]{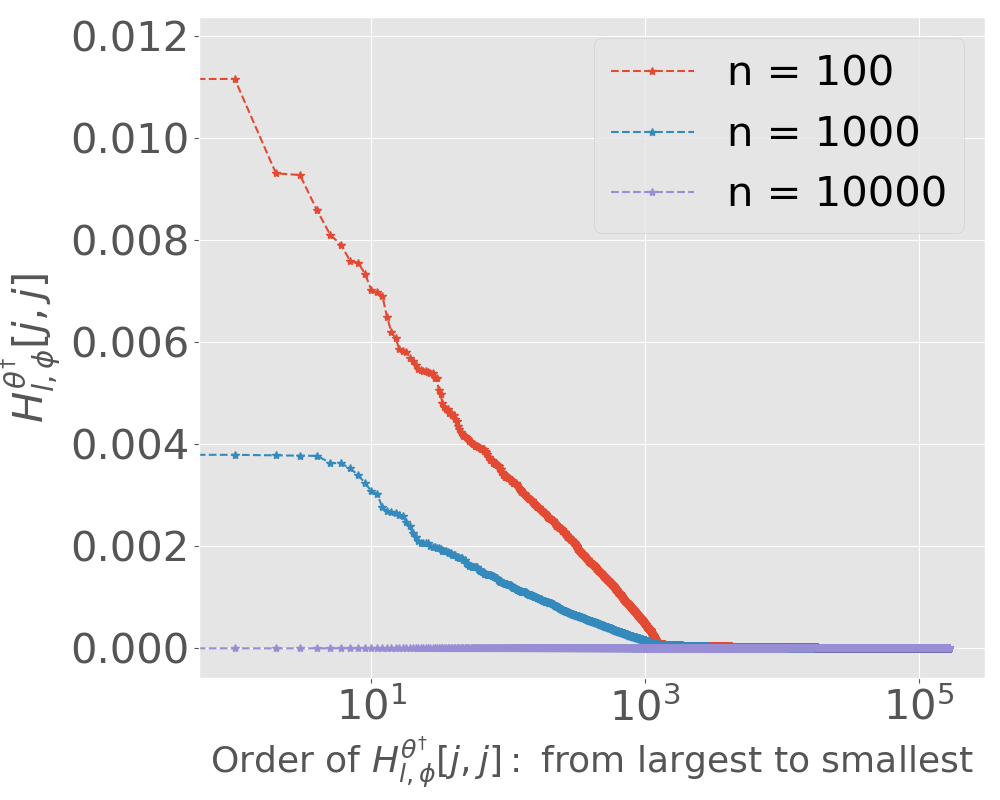}
 } 
 \subfigure[Effective Curvature.]{
 \includegraphics[width = 0.31 \textwidth, height =0.23\textwidth]{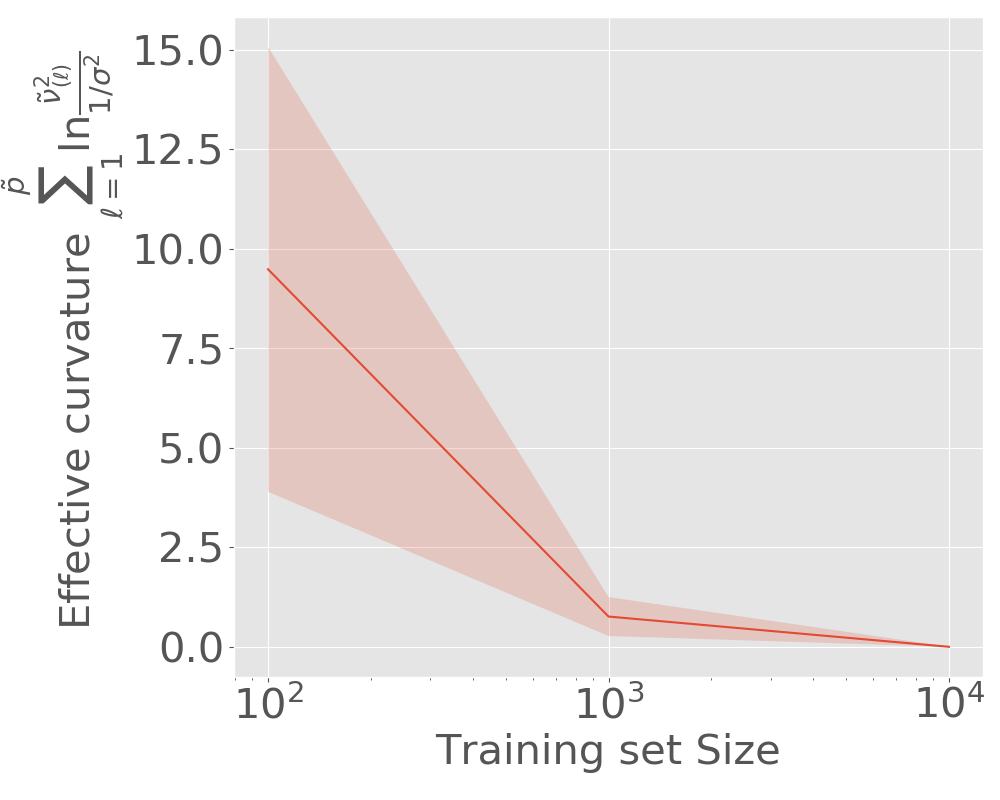}
 } 
 \subfigure[$L_2$ norm / no. sample.]{
 \includegraphics[width = 0.31 \textwidth]{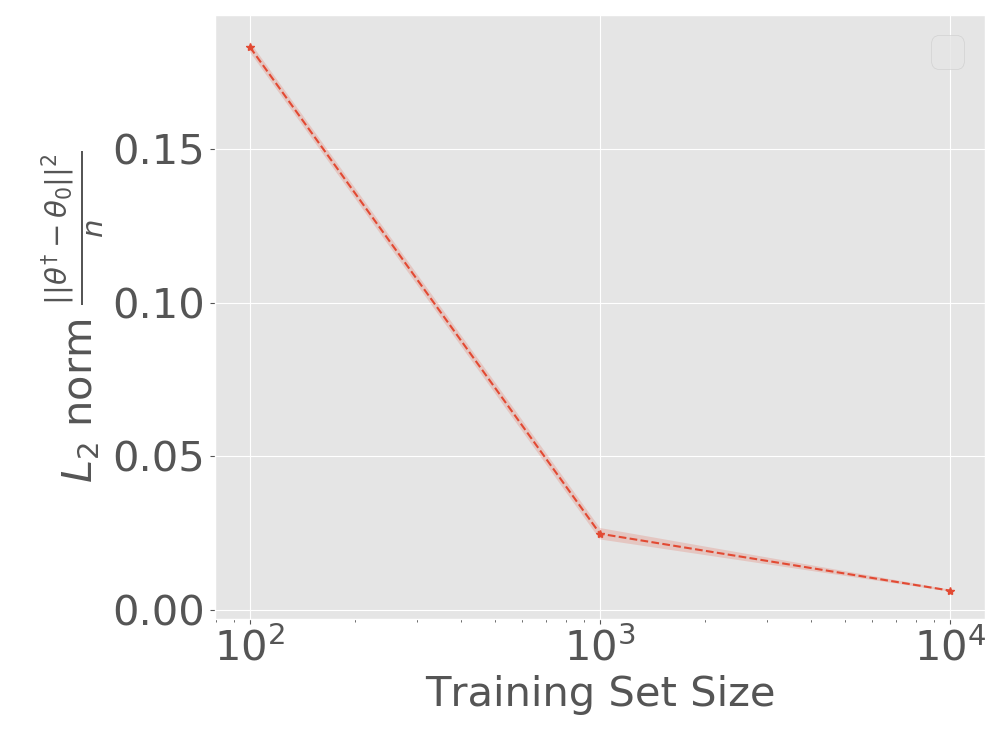}
 } 
  \subfigure[Margin Loss.]{
 \includegraphics[width = 0.31 \textwidth]{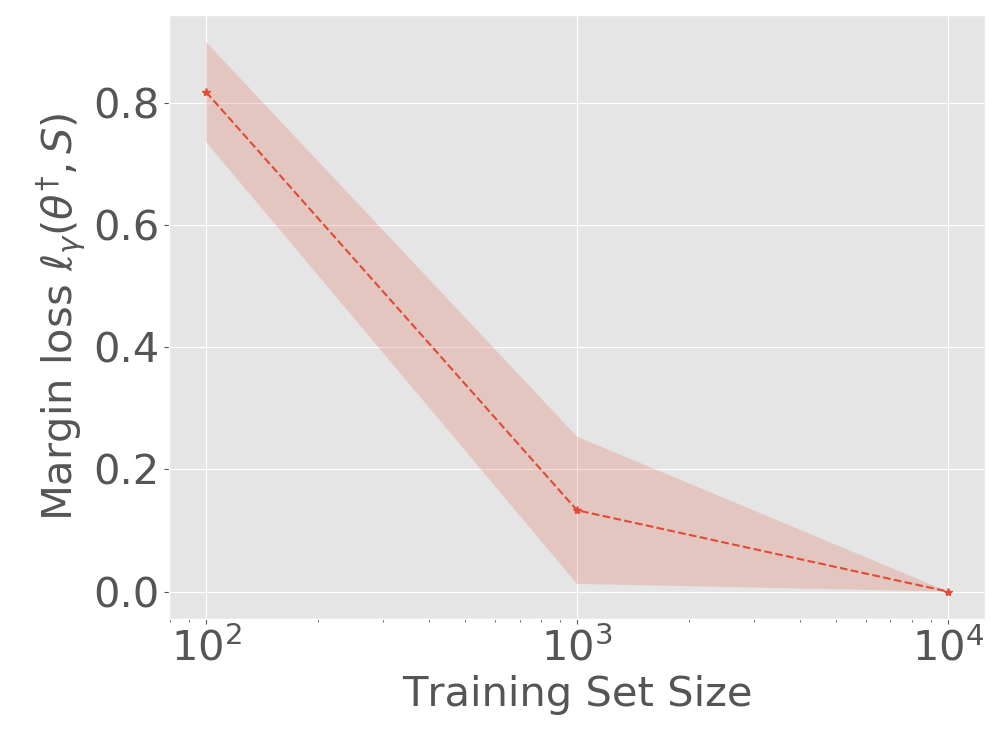}
 } 
  \subfigure[Scale-invariant Generalization Bound.]{
 \includegraphics[width = 0.31 \textwidth, height =0.23\textwidth]{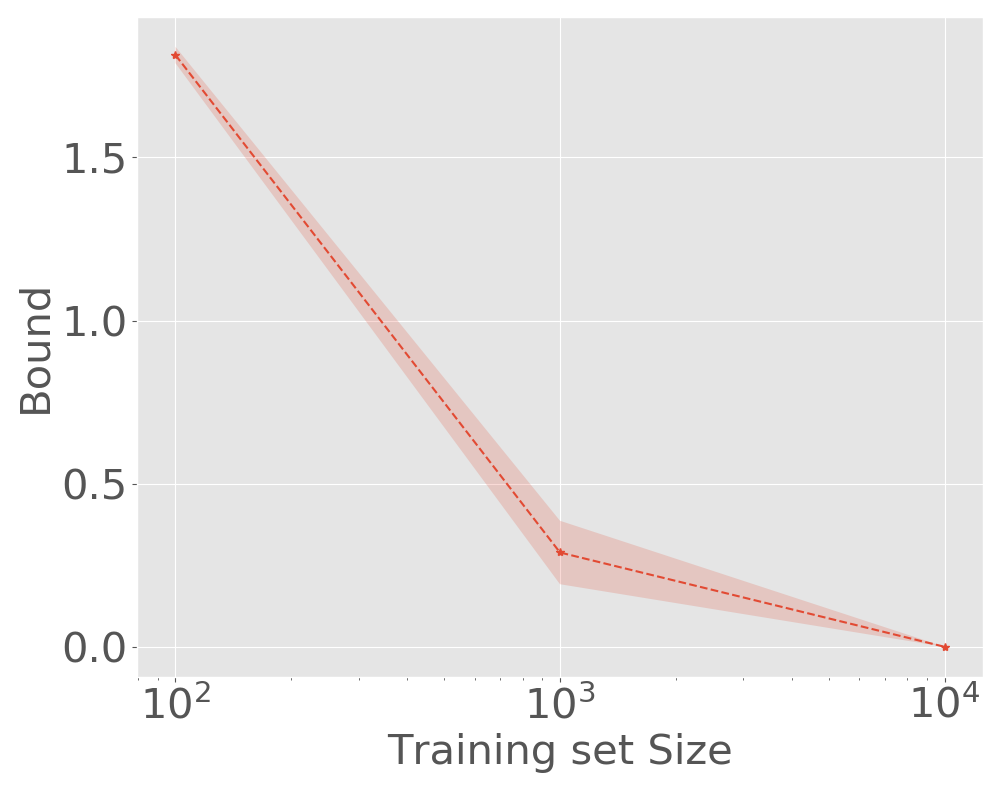}
 } 
\vspace{-2mm}
\caption[]{
Results for ReLU-nets with depth = 4, width =128 total 150,784 parameters, trained on MNIST with batch size = 16.
(a-f) refer to Figure \ref{fig:mnist_sample_d4_l128}.
The bound and all its components decrease with increase in $n$ from 100 to 10,000.
}
\label{fig:mnist_sample_d4_l128_bs16}
\vspace*{-4mm}
\end{figure*}

\noindent {\bf MNIST.} The results for MNIST with ReLU network with depth = 4 and depth = $\{128, 256, 512\}$ are presented in Figure \ref{fig:mnist_sample_d4_l128}, \ref{fig:main_mnist_sample}, and \ref{fig:mnist_sample_d4_l512}, suggesting that our bound holds for ReLU network with different width. The results for MNIST width = 256 and depth = $\{2, 4, 8\}$ are presented in Figure \ref{fig:mnist_sample_d2_l256}, \ref{fig:main_mnist_sample}, and \ref{fig:main_mnist_sample_d8_l256}. The results suggests that our bound is also valid for ReLU network with different depth. We also present the result when training the ReLU network with micro-batch 16 in Figure \ref{fig:mnist_sample_d4_l128_bs16} and \ref{fig:main_mnist_sample_d8_l128_bs16}, which shows that our bound holds for micro-batch training as well.


\begin{figure*}[t] 
\centering
 \subfigure[Test Error Rate]{
 \includegraphics[width = 0.31 \textwidth]{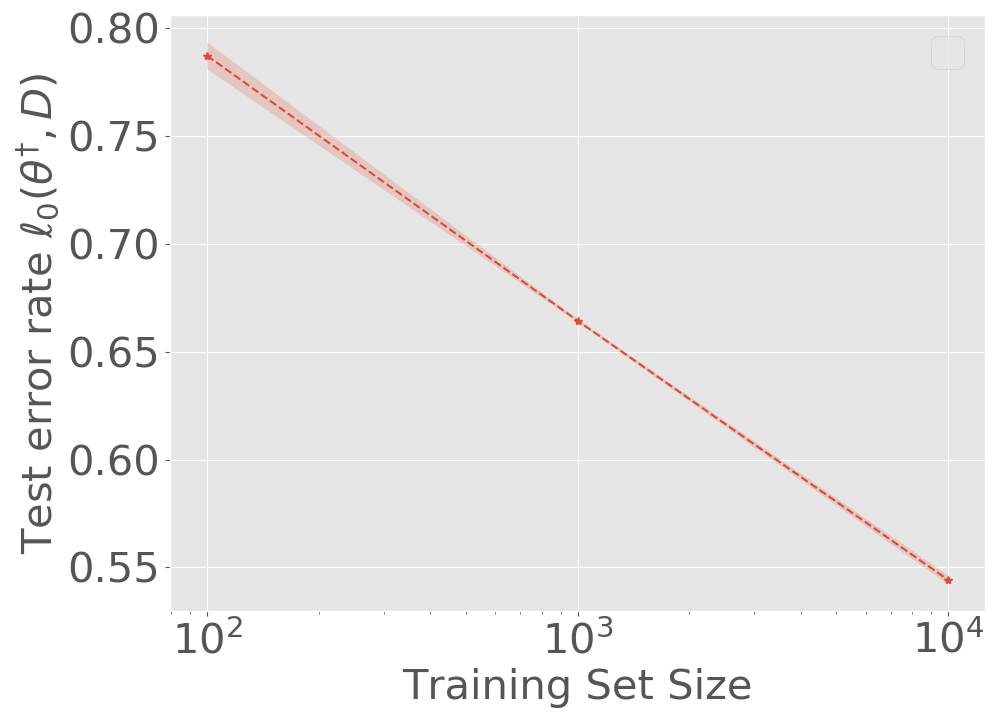}
 } 
 \subfigure[Diagonal Elements of  Hessian.]{
 \includegraphics[width = 0.31 \textwidth, height =0.23\textwidth]{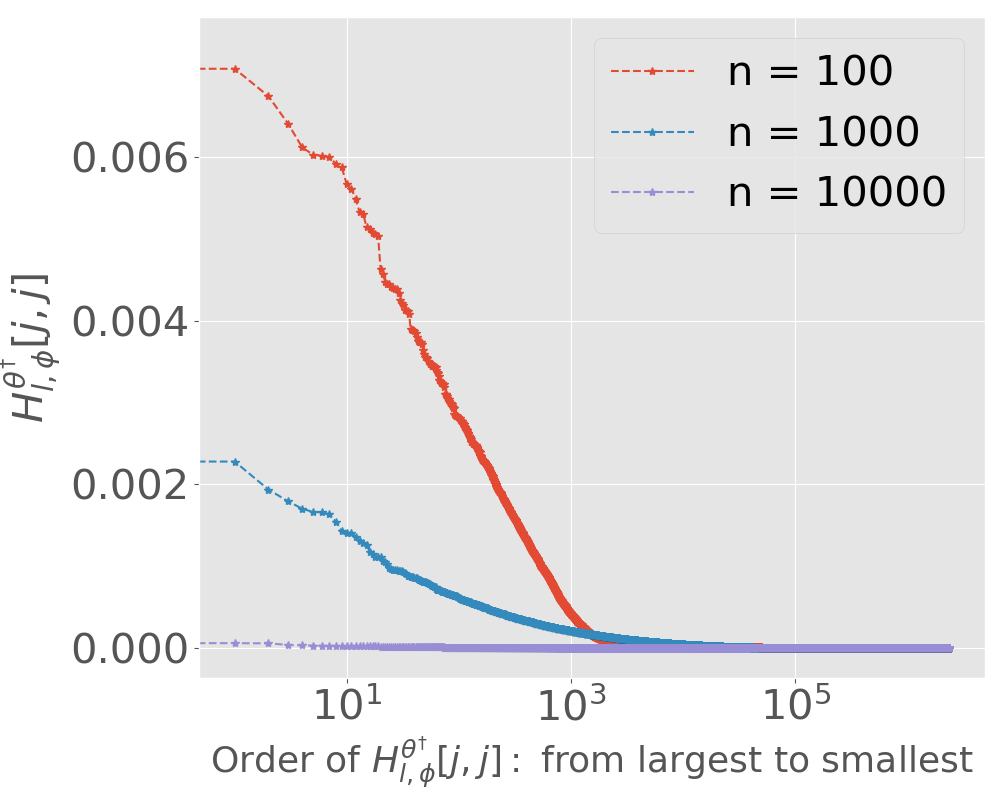}
 } 
 \subfigure[Effective Curvature.]{
 \includegraphics[width = 0.31 \textwidth, height =0.23\textwidth]{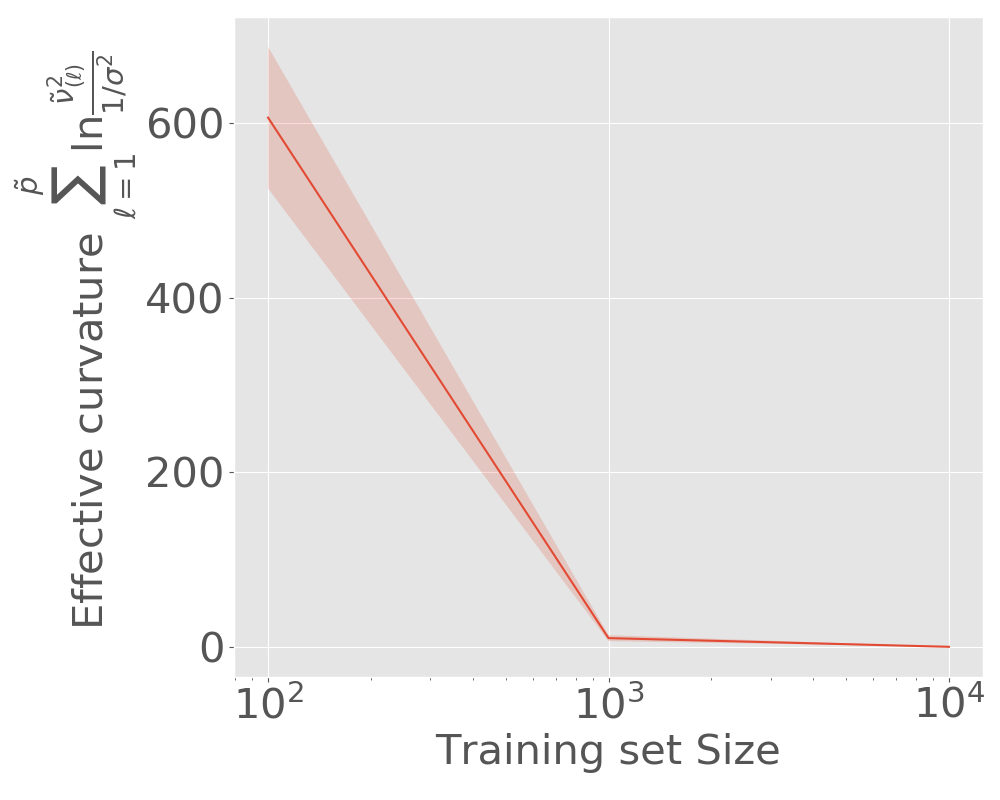}
 } 
 \subfigure[$L_2$ norm / no. sample.]{
 \includegraphics[width = 0.31 \textwidth]{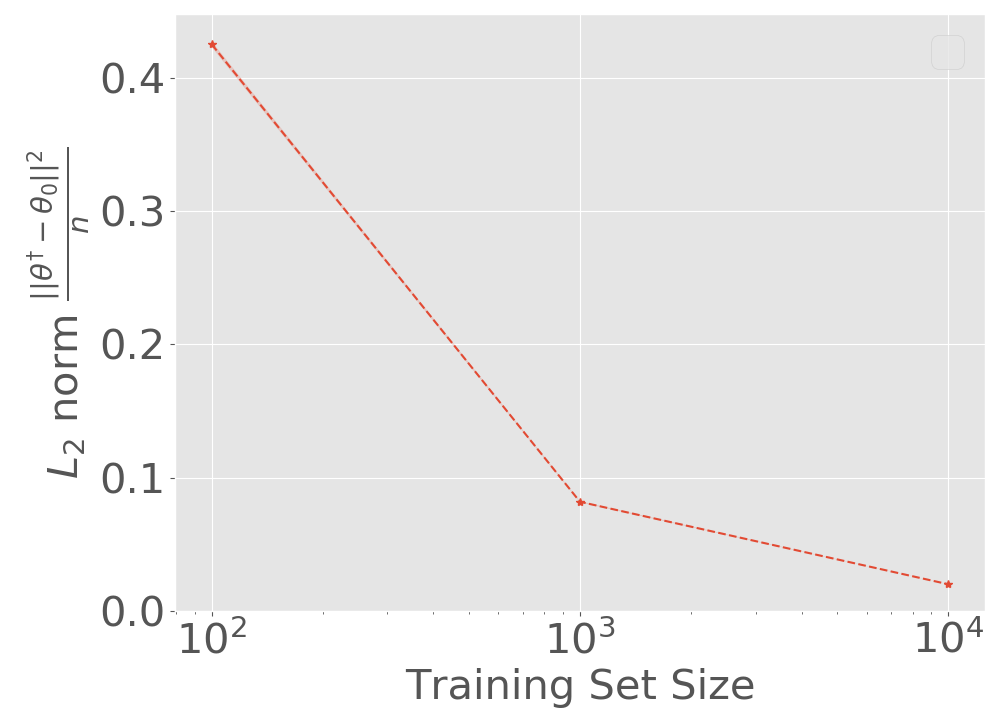}
 } 
  \subfigure[Margin Loss.]{
 \includegraphics[width = 0.31 \textwidth]{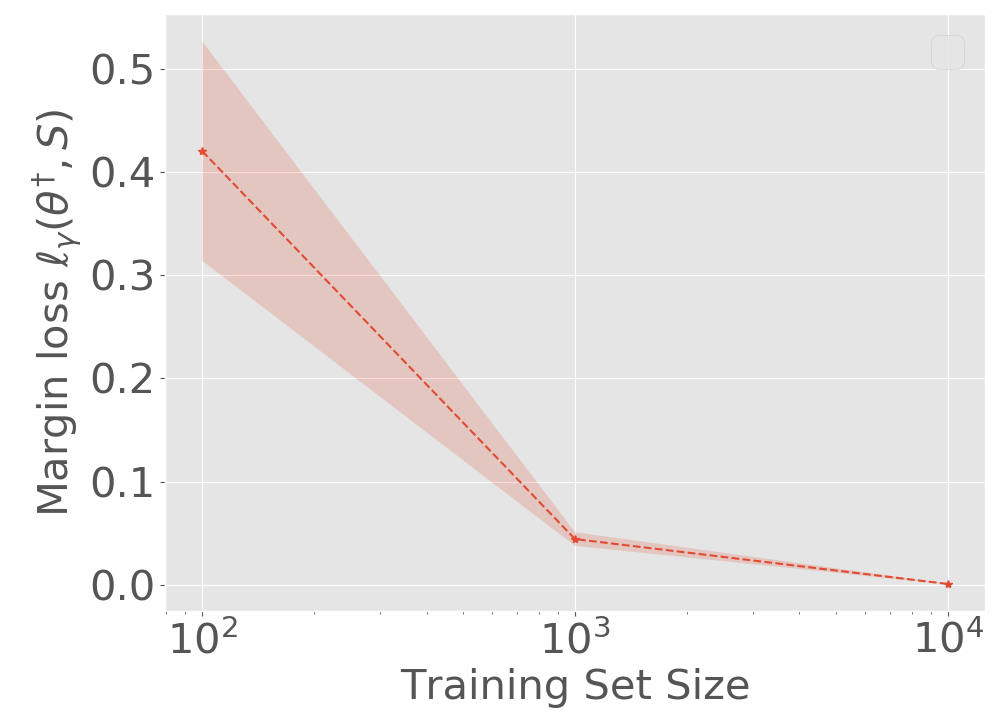}
 } 
  \subfigure[Scale-invariant Generalization Bound.]{
 \includegraphics[width = 0.31 \textwidth, height =0.23\textwidth]{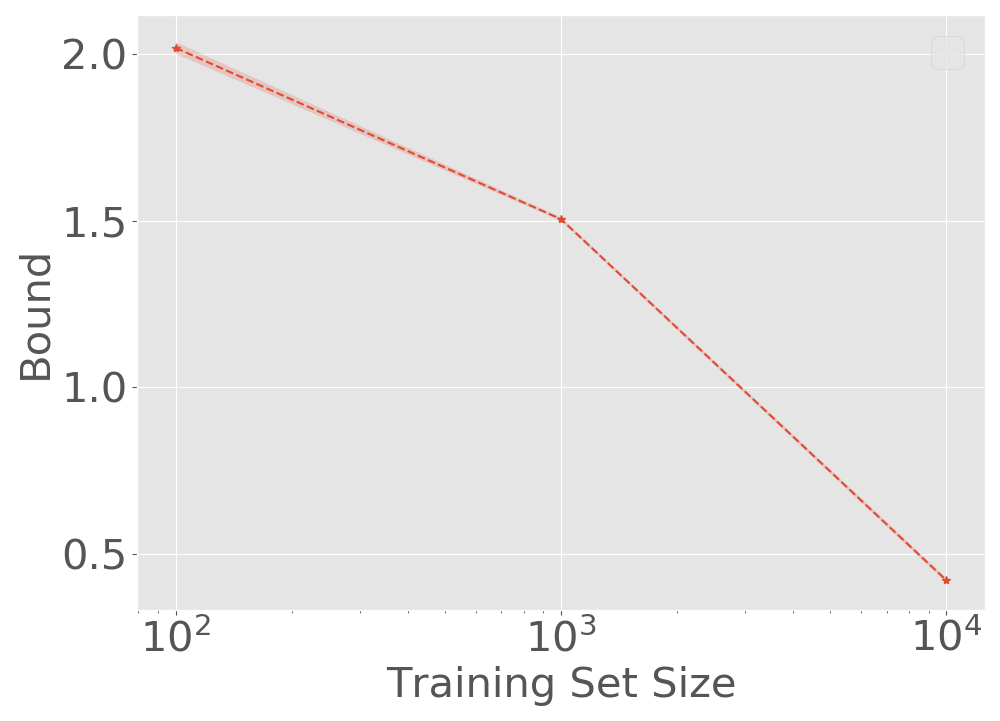}
 } 
\vspace{-2mm}
\caption[]{
Results for ReLU-nets with depth = 8, width =384, trained on CIFAR-10 with batch size = 128.
(a-f) refer to Figure \ref{fig:mnist_sample_d4_l128}.
The bound and all its components decrease with increase in $n$ from 100 to 10,000.
}
\label{fig:cifar_sample_d8_l384}
\vspace*{-4mm}
\end{figure*}

\begin{figure*}[h] 
\centering
 \subfigure[Test Error Rate]{
 \includegraphics[width = 0.31 \textwidth]{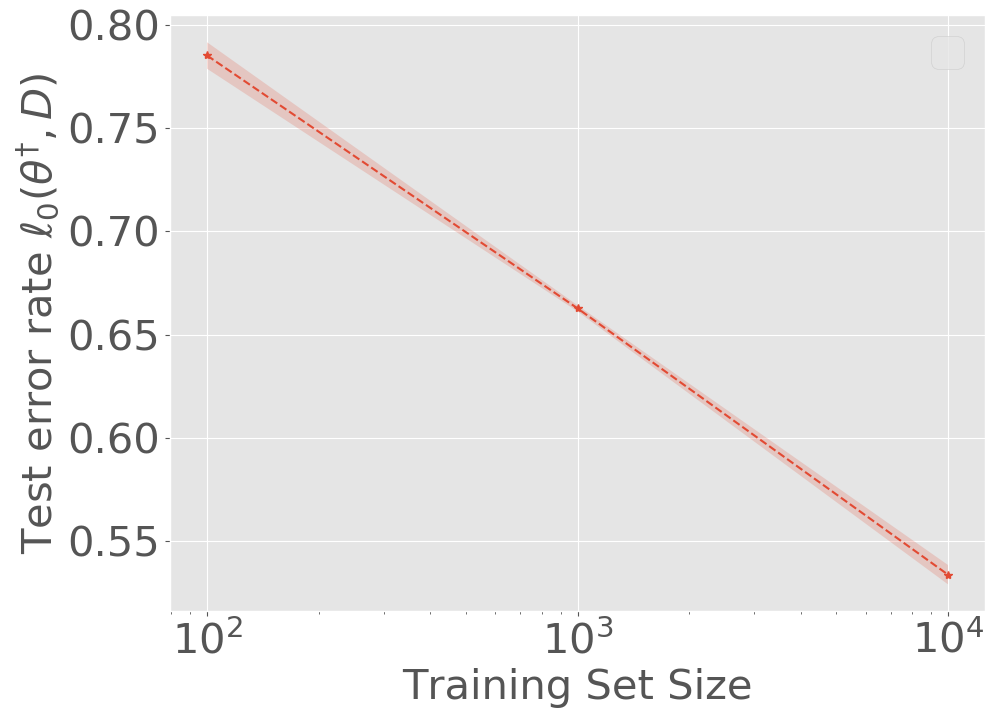}
 } 
 \subfigure[Diagonal Elements of  Hessian.]{
 \includegraphics[width = 0.31 \textwidth, height =0.23\textwidth]{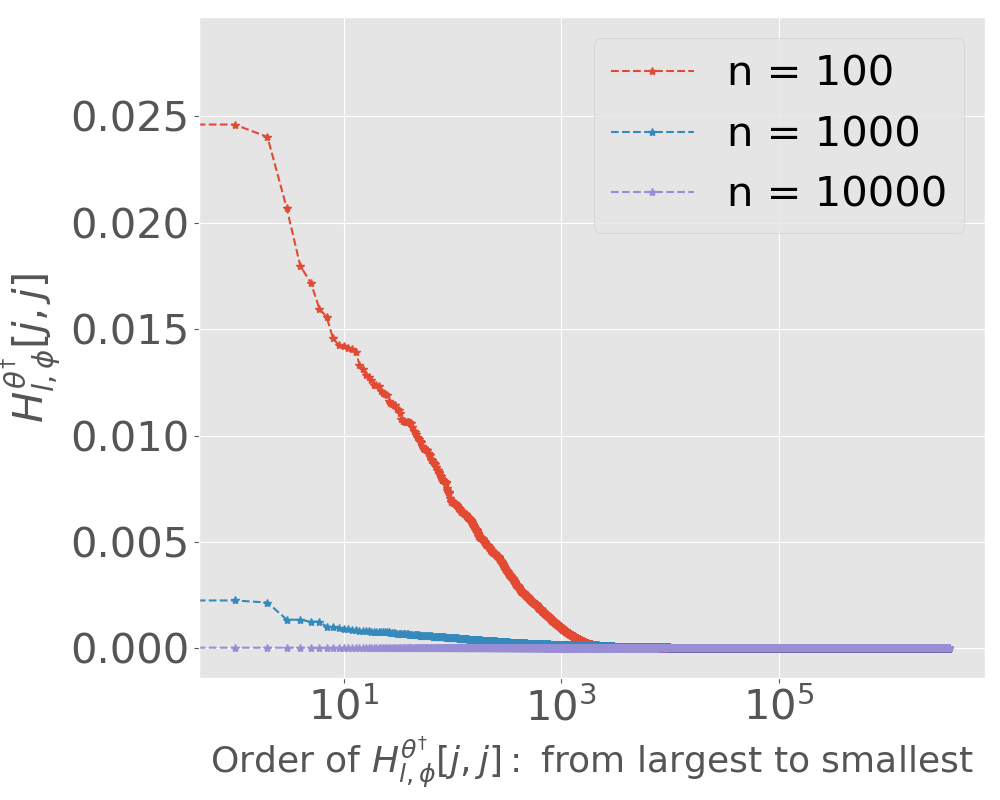}
 } 
 \subfigure[Effective Curvature.]{
 \includegraphics[width = 0.31 \textwidth, height =0.23\textwidth]{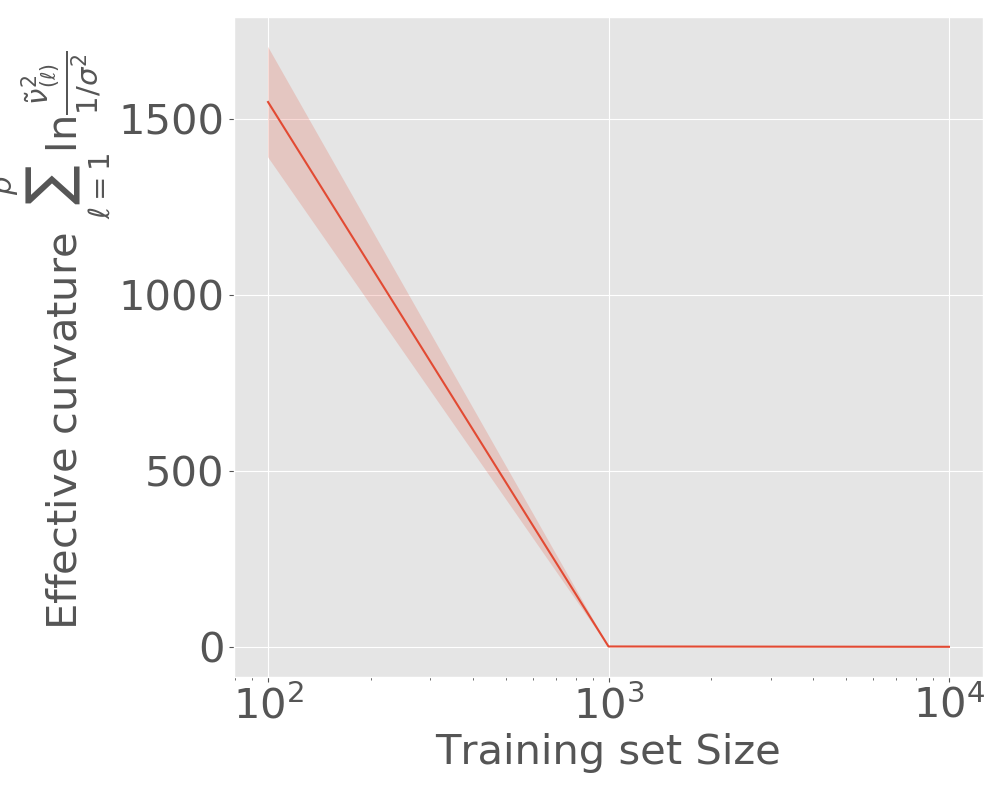}
 } 
 \subfigure[$L_2$ norm / no. sample.]{
 \includegraphics[width = 0.31 \textwidth]{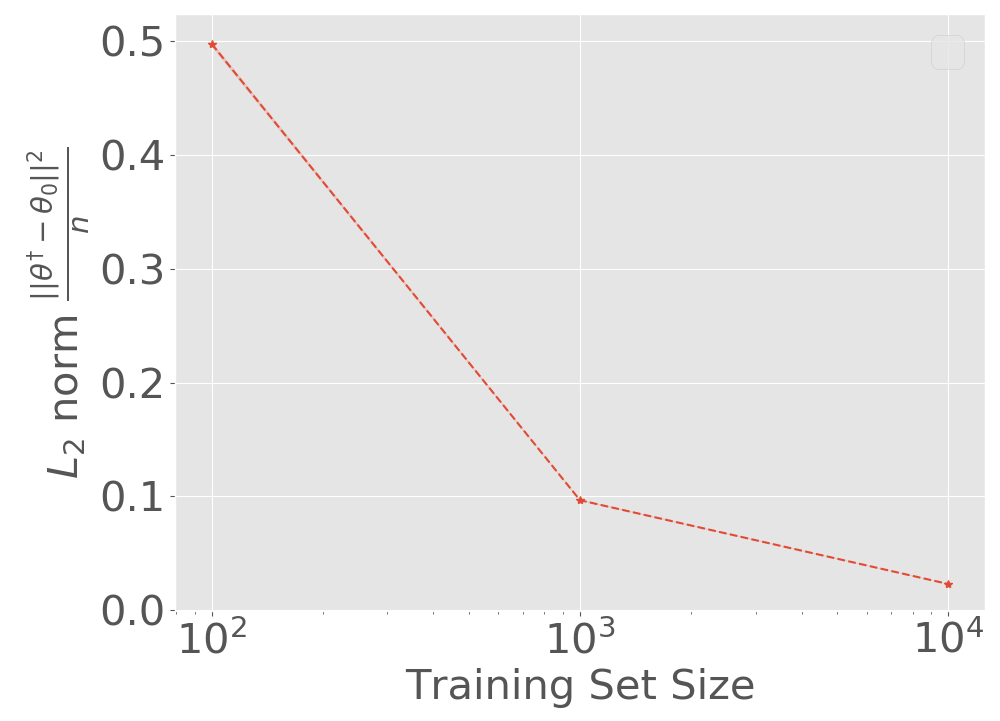}
 } 
  \subfigure[Margin Loss.]{
 \includegraphics[width = 0.31 \textwidth]{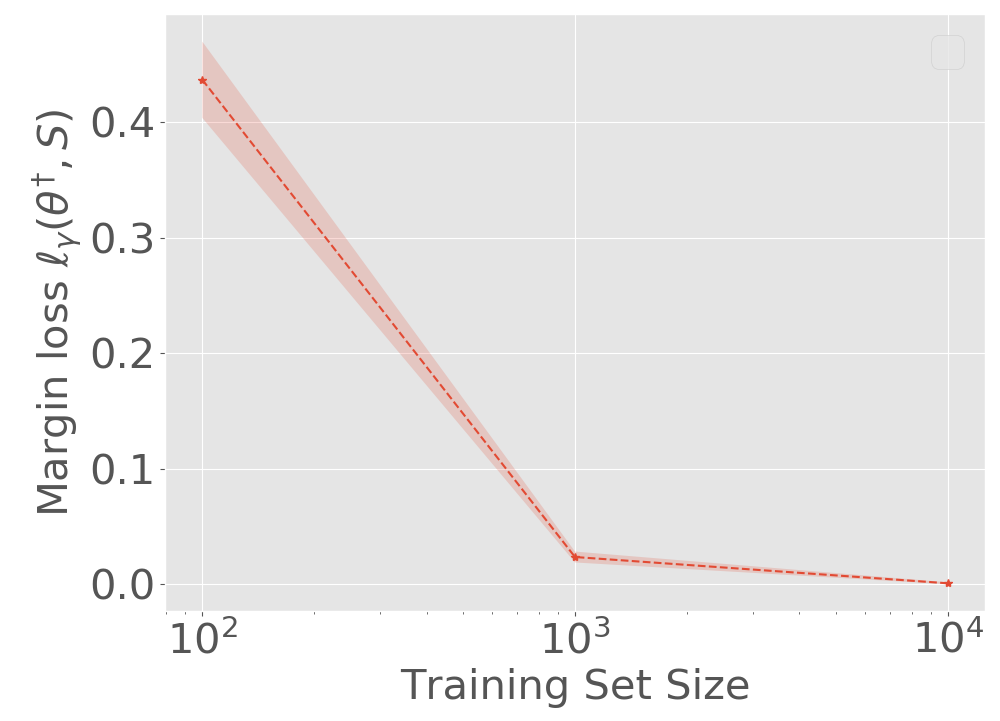}
 } 
  \subfigure[Scale-invariant Generalization Bound.]{
 \includegraphics[width = 0.31 \textwidth, height =0.23\textwidth]{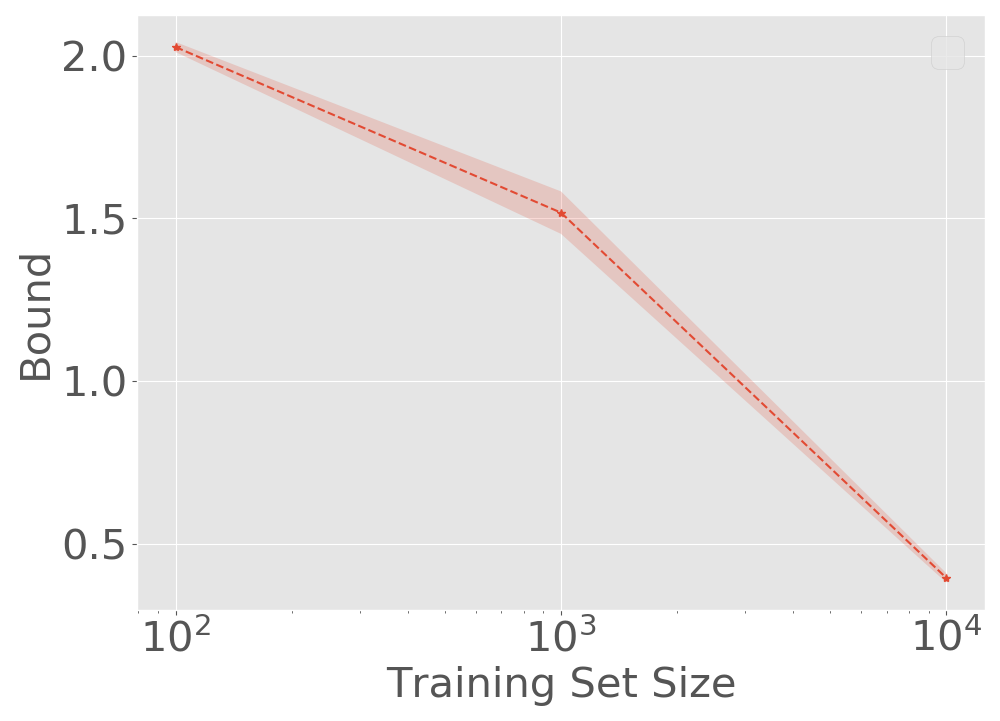}
 } 
\vspace{-2mm}
\caption[]{
Results for ReLU-nets with depth = 8, width =512, trained on CIFAR-10 with batch size = 128.
(a-f) refer to Figure \ref{fig:mnist_sample_d4_l128}.
The bound and all its components decrease with increase in $n$ from 100 to 10,000.
}
\label{fig:cifar_sample_d8_l512}
\vspace*{-4mm}
\end{figure*}

\begin{figure*}[t] 
\centering
 \subfigure[Test Error Rate]{
 \includegraphics[width = 0.31 \textwidth]{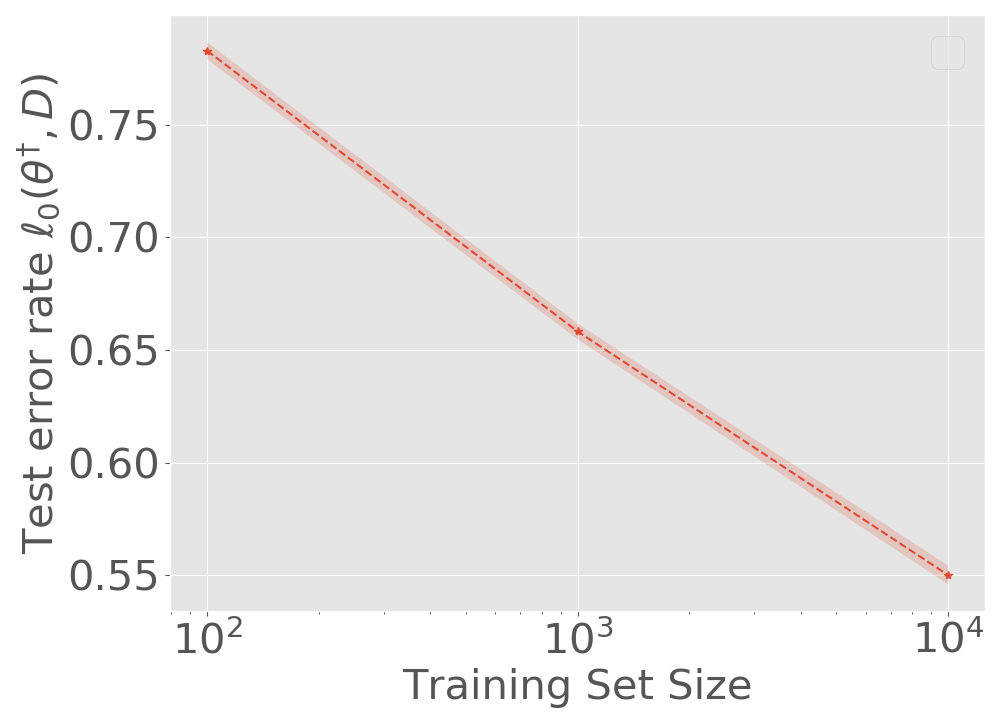}
 } 
 \subfigure[Diagonal Elements of  Hessian.]{
 \includegraphics[width = 0.31 \textwidth, height =0.23\textwidth]{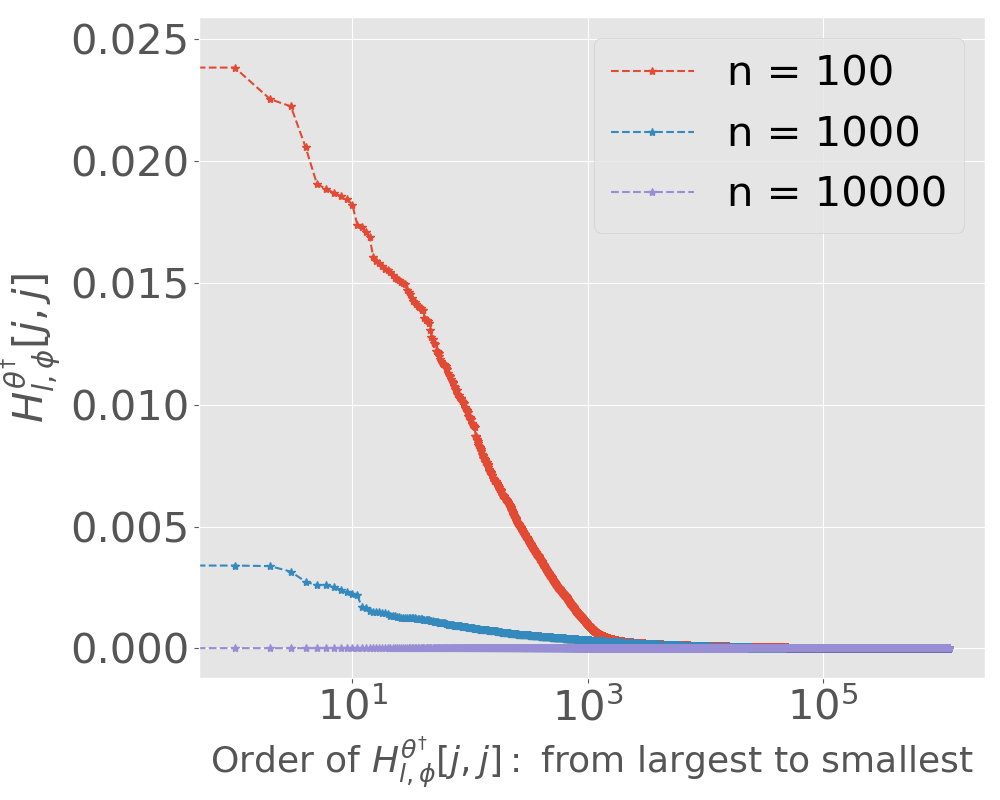}
 } 
 \subfigure[Effective Curvature.]{
 \includegraphics[width = 0.31 \textwidth, height =0.23\textwidth]{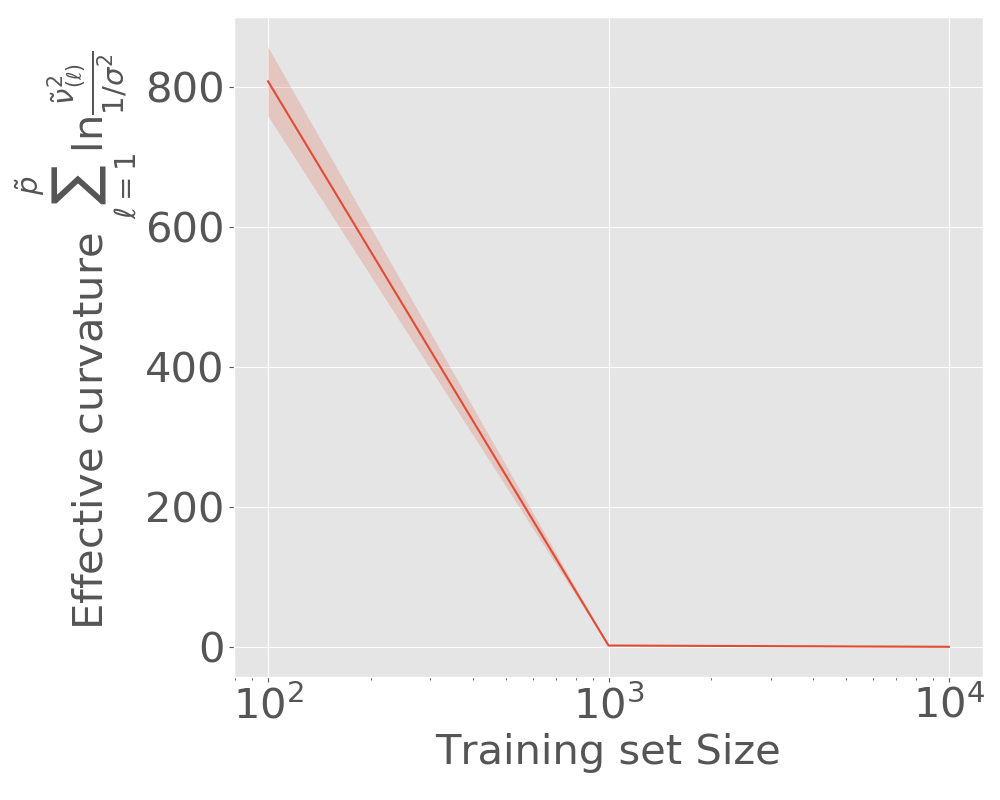}
 } 
 \subfigure[$L_2$ norm / no. sample.]{
 \includegraphics[width = 0.31 \textwidth]{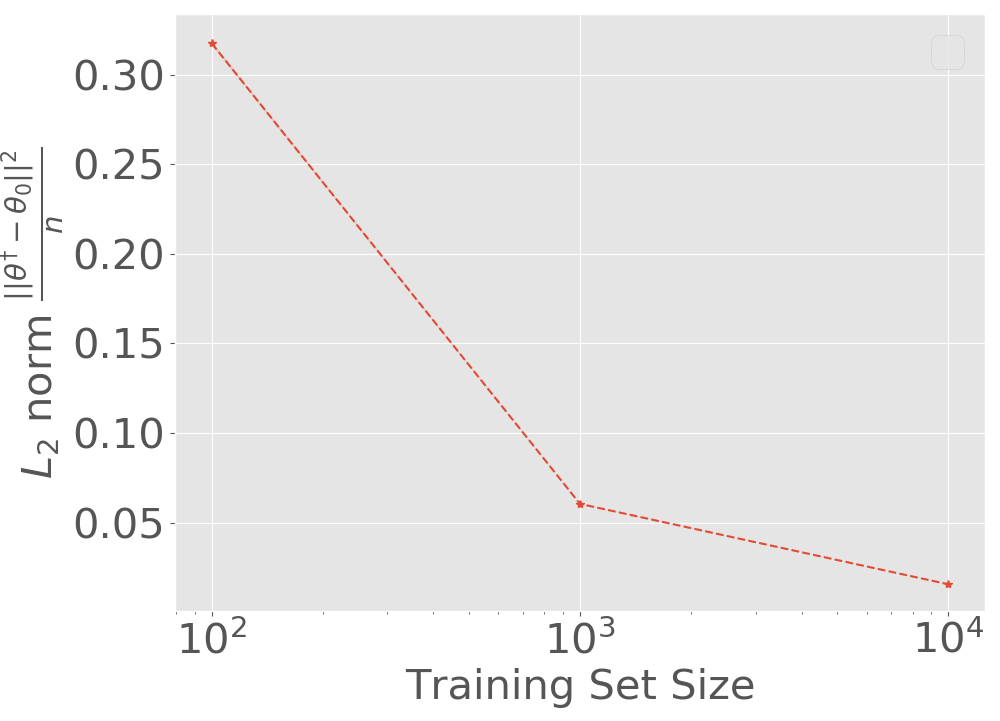}
 } 
  \subfigure[Margin Loss.]{
 \includegraphics[width = 0.31 \textwidth]{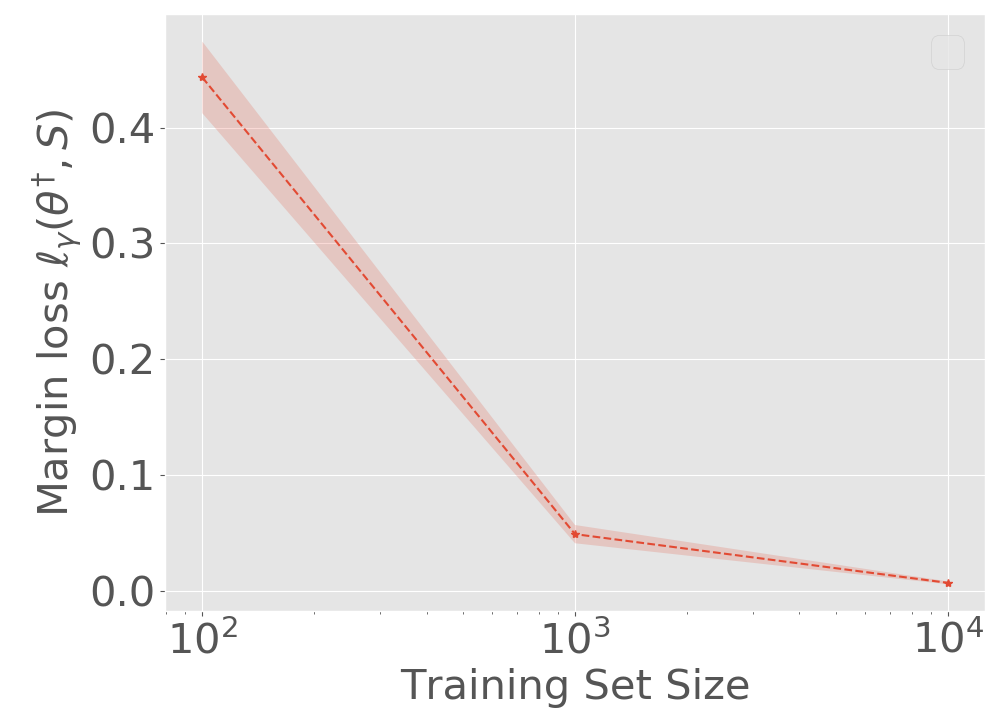}
 } 
  \subfigure[Scale-invariant Generalization Bound.]{
 \includegraphics[width = 0.31 \textwidth, height =0.23\textwidth]{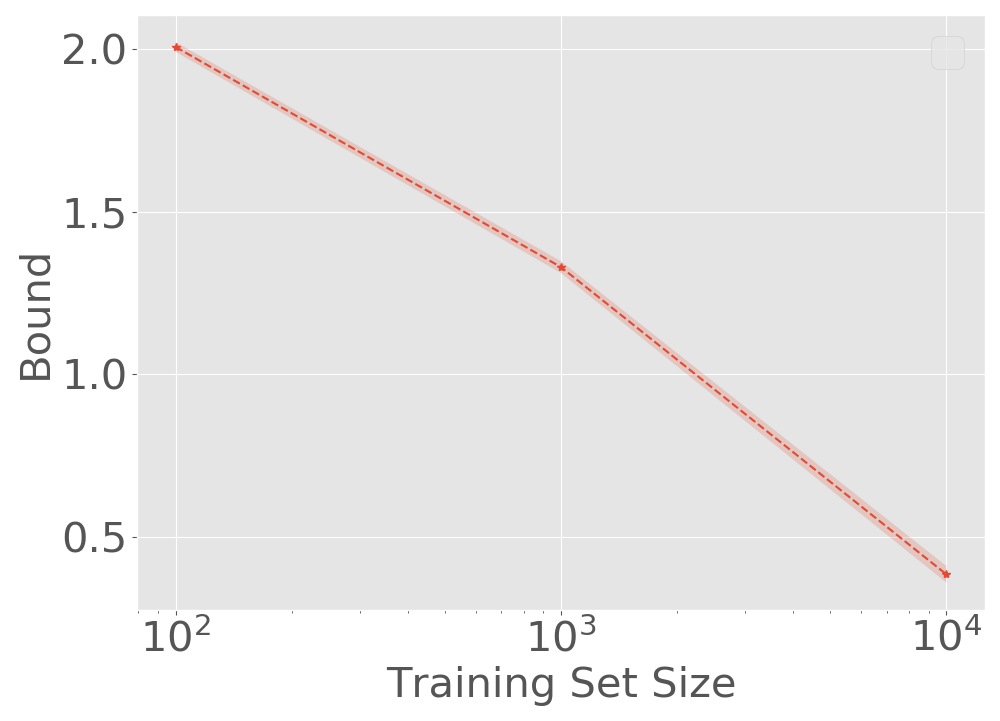}
 } 
\vspace{-2mm}
\caption[]{
Results for ReLU-nets with depth = 6, width =256, trained on CIFAR-10 with batch size = 128.
(a-f) refer to Figure \ref{fig:mnist_sample_d4_l128}.
The bound and all its components decrease with increase in $n$ from 100 to 10,000.
}
\label{fig:cifar_sample_d6_l256}
\vspace*{-4mm}
\end{figure*}

\begin{figure*}[t] 
\centering
 \subfigure[Test Error Rate]{
 \includegraphics[width = 0.31 \textwidth]{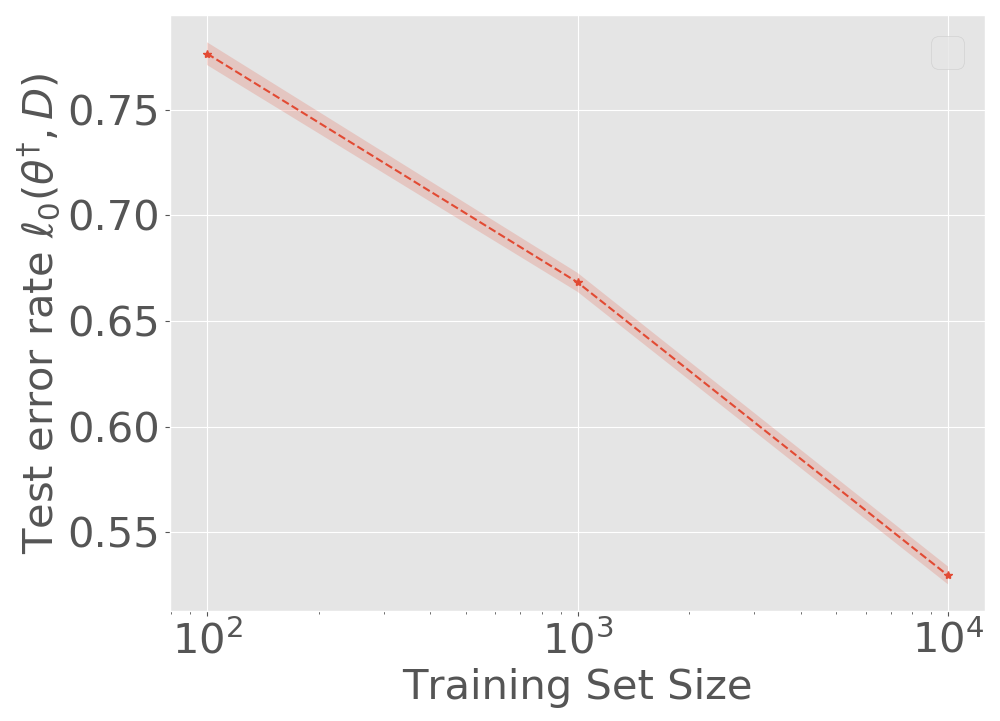}
 } 
 \subfigure[Diagonal Elements of  Hessian.]{
 \includegraphics[width = 0.31 \textwidth, height =0.23\textwidth]{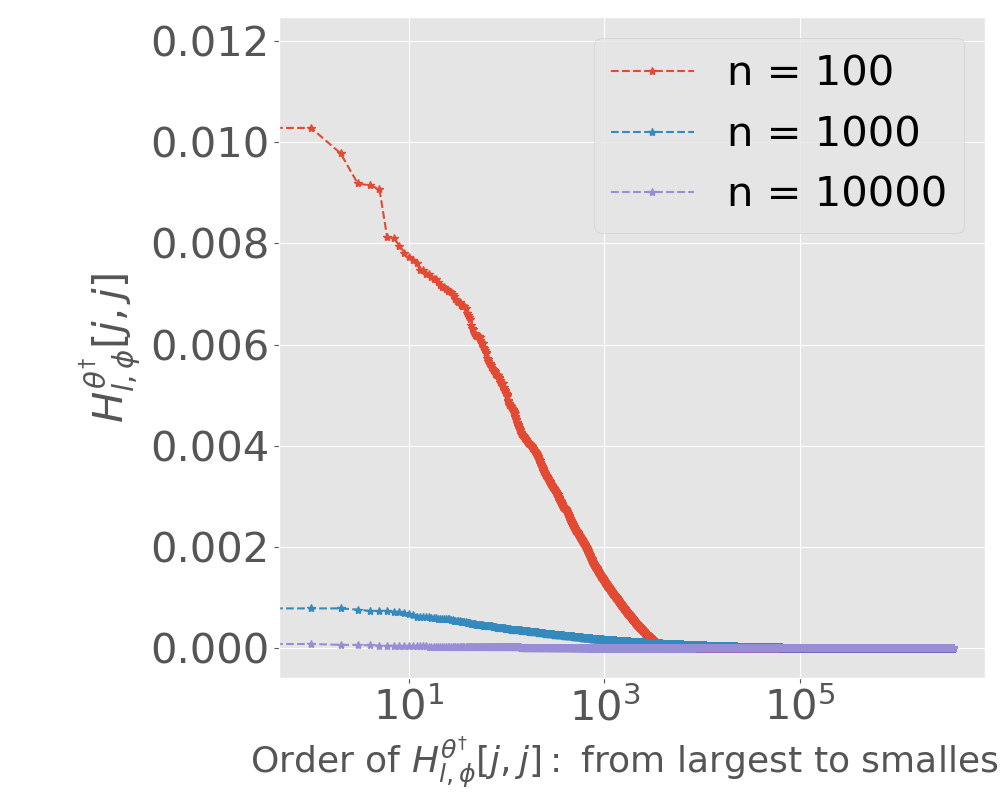}
 } 
 \subfigure[Effective Curvature.]{
 \includegraphics[width = 0.31 \textwidth, height =0.23\textwidth]{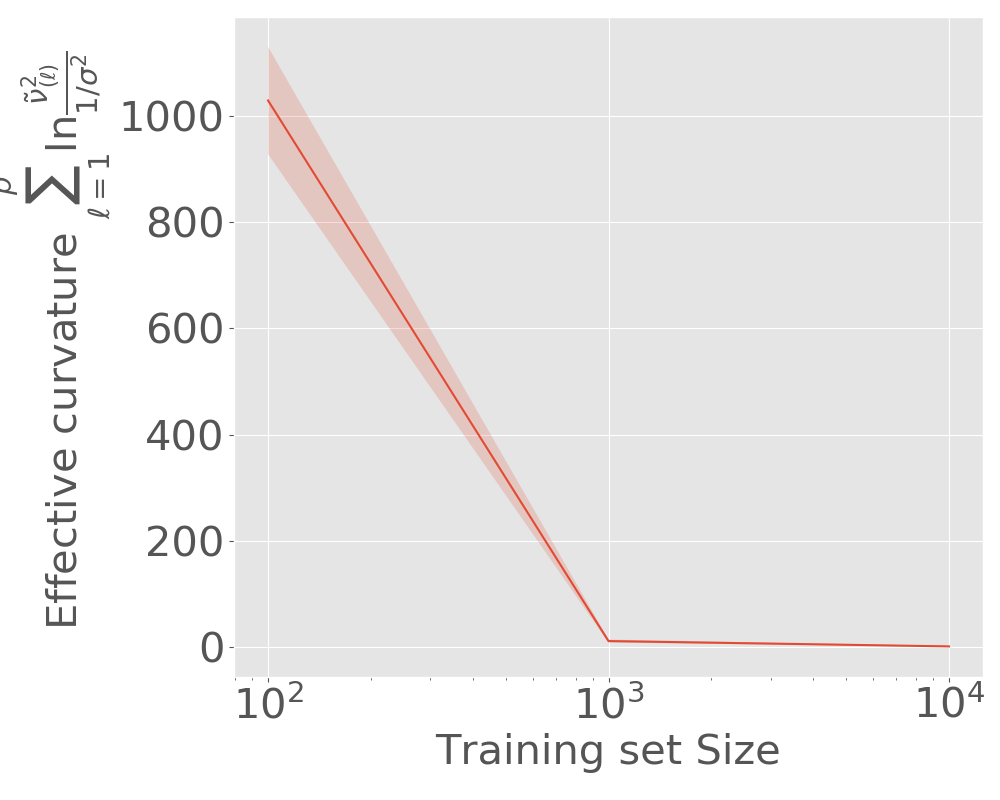}
 } 
 \subfigure[$L_2$ norm / no. sample.]{
 \includegraphics[width = 0.31 \textwidth]{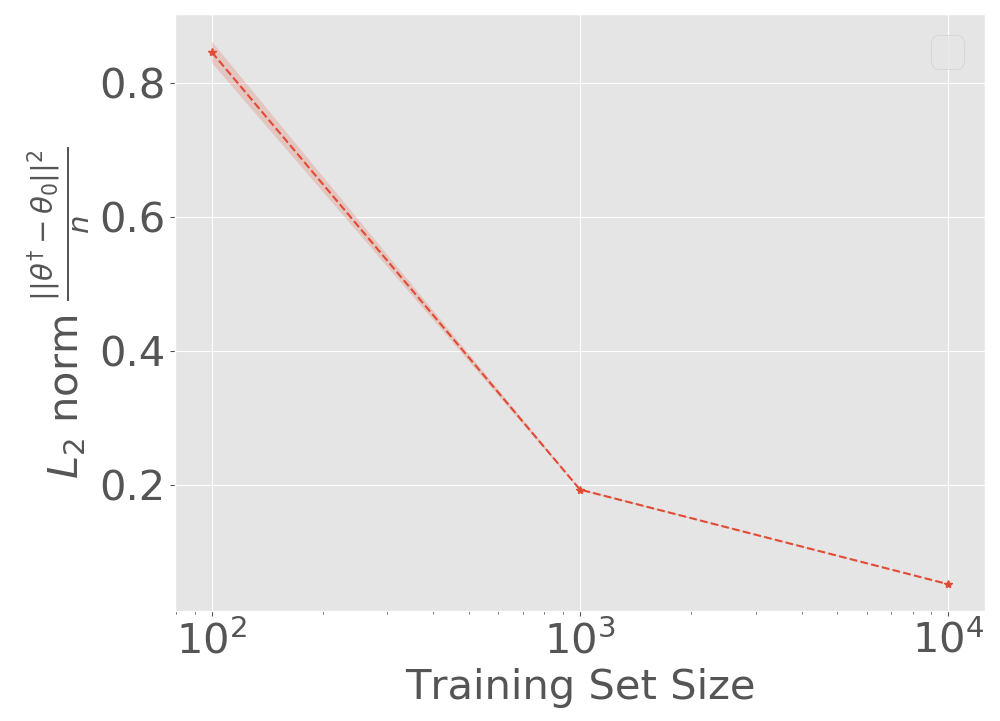}
 } 
  \subfigure[Margin Loss.]{
 \includegraphics[width = 0.31 \textwidth]{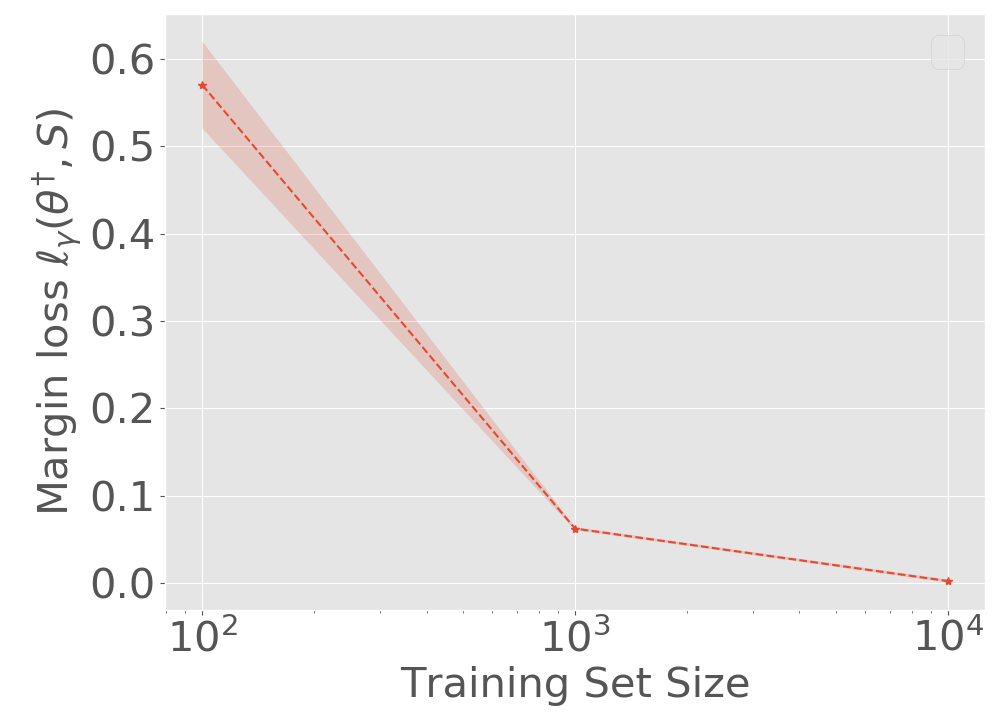}
 } 
  \subfigure[Scale-invariant Generalization Bound.]{
 \includegraphics[width = 0.31 \textwidth, height =0.23\textwidth]{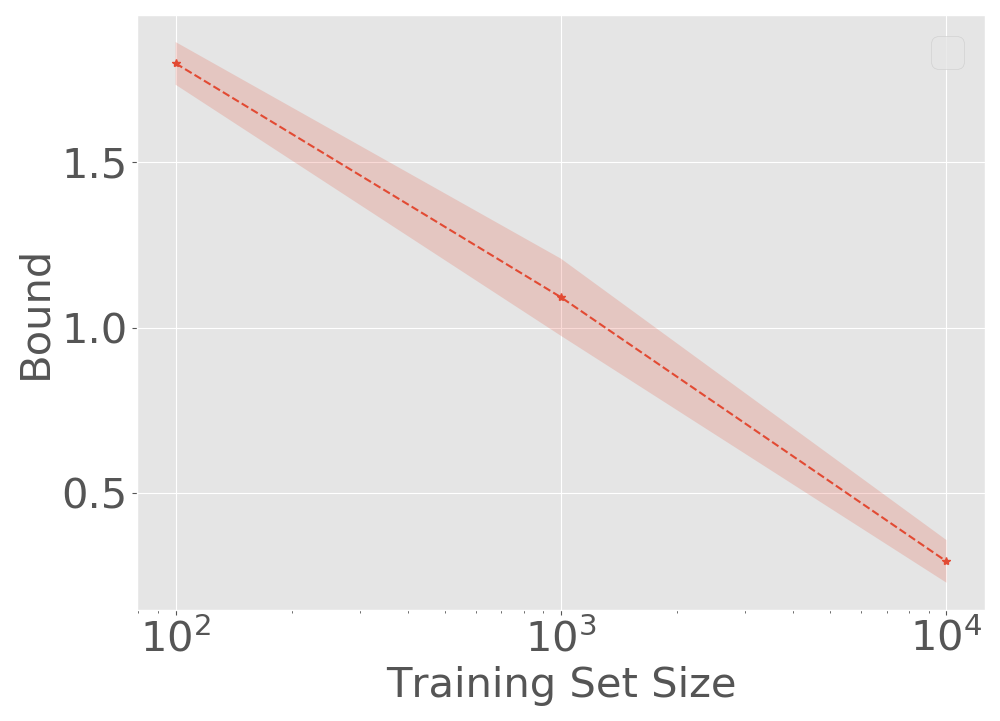}
 } 
\vspace{-2mm}
\caption[]{
Results for ReLU-nets with depth = 8, width =512, trained on CIFAR-10 with batch size = 16.
(a-f) refer to Figure \ref{fig:mnist_sample_d4_l128}.
The bound and all its components decrease with increase in $n$ from 100 to 10,000.
}
\label{fig:cifar_sample_d8_l512_bs16}
\vspace*{-4mm}
\end{figure*}

\noindent {\bf CIFAR-10.} The results for CIFAR-10 with ReLU network with depth = 8 and depth = $\{ 256, 384, 512\}$ are presented in Figure \ref{fig:main_cifar_sample_d8_l256}, \ref{fig:cifar_sample_d8_l384}, and \ref{fig:cifar_sample_d8_l512}, suggesting that our bound holds for ReLU network with different width. The results for CIFAR-10 with width = 256 and depth = $\{4, 6, 8\}$ are presented in Figure \ref{fig:main_cifar_d4_l256}, \ref{fig:cifar_sample_d6_l256}, and \ref{fig:main_cifar_sample_d8_l256}. 
Those results suggests that our bound is also valid for ReLU network with different depth. We also presented the result when training the ReLU network with micro-batch 16 in Figure \ref{fig:main_cifar_sample_d8_l256_bs16} and \ref{fig:cifar_sample_d8_l512_bs16}, which shows that our bound stays valid for micro-batch training as well.


\end{document}